Master’s Thesis

# Developing a Strong Pre-Trained Base Model for Plant Leaf Disease Classification

Department of Artificial Intelligence Convergence

Graduate School, Chonnam National University

David Jona Richter

August 2025

# Developing a Strong Pre-Trained Base Model for Plant Leaf Disease Classification

Department of Artificial Intelligence Convergence

Graduate School, Chonnam National University

David Jona Richter

Supervised by Professor Kyungbaek Kim

A dissertation submitted in partial fulfillment of the requirements for the
Master of Science in Artificial Intelligence Convergence

Committee in Charge:

Yeong-Jun Cho ____________________

Seok-Bong Yoo ____________________

Kyungbaek Kim ____________________

August 2025

# Preface

This thesis was written as part of a Digital Agriculture project that our laboratory (DNS Lab at Chonnam National University) was tasked with. As part of the multi-year project, this thesis covers the first year of this project's work and will therefore be continued in future years. As such, in this first year, we set ourself the goal of finding the most suitable datasets, models and methods for plant leaf disease classification as well as build a foundation model for the field.

# Contents

# List of Figures

# List of Tables

# List of Abbreviations

**BM** Base Model

**CA** Channel Attention

**CBAM** Convolutional Block Attention Module

**CNN** Convolutional Neural Network

**DL** Deep Learning

**DSC** Depthwise Separable Convolution

**DwC** Depthwise Convolution

**FAO** Food and Agriculture Organization of the United Nations

**FtC** Factorized Convolutions

**FL** Federated Learning

**FM** Foundational Model

**FSL** Few-Shot Learning

**FT** Fine-Tuning

**GAN** Generative Adversarial Network

**HSI** Hyperspectral Imaging

**MIB** Mobile Inverted Bottleneck

**ML** Machine Learning

**MSP** Multi-Scale Processing

**OSL** One-Shot Learning

**PwC** Pointwise Convolution

**RL** Reinforcement Learning

**SA** Spatial Attention

**SE** Squeeze & Excitation

**SVM** Support Vector Machine

**TL** Transfer-Learning

**UAV** Unmanned Aerial Vehicle

**ViT** Vision Transformer

# List of Terms

**Base Model** A pre-trained model that can be used as a baseline to be trained on a new dataset, allowing the model to benefit from the pre-trained weights and learn faster and more robust.

**Channel Attention** An attention mechanism for Convolutional Neural Network (CNN) models that operates on the channel axis. This method uses Global Average Pooling and Global Max Pooling, before feeding them into separate Autoencoders, adding the results and creating importance weights via sigmoid to weigh channel by their importance.

**Convolutional Block Attention Module** This block combines Channel Attention (CA) and Spatial Attention (SA) by stacking them on top of each other, CA first then SA. This allows the Convolutional Block Attention Module (CBAM) to identify the importance of feature maps across the channel axis as well as the width and height dimensions.

**Convolutional Neural Network** A Deep Learning Model type that is commonly used for image data, as it uses spatially aware and parameter efficient filters to iterate over the pixels.

**Deep Learning** A Machine Learning method that uses Deep Neural Networks to learn. Deep Learning models can handle large and high dimensional data and generalize upon it. These models can also handle numerous tasks, from simple classification and clustering, to fully generative models that can produce high quality imagery for example.

**Depthwise Convolution** Depthwise convolutions are a type of convolutional layer that applies a filter to only 1 channel at a time. While regular 2D convolution kernels are as deep as the input layer, Depthwise Convolution (DwC) are only 1 channel deep and only applied to one. These layers are often followed by Pointwise Convolution (PwC).

**Depthwise Separable Convolution** Depthwise Separable Convolution combine the ideas of Depthwise Convolution and Pointwise Convolution by stacking them on top of each other. Like this, they form a parameter efficient alternative to regular convolutions, that still consider information across height, width, and channels by combining DwC (height and width) and PwC (channel).

**Factorized Convolutions** A cconvolutional method that splits convolutional kernels into two 1D kernels instead of a 2D kernel when considering a single channel. A 3x3 kernel would for example be replaced by a 3x1 and a 1x3 kernel. This is done to reduce parameter count.

**Federated Learning** A Distributed Methodology to Deep Learning. Is uses decentralized data which provides high security and privacy. Local models get trained on local data and are then aggregated on a central server to a global model and redistributed.

**Few-Shot Learning** A way of domain adaptation using limited data per class to generate class embeddings, which can then be used to compare to future test images and predict classes. No backpropagation is needed, embeddings are generated in forward runs after the global average pooling layer.

**Fine-Tuning** Fine-Tuning refers to the process of Transfer-Learning with un-frozen weights. This can be carried out on its own, but also often sees application after frozen TL has already concluded.

**Food and Agriculture Organization of the United Nations** The United Nations agency tasked with defeating world hunger and improving food security.

**Foundational Model** A large scale Deep Learning (DL) model with a large number of parameter that was trained on a large scale dataset. Foundational Models are often used for TL tasks.

**Generative Adversarial Network** A Generative Adversarial Network is a generative DL method that train a generator network that generates synthetic images and a discriminator network that is trained to classify these generated images against a real dataset, which leads to both networks competing, through which the generator will be able to generate images that are almost indistinguishable from the dataset.

**Hyperspectral Imaging** Hyperspectral Imaging describes the imaging process in which hundreds of wavelengths (in the visible and near-infrared spectrum) are captured and stored for each pixel (unlike the usual three (RGB) in regular photography). These extra channels contain a lot of extra information, especially relevant in the field of plant diseases.

**Machine Learning** Statistical Methods or Algorithms that allow computers to learn without implicit instructions from data in a way that generalizes to before unseen data.

**Mobile Inverted Bottleneck** A DL method used in CNN that uses PwC to inflate the number of channels, before using a a depthwise separable block of DwC followed by PwC.

**Multi-Scale Processing** A DL method used in CNN where different sized kernels are present in an architecture, often in parallel, to observe the image and its features at different scales and receptive fields.

**One-Shot Learning** A method for domain adaptation using extremely limited data, with only 1 image per class to generate class embeddings, which are then used to compare to future test image embeddings to predict classes. No backpropagation is needed, embeddings are generated in forward runs after the global average pooling layer and remebered by the model.

**Pointwise Convolution** A convolutional layer method where the kernel is of size 1x1 but parses over each channel. Often used to either bottleneck CNN, re-arrange channel features, or after DwC to take into account information across channels.

**Reinforcement Learning** A branch of deep learning. Reinforcement Learning (RL) agents are not instructed how to operate, but are rather given an objective through a reward function that they need to optimize by choosing actions based on the state and rewards that influence the environment to reach said goal.

**Spatial Attention** Spatial Attention is a attention mechanisms that focuses on spatial features on the dimensions of width and height. To do that it uses Max Pooling reduction and Average Pooling reduction across all channel to condense them to 1 channel each, before a sigmoid activated convolutional layer estimates importance weights for each pixel that will be multiplied to the features maps.

**Squeeze & Excitation** Squeeze & Excitation is a attention block that operates on the channel dimension. It uses Global Average Pooling to reduce the channels to one value per channel, before using an Autoencoder setup to reduce layers to their essential information and then a sigmoid layer to rank layers by importance.

**Support Vector Machine** A Support Vector Machine is a Machine Learning method that classifies data by optimizing a line or hyperplane in an N-dimensional space, by maximizing the distance between the classes.

**Transfer-Learning** Transfer-Learning is a Deep Learning technology, often used with image data and CNN, in which a pre-trained network architecture and its weights are downloaded and then applied to train a model using said weights to learn on a new and different dataset. Often this refers to training with frozen CNN weights.

**Unmanned Aerial Vehicle** A Unmanned Aerial Vehicle is a type of aircraft that is capable of flying without an pilot on board. Unmanned Aerial Vehicle are often either controlled through remote control or even capable of autonomous flight. Unmanned Aerial Vehicle can be drones (quadrotor, multi-rotor), helicopters or fixed-wing aircraft.

**Vision Transformer** A Vision Transformer is a DL model for vision tasks based on the idea of transformers, which utilize multi-head attention. ViT comply with that idea by embedding images into patches which can then be turned into vectors that work well with transformers.

# Developing a Strong Pre-Trained Base Model for Plant Leaf Disease Classification

## David Jona Richter

Department of Artificial Intelligence Convergence

Graduate School, Chonnam National University

Supervised by Professor Kyungbaek Kim

(Abstract)

Plants, crops and their yields are essential to our very existence. Our society heavily relies on agriculture and farmers for food, but diseases and pests cause large losses every year. As such it is vital to ensure that diseases can be spotted early and treated accordingly and stopping the spread while still possible. Manual and traditional methods require knowledgeable personal to walk through the field and check for symptoms ”by hand” or using laboratory equipment such as microscopes to analyze the plants for diseases. This is, obviously, very laborious and very time consuming and also requires trained manpower. This resulted in the desire and need to automate this process. Machine Learning (ML) methods have been applied as a result and they have garnered promising results, but ML has its shortcomings. It needs researchers to extract features from the data, especially for image data (regular images, multispectral images, and hyperspectral images alike). Deep Learning (DL) methods can not only overcome said problem, but also tend to perform better than traditional ML. Convolutional Neural Network (CNN) models are especially efficient as they can automatically extract features from images without any manual feature construction before then feeding the features to a classifier (both ML or DL). Many different DL models have since been proposed and tested to identify plant diseases to varying degrees of success, most of them being CNN architectures using image data as input. CNN, and all other DL methods, require large amounts of data to train properly, making datasets largely influential to the final performance of the model. Despite the importance that datasets pose to the field, there still seems to be somewhat of a discrepancy between what is publicly available for use and what would be required to sufficiently train fully capable models. To overcome these shortcomings, as part of this thesis open datasets for the field of plant leaf disease classification have been identified as well as models that can be trained on them and extensive benchmarks have been carried out to identify their suitability. Then a new dataset was constructed based on those findings as well as on the findings of a augmentation applicability study, which will be used to train a new Base Model based on the DenseNet201 architecture, which managed to outperform the baseline model on said new dataset as well as outperforming it on plant leaf disease classification domain specific Transfer-Learning experiments on another new dataset. This new model manages to train models through Transfer-Learning (TL) faster, more robust, more stable, and with less data than general Foundational Model (FM) would, overcoming a large number of issues that the field still suffers from.

# 1. Introduction and Motivation

## 1.1 Introduction

One of the most important issues that still very much impacts humankind, even to this day, is hunger. Not only is the problem still relevant today, it also trending in the wrong direction, with recent reports projecting that world hunger will only worsen [1]. These reports by the Food and Agriculture Organization of the United Nations (FAO) have reported an increase of 7.5% to 9.1% of to the number of people suffering from undernourishment worldwide in only a 5 year span from 2019 to 2024 [1]. These numbers correlate to about 750 million people that were affected by undernourishment in 2023 with the trends only increasing. While world-wide the numbers currently measure in at the afore mentioned 9.1%, regional number are much mire dire in certain parts of the world, with Africa having a 20.4% share of the population that is affected [1]. On top of that, the recent COVID-19 pandemic also negatively impacted this trend and contributed to another 130 million potentially being subject to undernourishment [1]. So while many people in developed countries might not feel it first hand, hunger, to this day, remains a problem affecting hundreds of millions of people world-wide and therefore deserves and requires attention to solving it.

Crops play a pivotal role in that. Crops, according to the Food and Agriculture Organization of the United Nations (FAO), contribute heavily to the food supply of humans everywhere. Plant based products (cereals, fats and oils, sugar, fruit and vegetables, roots, tubers and pulses) are the biggest source of calorie consumption with over 80% while also being responsible for about 60% of the protein supply [2] with numbers being even more significant in Africa (nearly 80% of proteins and over 90% of calories). Crop production has increased by over 50% between 2000 and 2021 [2] and is expected to further increase by another 50% from its 2012 numbers to 2050. All this proves how pivotal and important agriculture, crop production and and its yields are to food security world-wide and how important it is to make sure that crop production remains as efficient and reliable as possible.

FAO reports, however, showcase that about 40% of planted crops are lost to some kind of pests or diseases every year [3]. With prognoses of diseases continuously spreading to new locations, the impact of diseases and the resulting losses are likely to increase [3]. Plant disease pandemics are also said to be increasingly more likely to spread and be more severe than in the past [4]. To try and limit the impact felt by disease infections to the planted crops, detecting them early is of high importance [5]. Because of this accurate, fast and reliable detection of plant diseases is becoming more and more important. Field disease monitoring is traditionally carried out manually through field scouting [6], where trained staff would go out into the field where the crop seeds have been sown, to check the plants for signs of diseases. Staff needs to be well trained to make sure they can properly identify the symptoms accurately and in a timely manner to prevent negative results [7], and since plants need to be checked periodically, multiple times during the growing stages [8], the whole process is rather laborious.

One way of accelerating this process is through the application of Deep Learning (DL) models. Image classification models can be utilized to identify whether or not plants are infected and if they are they can classify what disease they are infected with [9, 10, 11, 12, 13]. Through advances in computer imaging and computer vision, new methods are becoming increasingly stronger and more capable of handling tasks such as this one. Recent years have brought forth better and more advanced models that exceed in the field of categorizing images into their respective classes. Complex models such as that do, however, rely on high quality datasets that are: big enough, diverse enough, high resolution, and representative of the task that they are supposed to be trained for.

Such datasets are, however, sparse and especially in the field of plant leaf disease detection ideal datasets are not a given [14, 11]. The most popular dataset, PlantVillage, is a lab dataset taken in ideal conditions, which tends to train models that are not as capable when used in field [15]. PlantVillage being this popular is understandable, considering that not everyone has the opportunity to create their own dataset, and even if they do, doing so is very time consuming. On top of that, PlantVillage comes with over 50,000 images and with 38 classes including multiple plants and multiple diseases per plant. All images are of high quality and so is the dataset as a whole. The only downside one can bring up is that it is taken in lab.

Most new datasets collected by researchers are kept private (not open to access by the public) [10], further amplifying this limitation currently faced in the field. With most researchers using PlantVillage as the dataset in their work (see Figure 2.1), a certain level of comparability is available across studies (although many studies only use subsets). However, with PlantVillage being a lab dataset and therefore rather easy for DL models and CNN to learn, most models achieve very high scores, making said comparability somewhat tricky, since all models score incredibly well. This means that even weaker models will perform well on this data and that most models only differentiate themselves from each other by very small margins.

To improve the current state of plant leaf disease classification using DL methods, in this thesis a number of open datasets and powerful CNN models have been identified and underwent an extensive set of benchmarking experiments in order to identify the validity of each model and dataset for this specific field. Based on the findings of this benchmark, along with findings of another set of experiments to identify the applicability of image augmentation methods for plant leaf classification task (to overcome the issue of limited data availability), a new dataset was constructed. This new dataset, which was built by choosing 9 pre-existing datasets, now has 80 classes across 25 different plants with over 300,000 images (of which 280,000 are used for training and validation and are augmented, and 24,507 are un-augmented test images). This dataset is perfectly balanced on the training side with each class having exactly 3,500 images per class before train-val splitting. This dataset was then used to train the best performing models (according to the benchmarks) to find the most powerful model available and improve upon it to create a new Base Model (BM) specifically for this field. The model chosen as a baseline was DenseNet201, which was improved upon by adding advanced activation functions as well as adding Channel Attention (CA) mechanisms to it. Lastly, this new pre-trained architecture was compared to a ImageNet pre-trained DenseNet201 to showcase the validity, need, and power of this new Base Model.

FM are large scale models trained on a large scale datasets that can then be used as BM to train (using Transfer-Learning (TL)) new models in related, or unrelated areas [16]. A new and domain-related Base Model, a model trained on data that is domain-specific or at least domain-related data, comes with many benefits, as it can generate potentially better Transfer-Learning models, do that faster, less computationally expensive and in a more robust manner [17]. The resulting model manages to outperform general FM in a TL task significantly, performing better while also being faster, more stable, more robust, and also capable of learning robust features on significantly less data. This proves that the idea of designing, creating, training and using a domain informed Federated Learning (FL) model for the field of plant leaf disease classification is a step that can be immensely helpful and that the methods proposed in this thesis manage to perform well and as a result manage to overcome many of the issues this field currently faces, when compared to current methods.

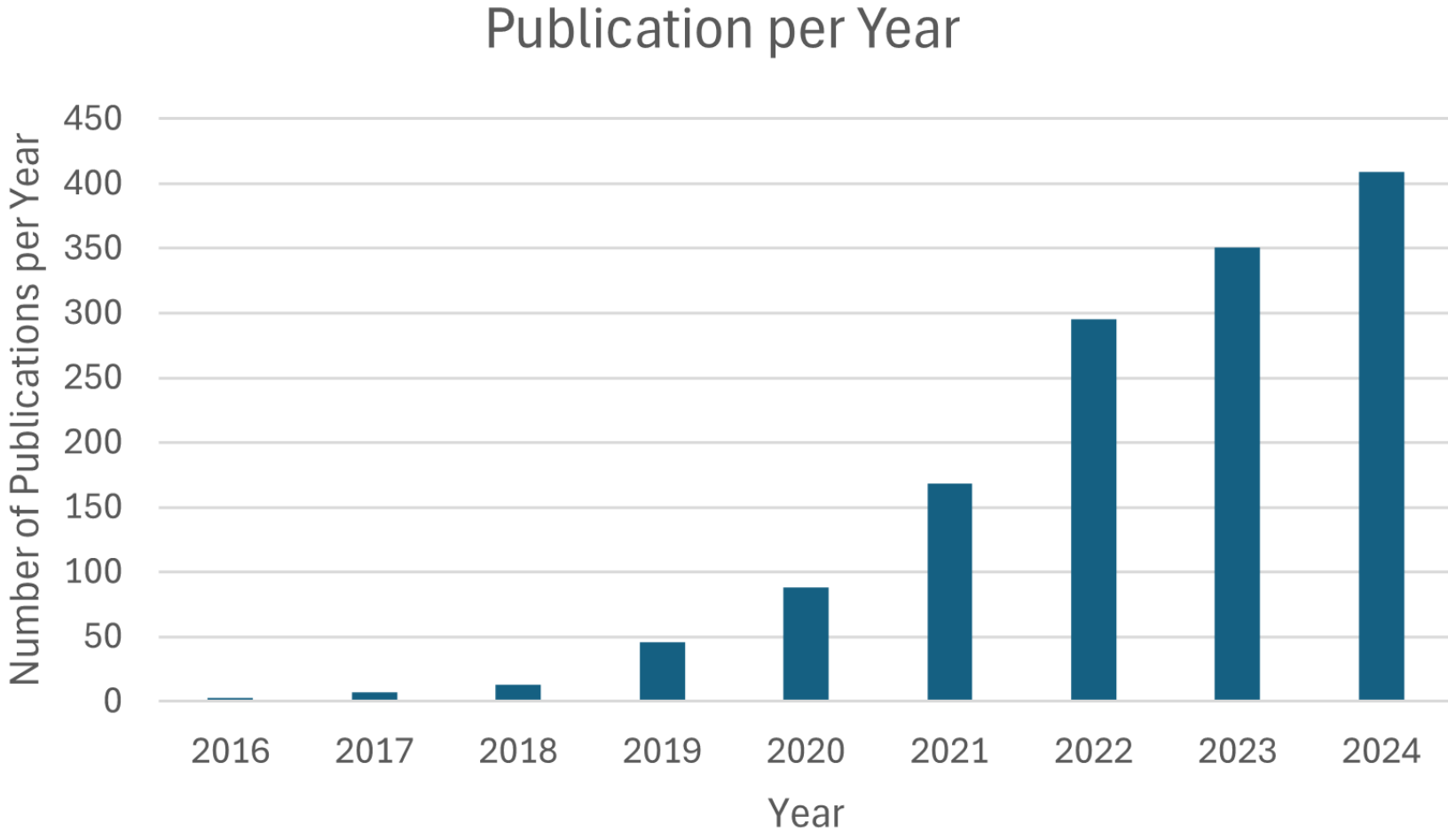

Figure 1.1: Publications per year for ”deep learning AND plant leaf disease” papers based on Web of Science data in recent years.

As part of this thesis, the following contributions were made to the field of plant leaf disease classification using DL techniques.

- Current literature of using CNN FM for plant leaf disease classification is reviewed
- Open and relevant datasets were collected, introduced, described and analyzed
- Current image classification CNN FM are introduced, explained, and analyzed
- These datasets and models are benchmarked in a extensive set of experiments and the results are analyzed
- Using the benchmarking results, the best performing models and datasets are identified
- Augmentation methods to improve plant leaf disease classification datasets to better perform are tested

- Based on the models identified in the benchmark and the augmentation techniques a new large scale dataset is created
- A new model is constructed based on DenseNet201 (which performed well in benchmarks) that outperforms baseline models
- That model is trained on the newly constructed dataset to create a new plant leaf disease classification field specific BM
- That new BM is compared to a baseline FM in a TL experiments on a different plant leaf disease dataset with unseen classes

The rest of this thesis is structured as follows: First the current literature in the field of plant leaf disease classification will be review to identify current trends, best practices, and weaknesses in Chapter 2. Chapter 3 will explain basic methods and technologies relating to both Deep Learning (DL) techniques with a focus on Convolutional Neural Network (CNN) and also looks at plant leaf disease classification in relation with DL. Chapter 4 will introduce current open plant leaf disease datasets and quantify them for better understanding, comparison, and for later usage in building a large scale dataset for training a new model. Chapter 5 introduces and explains foundational CNN models found frequently in current literature. These models will be used in the benchmark to study their applicability in the field and to use them as a baseline for improvements. Chapter 6 explains the methodology used to run the benchmarking experiments and analyzes the results, showcasing how different models and datasets are more or less suited to be applied to the field of plant leaf disease classification. In Chapter 7 different augmentation techniques will be tested to see if they can be beneficial to not only overcome the low number of data availability, but if they can also improve performance. Next, in Chapter 8 the new dataset used to train the Foundational Model will be presented and explained, along with the requirements to recreate that dataset. Then, in Chapter 9, the architecture of the new DenseNet201 based model will be presented along with the experimental results validating it (as well as results with other architectures that were considered), before exposing the now pre-trained model to new data to test it for TL capabilities, comparing it to a DenseNet201 that is pretrained on ImageNet data. Chapter 10 discusses the results and findings obtained in the making of this thesis. In the next chapter, Chapter 11, possible future directions will be presented and discussed, before Chapter 12 concludes this work. The appendices offer additional information and results corresponding to the related chapters.

# 2. Literature Review

This chapter will review relevant review papers for the field of plant leaf disease classification, as well as a few papers that use distributed methods and also numerous papers that are focusing on CNN TL approaches for plant leaf disease classification.

## 2.1 Previous Work

In this section previous work will be summarized and then analyzed. A more extensive review can be found in [18]. The review in this thesis is based upon said referenced review.

### 2.1.1 Literature Review Papers

The survey paper by Demilie [13] summarizes a large number of papers in the field of plant disease detection using both ML and DL methods. The paper showcased that this field experienced a huge uptick in publications starting in 2019. In this work the author found that DL methods generally outperformed ML methods, but it was noted that a hybrid using a CNN as a feature extractor and then a ML method as a classifier, Support Vector Machine (SVM) for example also performs well. Most papers managed to achieve scores of 95% and more, with various modern and advanced CNN models being among the best performing models (some of which used Transfer-Learning). As for plants, tomato plants and their diseases are by far the most researched plants (around 39% of the surveyed papers) with rice plants being the second most popular (16%). Additionally, it was found that most of the surveyed studies used the PlantVillage [19] dataset, due to its availability and size. The dataset is, however, created under lab conditions, and future work would benefit from a dataset recorded in field.

Nagaraju et al. [9] discuss papers published between 2015 and 2019 in their review, with a focus on Hyperspectral Imaging (HSI). A total of roughly 80 papers were introduced in the review and among them were many papers that also introduced datasets. Here too it was brought up that networks using the CNN methodology are the most powerful networks (and applicable to many different plant types) for image based plant leaf disease detection and that CNN SVM hybrid models also perform well. Image pre-processing is also discussed in the form of mainly image resizing. It is noted that 60x60 pixels, 96x96 pixels, 128x128 pixels and 256x256 pixels are the most commonly used sizes.

Li et al. [10] review Deep Learning for plant disease detection and classification in their work. The authors state that most symptoms are visible on the leafs, which is why researchers often use leaf images as input for their models. They also argue that the switch from ML to DL can be attributed to ever larger datasets, more and more powerful computers and graphics cards, and ever better and more functional code libraries for DL. This paper puts a focus on TL and on working with small sample datasets, where methods like Few-Shot Learning are well suited and often applied, which allow the model to train with very small training data. When discussing data augmentation the authors note that color should not be changed when using traditional augmentation, but also mention the power and validity of Generative Adversarial Network (GAN). Visualization methods were also compiled and presented, as were full disease recognition systems and HSI (which are well suited for early disease detection).

Lu et al. [11] summarize DL for the use in plant disease classification task. The paper lists common fungal, bacterial and viral infections for a set of common plants. According to the authors, the major disadvantages that traditional ML methods pose are: low performance, lack of data, manual feature extraction, and that plants are separated from their roots to be classified. problems that DL does not suffer from. The paper discusses several classification networks in detail, listing their architecture, pros, cons, novel aspects and other interesting details. Other facts are also disclosed, such as the fact that making CNNs deeper only works until a certain depth, at which the network will not benefit from the additional layers, but rather suffers from them in the form of slower training. This paper also intensively covers augmentation methods. As for evaluation metrics, the author note the popularity of accuracy across the field. The main issues of DL in plant disease identification are the lack of data, especially data taken in the field as data taken in the lab does not tend to generalize well to real field conditions. With PlantVillage, a dataset with lab condition images, being the biggest publicly available dataset, and most in field datasets being closed source or too small, there is a lack for a sufficiently sized in-field dataset. In cases where the background of an image is too busy, segmenting images first in an effort to detach the leaf from the background could be useful. The paper also explains various factors that impact the disease symptoms, those being: the stage the disease is in, whether or not the plant is affected by multiple diseases whose symptoms would them impact each other, as well as extrinsic factors such as humidity, temperature, soil conditions, sunlight and wind.

Sarkar et al. [12] summarize the field of DL for plant leaf disease identification from 2010 to 2022. The paper goes in depth on the root causes to many different plant diseases for many different plants. It also covers different types of noise and how they impact the performance. The paper also offers many different graphs and statistics analyzing previous publications and also provides an in-depth overview over many different ML and DL algorithms alongside their strengths and weaknesses. One method mentioned specifically is TL, as it outperforms many models that are trained from scratch. One interesting point brought up by the authors is the fact that climate change has an impact on plant leaf diseases as the change in temperature accelerates infection rates. Another topic that is covered in depth in this work is the explanation of more advanced image pre-processing techniques such as Sobel edge detection, CIELAB, top-hat transformations, and more. On top of that it also covers a number of different data augmentation methods as well as methods for image segmentation. Additionally feature selection and extraction is also included. Different classification methods are also explained, ranging from traditional ML to more modern DL methods. The paper also mentions the current interest in hybrid and ensemble methods.

### 2.1.2 Using Distributed Methods

In [20] Mamba et al. present their findings following a set of experiments for crop disease detection using Federated Learning (FL) methods on the PlantVillage dataset [19]. The authors set out to compare and contextualize how different approaches, that are common outside the FL scope, perform once they are applied with FL. They ran a multitude of experiments, in which they iterated over: the dataset (different plant types with different amounts of diseases/classes), the communication rounds, the local agent's epochs and the number of clients. When iterating over the number of agents (3, 5, 7), they found that less agents (3) performed better than when adding more, this was partially due to the fact that the overall amount of data remained the same. So once there were more agents data had to be split into smaller subsets. ResNet [21] (or ResNet50 to be specific) did the best, with MobileNet [22] (MobileNet-v2) not being far off while being a much smaller model. When iterating over the number of communication rounds (10, 30, 50, 100) ResNet again performed best (with MobileNet again doing very well too), while some models did best with 30 while other models tended to do better with more rounds. The local epoch experiment (1, 5) showed that more local epochs improved performance with ResNet and MobileNet yet again standing out. The Dataset experiment showed that the models performed differently on different set, with grapes being the easiest and tomatoes being the toughest (tomatoes have more classes than the rest). Overall this has shown that ResNet and MobileNet translate the best to Federated Learning in this case and that it is important to consider communication rounds, local epochs, number of agents and dataset carefully as they all impact the performance of the models.

Mehta et al. apply Federated Learning to predict wheat disease severeness [23]. A CNN was used that took 256x256 pixel Greyscale images as input, that were normalized to be in the range of 0 to 1. After applying data augmentation on the collected image dataset (spinning and turning) the set contained 9,967 images of which 1,000 were used for validation with the remaining 8,967 making up the training set. The dataset contained 5 diseases and therefore 5 classes, with those being leaf rust, Fusarium head blight, wheat stripe mosaic virus, powdery mildew, and Septoria leaf blotch. The global model converged at around 90% accuracy, although the paper fails to explain how the severity is actually being calculated or applied to the model. Hyperparameters as well as the model architecture are also not included in the paper.

In [24], Dhiman et al. train a novel DL model (a VGGNet model with Transfer-Learning (TL)) to predict the disease severity of citrus fruits. Here the fruit itself, post harvesting, is analyzed, not the plant it grew on. The image data was gathered from publicly available datasets (PlantVillage, Kaggle) and was then annotated by an expert (horticulture scientist). The expert used only exterior features such as size, color and shape, not considering internal features. The resulting dataset contains bounding boxes for any imperfections that state the severity (healthy, low, medium, high) of said imperfection and it consists of about 4,000 annotated images.

[25] explores the application of Unmanned Aerial Vehicle (UAV) in combination with FL for pest detection. The experiment conducted here was carried out at 4 different sites (4 clients), where each site had a UAV setup that carried out real-time pest analysis. The model is capable

of detecting pest such as mosquito's, grasshoppers, sawflies, mites and beetles, among others. A total of 9 pest classes was present in the dataset. Data would be collected on site by the UAV in the proposed method, but for this work data was collected via Kaggle and Google image search queues and then pre-processed (resized) and augmented (mirroring and rotation) which resulted in a dataset in 5,400 images overall. This data was then used for training on an EfficientNetB3 architecture. The proposed methodology outperformed previous works and reached a accuracy score of 99.55% according to the authors.

Idoje et al. also utilize Federated Learning in [26]. In this paper the objective of the FL model is the classify the crop types in smart farm environments. One big difference here is that the data is tabular climatic data (rainfall, potential of Hydrogen (pH), humidity and temperature) as opposed to image data. The federated learning model was tested across different optimizes (Adam and SGD). The models using SGD did not perform well and the model using Adam with a rather large starting learning rate also failed to reach high accuracy. The model using Adam and a smaller stating learning rate of 0.001 however managed to reach 90%.

### 2.1.3 Transfer-Learning CNN

Hammou et al. [27] use DenseNet169 and InceptionV3 on the PlantVillage dataset, using only the classes that were dedicated to tomato plant leaves. This resulted in a subset of roughly 18,000 imaes across 10 tomato leaf classes. Both models were pretrained on the ImageNet dataset. The models were trained for 100 epochs and both models managed 100% accuracy.

Tej et al. [28] implemented ResNet50 as well as ResNet152 in their study. Said study introduces a new dataset which was collected by the authors in Tunisia. The dataset contains tomato and pepper images and totals 488 images, which were the augmented (using rotation, brightness, contrast, saturation, flip, gaussian blur) to increase the dataset to 8,400 images, belonging to 7 classes (1,200 each). The trained models (after 100 epochs) managed scores of: 92.7% / 83.3% (ResNet50) and 98.85% / 92% (ResNet152) with the first value being the augmented data and the second one being trained without the augmented data.

Nagi et al. [29] made use of MobileNet, AlexNet, VGG16, and VGG19 in their study, in which they trained all 4 models on the PlantVillage dataset. They did, however, not use all classes, but rather only classes that contained grape plant images. As such, the final dataset contained 4 classes and 3,423 images. After training for 30 epochs the AlexNet reached roughly 95.5%, VG16 managed roughly 97%, VGG19 managed roughly 97.5%, and MobileNet reached roughly 98%.

Li et al. [30] use RegNet, ShuffleNet, MobileNetV3, and EfficientNet-B0 in their work, training them on a apple plant leaf dataset that they themselves collected in China. The dataset consists of 5 classes with 2,141 images before augmentation (rotation, translation, brightness, scaling). ShuffleNet (90.2%), MobileNetV3 (93.4%), and EfficientNet-B0 (94.3%) managed their highest accuracy with the Adam optimizer, while RegNet managed its best score (99.9%) with Ranger (Adam 99.8%).

Naik et al. [31] compared a large number of models, namely AlexNet, DarkNet53, DenseNet201, EfficientNet-B0, InceptionV3, MobileNetV2, NasNetLarge, ResNet101, ShuffleNet, SqueezeNet, VGG19, and XceptionNet. The dataset used in this study was collected by the authors in India. It contains chili leaf images which make up 5,100 images in 6 classes. Training was carried out for 30 epochs and the resulting models managed: AlexNet 97.65%, DarkNet53 98.82%, DenseNet201 97.84%, Efficient-B0 97.55%, Inceptionv3 96.86%, MobileNetV2 97.35%, NasNetLarge 96.18%, ResNet101 97.55%, ShuffleNet 97.45%, SqueezeNet 97.35%, VGG19 97.84%, and XceptionNet 98.63% on the augmented dataset (rotation, brightness, contrast, saturation, flip, gaussian blur).

Pradhan et al. [32] made use of DenseNet201, DenseNet169, InceptionV3, InceptionResNetV2, MobileNet, MobileNetV2, ResNet50, VGG16, VGG19, and Xception in their work and train the models on the PlantVillage dataset apple classes (2,537 images in 4 classes). Models were trained for 30 epochs and managed 98.75% (DenseNet201), 97.52% (DenseNet169), 94.87% (InceptionV3), 95.27% (InceptionResNetV2), 98.66% (MobileNet), 98.29% (MobileNetV2), 74.29% (ResNet50), 98.26% (VGG16), 97.00% (VGG19), and 95.11% (Xception).

Reda et al. [33] focus on smaller and more lightweight models in MobileNet, MobileNetV2, NasNetMobile, and EfficientNetB0. The dataset of choice in this paper was again PlantVillage. In this case PlantVillage was not cut to a subset, but rather used over 60,000 images belonging to 39 classes. Training was carried out for a total of 30 epochs, after which the EfficientNetB0 without any frozen layers achieved over 99%, which made it the best performing model.

Mafukidze et al. [34] use InceptionV3, DenseNet121, DenseNet201, MobileNetV2, VGG16, EffcientNetB0, ResNet50, and LeNet to classify the 4 maize classes of the augmented PlantVillage dataset. Through shearing, rotation, flipping, saturation changes, mean filtering, and crop-ping the dataset was enlarged from 4,188 to 46,898 images. After 500 epochs, the models managed 99% accuracy, except for LeNet, which reached 94%.

Latif et al. [35] utilize GoogleNet, VGG16, VGG19, DenseNet201, and AlexNet to then train them on a rice leaf dataset that they obtained from Kaggle. 2,164 images were originally included in the dataset which was then enlarged via augmentation methods. GoogleNet managed 89.63%, VGG16 95.62%, VGG19 96.08%, DenseNet201 94.24%, and AlexNet reached 92.63%.

Bruno et al. [36] use different EfficientNet architectures (B0 to B7) to categorize 39 classes of PlantVillage data. Augmentation was not applied in this work and the training was concluded after the early stopping callback with patience determined that the model had converged.

Eunice et al. [37] opted for 4 separate pretrained CNN models (DenseNet121, ResNet50, VGG16, and InceptionV4) for their experiments. The dataset used was PlantVillage, again. 38 classes with 54,305 images were used in this work and the dataset was also exposed to augmentation. The models were traiend for 30 epochs, which led to 99.78% (InceptionV4), 84,27 (VGG16), 99.82 (ResNet50), and 99.87% (DenseNet121).

Albahli et al. [38] implemented AlexNet, VGG19, InceptionV3, ResNet50, DenseNet201, MobileNetV2, EfficientNet, and EfficientNetV2 to train them on 3 datasets, those being PlantVillage, PlantDoc, and CD&S. The datasets were filtered to only include corn classes and augmented with rotation, flipping, translation, scaling, and color jittering. As a result of that the final dataset spanned 8,231 images across 5 classes. The models were trained for a baseline of 25 epochs, but with early stopping (patience 10) enabled. Results show test accuracies of 87.12% (AlexNet), 90.34% (VGG19), 92.82% (InceptionV3), 87.58% (ResNet50), 94.17% (DenseNet201), 84.38% (MobileNetV2), 93.52% (EfficientNet), and 94.37% (EfficientNetV2).

Novtahaning et al. [39] tested EfficientNet-B0, ResNet152, VGG16, InceptionV3, Xception, MobileNetV2, DenseNet 201, InceptionResNetV2, and NasNetMobile to then build and ensemble a model from among them. A open coffee plant leaf image dataset from Kaggle was used for training which included 6 classes of 260 images each (1,300 total images). Training was carried out for 30 epochs after which VGG16 reached 94.2%, InceptionV3 83.9%, ResNet152 93.8%, Xception 85.4%, MobileNetV2 74.6%, DenseNet201 83.8 InceptionResNetV2 86.9%, NASNetMobile 83.8, and EfficientNet-B0 95.0%, all using the Adam optimizer. Due to these results the authors decided to build an ensemble network using EfficientNet-B0,

ResNet152, and VGG16.

Alirezazadeh et al. [40] use the DiaMos pear leaf dataset in their work to train EfficientNetB0, MobileNetV2, ResNet50, InceptionV3, and VGG19 in combination with CBAM. The dataset, augmented with mirroring, rotation, and color variation, contained 4 classes and 3,006 images and the models were trained on it for 100 epochs. The results (ResNet50: 69.90%, ResNet50+CBAM: 71.36%, VGG19: 73.09%, VGG19+CBAM: 73.42%, InceptionV3: 76.62%, InceptionV3+CBAM: 77.87%, MobileNetV2: 82.06%, MobileNetV2+CBAM: 83.99%, EfficientNetB0: 85.82%, EfficientNetB0+CBAM: 86.89%) that the CBAM block managed to increase performance for all models.

Fulle et al. [41] train MobileNetV3-Small, NASNetMobile, MobileNetV2, and DenseNet121 on a dataset which the authors themselves collected in field in Ethiopia and then shared online. The dataset is a Enset plant leaf dataset consisting of 2 classes and 1,228 images before augmentation which were enhanced to a total of 7,580 after augmentation (shearing, flipping, and brightness changes). The models were in training for 30 epochs and the performed with test accuracies of 76.68% (NASNetMobile), 94.33% (MobileNetV2), 99.28% (DenseNet121), and 99.93% (MobileNetV3-Small).

Mavaddat et al. [42] make use of the VGG16, InceptionV3 and Resnet152v2 models and train them on an open rice leaf dataset. The dataset is made up of 5,932 images in 4 classes. The 3 models were trained for 30 epochs each and achieved 100% after the application of Fine-Tuning (FT).

Paiva et al. [43] present a large scale comparison of 12 different models on the iCassava dataset. VGG16, VGG19, ResNet50, ResNet50V2, ResNet101V2, ResNet152V2, InceptionV3, InceptionResNetV2, MobileNetV2, DenseNet121, DenseNet169, and DenseNet201 were chosen to be the model being compared in this study. The cassava leaf dataset was augmented before training by applying rotations, brightness adjustments, zooming, and flipping. The resulting set has 5 classes and 21,397 images. 20 epochs were set for training with early stopping with the patience set to 7. The results were: VGG16 with 60.63%, VGG19 with 62.90%, ResNet50 with 73.46%, ResNet50V2 with 72.34%, ResNet101V2 with 71.64%, ResNet152V2 with 70.14%, InceptionV3 with 69.91%, InceptionResNetV2 with 69.67%, MobileNetV2 with 71.17%, DenseNet121 with 73.27%, DenseNet169 with 74.77%, and DenseNet201 with 70.84%.

Mohapatra et al. [44] make use of AlexNet in combination with a rice plant leaf disease dataset from Kaggle. The dataset contains 3 classes with 1,732 images and AlexNet will be trained on it for 50 epochs after which it managed 98.80%.

Diker et al. [45] combine pretrained ResNet101 and MobileNet with ML classifiers like Random Forrest (RF) and SVM, as well as also using regular neural networks as classifiers. The dataset used here is a dataset that was collected by the authors and they decided to make it public for further use. The images were collected in turkey and are of olive tree plant leaves. Classes are set to be 2 (healthy and unhealthy) and the dataset accounts for a total of 954 images. The best performing combination was achieved using ResNet101 with k-fold set to 10, and while using SVM as the classifier, where the model managed to reach 98.63%.

Lanjewar et al. [46] use a Kaggle dataset for potato plant leaf diseases with NASNetMobile, VGG19, and DensNet169, ResNet50V2, InceptionV3, and Xception. The 1,500 image dataset was augmented through rotation and flipping and it contains 3 classes. The resulting models managed, on the test set, 95.33% (NASNet), 95.67% (VGG19), 99.67% (DenseNet169), 97.00% (ResNet50V2), 97.00% (InceptionV3), 96.00% (Xception) after 100 epochs of training.

Ramya et al. [47] use AlexNet on a dataset combined from different online sources. The dataset was augmented which resulted in 22,411 images across 10 classes. Training was only carried out for 15 epochs, after which the model managed an accuracy score of 99.7% on the validation dataset.

Shaheed et al. [48] apply the PlantVillage dataset to ResNet50, VGG16, Xception, and InceptionV3. PlantVillage was reduced to only include 3 classes, those being of potato images, and the dataset was augmented with rotating, shearing, and shifting images after augmentation. 100 epochs were set to be the training period and the best pretrained model (Xception) ended on accuracy score of 98.84%.

An et al. [49] also uses AlexNet as the model of choice selected for training in their paper. In this paper, however, the dataset was reduced to include only classes containing corn images. As such, the final dataset was made up of 4 classes totaling 2,013 images (augmentation was applied). Training was carried out over 20 epochs after which the model ended up with 95.7% using SGD and 95% using Adam.

Salam et al. [50] opted to use MobileNetV3Small, ResNet50, and VGG19 for the research.

The mulberry leaf dataset used was one that was collected by the authors in field in Bangladesh. The dataset was increased from 1,091 images to 6,000 images post augmentation (rotating, flipping, transformations, translation, zoom). Results show that MobileNet reached an average of 92.5% on 5 folds k-fold, VGG19 reached 90.89% and ResNet50 achieved 94.4%.

Bania et al. [51] use ResNet101 and a potato leaf dataset with 3 classes and 2,152 images after augmentation (flipping, rotation, and translation). The resulting model performed with 95.13%.

Hang et al. [52] utilized Xception, MobileNetV2, NASNetMobile, and MobileNet to train them on their own dataset, which was captured in China. It contains soybean images and was enhanced using CycleGan. CycleGan generated synthetic images that increased the dataset size from 2,820 to 8,456 (7 classes). After 1,000 epochs of training the model reached test accuracies 91.00% (Xception), 91.21% (MobileNetV2), 90.39% (NASNetMobile), 80.57% (MobileNet).

Haikal et al. [53] applied GoogleNet, ResNet34, and MobileNetV3 to the paddyDoc dataset (post augmentation). The dataset contains 10,799 images which are split into 4 classes. Additionally it was split in 2, one version containing 3,000 images in lab and the other version contained the remaining 7,799 images taken in field. The models were trained for 100 epochs each, with a focus on augmentation mehtods, before being evaluated. These results showed that models using augmentation were performing better then ones that did not.

Shah et al. [54] employ VGG16, VGG19, InceptioV3, InceptionRes-NetV2, Xception, and ResNet50V2 in their study, by training them on a field grape leaf dataset that the authors collected in India. Said dataset contains 300 original images which were extended to 1,600 images via augmentation (cropping, rescaling, transformation, resizing, rotation, adjusting brightness, contrast, sharpness). Training was only carried out for 10 epochs and the models ended up with accuracy scores of VGG16: 95%, VGG19: 96%, InceptioV3: 94%, InceptionRes-NetV2: 94%, Xception: 96%, ResNet50V2: 98%.

Mazumder et al. [55] inspect MobileNetV2, DenseNet121, ResNet152V2, DenseNet169, DenseNet201, InceptionV3, NASNetLarge, InceptionResNetV2, EfficientNetV2S, EfficientNetV2L, and a DenseNet201 with attention modules. The dataset contains images of banana and black gram leaf images with roughly 2,000 images reserved for training (3 classes for banana leaves and 5 for black gram). The models were trained for 50 epochs with early stop-ping enabled and MobileNetV2 was best with 93% (besides the modified DenseNet201 with 99.50%) on the black gram data, while DenseNet201 was the best regular model with 82.3% (modified DenseNet201 90.12%).

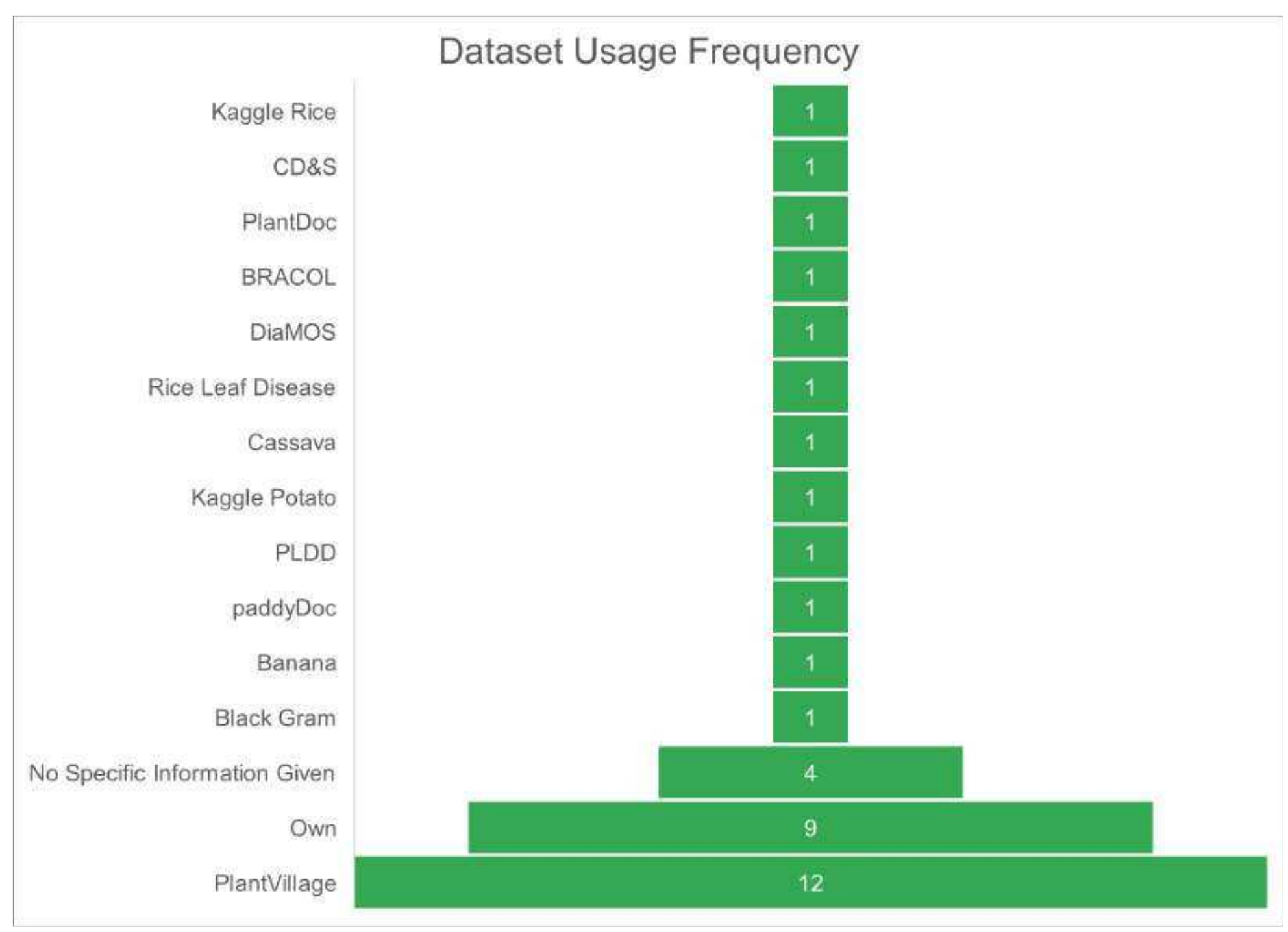

Figure 2.1: The frequency of datasets used in reviewed papers.

Winiarti et al. [56] explore MobileNetV2, and VGG16 on a combined dataset of Kaggle chili leaves and leaves that the authors collected by themselves in field in Indonesia. The resulting full dataset spans 250 images across 5 classes. VGG16 reached 88% while MobileNetV2 reached 94%.

Batool et al. [57] experiment with InceptionV3, AlexNet, and VGG16 on the PlantVillage dataset. They limited themselves to only using the tomato classes, of which there are 10 with roughly 16,000 images. After 80 epochs of training, the models reached 95% (AlexNet and InceptionV3) and 96% (VGG16).

### 2.1.4 Analysis

When looking at the 31 papers that were review in this work that used TL CNN for plant leaf disease classification, then one can get a good understanding of the recent trends and practices used in the field. As becomes very obvious in Figure 2.1, PlantVillage is by far the most used dataset among the review papers. While PlantVillage is used in 12 of the review works (over 1/3), no other dataset is used more than once. It is technically not a bad thing to have dataset consistency across different studies, but PlantVillage is a lab condition dataset, which is much easier for models to handle and therefore makes comparability harder since all models tend to over-perform on such lab data. Also, la data trained models do not generally perform well in real field conditions [15]. This trend towards lab data can also be observed across all datasets. Figure 2.2 shows that among publicly downloaded datasets a total of 15 were taken in lab conditions (12 are PlantVillage) and only 5 were in field conditions, with another 2 being hybrid. Out of the datasets that were specifically collected for the studies a total of 6 were field datasets and only 3 lab (a much better ration in terms of field images), but sadly out of those 9 datasets that were collected as part of the studies only 2 were shared with the public and 7 were kept private (see Figure 2.3). This shows that datasets are indeed sparse, since most researchers use PlantVillage (which is a lab set and therefore not ideal) or other lab datasets, and most of the newly created sets are not even shared with the public. This further showcases the need for a large scale dataset in field or hybrid conditions that can challenge models and offer the grounds for true comparability.

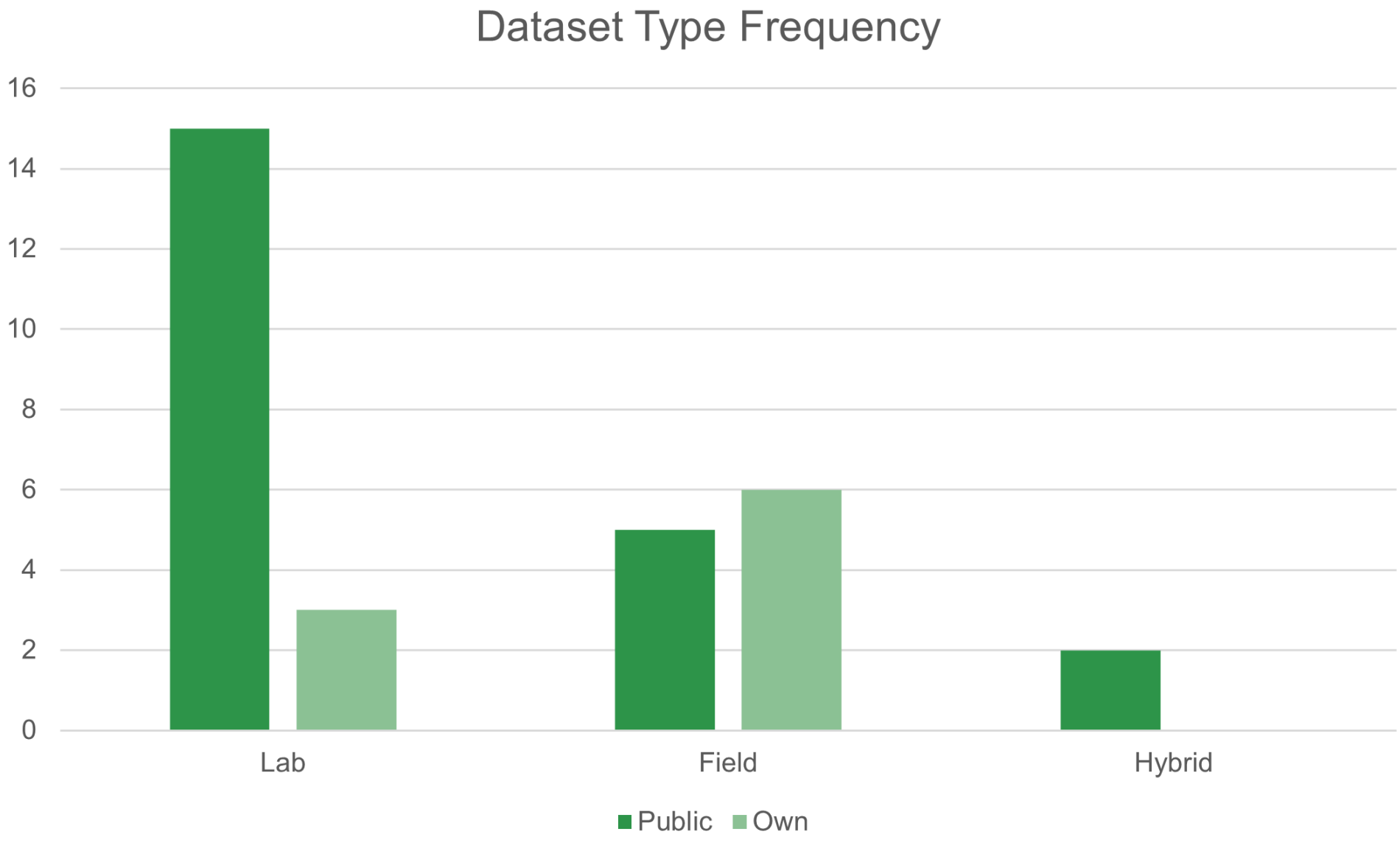

Figure 2.2: Frequency of used dataset types in papers review for this work.

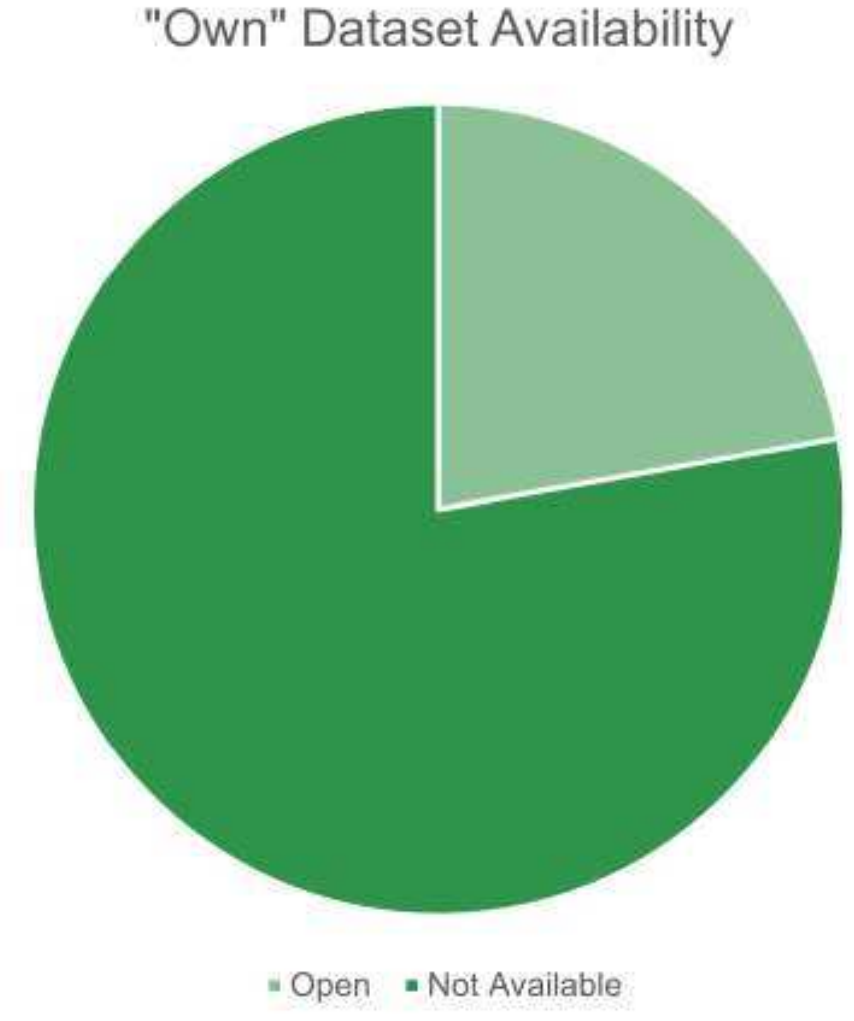

Figure 2.3: Availability of datasets newly introduced in review papers.

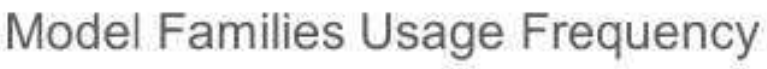

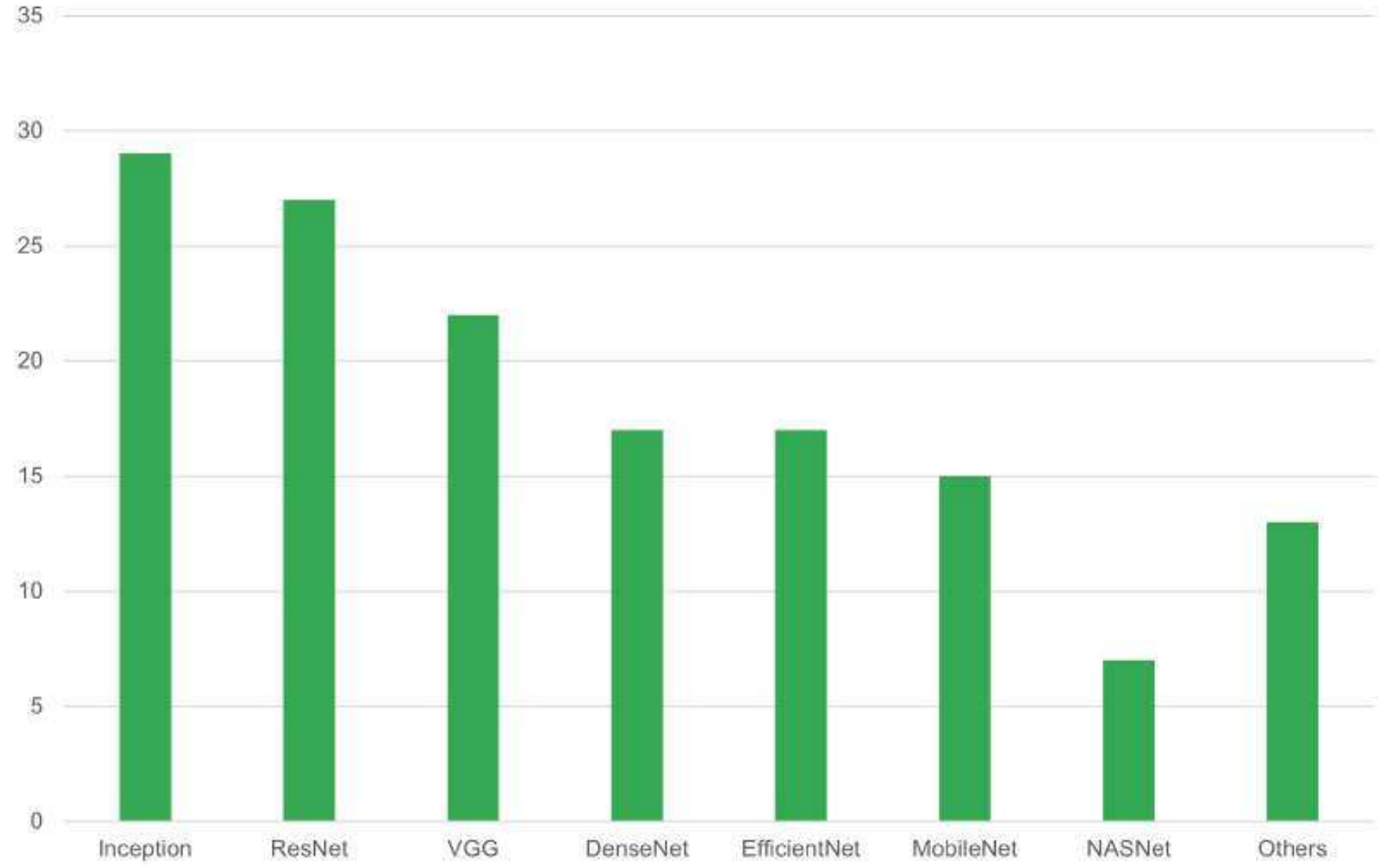

Figure 2.4: The most used model families that were used in the review papers.

When looking at the model usage, the picture is quite different. Where in terms of datasets most researchers worked with the same option (all though it comes with drawbacks), in model selection there seems to be no consensus as to which model or even model family to gravitate towards. A large number of different models from different architecture families see frequent use (see Figure 2.5). InceptionV3 sees the most usage with VGG16, VGG19, ResNet50, Xception, MobileNetV2, EfficientNetB0, DenseNet201, and AlexNet also seeing a lot of application. Numerous other models also saw repeated usage in the papers studied. Even when looking at the model family usage in Figure 2.4, it becomes apparent that there is no clear agreement about which models are better suited to be applied and tested in plant leaf disease classification tasks.

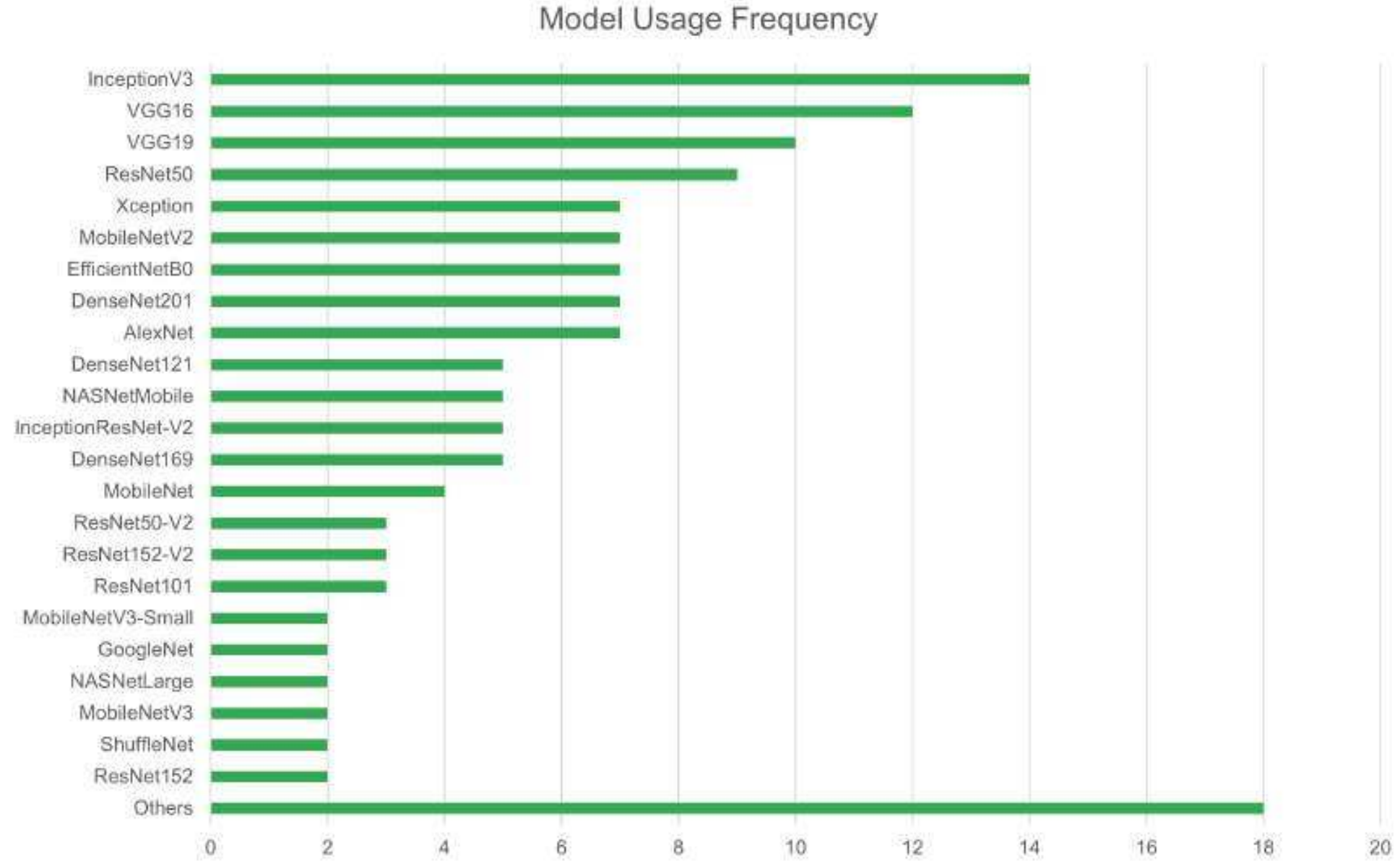

Figure 2.5: The most used models in review work.

Overall, through this review is becomes apparent that data is an issue in plant leaf disease classification that needs solving. Ideally a new large scale dataset that does not only contain lab condition images would help solve this issue. Such a field or hybrid dataset could be used to test, compare, benchmark, and pre-train future models, which could lessen the reliance on the PlantVillage dataset. Additionally, a better understanding of which models are inherently better suited to be applied to plant leaf disease classification tasks would be desirable. If such a understanding could be reached, then future work could focus more on using models that are better suited and spend less time working with models that are generally subpar when applied to this field. Additionally, all these models are general CNN model. A dedicated model maybe even one with pre-trained weights and knowledge in the field of plant leaf disease classification, could help future research by providing a dedicated model architecture to work with and by allowing future work to be based in said model, potentially even use it as a pre-trained BM for TL could save future work time, and computational resources, since it could help models to train faster, more stable, more robust and even with less new data.

# 3. Background

## 3.1 Image Classification using Deep Learning

Deep Learning and Neural Networks have seen a ton of usage, work, development, and application in recent years. With the rapid development and increase in computational power, of newer hardware, GPU in particular, neural networks have not only become feasible, but in this day and age they have become very powerful and they have become common place in many areas and also in many people's everyday life. One area that has benefited heavily from the increase in computational power and has subsequently seen a lot of recent work is the field of image processing using AI methods. One popular, efficient and effective method to build AI models that handle images are CNN. Before CNN and DL became computationally viable, traditional methods were employed, where feature extractors had to be built by hand. While the resulting models can be very strong and efficient, they require a lot of work, expert knowledge and need to be redone for any new domain-specific task that one might need a model for. This means that they were often laborious, time consuming and expensive in making. CNNs on the other hand, can be trained automatically, as long as sufficient (and often labeled) data is present. CNN can learn feature extractors through the learning process, without developers having to design any of them on their own. Only the model needs to be constructed, but there are numerous popular and proven CNN model architectures readily available for use in publications and online. Fundamentally speaking, they are all somewhat based on the initial idea of convolutional layers paired with pooling layers to automatically train feature maps , which can then later be used for classification by adding a classifier top to the end of the convolutional block.

These convolutional layers are made up of so called kernels, which parse over the image. Kernels are usually square, often of the size of 3x3 pixels, and parse over the image from the top left to the bottom right pixel by pixel (see Figure 3.1 and Figure 3.2), multiplying their own values with the image pixel values to generate automated feature maps. Early convolutional layers generally learn basic features, with deeper layers learning features that are more nuanced. Deeper layers also learn features on a more global level, and across a larger area of the image, whereas early layers usual learn more local features over a smaller part of the image. This is due to the nature of the pooling layers that often appear in between stacks of convolitional layers. Pooling layers, max-pooling being the most commonly used one for this task, reduce the image in size and only keep the values one chooses to keep. In a 2x2 max-pooling layer, which is the standard, an image gets reduced to half its size, with each 2x2 block of original pixels in the image being reduce to a single pixel in the resulting pooled image, with the larges of the 4 values in the 2x2 square being kept (see Figure 3.3).

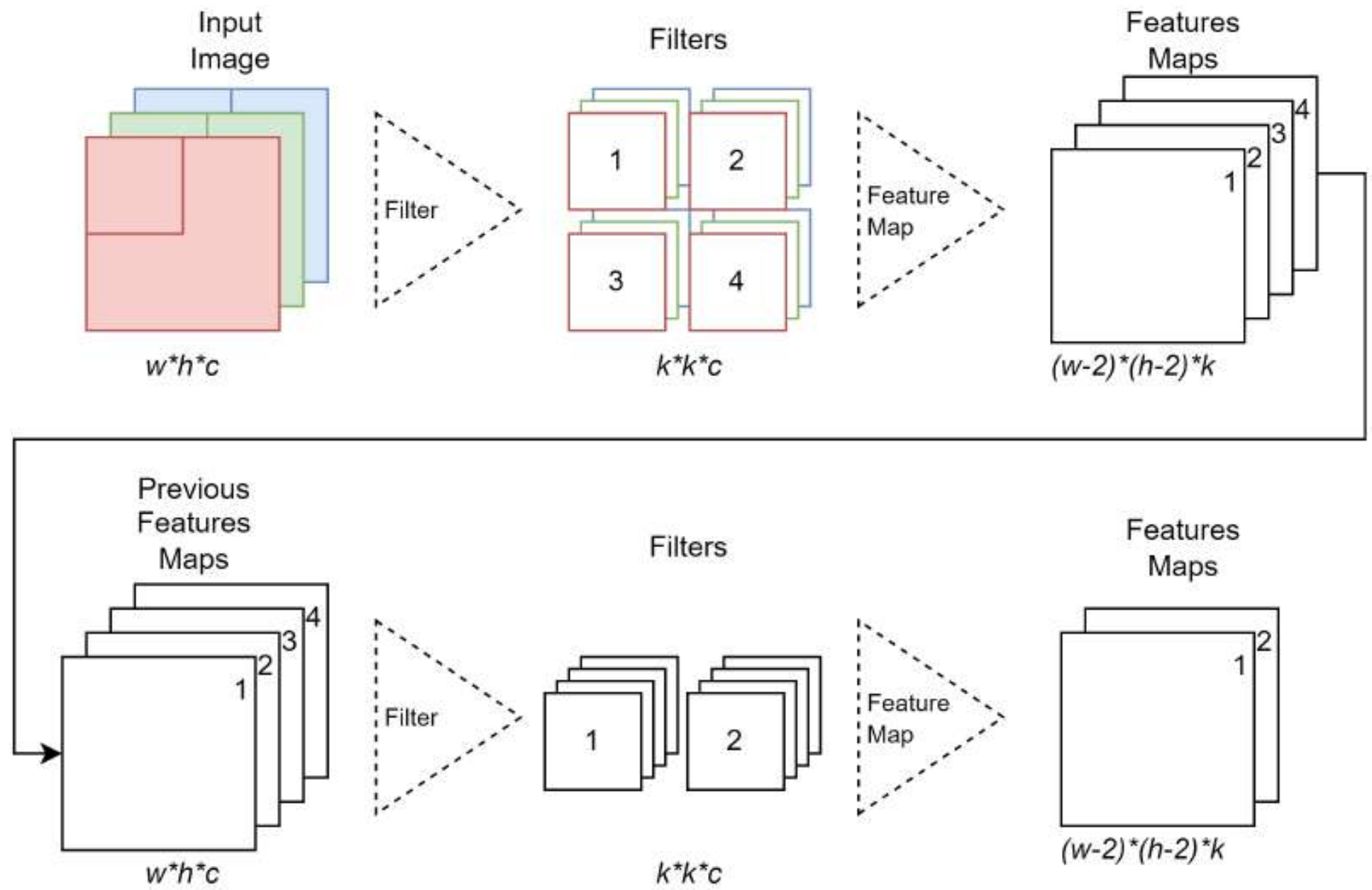

Figure 3.1: This figure explains how kernels parse over the image and extract feature maps from them.

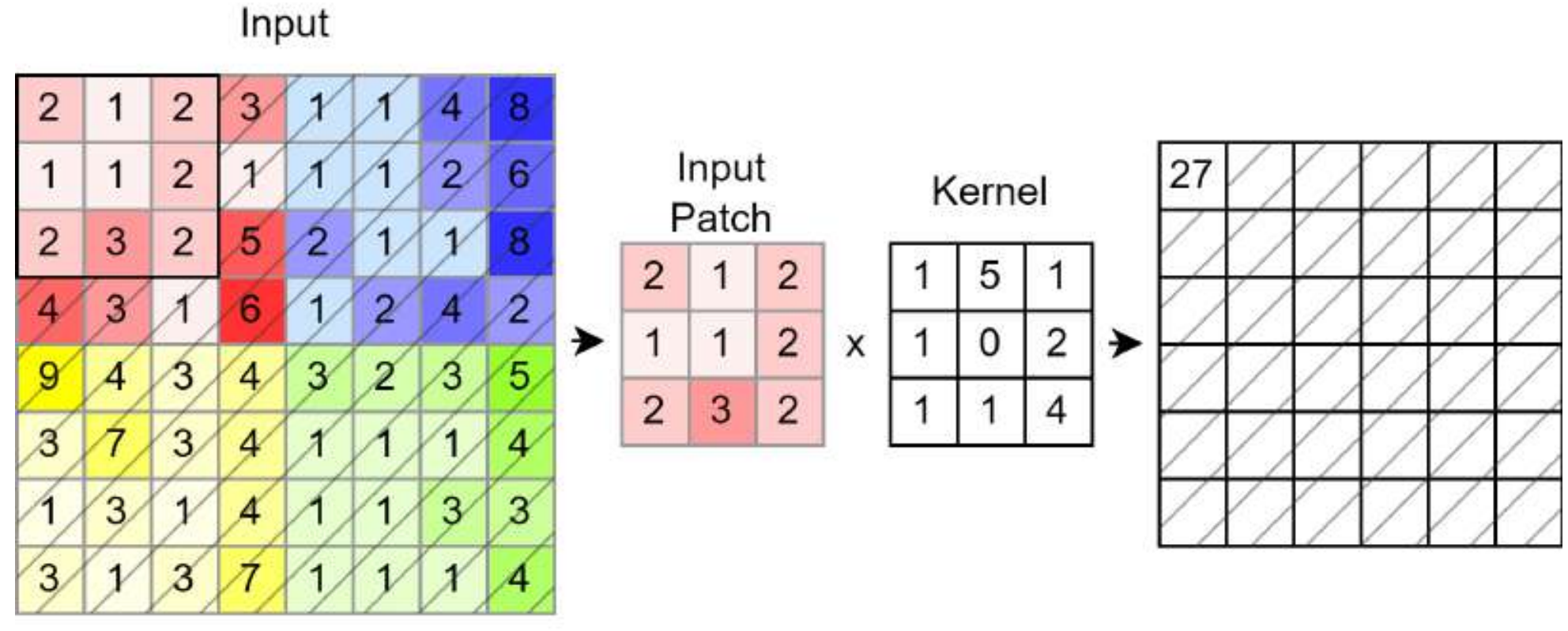

Figure 3.2: Kernel operation visualized.

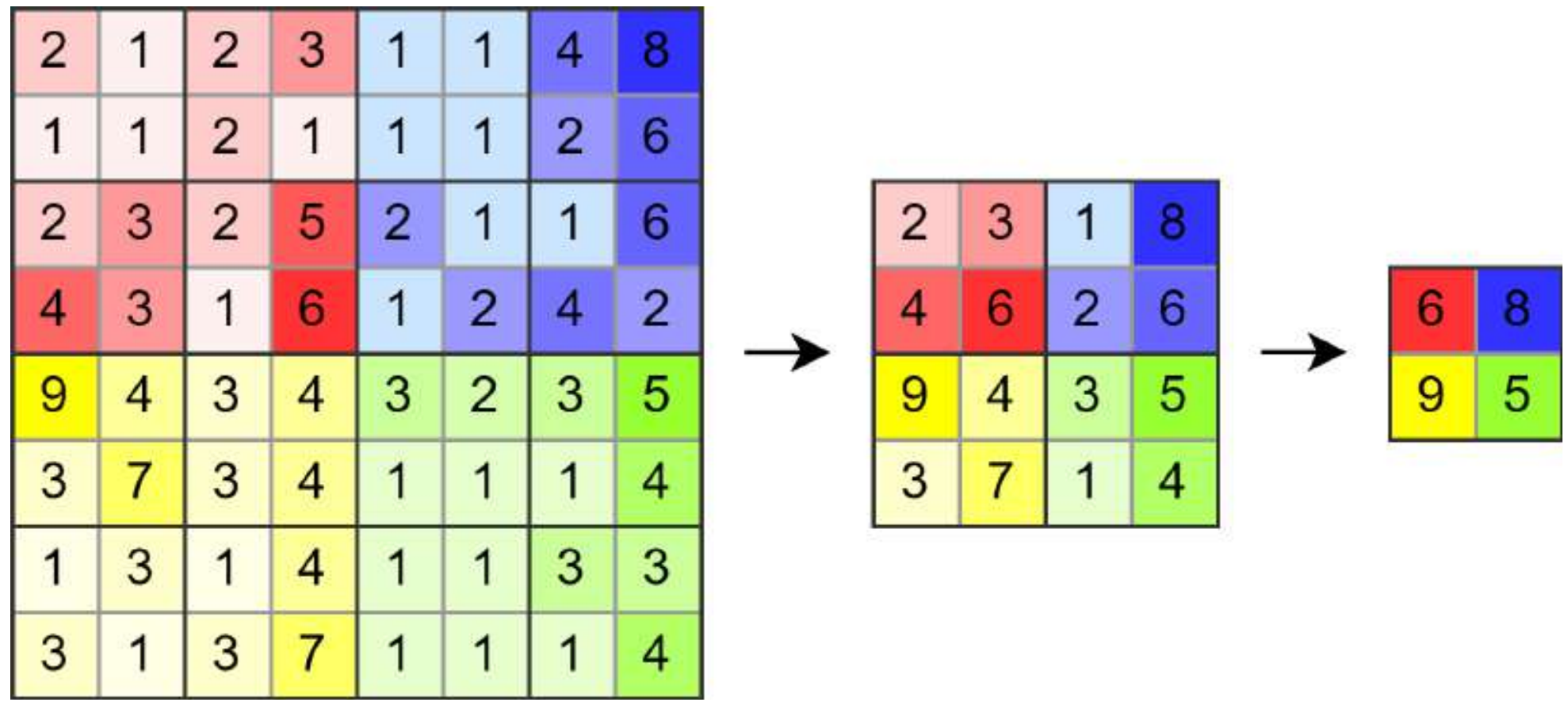

Figure 3.3: Max Pooling process explained and visualized. 2 max pooling operations are shown in this figure.

After the CNN feature extractor has learned and constructed all features maps, a classifier is needed to take the provided features and turn them into class predictions. There are various different classifiers that one can choose from, ranging from ML classifiers such as random forest or SVM classifiers to neural network based classifiers. Generally speaking, all classifiers need the features to be condensed into a 1 dimensional vector, as they are not meant to be trained with data of higher dimensionality. Feature vectors are provided in 3D tuples, so they need to be converted into 1 dimensionality first. The two most common ways of doing this are flattening and global average pooling. Flattening simply takes all the pixels of all channels and orders them into a long 1 dimensional vector where all pixels are simply a long list. Global average pooling takes the average value per channel and returns that, creating a 1D list of one

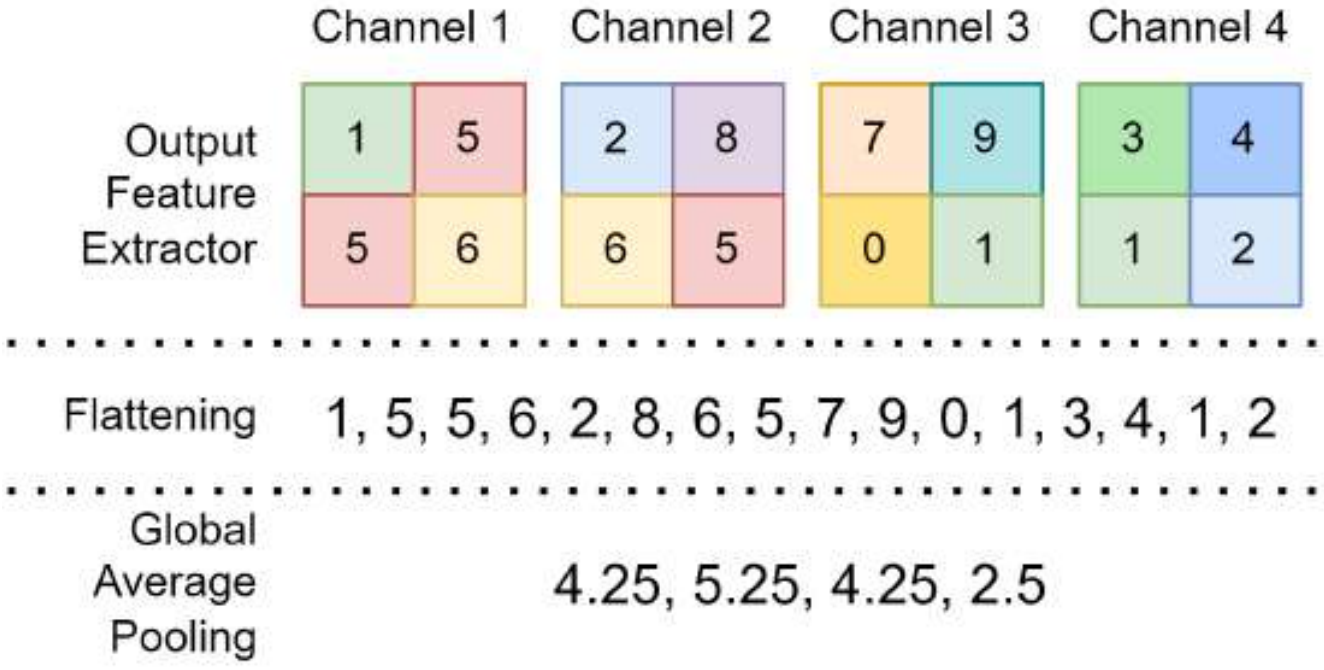

Figure 3.4: Flatten and Pooling visually explained.

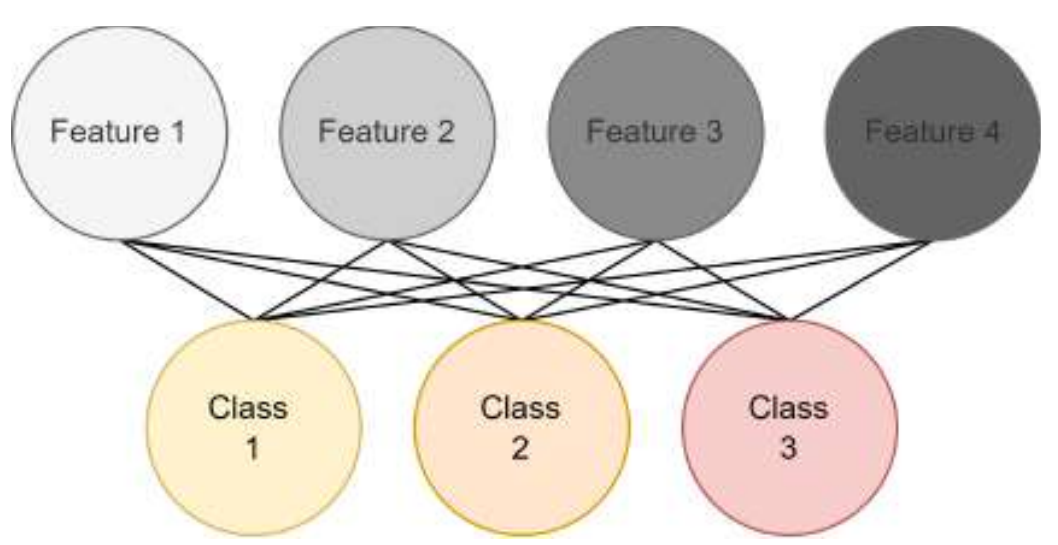

Figure 3.5: A fully connected neural network classifier top.

single value per channel (see Figure 3.4).

Then the converted 1D vector can be forwarded to the classifier. A neural network based classifier which is most often used would then for example be a fully connected network with the 1D vector as input and a final layer with softmax activation that has the number of classes as neurons.

### 3.1.1 Domain Adaptation

#### 3.1.1.1 Transfer-Learning

One way, which could be classified as the standard way, to train image classification models is to just initialize the models weights and other parameters at random. This means that before training, the model has no knowledge of this task or any other domain and starts of by essentially guessing at random and starting completely from scratch. This can generate well performing models that are only trained on the domain specific data. If, however, data is limited, or there is not enough time available for training, or if the computational resources are limited, then one might want to pivot to Transfer-Learning (TL) instead of training the model from scratch. This process uses a pre-trained model, a model that has previously been trained on a different dataset, and then trains that model, with the old weights still intact, to now work on the new domain specific data. Pre-trained models are usually trained on the ImageNet [58] dataset, a huge scale dataset with over 1.2 million images belonging to 1,000 classes. Through this training, the model will already learn many general features that are applicable to a wide range of topics which can be applied to the new domain-specific topic through Transfer-Learning. Pure TL takes the pre-trained feature extractor and freezes all parameters (they become not trainable, meaning they will not be updated during training on the new domain-specific data) and trains only the classifier top. Because of freezing all the CNN layers, the computational cost during training is heavily reduced, allowing models trained in this manner to be trained on machines much weaker than the ones that would be needed for train from scratch. Additionally, because all feature extractor layers are already intact, training will also be much quicker and because the features already exist, the change of overfitting is much reduced (see Figure 3.6).

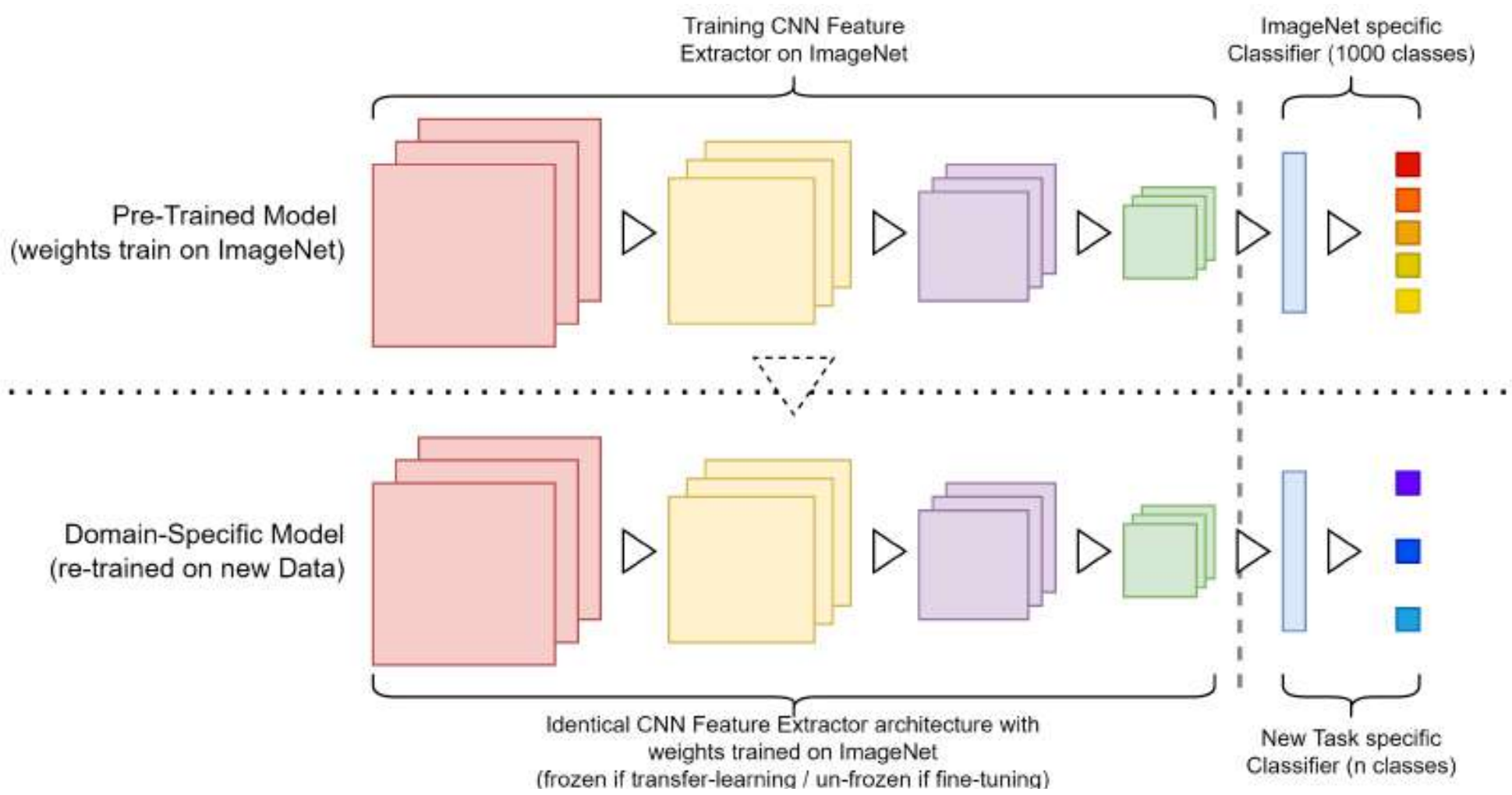

Figure 3.6: Transfer-Learning method visualized

#### 3.1.1.2 Fine-Tuning

If the results are not strong enough, or if one wants to improve the model's results as much as possible, then FT on top of the TL model is possible. To do this, the model would, after all the updates made to it via TL, be un-frozen, meaning that all layers and parameters are now trainable. This does increase computational costs again and it will also add additional training time, but it will allow each parameter to be adjusted slightly to better match the target domain-specific data from the original ImageNet features. This can extract extra accuracy percentages as the model will be even better adjusted to the target data and the resulting model will benefit from the pre-trained weights based on ImageNet, the TL adjustments, and lastly the improvements made through FT (see Figure 3.6).

#### 3.1.1.3 One-Shot Learning

In One-Shot Learning (OSL), one does not train the model in the standard sense of deep learning. Rather than that, one will use the pre-trained model, use it as a feature extractor only, and use said features as class embeddings to compare future feature embeddings to. The features are collected after the global average pooling layer, with the rest of the model's classifier top being disregarded. To do this, the pre-trained model (with the classifier top removed) is taken, and then each image for each class, one image per class, is fed into the network in inference mode to obtain the feature embeddings, which are saved for each class. This feature embedding is now the representation for said class. Future test images will be fed into the image and then their distance will be calculated for all class embeddings. The shortest distance from the test image's embeddings to any class embedding will be the projected class. As such, the model is more light weight than before (since the classifier top is removed) and no training is needed at all, only one inference run per class is required during ”training”.

#### 3.1.1.4 Few-Shot Learning

Few-Shot Learning (FSL) works essentially the same way One-Shot Learning does, with one difference. Instead of a single image per class, here $n$ images, say 5, are present during training to generate the embeddings. Each image is fed into the network, embeddings are averaged per class (if 5 images per class are used then all 5 embeddings are generated and then averaged, with the averaged embeddings being used for predictions) and used for future distance calculations and predictions.

### 3.1.2 Base Models & Foundational Models

In this work the term Foundational Model (FM) refers to large scale CNN Base Model that are trained on a large dataset to later be used to be potentially adapted, transferred, and retrained for other (potentially) related tasks [16] (but FM can generally also refer to more general AI models). Examples of such CNN architectures commonly used as Foundational Model include models such as ResNet, DenseNet, ConvNeXt, MobileNet, and more, just to name a few. These model architectures are then most often pre-trained on the ImageNet dataset, with these pre-trained weights being available to download by the public, which can then use these pre-trained models to apply them to their area of need (as mentioned above through TL and FT). ImageNet as the base dataset to pre-train the model on comes with certain advantages and disadvantages. Firstly, ImageNet is a huge dataset with over 1.2 million images across 1,000 classes. This offers the models a diverse, large scale and high quality dataset to train on, with low risks of overfitting and low risks of being too centered on a single domain. Resulting models trained on ImageNet therefore boast feature maps with a knowledge base that encompass a very large variety of different image types and classes, and therefore offers a great base on which one could build models for a wide variety of different target domain topics. However, ImageNet does, as a result of the wide variety offered, contain a large number of classes and images that will inevitably not be related to the target task, and only contain a small number of images and classes that are directly related to said task, if any at all. While the features learned on other images are still valuable and can still be used to retrain models to a new task area, they are most likely not fine-tuned for specific tasks that rely on very specific features, that one might not obtain when differentiating between bookcases and brooms on ImageNet for example. As a result, to achieve the best performance, one would have to fully fine-tune these models which is computationally expensive [17]. A model already trained on a large dataset that is related, or even identical, to the target domain, can overcome these shortcomings of potentially having to re-train the entire model in FT mode, and learns domain-specific knowledge during pre-training already. As such, the resulting model does not necessarily need to be entirely fine-tuned before being applied to the domain-specific task, because it was trained on related data, and does perform better on said data without full FT due to the presence of domain-specific knowledge in the FM weights [17]. This means that a domain-specific Foundational Model can potentially learn faster, more robust, more consistent, more powerful models on less data, while not necessarily requiring full FT.

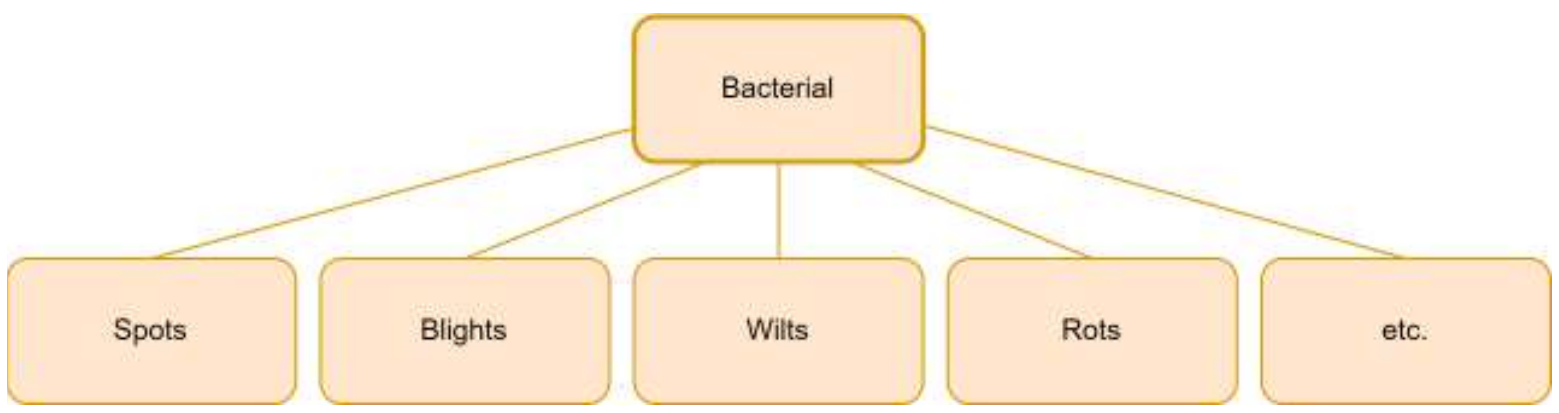

Figure 3.7: Bacterial Diseases

## 3.2 Plant Pathology using Deep Learning

These DL methods have been applied to many different fields in recent years and due to the capabilities of CNN to automatically learn features based on the labeled data, CNN are perfectly capable of being applied in many different fields successfully, granted the availability of sufficient data. For the field of plant leaf disease classification this equates to large image datasets depicting images of leaves both with and without disease symptoms that have been classified into healthy and diseased (one class per disease). This classification needs to be carried out by experts that can reliably differentiate between the different plant diseases without error to make sure that the dataset is accurate, because a faulty dataset will lead to faulty models.

### 3.2.1 Disease Types

Plant diseases can generally be grouped into 3 subcategories, based on which pathogen has infected the plant. These three categories are viral, bacterial, and fungal diseases.

#### 3.2.1.1 Bacterial

When bacteria enters a plant through openings in the plant, they can cause the plant to suffer from bacterial plant diseases. The opening through which the bacteria enter the plant can be caused by external damage (hail, heavy rain, animals, humans, etc.) or they can be natural opening of the plant for water and gas exchange for example [59, 60, 12, 61]. They then infect seeds which can spread the disease, but it can also spread from plant to plant, for example through water. Some examples of bacterial diseases are spots, scabs and rots (see Figure 3.7). Bacterial diseases are often not curable, and the plants need to be removed from the field to prevent further spreading [12].

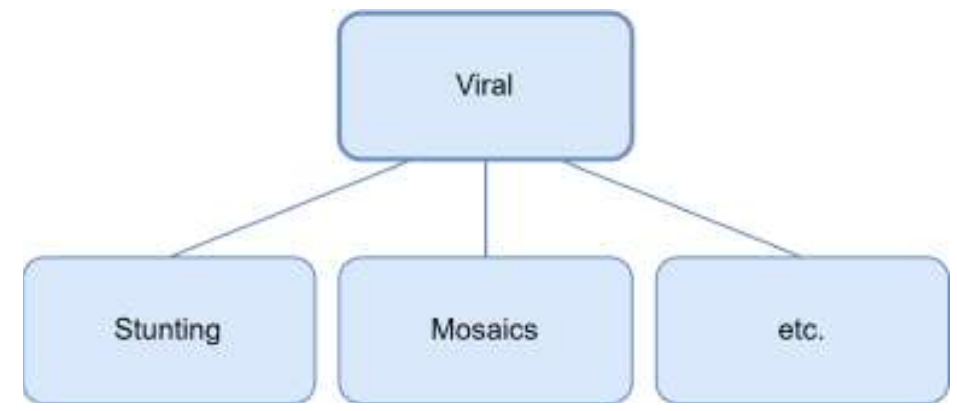

Figure 3.8: Examples of Viral Diseases

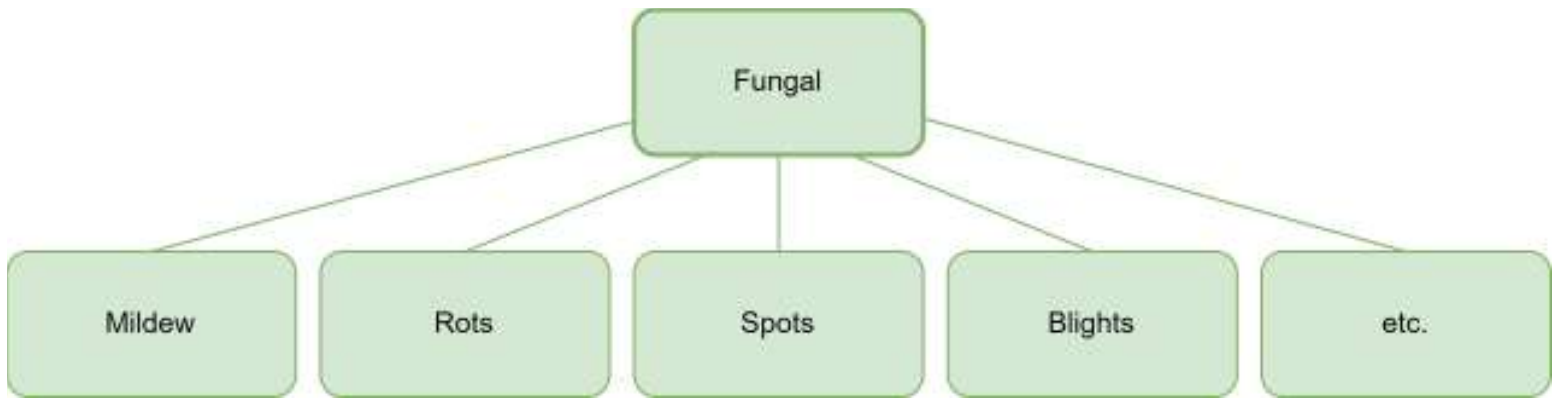

Figure 3.9: Fungal Disease Examples

#### 3.2.1.2 Viral

Viral diseases can modify the DNA of plant host cells, which will make them produce new viruses, helping them reproduce. Viral diseases can be transmitted through water, but also via insects that move from one plant to the next [60]. Some viral diseases are mosaic, mop-top, etc. [60, 12] (see Figure 3.8). Viral diseases tend to be very hard to identify [61].

#### 3.2.1.3 Fungal

Fungal diseases are often contracted through air, water or soil and can cause diseases such as blights, mildew, etc. [60] (see figure 3.9). They are often transmitted through spores that can travel quite far. Plants infected with fungal diseases can be treated by removing the infected parts of the plant and only leaving behind parts of the plants that are still healthy [12].

### 3.2.2 Datasets

Datasets, in this field, are collections of images depicting healthy images as well as images containing a set of different diseases. Datasets, disregarding the types of diseases, can be divided into to categories: field and lab datasets. As the name suggests, field datasets are taken in field (in real field conditions), while lab datasets are taken in lab (in ideal conditions). These two different capturing methods create widely different datasets with different different characteristics.

#### 3.2.2.1 Lab Datasets

Lab datasets, datasets that were taken in laboratory like conditions, are a type of datasets in which leaves and backgrounds are prepared before images are taken. This means that leaf images are usually taken one leaf at a time, with the leaf separated from the plant itself on a single color (often white) background. As such, these datasets are rather simple in terms of complexity. Where a real plant leaf is surrounded by other plants, leaves, soil, etc. the isolated leaves in the images in lab plant leaf datasets have no such distracting noise. Models trained on such data are, as a result, facing a much smaller challenge during training. They do not have to learn how to differentiate between background and foreground. This leads to models trained exclusively on lab data achieving high results on said lab data, in some cases making it hard to differentiate the trained models, since they all reach high score close to one another. Such models are, however, said to be less capable when used in real field conditions [15]. The most used dataset in the field, PlantVillage [19] is a lab dataset. Sample image of a lab dataset can be seen in Figure 4.1.

#### 3.2.2.2 Field Datasets

Field datasets, in contrast to the afore mentioned lab datasets, are taken in real in field conditions. Field datasets are taken with the leaf still attached to the plant and therefore with a lot of noise in the vicinity of the leaf that one wishes to analyze. This noise makes it quite a lot harder for DL models to work with the data, since unlike perfect lab data, here the model needs to not only analyze the leaf for symptoms, but also the image itself for the location of the leaf within in. Sadly, there is no field dataset of the size, variety, and quality of PlantVillage, so often when researchers pick open datasets, they gravitate towards the lab dataset PlantVillage. Many datasets collected by researchers are field datasets, but they are often not released to the public, leading to somewhat of a shortage of field data. Field data being more challenging makes it easier to differentiate the performance of different architectures and also train more robust and generalizable models that can also be used in field conditions. Sample images of a field dataset can be seen in Figure 4.2.

#### 3.2.2.3 Hybrid Datasets

Hybrid datasets combine image data taken in field and image data taken in lab. This can take the form of one class being exclusively field data and another class being only lab data, which can cause problems, as the model might focus more on learning the difference between data types and less on the actual disease symptoms. Other hybrid datasets are made up in a way in which each class has data of both types. This can lead to the model learning how to handle both types of data, and this can alleviate some of the drawbacks of lab data and create robust systems that can functions in different scenarios [62, 63, 64].

# 4. Datasets

As is true with every kind of Deep Learning, a large amount of high quality data is also needed for the problem of plant disease classification as well. If the data is too small, not diverse enough, of too low quality, or has other problems, then the model can not learn properly. A lack of sufficient data of good quality, for example, can lead to the model overfitting to the training data and not generalizing to the actual problem being studied, which will render the model useless, although the training results might initially sound promising. This requirement needs to be met for all classes being trained, and ideally all classes should be well balanced to encourage the model to actually analyze all images in detail to make it's predictions, instead of exploiting issues with the training data, as heavily imbalanced data could make the model lean and bias towards one class per default, simply because it has a much higher probability of being correct due to the possible imbalance. Data should also be uniform across all classes, so that the model needs to solve and discriminate based on the actual problem being studies, instead of learning to simply predict based on other differences across classes (e.g. if one class has all images taken in front of a red background and the second class only features blue backgrounds, then the model, in all likelihood will learn only that, instead of focusing on the actual foreground of the image). All this proves that the dataset is a incredibly important factor in the training of a successful model and therefore needs to be well chosen and constructed, maybe also pre-processed and augmented, to end up with a dataset that allows the model to train properly and well.

Creating datasets for this plant disease use-case is not easy, as one needs access to a large enough amount of plants, invest a lot of time and effort to take pictures of adequate quality, and then have an expert label said image data by disease type. This is why, as part of this thesis, no new data is collected. Instead this work falls back on image datasets that are available online and adjust and modify them to fit the goals and scope of this work. In this chapter the different datasets chosen will be discussed. First all datasets will be introduced. All datasets will be explained through a table of key information, sample images for each class, as well as a plot showcasing the distribution of images across all classes. These datasets will be used to test a wide variety of models to find a suitable baseline model to build a new architecture on, and then a selection of these datasets will be combined to train said new model on. When discussing image datasets for plant disease identification, there are 2 major distinct groups among them. One group contains plant images taken in the lab, while the other group includes image datasets taken in field. Lab images are inherently easier for DL models to handle, as the leaf is completely isolated in the image and the background is uniform and without any complexities (see Figure 4.1).

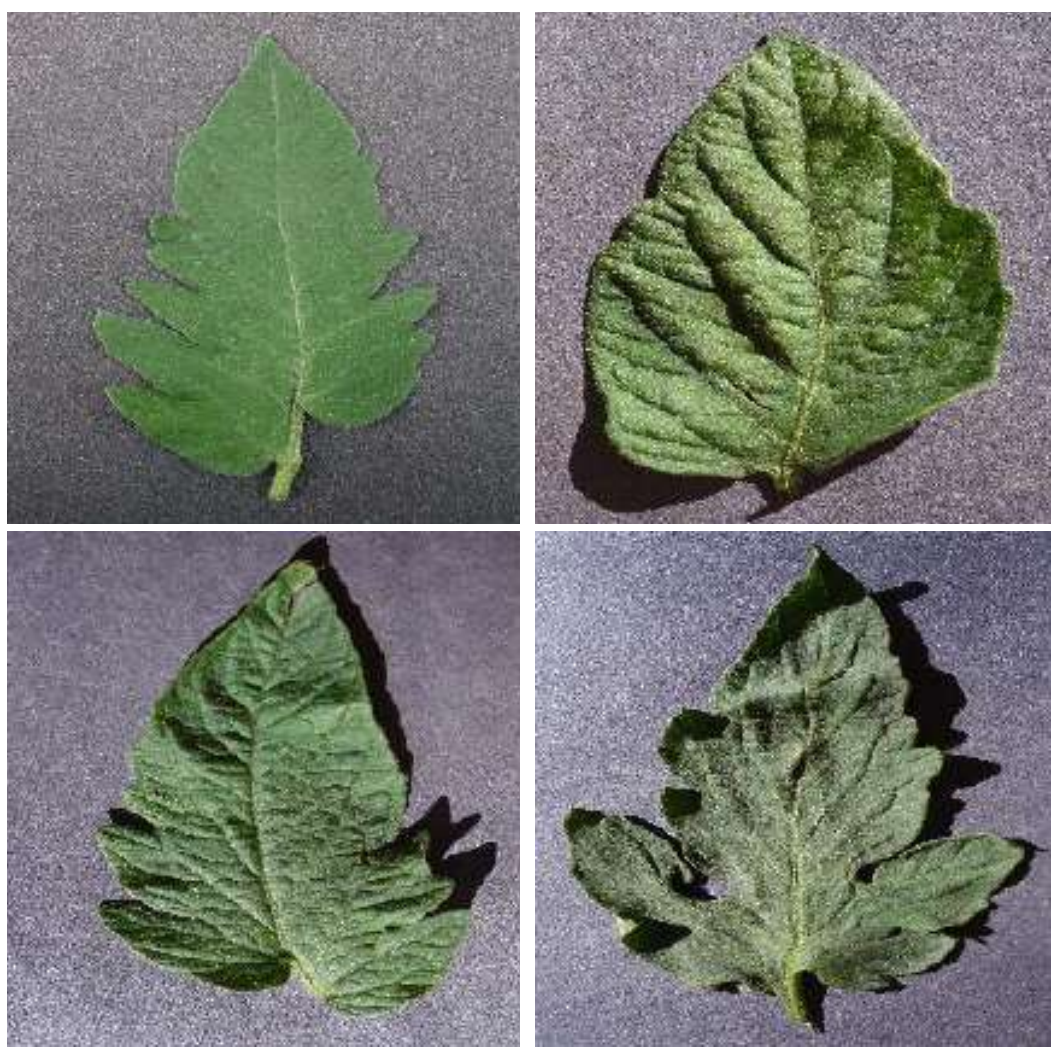

Figure 4.1: Sample of Tomato Leaf images taken under Lab conditions from the Plant Village dataset [19].

Images taken in field are much more complex and difficult to predict since the background, instead of being just one uniform color, they often feature soil, other plants, and other more complex features, that the model needs to handle. But while these images are more complex and harder to train, they also produce better performing classifiers that can be used in real world applications in the end [11]. While lab image trained networks work well as proof of concepts, they fail when one tries to use them with in field images, as would be the case if farms would move to DL models for practical use. As such, it is better to use field image datasets or hybrid sets that include or combine the two, because field images datasets essentially mirror the way images would look like if someone were to take them in field when inspecting their field for diseases with a camera or phone (see Figure 4.2) and should therefore translate to real world application much more easily [62, 63, 64].

Datasets also vary by which plants are included as well as which diseases are present in it. Multi-plant datasets, as well as datasets that only contain a single plant can be found publicly available online. Datasets also very in the number of diseases per plant type that are included within, ranging from simple binary classification to multi-class problems. Lastly, the amount of images per class also heavily fluctuates between datasets, with some datasets having images in the hundreds, while other datasets offer tens of thousands of images, both to varying degrees in terms of per-class-distribution.

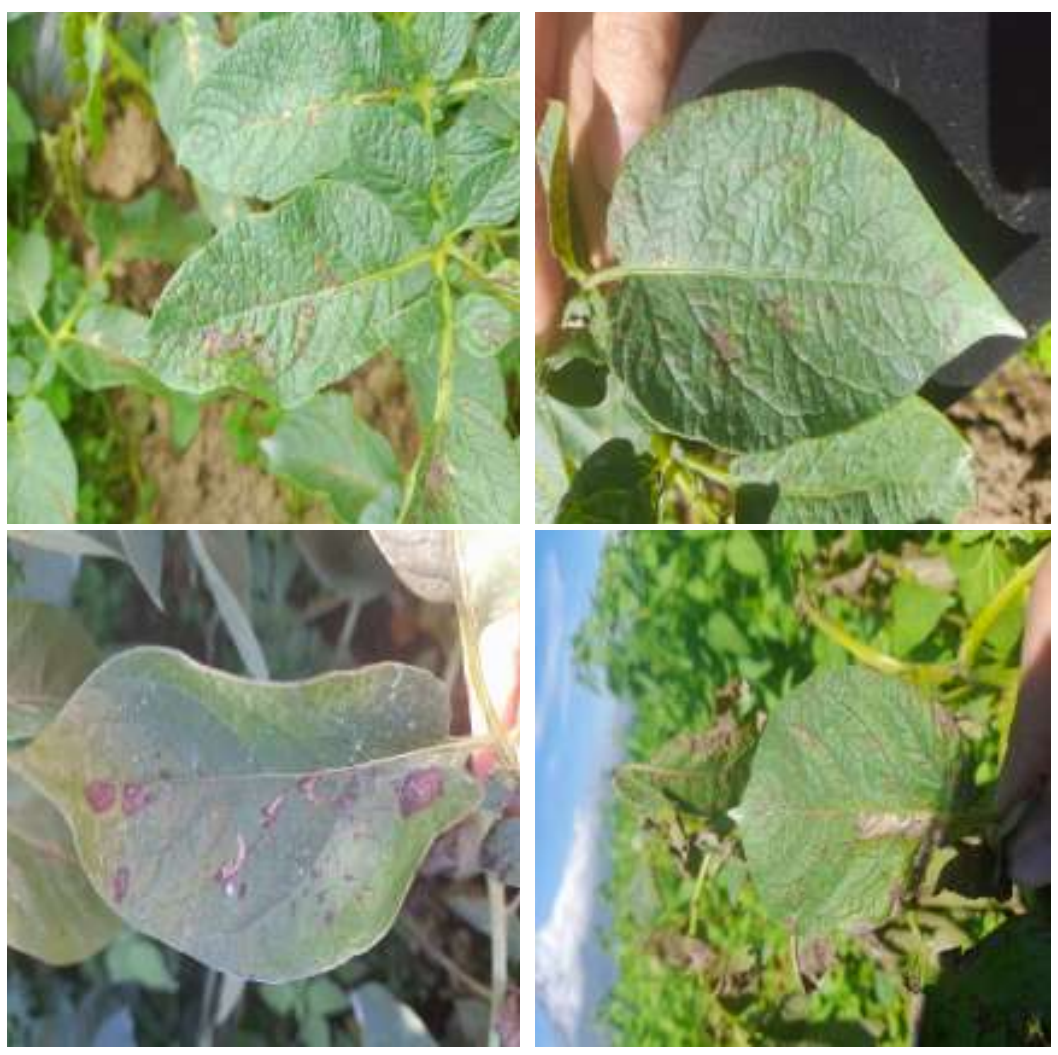

Figure 4.2: Sample of Potato Leaf images taken taken in the field conditions from the novel Potato dataset [65].

With all that, as part of this thesis, openly available datasets will be presented, categorized and analyzed in this chapter. First lab image datasets will be introduced, then field datasets and after that hybrid sets will be listed. More detailed information for each dataset can be found in Appendix A.

## 4.1 Lab Datasets

### 4.1.1 Plant Village (plantVillage)

The Plant Village Dataset [66, 19, 67] might currently be the most popular and commonly used dataset in plant disease classification. It contains various different plants and their corresponding diseases all taken under lab conditions (see Figure A.2) at different research stations across the USA. The distribution of images per dataset is rather imbalanced (see Figure A.1). It is a also one of the biggest openly available datasets in terms of image numbers. Classes per plant range from 1 to 10 with a healthy class always included (see Table A.1). It has been used in numerous papers [68, 69] and tends to produce very strong results, with the caveat that lab images are much easier to handle.

### 4.1.2 Tomato Village (tomatoVillage)

The Tomato Village dataset [70, 71] is a single plant dataset focusing on tomato plant leaf diseases. It includes 7 diseases as well as a healthy class (see Table A.2), all of which are taken under laboratory conditions (see Figure A.4). Classes are mostly imbalanced in this dataset (see Figure A.3). This dataset, when compared to the tomato images in the Plant Village dataset [19], does have some overlap in diseases, but does also introduce new diseases as well. This is due to the nature of this dataset being captured at three different locations in India, where other diseases are prominent. Tomato Village has seen prior use in research projects [72].

### 4.1.3 Potato Disease Leaf Dataset (pld)

The PLD Potato Leaf Disease dataset [73] is a dataset of potato leafs collected in Pakistan under lab conditions as can be seen in Figure A.6. The dataset contains healthy potato leaf images alongside 2 more classes of diseases, totaling 4,062 images across all 3 classes (see Table A.3) which are fairly well balanced (see Figure A.5). PLD has been used in research before as well [74].

### 4.1.4 Tea Sickness Dataset (tea)

The Tea sickness dataset [75] includes tea leaf images from 8 different classes (see Table A.4) captured in eastern Africa. All images were taken in laboratory like conditions (see Figure A.8), resulting in a dataset with 885 total images, which have proven to train strong models in prior work [76]. When looking at the distribution of those images across all classes, one can see that the dataset is rather well balanced (see Figure A.7).

## 4.2 Field Datasets

### 4.2.1 Cassava Leaf Disease Classification (cassava)

The Cassava Leaf Disease Classification dataset [77, 78] was originally used for a Kaggle coding competition, but has since also found use in various research papers [79, 80]. All images are taken in field (see Figure A.10) in Uganda. It contains 5 classes, 4 of which are diseased leaves with the fifth being the healthy leaf class, totaling 21,397 images (see Table A.5). While the dataset is heavily imbalanced (see Figure A.9), it is one of the bigger and high quality datasets of infield plant disease images.

### 4.2.2 CD&S Dataset (cds)

The CD&S dataset [81, 82] is a corn plant leaf disease dataset. Images were acquired in West Lafayette, IN, USA at Purdue University and all images were taken in field (see Figure A.12). The dataset contains a total of 3 classes (see Table A.6), all of which are disease classes. One thing of note is the near perfect distribution of images per class, as can be seen in Figure A.11. Prior work has shown good results with this dataset [83].

### 4.2.3 Cucumber Disease Recognition Dataset (cucumber)

The Cucumber Disease Recognition Dataset [84, 85] contains images of both cucumber fruits as well as cucumber plant leaves. Since this thesis focuses on leaf images only, the fruit image classes have been discarded and not considered. They will also not be present in the tables and images below. The dataset contains 5 leaf classes (see Table A.7), one of which is healthy leaf images (see A.14). The distribution across classes is near perfect (see Figure A.13). The dataset was successfully used in [86] fro example.

### 4.2.4 DiaMOS Plant (diaMos)

The DiaMOS Plant Dataset [87, 88], just like the aforementioned Cucumber Dataset, also includes fruit images alongside leaf images, and just like with the Cucumber Dataset, here too the fruit images will be ignored and, therefore, not listed in this work. Severity data is also present in this dataset, but that data too will not be considered. Pear leaf images are divided into 4 heavily imbalanced (see Figure A.15 and Table A.8) distinct classes, on of which includes healthy images. All images were taken in field (see Figure A.16) in Italy and the dataset has already been applied in plant disease related research [89].

### 4.2.5 Paddy Doctor (paddy)

The Paddy Doctor Dataset [90, 91] is a dataset of in field images of rice leafs taken in paddies in India, for samples see Figure A.18. The dataset contains 10,407 images over 10 classes (see Table A.9), which are somewhat imbalanced as can be seen in Figure A.17. [92] is an example of the Paddy Doctor dataset being used in academic work.

### 4.2.6 Plant Pathology 2020 - FGVC7 (fgvc7)

The Plant Pathology 2020 - FGVC7 Dataset [93, 94] is a dataset comprised of apple leaf images including a total of over 1,500 apple leaf images, all of which were captured in field in the US (see Figure A.20). The dataset was originally used for the Plant Pathology Challenge at the Fine-Grained Visual Categorization (FGVC) workshop at the 2020 CVPR conference. The dataset includes two distinct diseases and a multi-disease class (see Table A.10). When looking at Figure A.19, one can see that the FGVC7 dataset's distribution is pretty even, except for the multi-disease class. And due to its high popularity and exposure at the conference, it has seen a vast amount of research interest ([95, 96]).

### 4.2.7 Plant Pathology 2021 - FGVC8 (fgvc8)

One year after the above mentioned Plant Pathology 2020 - FGVC7 challenge, the challenge was renewed with an extended dataset as the Plant Pathology 2021 - FGVC8 challenge and dataset [97, 98] (see Figure A.22 for sample images). This newer dataset now features over 18,000 images divided into 12 classes (see Table A.11) that overlap and also vary in complexity, which are unbalanced, as can be seen in Figure A.21. This popular dataset has seen use in research work, e.g. [99, 100].

### 4.2.8 PDD271: Plant Disease Recognition Dataset (pdd271)

The PDD271 dataset [101, 102] is a very large dataset originally containing a total of 271 different classes across 220,592 total images, all taken in field. This dataset, however, is proprietary and not open to the public. Only a sample containing 10 images per class can be accessed online, except for 10 classes that seem to be unaltered and still seem to contain all images. In this thesis only these 10 classes will be considered for training. This subset of PDD271 now contains a total of 7,555 field images (see Figure A.24), of 5 different plants (see Table A.12) with reasonably well balanced classes, as shown in Figure A.23.

### 4.2.9 Strawberry Disease Detection Dataset (sms)

The Strawberry Disease Dataset [103, 104] contains images of leaves, fruits as well as blossoms. Due to the nature of this thesis working with leaf images exclusively, the blossom images as well as the fruit images have been discarded. As such, these classes and their images will also not be listed here. One more thing to note, it seems as if the dataset has already been enlarged by a lot of augmentation. As is, the dataset contains a total of 3 classes, all diseases, after removing the blossoms and fruits, as can be seen in Table A.13. Images are all taken in field (see Figure A.26), resulting in a rather well balanced dataset (see Figure A.25). This data has been utilized in a number of studies (e.g. [105, 106]).

## 4.3 Hybrid Datasets

### 4.3.1 plantDoc

The plantDoc dataset [15, 107] is a multi-plant dataset that features a wide variety of plant across 28 total classes (see Table A.14). It includes images are taken in field, but when looking at figure A.28, it becomes clear that the dataset also includes images that realistically should have been excluded from the dataset (e.g. PowerPoint slides). Class distribution is not too bad, except for some outlier classes (see Figure A.27). Regardless, planDoc has seen researcher attention [108].

### 4.3.2 Taiwan Tomato (taiwanTomato)

The Taiwan Tomato Dataset [109] is a dataset that includes images for 5 different classes of tomato diseases as well as a class for healthy images. The classes contain both, images taken in the field as well as images taken under lab conditions (see Figure A.30). The dataset also includes images taken from the Plant Village dataset [19], but these were not included in this work. Augmented images were also ignored, as this work applies its own set of augmentations. As can be seen in Table A.15, the dataset is rather small but the distribution is not as imbalanced as it is in other datasets. The Taiwan Tomato dataset has already seen use in research in the past [110].

### 4.3.3 Potato Leaf Disease Dataset in Uncontrolled Environment (novelPotato)

This novel Potato image dataset for leaf disease classification [65, 111] contains a total of 7 classes, one of which is healthy, taken in field and lab-like conditions (see Table A.16 and Figure A.32) in Indonesia. The per-class distribution is unbalanced, especially when looking at the ”Nematode” class. This dataset had seen prior use in research (e.g. [112]).

### 4.3.4 Rice Leaf Diseases Dataset (rldd)

The Rice Leaf Diseases Dataset [113] by Antony and Prasanth is a dataset that contains 3 classes, all of which are diseased (see Table A.17). The dataset is rather small in size and it seems as if some amount of augmentation has already been used to expand the bacterial blight class, which as a result of that is now much larger than the other 2 (see Figure A.33). Images seem to be taken under varying conditions, some taken in conditions somewhat similar to the ones found in lab datasets, while others are taken in field, as can be seen in Figure A.34.

### 4.3.5 Sugarcane Leaf Dataset (sugar)

This Dataset containing sugarcane images [114, 115] originally contained a total of 11 classes, one of which is dried leaves. This work will not consider the dried leaf class, so the modified dataset totals 10 classes (see Table A.18), containing a healthy class. Images were acquired in India and the dataset includes both field images and also images taken under lab-like conditions (see Figure A.36). The number of images per class varies rather heavily (see Figure A.35). This dataset has proven to work well in recently published papers [116, 117].

# 5. Models

Datasets are essential to assure good performing models. But just like the datasets, choosing the right model architecture has large impact on the final result. As such, one must consider which model to use wisely. CNN models have come a long way in recent years, with new FM being published that introduced new methods. However, different models perform differently depending on the way they are applied but also depending on which domain they are being applied to. In order to build a new dedicated plant leaf disease classification model or to improve an existing model to perform better, one first needs to learn which models are inherently potent in the field. This chapter will introduce and briefly explain some of the FM that see usage these days which will be tested on their performance on all the datasets mentioned before in a Transfer-Learning fashion, in order to gain insights and learn which model architecture and methods can be built upon for the new model.

## 5.1 ResNet

The ResNet architecture [118] utilizes so-called residual connections (see Figure B.1). These residual connections pointwise add the input of a layer to its output to preserve features captured in earlier layers and to provide a more direct path to earlier layers which improves the gradient flow during learning. To enable such connections, the layers outputs need to be of the same dimensionality as the inputs so padding is set to same within the blocks and the dimensionality only changes through stride 2 convolutional layers in the beginning of each block.

## 5.2 VGG

The VGG [119] model architecture is a rather simple model when looking at the blocks is is build out of (see Figure B.2). The blocks are simply made up of regular convolutional layers and max pooling layers at the end of each block.

## 5.3 Inception

The Inception [120] family of CNN models makes use of Multi-Scale Processing (MSP), batch normalization, Factorized Convolutions (FtC), and label smoothing. Multi-Scale Processing is the method of using different sized kernels across the network (often in parallel) to recognize differently sized features, batch normalization normalizes the outputs of a layer to stabilize training. Factorized Convolutions split kernels into 2 (e.g. a 5x5 kernel into a 1x3 and into a 3x1 kernel) which minimizes the amount of parameters needed. Lastly, label smoothing improves generalization.

## 5.4 EfficientNet

EfficientNet [121] is a model architecture family that is based on the ideas of dropout layers and stochastic depth, swish activation, and Mobile Inverted Bottleneck (MIB) convolution. The dropout layers ignore certain connections per layer to improve generalization, whereas stochastic depth ignore entire layers for the same effect. Swish is a activation function that improves the normally used ReLU and Mobile Inverted Bottleneck convolutions use Pointwise Convolution (PwC) and Depthwise Convolution (DwC) to reduce parameter counts. Depthwise Convolution only operate on a single channel (regular convolutions operate on all) and Pointwise Convolution are essentially 1D convolutions that only look at 1 pixel per channel but across all channels. In MIB convolutions, first a PwC layer increases the number of channels, before a DwC, followed by another PwC (in a Depthwise Separable Convolution (DSC) block) layer parse the image for features.

## 5.5 NASNet

NASNet [122] is a architecture that was made with optimization algorithms including Reinforcement Learning (RL). It is made up of different blocks and cells with max pooling, average pooling, regular convolutions, and Depthwise Separable Convolution.

## 5.6 MobileNet

MobileNet [22] is a family of models mainly concerned with building models that are efficient and light-weight. To achieve these goals methods such as the above mentioned Mobile Inverted Bottleneck, and Squeeze & Excitation (SE) were employed. Squeeze & Excitation is a attention mechanism that focuses on channel attention by using a global average pooling layer with an MLP to identify the importance of the given channels.

## 5.7 ConvNeXt

ConvNeXt [123] is a model architecture that is based on the principles of Vision Transformer (ViT), since they have managed to achieve high levels of performance in recent works. As such, the creators of ConvNeXt built a model with similar characteristics based on CNN to rival these Vision Transformer and to combine the advantages of both. It does that by including techniques such as large kernels (7x7), layer normalization, residual connections GELU activations and inverted bottleneck (see Figure B.3).

## 5.8 DenseNet

The DenseNet [124] family of networks is also based on residual type connections (see Figure B.4), but unlike ResNet, here dense connections are used. They work differently where, instead of adding layers, they append the channels to one another, also preserving earlier features and reducing the vanishing gradient problem in the mean time.

# 6. Benchmarking

To this day there seems to be no real consensus on which model is best suited to handle the task of plant leaf disease classification. This understanding is helpful and even somewhat essential when trying to chose a model for a new project in the field. But not just for picking already existing models does one require this knowledge, but also to build a new and better model, one best first find out which model is inherently well suited to be used as a baseline, and also to see which techniques might be better applied to the task than others. Additionally, as mentioned in earlier chapters, datasets are a part of this field that requires improvement. So, while benchmarking the models, a number of datasets will be tested at the same time. This will eliminate the bias any given model might have to a certain dataset and therefore generate much more generalizable and robust data, while also giving insight into the currently available datasets. In this chapter the methods of benchmarking, in an effort to find suitable data for training and a suitable baseline model, will be explained and the results gathered will be presented and contextualized. Based on these results, in later chapters, a new dataset and model will be built with a sole focus of improving performances in the field of plant leaf disease classification.

## 6.1 Benchmarking Methods

In this benchmark, to find a suitable baseline model to build upon and datasets to combine and train on, a total of 23 foundation models have been trained on 18 different plant leaf disease datasets. Each model was trained for up to 200 epochs (unless early-stopping triggered) for both TL and FT on each dataset (models are pre-trained on ImangeNet). Each combination was trained 5 times each to get more comparable and representative results. A full list of all datasets can be found in Table 6.1 and all models are listed in Table 6.2, with key information being provided for both. The workflow can be seen in Algorithm 1 and in Figure 6.1. This chapter is based on [125]. The experiments in the paper are run for 5 iterations, while in this thesis they were only run for 2. For more generalized results, read [125].

### 6.1.1 Datasets

As pointed out in prior work [14, 11, 126], datasets for plant leaf disease classification are still at a point where they are far from being abundant. New datasets are still often kept private, many datasets are rather small in size, popular datasets are often taken only in lab conditions, and often datasets are somewhat limited in the selection of plants, or diseases, or classes that they contain. In an effort to learn which datasets are better equipped to be used to train a large scale model on them, a set of 18 datasets (see Figure 6.1) that are openly available to the public have been selected for this benchmarking. Based on these results a new dataset can be constructed that contains images from datasets that show good characteristics and based on their performance in this benchmark.

Table 6.1: List of all the datasets considered in this benchmark with all key-metrics.

| Ref. | Dataset | Type | Plant | Num. Class | Total Img. |
|---|---|---|---|---|---|
| [77] | cassava | Field | Cassava | 5 | 21,397 |
| [81] | cds | Field | Corn | 3 | 1,571 |
| [85] | cucumber | Field | Cucumber | 5 | 800 |
| [87] | diaMOS | Field | Pear | 4 | 3,006 |
| [93] | fgvc7 | Field | Apple | 4 | 1,821 |
| [97] | fgvc8 | Field | Apple | 12 | 18,632 |
| [65] | novelPotato | Hybrid | Potato | 7 | 3,076 |
| [90] | paddy | Field | Rice | 10 | 10,407 |
| [101] | pdd271 | Field | Multi | 10 | 7,555 |
| [15] | plantDoc | Hybrid | Multi | 28 | 2,577 |
| [19] | plantVillage | Lab | Multi | 38 | 54,304 |
| [73] | pld | Lab | Potato | 3 | 4,062 |
| [113] | rldd | Hybrid | Rice | 3 | 524 |
| [104] | sms | Field | Strawberry | 3 | 1,583 |
| [114] | sugar | Hybrid | Sugarcane | 10 | 6,405 |
| [109] | taiwanTomato | Hybrid | Tomato | 6 | 622 |
| [75] | tea | Lab | Tea | 8 | 885 |
| [70] | tomatoVillage | Lab | Tomato | 8 | 4,526 |

### 6.1.2 Models

Since this benchmarking is based on the idea of Transfer-Learning, to both extract high level performance out of the models while also doing so in less time, models have been picked based on their performance potential, reputation, renown, their usage in recent works, and their availability of pre-trained ImageNet weights. Following these criteria, the models picked to be trained in this benchmark are listed below. A total of 23 models (see Table 6.2) belonging to various different Foundational Model architectures have been selected to be trained in order to find which models are better suited to be used in plant leaf diseases classification and which ones are not so suitable. Based on these findings, the later process of building/improving a new BM will be more streamlined and better informed.

Table 6.2: A list of all the models that will be considered for the benchmark in this work. The metrics were obtained from the official TensorFlow implementations via the model summary [127].

| Ref. | Model | Total Params | Train. Params | Total Mem. | Train. Mem. |
|---|---|---|---|---|---|
| [123] | ConvNeXtSmall | 49.45M | 49.45M | 188.65 | 188.65 |
| [123] | ConvNeXtTiny | 27.82M | 27.82M | 106.13 | 106.13 |
| [124] | DenseNet121 | 7.04M | 6.95M | 26.85 | 26.53 |
| [124] | DenseNet169 | 12.64M | 12.48M | 48.23 | 47.62 |
| [124] | DenseNet201 | 18.32M | 18.09M | 69.89 | 69.02 |
| [121] | EfficientNetV2B0 | 5.92M | 5.86M | 22.58 | 22.35 |
| [121] | EfficientNetV2B1 | 6.93M | 6.86M | 26.44 | 26.17 |
| [121] | EfficientNetV2B2 | 8.77M | 8.69M | 33.45 | 33.14 |
| [121] | EfficientNetV2B3 | 12.93M | 12.82M | 49.33 | 48.91 |
| [121] | EfficientNetV2M | 53.15M | 52.86M | 202.75 | 201.64 |
| [121] | EfficientNetV2S | 20.33M | 20.18M | 77.56 | 76.97 |
| [120] | InceptionV3 | 21.80M | 21.77M | 83.17 | 83.04 |
| [128] | InceptionResNetV2 | 54.34M | 54.28M | 207.28 | 207.05 |
| [129] | MobileNetV3Large | 2.99M | 2.97M | 11.43 | 11.34 |
| [129] | MobileNetV3Small | 0.94M | 0.93M | 3.58 | 3.54 |
| [122] | NASNetLarge | 84.92M | 84.72M | 323.93 | 323.18 |
| [122] | NASNetMobile | 4.27M | 4.23M | 16.29 | 16.15 |
| [118] | ResNet50V2 | 23.56M | 23.52M | 89.89 | 89.72 |
| [118] | ResNet101V2 | 42.63M | 42.53M | 162.61 | 162.23 |
| [118] | ResNet152V2 | 58.33M | 58.19M | 222.52 | 221.97 |
| [119] | VGG16 | 14.71M | 14.71M | 56.13 | 56.13 |
| [119] | VGG19 | 20.02M | 20.02M | 76.39 | 76.39 |
| [130] | Xception | 20.86M | 20.81M | 79.58 | 79.37 |

### 6.1.3 Benchmark Workflow

To identify which models and datasets are indeed better performing, each model has been trained on each dataset once only using TL and once with FT added on top. Each model was trained for 200 epochs each (TL & FT) with early stopping enabled with a patience of 50 to eliminate overfitting. After training is completed, the model will be reset to its best performing epoch based on the checkpoint callback considering the validation loss. Then that best model will be tested on the train set, the validation set, and the test set one more time and metrics such as accuracy, F1-score, etc. will be recorded, compared, contextualized, and analyzed. Each model is run for 2 iterations, and results are the average across both runs, in order to obtain and present more representative results.

**Algorithm 1** Loop used to train all models on all datasets for benchmarking based on [125]. This loop is carried out twice for each model-dataset combination and then results are averaged.

```
 1: for each pre-trained CNN model F do:
 2:     for each Dataset D do:
 3:         Init Dataset D
 4:         Init Hyperparameters
 5:         Init Checkpoint C & Early-Stopping e with patience = 50
 6:         Init Feature Extractor CNN F with ImageNet weights θ
 7:         Init Classifier C with random weights γ
 8:         Init Model M by merging F with C
 9:         # Transfer-Learning
10:         Freeze F
11:         Train M with D until num_epochs or until e
12:         Load Best weights for M from c
13:         Eval Model M for D(train), D(val), D(test)
14:         # Fine-Tuning
15:         Un-Freeze F
16:         Train M with D until num_epochs_tuning or until e
17:         Load Best weights for M from c
18:         Eval Model M for D(train), D(val), D(test)
19:         Log Results
20:     end for
21: end for
```

## 6.2 Benchmarking Results

When looking at the results obtained during the benchmarking experiments some trends become apparent. Table 6.3 shows the weaknesses that some of the datasets currently available pose. Datasets such as plantDoc and taiwanTomato show rather weak overall performances when looking at the test results. PlantDoc as a dataset is made up of images that are well suited to the task, but also of images that are far from the domain (e.g. PowerPoint slides). With these suboptimal images being part of the dataset, it is hardly possible to train a robust model with strong feature maps. Datasets such as taiwanTomato, novelPotato, fgvc8, cassava, and fgvc7 show low F1-Scores, which implies imbalanced data which leads to overfitting and biasing to larger classes, while not properly learning the features that make up less represented classes. One the other end of the spectrum datasets such as pld, plantVillage, and pdd271 stick out due to their very high performance scores. PlantVillage and pld are lab datasets with adequately large classes (images per class). Lab datasets are easier to learn, making these datasets inherently less challenging, but the well balanced and well sized classes most likely further help to reach these levels of performance. However, lab datasets are not very robust to training in real world applications in field [15], so purely lab data trained models are not, while the results might suggest otherwise, the ideal choice. The pdd271 dataset on the other hand is taken in field, but does come with the caveat that only 10 classes are available with only 5 plant types (out of which 2 only have a single class). This showcases that a model trained on well balanced and high quality data, even if it is taken in field, can achieve results that are outstanding, signifying the importance for a solid high quality dataset. These findings will be considered when creating a new dataset for training and benchmarking.

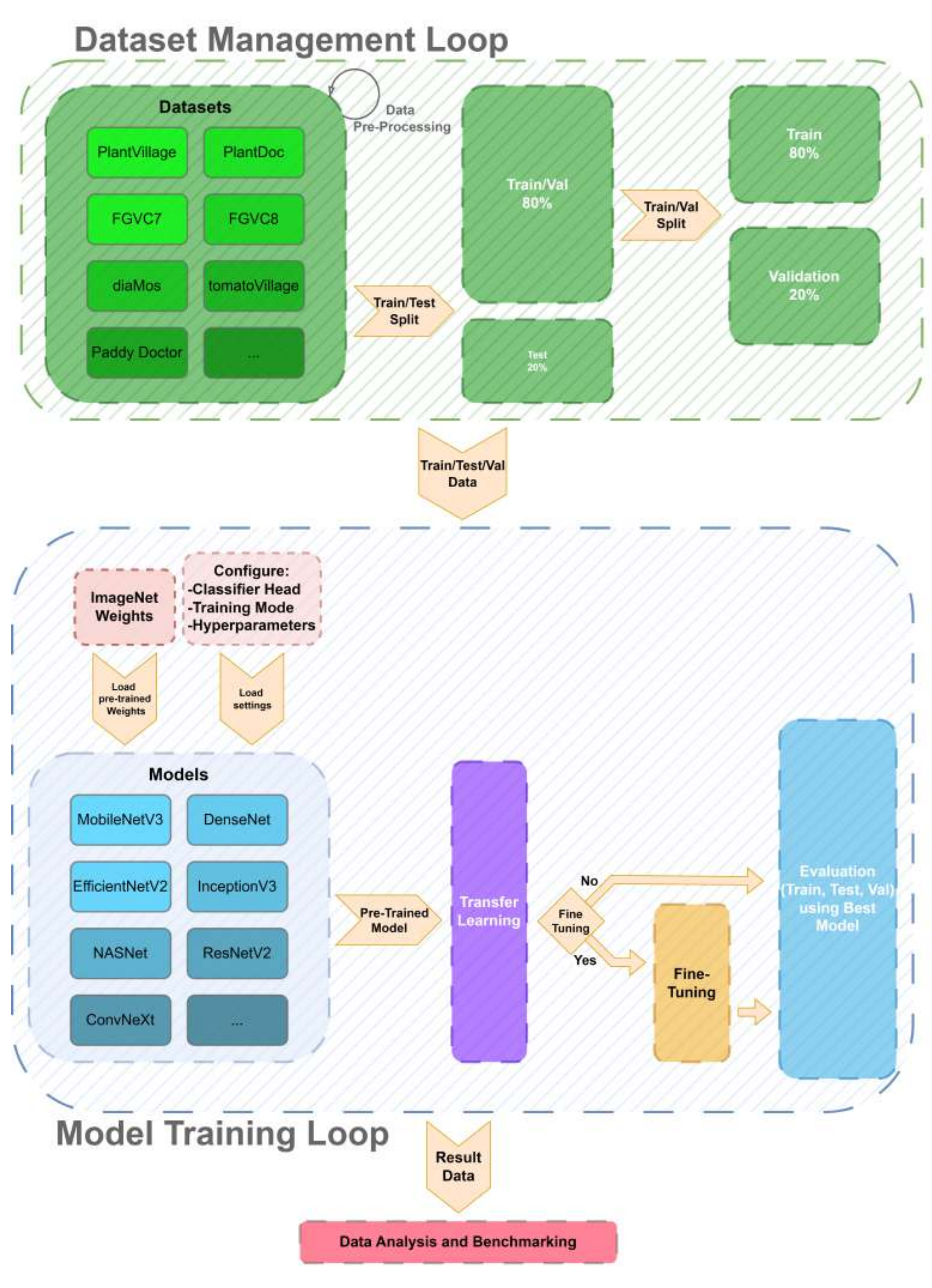

Figure 6.1: The workflow used during benchmarking

Table 6.3: The average performance metrics for each models averaged per dataset. All values are the 5 run average and all of the metrics are based on the test dataset, except for epochs which are taken during testing. FT stands for fine-tuned.

| **Dataset** | **Acc.** | **F1** | **Epochs** | **Acc. FT** | **F1 FT** | **Epochs FT** |
|---|---|---|---|---|---|---|
| pdd271 | 0.9748 | 0.9739 | 84.2609 | 0.9904 | 0.9903 | 88.6522 |
| plantVillage | 0.9747 | 0.9662 | 68.1739 | 0.9934 | 0.9906 | 60.3478 |
| pld | 0.9721 | 0.9707 | 48.0217 | 0.9872 | 0.9865 | 13.5652 |
| rldd | 0.9584 | 0.9432 | 130.0870 | 0.9685 | 0.9559 | 27.7609 |
| cds | 0.9340 | 0.9331 | 90.0652 | 0.9723 | 0.9719 | 64.5435 |
| sugar | 0.9166 | 0.8736 | 68.5652 | 0.9171 | 0.8726 | 4.4565 |
| sms | 0.9081 | 0.9075 | 116.9130 | 0.9310 | 0.9309 | 63.8261 |
| paddy | 0.8779 | 0.8665 | 113.1522 | 0.9209 | 0.9109 | 17.8043 |
| tomatoVillage | 0.8474 | 0.8213 | 85.1957 | 0.8722 | 0.8491 | 6.1087 |
| tea | 0.8418 | 0.8372 | 125.3913 | 0.8852 | 0.8809 | 30.3478 |
| cucumber | 0.8322 | 0.8317 | 72.0217 | 0.8586 | 0.8578 | 6.4565 |
| diaMos | 0.8252 | 0.7539 | 38.7391 | 0.8544 | 0.8227 | 2.9348 |
| fgvc7 | 0.8234 | 0.6681 | 49.2826 | 0.8794 | 0.7627 | 6.3696 |
| cassava | 0.7521 | 0.5780 | 20.8261 | 0.7911 | 0.6447 | 3.9348 |
| fgvc8 | 0.7480 | 0.4042 | 44.1304 | 0.8121 | 0.4715 | 5.6304 |
| novelPotato | 0.7196 | 0.6809 | 53.9565 | 0.7650 | 0.7350 | 5.6739 |
| taiwanTomato | 0.6450 | 0.6061 | 59.1739 | 0.6705 | 0.6315 | 4.1304 |
| plantDoc | 0.4959 | 0.4669 | 75.8043 | 0.5151 | 0.4915 | 3.3696 |

When inspecting the results grouped by model (Table 6.4 and Table 6.5), one can again see large deviations between the performances near the top and the performances near the bottom of the models compared. Even after Fine-Tuning, which helped many of the less performing models to ”catch-up” so to speak, the difference in performance is still very noticeable. Models such as the models of the NASNet family, the Inception family, and the more traditional

ResNet family models can generally be found towards the bottom in terms of training accuracy. This leads to the assumptions that technologies such as the residual connections, as they are present in ResNet, or the FtC and the Multi-Scale Processing, of Inception models tend to not work as well in the field of plant leaf disease classification. Even models such as VGG19 and VGG16, which are built solely without any such advanced features have managed to consistently outperform the above mentioned models. On the other hand, models such as EfficientNet models, DenseNet models, MobileNet models and ConvNeXt models are found near the top, showcasing that the methods that are employed in their architectures (e.g. advanced activation functions, dense connections, Depthwise Separable Convolution, larger kernels, etc.) seem to be more powerful and well suited in this field. These results are helpful and will therefore be considered when constructing the Federated Learning. As mentioned above, all these results are based on [125].

Table 6.4: This list displays the average performance metrics of each model averaged across all datasets and iterations. Again, all of the metrics are based on the test dataset, except for epochs which are taken during testing. FT stands for fine-tuned.

| **Model** | **Acc.** | **F1** | **Epochs** | **Acc. FT** | **F1 FT** | **Epochs FT** |
|---|---|---|---|---|---|---|
| EfficientNetV2B2 | 0.8699 | 0.8223 | 76.3889 | 0.8842 | 0.8420 | 26.2778 |
| MobileNetV3Large | 0.8698 | 0.8180 | 79.5000 | 0.8841 | 0.8363 | 35.4722 |
| EfficientNetV2S | 0.8680 | 0.8139 | 78.4722 | 0.8839 | 0.8395 | 25.0278 |
| EfficientNetV2B0 | 0.8670 | 0.8166 | 77.5000 | 0.8844 | 0.8442 | 28.2778 |
| EfficientNetV2B1 | 0.8655 | 0.8181 | 76.1389 | 0.8829 | 0.8420 | 28.5556 |
| EfficientNetV2B3 | 0.8648 | 0.8134 | 72.8056 | 0.8822 | 0.8357 | 30.3333 |
| ConvNeXtTiny | 0.8611 | 0.8167 | 108.7500 | 0.8862 | 0.8419 | 37.1944 |
| DenseNet201 | 0.8580 | 0.8108 | 64.2778 | 0.8775 | 0.8372 | 16.8333 |
| ConvNeXtSmall | 0.8568 | 0.8085 | 112.3889 | 0.8791 | 0.8312 | 28.0833 |
| DenseNet169 | 0.8534 | 0.8024 | 75.4722 | 0.8761 | 0.8342 | 20.2222 |
| DenseNet121 | 0.8530 | 0.8047 | 96.2778 | 0.8761 | 0.8339 | 21.2222 |
| MobileNetV3Small | 0.8491 | 0.7935 | 96.2778 | 0.8650 | 0.8131 | 35.3056 |
| EfficientNetV2M | 0.8424 | 0.7894 | 85.6111 | 0.8784 | 0.8352 | 20.0278 |
| ResNet101V2 | 0.8215 | 0.7623 | 45.6389 | 0.8536 | 0.8076 | 13.3056 |
| ResNet50V2 | 0.8188 | 0.7646 | 44.9722 | 0.8450 | 0.7986 | 16.3333 |
| VGG16 | 0.8133 | 0.7636 | 94.3611 | 0.8667 | 0.8193 | 24.3333 |
| ResNet152V2 | 0.8130 | 0.7496 | 46.5556 | 0.8511 | 0.8042 | 14.8611 |
| VGG19 | 0.8130 | 0.7590 | 84.7778 | 0.8628 | 0.8144 | 25.0556 |
| InceptionResNetV2 | 0.8050 | 0.7512 | 85.0278 | 0.8427 | 0.7979 | 11.8056 |
| Xception | 0.8011 | 0.7370 | 59.6111 | 0.8399 | 0.7902 | 17.3611 |
| NASNetMobile | 0.7891 | 0.7298 | 69.3611 | 0.8387 | 0.7877 | 25.7500 |
| NASNetLarge | 0.7871 | 0.7255 | 40.5556 | 0.8385 | 0.7879 | 14.1111 |
| InceptionV3 | 0.7862 | 0.7242 | 46.3056 | 0.8347 | 0.7820 | 15.6944 |

Table 6.5: The model average accuracy rankings represent the models average accuracy across all datasets and rank them accordingly, while the average model rank takes the model rank per dataset (based on accuracy per dataset) and averages that value.

| **Model Rank By Avg. Acc after FT** | | | **Model Rank By Avg. Rank after FT** | | |
|---|---|---|---|---|---|
| **Rank** | **Model** | **Average Acc.** | **Rank** | **Model** | **Average Rank** |
| 1 | ConvNeXtTiny | 0.8862 | 1 | EfficientNetV2B2 | 6.4444 |
| 2 | EfficientNetV2B0 | 0.8844 | 2 | EfficientNetV2B0 | 6.5556 |
| 3 | EfficientNetV2B2 | 0.8842 | 3 | EfficientNetV2S | 6.7222 |
| 4 | MobileNetV3Large | 0.8841 | 4 | DenseNet201 | 6.9444 |
| 5 | EfficientNetV2S | 0.8839 | 5 | EfficientNetV2B1 | 7.0556 |
| 6 | EfficientNetV2B1 | 0.8829 | 6 | ConvNeXtTiny | 7.1667 |
| 7 | EfficientNetV2B3 | 0.8822 | 7 | EfficientNetV2M | 7.3333 |
| 8 | ConvNeXtSmall | 0.8791 | 8 | MobileNetV3Large | 7.8333 |
| 9 | EfficientNetV2M | 0.8784 | 9 | EfficientNetV2B3 | 8.0556 |
| 10 | DenseNet201 | 0.8775 | 10 | ConvNeXtSmall | 8.3889 |
| 11 | DenseNet169 | 0.8761 | 11 | DenseNet121 | 8.5000 |
| 12 | DenseNet121 | 0.8761 | 12 | DenseNet169 | 9.5556 |
| 13 | VGG16 | 0.8667 | 13 | VGG19 | 11.3333 |
| 14 | MobileNetV3Small | 0.8650 | 14 | VGG16 | 13.3889 |
| 15 | VGG19 | 0.8628 | 15 | MobileNetV3Small | 14.0000 |
| 16 | ResNet101V2 | 0.8536 | 16 | ResNet152V2 | 15.8333 |
| 17 | ResNet152V2 | 0.8511 | 17 | ResNet101V2 | 16.1111 |
| 18 | ResNet50V2 | 0.8450 | 18 | InceptionResNetV2 | 17.6667 |
| 19 | InceptionResNetV2 | 0.8427 | 19 | ResNet50V2 | 18.5000 |
| 20 | Xception | 0.8399 | 20 | NASNetLarge | 18.8889 |
| 21 | NASNetMobile | 0.8387 | 21 | NASNetMobile | 19.4444 |
| 22 | NASNetLarge | 0.8385 | 22 | Xception | 20.1111 |
| 23 | InceptionV3 | 0.8347 | 23 | InceptionV3 | 20.1667 |

### 6.2.1 ViT comparison

Additionally, the CNN models were also compared to ImageNet pre-trained ViT (b16) [131, 132], to verify that CNN are the correct choice and indeed better performing than the ViT. The CNN were not trained again, so their results remain identical to above, but the ViT was trained under the same conditions and its results are compared with the CNN model's results. ViT results are kept separate and are not included in any average scores beyond this subsection.

When looking at the scores that ViT achieved in comparison to the CNN models, then we can see that it did achieve competitive results, but was constantly beat out by CNN models, such as EfficientNet models, ConvNeXt models, and DenseNet models. This showcases that overall, CNN models do seem to still generate better results in the field. One could still consider comparing ViT to other models for their use cases, since the results are still competitive and superior to models such as NASNet, ResNet, and Inception based models in this benchmark. However, based on these results, and based on the fact that transformers are computationally heavier and bigger than CNN and that they tend to require more data [131], here we will keep using CNN models. They performed better and with them being more lightweight and data efficient, they suit the field of plant leaf disease detection, with its limitations, better.

Table 6.6: Model ranking comparisons (same methods as in Table 6.5), including the comparison with ViT.

| **Model Rank By Avg. Acc after FT** | | | **Model Rank By Avg. Rank after FT** | | |
|---|---|---|---|---|---|
| **Model** | **Avg. FT Test Acc** | **Rank** | **Model** | **Average Rank** | **Rank** |
| ConvNeXtTiny | 88.62% | 1 | EfficientNetV2B2 | 6.78 | 1 |
| EfficientNetV2B0 | 88.44% | 2 | EfficientNetV2B0 | 6.89 | 2 |
| EfficientNetV2B2 | 88.42% | 3 | EfficientNetV2S | 7.11 | 3 |
| MobileNetV3Large | 88.41% | 4 | DenseNet201 | 7.33 | 4 |
| EfficientNetV2S | 88.39% | 5 | EfficientNetV2B1 | 7.33 | 5 |
| EfficientNetV2B1 | 88.29% | 6 | ConvNeXtTiny | 7.44 | 6 |
| EfficientNetV2B3 | 88.22% | 7 | EfficientNetV2M | 7.72 | 7 |
| ViT | 88.14% | 8 | MobileNetV3Large | 8.33 | 8 |
| ConvNeXtSmall | 87.91% | 9 | EfficientNetV2B3 | 8.44 | 9 |
| EfficientNetV2M | 87.84% | 10 | ConvNeXtSmall | 8.78 | 10 |
| DenseNet201 | 87.75% | 11 | DenseNet121 | 8.89 | 11 |
| DenseNet169 | 87.61% | 12 | DenseNet169 | 10.11 | 12 |
| DenseNet121 | 87.61% | 13 | ViT | 10.56 | 13 |
| VGG16 | 86.67% | 14 | VGG19 | 11.89 | 14 |
| MobileNetV3Small | 86.50% | 15 | VGG16 | 14.06 | 15 |
| VGG19 | 86.28% | 16 | MobileNetV3Small | 14.78 | 16 |
| ResNet101V2 | 85.36% | 17 | ResNet152V2 | 16.67 | 17 |
| ResNet152V2 | 85.11% | 18 | ResNet101V2 | 16.89 | 18 |
| ResNet50V2 | 84.50% | 19 | InceptionResNetV2 | 18.61 | 19 |
| InceptionResNetV2 | 84.27% | 20 | ResNet50V2 | 19.22 | 20 |
| Xception | 83.99% | 21 | NASNetLarge | 19.78 | 21 |
| NASNetMobile | 83.87% | 22 | NASNetMobile | 20.33 | 22 |
| NASNetLarge | 83.85% | 23 | Xception | 21.00 | 23 |
| InceptionV3 | 83.47% | 24 | InceptionV3 | 21.06 | 24 |

# 7. Data Augmentation Assessment

Datasets in the field are still considered to be sparse [14, 11, 126] and image collection is tedious and laborious and requires many prerequisites such as access to a large field of crops, equipment, hardware, expert staff, collection staff, labeling, and large amounts of time. This is not always possible due to different factors. However, if not enough data is available, training sufficient models can be difficult if not even impossible. One way to enlarge datasets artificially is augmentation. Data augmentation is the process of altering the images in a way where they are different enough from the original so that they pose an additional challenge and learning point to the model, while not loosing the important and classifying features. Like this, without the hassle of collecting new plant leaf data, one can extend the dataset and improve model performance. To ensure that this is true in this case as well, as set of experiments were carried out to validate these claims. The experiments, along with the results and analysis are based on the findings found in [133].

## 7.1 Augmentation Techniques

Augmentation comes in different types in which it can be applied. Here noise based augmentation, transformation based augmentation, and color based augmentation techniques will be applied, used, and compared to find the best possible augmentation methods for dataset enlargement and model training in the field of plant leaf disease classification. Color, noise and transformation based methods were used due to the fact that they are non invasive to the features of the images while also being fast, easily reproducible and adjustable.

Table 7.1: List of methods and values for color augmentation.

| Method | Value |
|---|---|
| Brightness | -0.75, 0.75 |
| Hue Shift | -20, 20 |
| Contrast | -0.25, 0.25 |
| Channel Shift | 75 |

Table 7.2: The noise augmentation method with the values used during augmentation.

| Method | Value |
|---|---|
| Gaussian Noise | mean: 0, stddev: 1 |

### 7.1.1 Color

Color based augmentation methods change the images by changing the colors of the image. These augmentations can simulate the way different camera equipment might differ in how it captures images or how the same plant might look different in different conditions (e.g. weather). Common and effective ways of making these changes are achieved by adjusting the brightness, contrast or by shifting the image's hue or channels. These methods all, if done within certain thresholds, do not disturb the image enough to a point where features would get lost, but change the image enough to where models would have to learn more robust features. The values with which we tested the effects of color augmentation on plant leaf disease classification are listed in Table 7.1 and the effects on the base image can be seen in Figure 7.1.

### 7.1.2 Noise

Noise augmentation adds noise to the image to add imperfections to the images. These imperfections simulate different imperfections that could appear during image capturing (e.g. camera, light conditions, compression, etc.). The added noise adds complexity to the image in a way that is not helpful to the classification task, and as such the model needs to learn how to form feature maps that are robust to such noisy imperfections. The values used can be seen Table 7.2 and the effect are shown in Figure 7.2.

### 7.1.3 Transformation

Transformation based augmentation methods change the image by transforming it. In this study rotation and flipping (mirroring) were applied to the image. This simulates the different ways of taking images could affect how they are represented in the dataset. Simply taking the image from another angle or another direction can easily happen during image capturing and through the augmentation methods listed below these differences can be simulated and applied to an already captured dataset. This will expose the model to more different data and can lead to better performance in more different scenarios. The methods and values are listed in Table 7.3 and the effects can be seen in Figure 7.3.

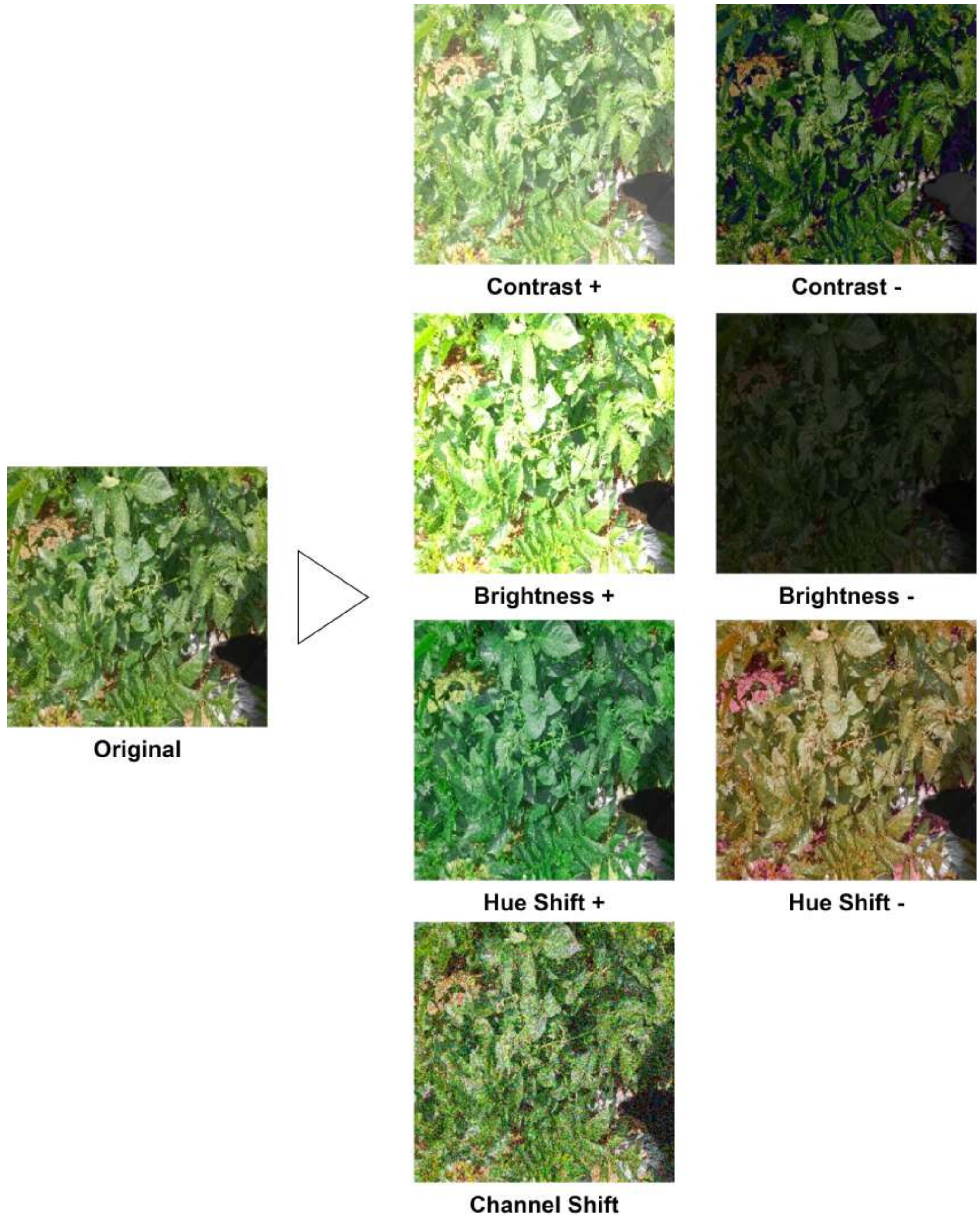

Figure 7.1: The different color based data augmentation techniques used in this paper visualized. Images are taken from the novelPotato [65] dataset.

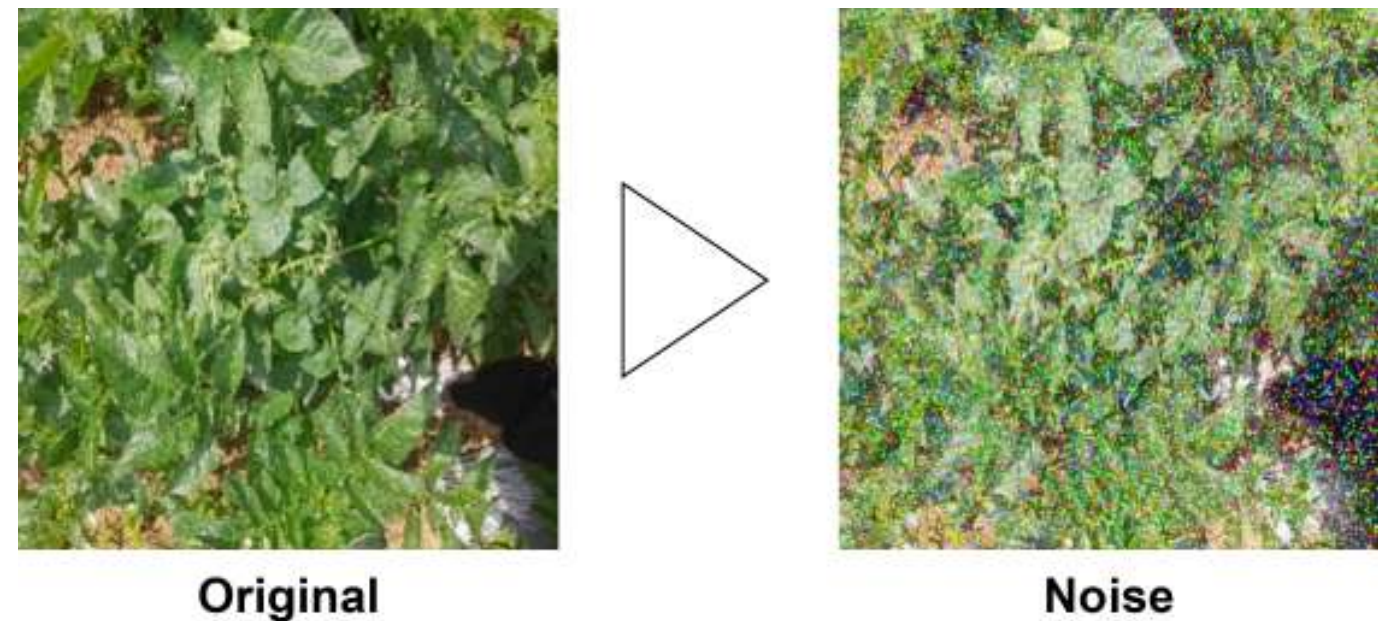

Figure 7.2: The noise based data augmentation technique used in this paper visualized. Images are taken from the novelPotato [65] dataset.

Table 7.3: Transformation augmentation methods.

| Method | Value |
|---|---|
| Rotation | 90 deg, 180 deg, 270 deg |
| Flip | horizontal, vertical |

## 7.2 Augmentation Methodology

To obtain representative results without having to run an equally, if not more, intensive, complex, and time consuming set of experiments, to obtain the results for augmentations only a subset of datasets and models has been selected. To keep other variable at a constant, 2 datasets have been chosen (novelPotato and pld, see Table 7.4), both of which are of potato leaves, but one is taken in lab with the other one having been taken in field. This allows an analysis of the effect on both different types of data. The models chosen here represent a wide variety of the families of models present in the benchmark, to obtain results for many different models, from which generalization will be possible. Each dataset will once be augmented with only color techniques, once with only noise techniques, once with only transformation techniques, and once with all three techniques combined (see Table 7.6 for dataset sizes post augmentation). Then each model will be trained on each of the 5 variations (no augmentation, color, noise,

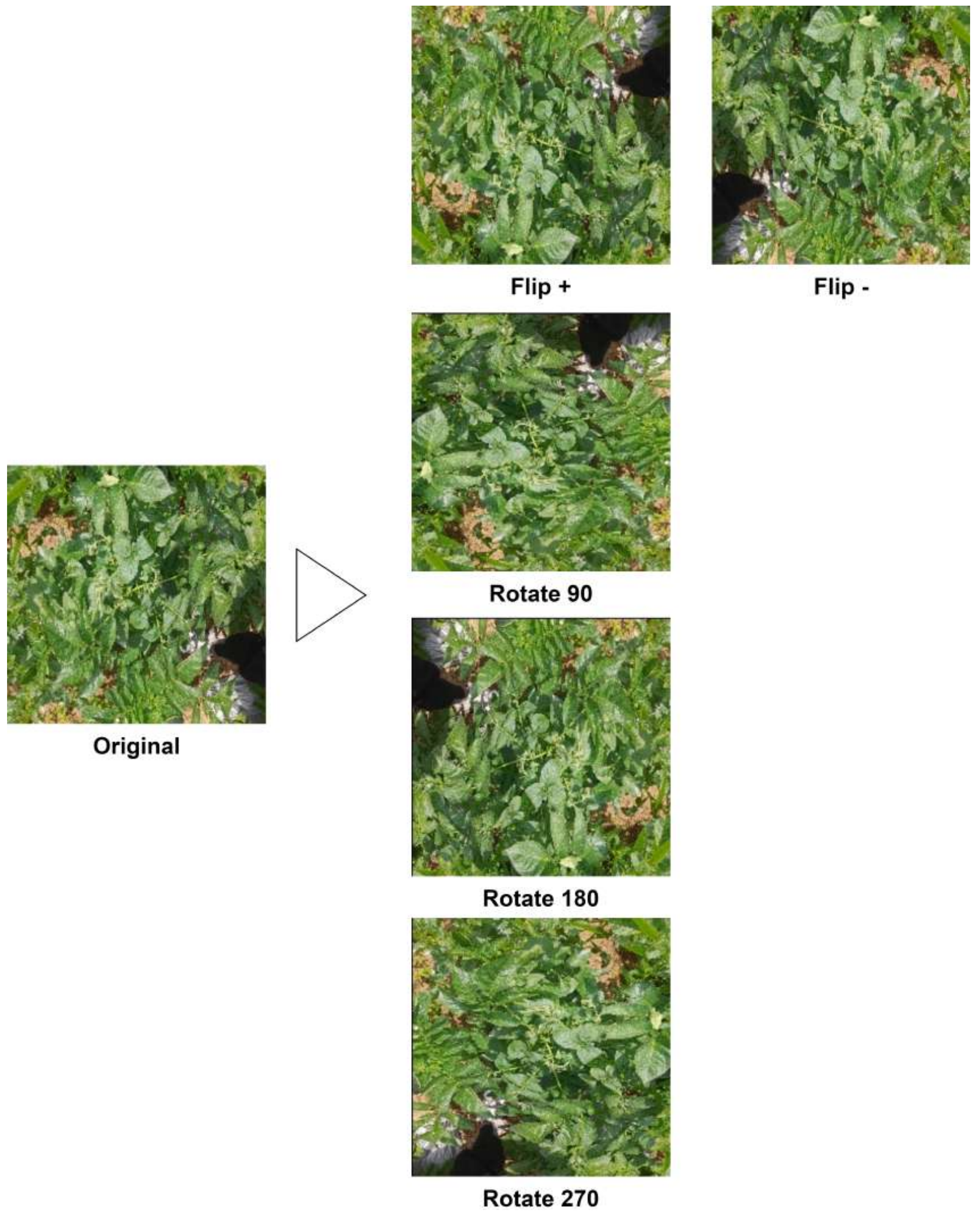

Figure 7.3: The different transformation based data augmentation techniques used in this paper visualized. Images are taken from the novelPotato [65] dataset.

Table 7.4: List of all the datasets considered in the augmentation experiments runs.

| Ref. | Dataset | Type | Plant | Num. Class | Total Img. |
|---|---|---|---|---|---|
| [65] | novelPotato | Hybrid | Potato | 7 | 3,076 |
| [73] | pld | Lab | Potato | 3 | 4,062 |

Table 7.5: Models used for the augmentation experiments.

| Ref. | Model | Total Params | Train. Params | Total Mem. | Train. Mem. |
|---|---|---|---|---|---|
| [129] | MobileNetV3Large | 2.99M | 2.97M | 11.43 | 11.34 |
| [124] | DenseNet121 | 7.04M | 6.95M | 26.85 | 26.53 |
| [124] | DenseNet201 | 18.32M | 18.09M | 69.89 | 69.02 |
| [121] | EfficientNetV2B0 | 5.92M | 5.86M | 22.58 | 22.35 |
| [121] | EfficientNetV2B2 | 8.77M | 8.69M | 33.45 | 33.14 |
| [118] | ResNet50V2 | 23.56M | 23.52M | 89.89 | 89.72 |
| [118] | ResNet152V2 | 58.33M | 58.19M | 222.52 | 221.97 |
| [123] | ConvNeXtSmall | 49.45M | 49.45M | 188.65 | 188.65 |
| [119] | VGG19 | 20.02M | 20.02M | 76.39 | 76.39 |

transformation, all augmentations) for both datasets.

## 7.3 Augmentation Results

The results, as can be seen in Tables 7.7 and 7.8, showcase that employing augmentation methods to extend datasets in the field of plant leaf disease classification, lead to better performing and more robust models. While no actual new data (as in new pictures taken in field) has been added to the datasets, the challenges provided by these augmented images, through image altering methods such as flipping, mirroring, contrast and brightness adjustments and noise, do lead to better results when compared to the results obtained by models without augmentation, and that is true for the lab and the field dataset. Combining all methods into one big augmentation pool and applying them to all images managed to yield the best results when combining the results for both datasets and it also did so for the novelPotato dataset taken in field. The pld lab dataset achieved slightly better results only using transformation, suggesting that transformation is the most powerful of the tested methods. However, since combining all 3 did perform best in the field+lab test and in the field only tests, moving forward, all methods will be combined to create a new dataset. Using transformation only would be slightly faster, but since that is not an issue, and since ”augmentation combined” was best on hybrid and field only data, which the new dataset will be, these results carry the most weight in this case.

Table 7.6: Datasets sizes for the different augmented datasets.

| Dataset | Healthy | E. Blight | L. Blight |
|---|---|---|---|
| pld | 918 | 1,466 | 1,273 |
| pld + Color | 7,344 | 11,728 | 10,184 |
| pld + Transform | 5,508 | 8,796 | 7,638 |
| pld + Noise | 1,836 | 2,932 | 2,546 |
| pld + Combined | 13,770 | 21,990 | 19,095 |
| **Dataset** | **Healthy** | **Fungal** | **Bacterial** |
| novelPotato | 160 | 598 | 455 |
| novelPotato + Color | 1,280 | 4,784 | 3,640 |
| novelPotato + Transform | 960 | 3,588 | 2,730 |
| novelPotato + Noise | 320 | 1,196 | 910 |
| novelPotato + Combined | 2,400 | 8,970 | 6,825 |

Table 7.7: The average test scores of each augmentation techniques across all models and datasets.

| Augmentation | Accuracy | F1-Score | Recall | Precision |
|---|---|---|---|---|
| Augmentation Combined | 98.82% | 98.64% | 98.82% | 98.82% |
| Augmentation Transform | 98.60% | 98.36% | 98.60% | 98.62% |
| Augmentation Color | 97.89% | 97.67% | 97.89% | 97.90% |
| Augmentation None | 97.58% | 97.08% | 97.58% | 97.63% |
| Augmentation Noise | 97.22% | 96.79% | 97.20% | 97.22% |

Table 7.8: The test scores for each augmentation method across all models based on the dataset.

| Augmentation | Dataset | Accuracy | F1-Score | Recall | Precision |
|---|---|---|---|---|---|
| Augmentation Transform | pld | 99.86% | 99.86% | 99.86% | 99.86% |
| Augmentation Combined | pld | 99.78% | 99.78% | 99.78% | 99.78% |
| Augmentation Color | pld | 99.42% | 99.42% | 99.42% | 99.45% |
| Augmentation None | pld | 99.34% | 99.28% | 99.34% | 99.34% |
| Augmentation Noise | pld | 99.04% | 99.04% | 98.98% | 99.04% |
| Augmentation Combined | novelPotato | 97.85% | 97.50% | 97.85% | 97.85% |
| Augmentation Transform | novelPotato | 97.34% | 96.85% | 97.34% | 97.38% |
| Augmentation Color | novelPotato | 96.36% | 95.92% | 96.36% | 96.36% |
| Augmentation None | novelPotato | 95.81% | 94.87% | 95.81% | 95.92% |
| Augmentation Noise | novelPotato | 95.41% | 94.53% | 95.41% | 95.41% |

All this is possible without having to spend any additional time or money on obtaining new images in field, a task that is not feasible, or even impossible for many. These findings will be implemented into the dataset construction. Results are an extension of the work in [133].

# 8. Dataset Construction

In order to construct a dataset that is large enough and diverse enough, the methods identified in Chapters 6 and 7 will be applied. Since it is not possible, as part of this work, to collect new data, due to the unavailability of fields and crops, existing datasets will be combined, merged, adjusted, and augmented in order to achieve a dataset with the desired characteristics. The final dataset will be a dataset with hybrid image sources (field and lab) to expose the model to as many different image types, in terms of diseases, plants, and conditions, in order to train a model that is as robust and generalizable as possible [62, 63, 64]. All classes will be set up in a way in which they will be perfectly balanced to support actual learning and to avoid biases in the trained weights. The dataset introduced here was submitted for publication as part of the following paper [134]. As such, chapter is based on the findings presented in this paper.

## 8.1 Dataset Construction Methods

Selected based on the benchmark, as well as on the original composition of each dataset, a set of 9 datasets have been chosen to be combined into a single large scale benchmarking dataset for this work. The datasets of choice can be found in Table 8.1. Other datasets have not been selected based on their weak performance in the benchmark (plantDoc, taiwanTomato, novelPotato), due to their redundancy (fgvc7 has been updated to fgvc8, tomatoVillage is a tomato dataset in lab conditions which plantVillage already covers, and pld is a lab potato dataset that again plantVillage already includes), and their small size (rldd, tea, cucumber). Also, during construction, an eye was kept on the distribution of lab to field images, to make sure that a good balance is kept. As a result the resulting dataset, after simply combining all the directories would end up being 95 classes with around 125,000 images. However, this does not consider overlap in classes between different datasets, where both the plant and the disease

Table 8.1: List of all the datasets that are considered to be part of the new benchmarking dataset. Given are all key metrics and performance metrics from the benchmarking [125].

| Dataset | Type | Plant | Num. Class | Total Img. | Acc. | F1 | Acc. FT | F1 FT |
|---|---|---|---|---|---|---|---|---|
| cds | Field | Corn | 3 | 1,571 | 93.338% | 93.248% | 97.323% | 97.286% |
| fgvc8 | Field | Apple | 12 | 18,632 | 74.810% | 40.618% | 81.754% | 48.848% |
| plantVillage | Lab | Multi | 38 | 54,304 | 97.492% | 96.654% | 99.377% | 99.117% |
| paddy | Field | Rice | 10 | 10,407 | 87.844% | 86.671% | 92.389% | 91.461% |
| sugar | Hybrid | Sugarcane | 10 | 6,405 | 91.582% | 87.267% | 91.634% | 87.139% |
| cassava | Field | Cassava | 5 | 21,397 | 75.175% | 57.688% | 79.427% | 64.935% |
| pdd271 | Field | Multi | 10 | 7,555 | 97.495% | 97.410% | 99.058% | 99.052% |
| sms | Field | Strawberry | 3 | 1,583 | 91.144% | 91.090% | 93.278% | 93.276% |
| diaMOS | Field | Pear | 4 | 3,006 | 82.412% | 75.049% | 86.201% | 84.002% |

Table 8.2: Classes that were deleted from the dataset and the reason they were.

| Dataset | Class | Reason for exclusion |
|---|---|---|
| fgvc8 | Apple Powdery Mildew Complex | too small |
| fgvc8 | Apple Rust Complex | too small |
| fgvc8 | Apple Rust Frog Eye Leaf Spot | too small |
| fgvc8 | Apple Frog Eye Leaf Spot Complex | too small |
| fgvc8 | Apple Scab Frog Eye Leaf Spot Complex | too small |
| fgvc8 | Complex | complex |
| fgvc8 | Scab Frog Eye Leaf Spot | multi disease |
| PlantVillage | Potato Healthy | too small |
| diaMOS | Pear Healthy | too small |
| diaMOS | Pear Curl | too small |

are the same and it does also not consider some of the weaknesses observed in the benchmark.

### 8.1.1 Class Cleanup

To ensure that each class truly represents a plant-disease combination that is not present in any other class and to make sure that all classes follow the same principles while also assuring class balance, some classes need to be combined while others need to be deleted from the dataset. Classes chosen for deletion (see Table 8.2) were picked based on a few criteria. On one hand classes with 200 or less images were deleted to preserve a adequate number of unique images before augmentation. Additionally, two more fgvc8 classes were deleted as they were complex and multi-disease classes, that did not fit with the rest of the classes that were all single disease.

With the deletion of 10 classes, the dataset now has 85 classes, not all of which are unique though. To ensure uniqueness a total of 10 classes need to be combined, as there are classes that contain the same plant-disease combination. These 10 classes will be combined into 5 corresponding classes ensuring uniqueness for the resulting dataset. The classes to be combined

Table 8.3: Classes that were combined in order to create unique classes for the final dataset.

| Dataset 1 | Class 1 | Dataset 2 | Class 2 |
|---|---|---|---|
| cds | Gray Leaf Spot | PlantVillage | Cercospora Leaf Spot Gray Leaf Spot |
| cds | Northern Leaf Blight | fgvc8 | Northern Leaf Blight |
| fgvc8 | Healthy | PlantVillage | Apple Healthy |
| fgvc8 | Rust | PlantVillage | Cedar Apple Rust |
| fgvc8 | Scab | PlantVillage | Apple Apple Scab |

Table 8.4: Metrics of the new dataset.

| | |
|---|---|
| Number of Different Plants | 25 |
| Number of Classes | 80 |
| Number of Images Total | 304,507 |
| Train-Test Split (before val split) | 80-20 |
| Number of Images Train (before val split) | 280,000 |
| Number of Images Test | 24,507 |
| Train-Val-Test Split (after val split) | 64-16-20 |
| Number of Images Train (after val split) | 224,000 |
| Number of Images Validation (after val split) | 56,000 |
| Image Size | 224x224x3 |

can be seen in Table 8.3.

Post deletion and combination the dataset is 80 classes in size, but the classes are heavily imbalanced, with some classes also being undersized. To ensure perfect balancing and, as part of that, to lower the risk of biasing during training, all classes will be set to be of the same size. First, to guarantee that robustness can be observed during training, the dataset will be split 80-20 into a train set and a test set (the test set will not be augmented). Then, each train class will be expanded through the augmentation techniques introduced and discussed in Chapter 7, due to their proven positive impact on training. Next, each class will be cutdown to exactly 3,500 images per class, creating a perfectly balanced dataset. Out of the augmented set, each class will randomly sample 3,500 images for the final dataset, creating a final train dataset of 304,507 images across 80 classes. Before training, the train set will be further split 80-20 into actual train data and validation data, to keep track of unseen image performance during training, before the final model will be tested on the un-augmented test set at the end. The key metrics for the new dataset, which is called PLDC-80, are listed in Table 8.4.

### 8.1.2 Dataset Method Availability

Due to the fact that the dataset is a merger of preexisting datasets, all of which come with different kinds of licenses, uploading the final dataset is not possible, since some dataset's licenses prohibit redistribution. Instead, a detailed method for how one can reconstruct the dataset is available at [135].

# 9. Model Construction

A dedicated Base Model for the field of plant leaf disease classification would potentially offer advantages such as faster, more robust, and more stable training, even in scenarios where data is sparse (as is often the case in this field). Such a model needs to first be constructed (based on an existing and already powerful architecture) before being trained on the PLDC-80 dataset to construct powerful and domain relevant feature extractors that contain feature maps that are relevant to the field specifically. This includes a multi-step process of first identifying a baseline model to base the architecture upon, then, if possible, improving said baseline to help it achieve even better results. Then that model, with the pre-trained weights, needs to be compared to the baseline model trained on general data, e.g. ImageNet, to see if it does indeed improve training and final results. This step requires yet another new dataset, that is unknown to the model, meaning data that is not part of the PLDC-80 dataset, and ideally also contains data that does not correspond to plant-disease class combinations found in PLDC-80 (as much as possible). The model and findings presented in this chapter will be submitted for publication in [136]. This chapter is based on the referenced paper, which is written as part of the work towards this thesis.

To build a new improved model architecture one needs to identifying potential baseline models and technologies that could improve the model and then implement them to assess their performance. To compare the models in a constant test environment with no outside inconsistencies, that allow for fair and accurate comparisons of the models performance only (to make sure that the model is truly the variable that is being tested and compared), the following settings were used and they remain constant across all tests for all models. The values kept constant can be seen in Table 9.1.

Table 9.1: Constant setup during model training and evaluation across all architectures.

| Parameter/Setting | Value |
|---|---|
| Batch Size | 128 |
| Epochs | 150 |
| Optimizer | Adam |
| Callbacks | Checkpoint (Val. Acc.) |
| Train Dataset | 224,000 |
| Val. Dataset | 56,000 |
| Test Dataset | 24,507 |
| Number of Classes | 80 |
| Input Size | 224x224x3 |
| Loss | Categorical Crossentropy |
| Machine | CPU: Intel Xeon Gold 6330, GPU: NVIDIA A100 SXM4 40GB, RAM: 125GB |
| Evaluation | Based on the best performing epoch Checkpoint |

Table 9.2: List of the models that were compared in the benchmark.

| Ref. | Model | Total Params | Train. Params | Total Mem. | Train. Mem. |
|---|---|---|---|---|---|
| [124] | DenseNet201 | 18.32M | 18.09M | 69.89 | 69.02 |
| [121] | EfficientNetV2B2 | 8.77M | 8.69M | 33.45 | 33.14 |
| [121] | EfficientNetV2S | 20.33M | 20.18M | 77.56 | 76.97 |
| [123] | ConvNeXtSmall | 49.45M | 49.45M | 188.65 | 188.65 |
| [123] | ConvNeXtTiny | 27.82M | 27.82M | 106.13 | 106.13 |

## 9.1 Baseline Model search

Based on these settings, different baseline models were trained (based on the top performing benchmarking models) to find the best performing model to then use it as the baseline model to built upon and compare to. The models used can be seen in Table 9.2. Models were trained on the new PLDC-80 dataset and then compared on their performance on the test set (which is both unseen to the trained model as well as un-augmented). After 150 epochs of training the model is reset to the best performing epoch (based on validation accuracy). This best performing model is then used for evaluation on all 3 sets (train, validation, test). As mentioned earlier, the models will be judged on their accuracy score on the test set, to evaluate which model is best performing. Results for these baseline model tests can be seen in Table 9.3.

Table 9.3: Baseline model performance comparison.

| Model | Train Accuracy | Validation Accuray | Test Accuracy |
|---|---|---|---|
| ConvNeXtTiny | 99.38% | 90.71% | 83.86% |
| ConvNeXtSmall | 99.44% | 91.43% | 84.07% |
| EfficientNetV2B2 | 99.47% | 94.62% | 89.35% |
| EfficientNetV2S | 99.45% | 94.64% | 89.57% |
| DenseNet201 | 99.38% | 95.07% | 91.10% |

Based on the results observed in Table 9.3, one can conclude that DenseNet201 is ideal to be used as the basis of the new model, as it is both top performing and also modular and block based. It managed to outperform even the best of the other models by over 1%, which is quite significant in a scenario where the dataset is this large and where all models already managed to achieve good results. As such, as mentioned earlier, the DenseNet201 model architecture has been chosen to be used as the baseline model to be built upon.

## 9.2 Model Improvement Methodology

With the new PLDC-80 dataset in place and through the findings of the benchmarks as well as the test, constructing the new model based on the DenseNet201 baseline is the next step. The improved model should be able to outperform other CNN baselines on the new PLDC-80 dataset and then, after training, manage to outperform the baseline in a Transfer-Learning task, where the trained model with the weights obtained during training will be compared with the baseline model trained on ImageNet when exposed to another plant leaf disease classification dataset. Ideally the new model can train TL model faster, more stable, and at the end of training perform better.

### 9.2.1 Model Architecture

Based on the findings of the benchmark, as well as after additional testing on the new dataset, the baseline model that this model will be based on will be the DenseNet201 architecture, which has proven to be very capable of training strong plant leaf disease classifying models during the benchmark and also during the experiments here. DenseNet201, as can be seen in Chapter 5 and Appendix B, is a model that features dense residual connection, which allow the model to carry attention maps into deeper layers, which improve gradient flow, lessen the vanishing gradient problem, and allow feature reuse. DenseNet201's architecture can be seen in Figure 9.1.

It feeds input images into the pre-amble which houses a padded convolutional layer with kernel size 7 and stride 2 to get a large kernel look at the image before downsizing the input for the actual dense blocks. Then a batch normalization layer is applied to the output of the convolutional layer and following that is the relu activation. Next is a padding layer before max pooling with stride 2. Then come the dense blocks, which are stacks of the dense layer blocks as seen in Figure B.4 (e.g. the first dense block as seen in Figure 9.1 would have 6 Figure B.4 layer blocks stacked on top of each other). Each layer block receives the concatenated input of all previous layer's outputs in said block. These dense blocks are divided by transition blocks. These transition blocks break up the dense connections, meaning they only exists within any given block but not across blocks, as they need to be of the same dimensions (width and height), which is not given after transition. The transition blocks first apply batch normalization to the input, then a relu activation before applying a Pointwise Convolution (a 1x1 convolution) that reduces large depth dimension of the input to a given new channel size. This is then followed by a average pooling layer with stride 2, reducing the width and height dimensions. After several stacks of these dense and transition blocks, follows the post-amble. The post-amble simply includes a batch normalization layer after which comes a relu activation, which completes the classifier. With these methods an un-edited model can already achieve high level results, as it is already fine-tuned to work in large scale, multi-class image classification tasks.

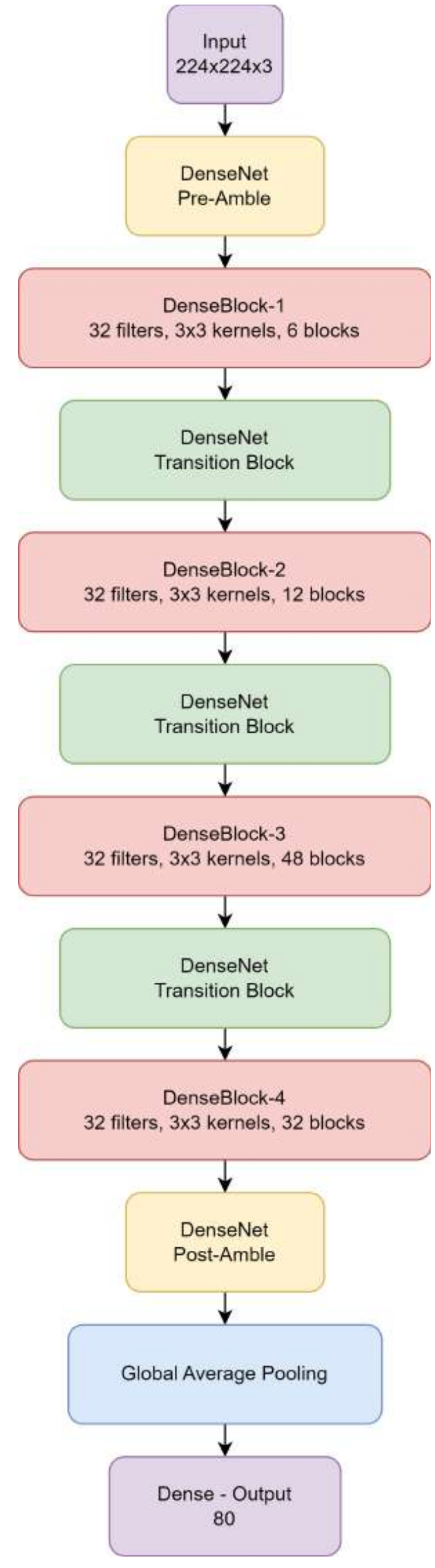

Figure 9.1: DenseNet201's architecture before modification

### 9.2.2 Possible Improvements

To improve models such as DenseNet201, a lot of different techniques can be implemented, be it from other Foundational Model that are based on different philosophies, or additional on techniques such as attention blocks, etc. However, since models like DenseNet are already finely tuned to work extremely well on large datasets for image classification, one needs to act thoughtfully and carefully when implementing such new concepts into the architecture, since small changes to the model are not guaranteed to improve the performance, but on top also run the risk of instead lowering the performance the model can achieve. As such, methods need to be well considered and tested, to make sure that they do indeed help the model achieve better results, as opposed to simply creating a different more complex model, that does however not manage to achieve any better results. Methods considered for improving the baseline model included, but were not limited to:

- GELU Activations
- Larger Kernels
- Layer Normalization

- ConvNeXt Blocks
- More filters
- More Dense Layers per block
- Dropout
- Progressive receptive field expansion
- Squeeze & Excitation (SE)
- Convolutional Block Attention Module (CBAM)
- Spatial Attention (SA)
- Large Kernel Attention
- Super Latent Connections
- Swish Activations
- Channel Attention (CA)

Many of these approaches were tested in different configurations. Large kernels were for example tested in 5x5 and 7x7, and also in progressive receptive field expansion setups, both per block and inside the blocks. Many of these methods were tested in such different setups, to find the ideal selection and combination of methods that improve the performance of Dense201 specifically for the field of plant leaf disease classification.

These methods and technologies now need to be applied to said DenseNet201 architecture and compared, in order to identify the ideal combination of model and additional methods to create the architecture for the Base Model. In testing the following results were observed which can be seen in table 9.4.

### 9.2.3 Successful Improvements

As seen in Table 9.4, the best performing model resulting from the numerous testing runs was DenseNet201 with Swish Activations in all blocks and a Channel Attention Block at the end of the CNN classifier. This model managed to outperform the baseline model on the test dataset with 91.46%, where the baseline model reached 91.10%. This might not seem like a large improvement at first, but with the dataset being this large, it does favor large scale models, meaning the baseline is already well suited as is (as is also proven by the benchmarks and the comparison on the dataset to other models). Also, with DenseNet201 already performing as well as it did with over 91% the room for further improvements is small and limited. As such, 0.36% does actually show fairly decent improvements and it is a 4.04% decrease of miss classifications over the whole test dataset. It does also equal around 88 more images that were classified correctly in the test set, which could make a difference in on field applications.

Table 9.4: Model modification comparison results. Here DN201 stands for DenseNet201. The Train, Val and Test columns list the accuracy for each set.

| Baseline | Added Methods | Train | Val | Test |
|---|---|---|---|---|
| DN201 | Swish + Channel Attention Block | 99.41% | 94.80% | 91.46% |
| DN201 | Channel Attention Block | 99.44% | 95.08% | 91.40% |
| DN201 | Swish | 99.41% | 94.72% | 91.40% |
| DN201 | Spatial Attention Block | 99.42% | 94.91% | 91.21% |
| DN201 | Channel Attention + Bigger Kernel per Block (1st Block 3x3, 2nd Block 5x5, 3rd Block 7x7) | 99.37% | 94.77% | 91.20% |
| DN201 | Kernel Size 5x5 all Blocks | 99.44% | 95.22% | 91.11% |
| DN201 | None | 99.38% | 95.07% | 91.10% |
| DN201 | Large Kernel Attention Block | 99.31% | 94.71% | 90.70% |
| DN201 | Swish + CBAM Attention Block + Bigger Kernel per Block (1st Block 3x3, 2nd Block 5x5, 3rd Block 7x7) | 99.41% | 95.02% | 90.66% |
| DN201 | Swish + Channel Attention Block (Swish also in Channel Attention Block) | 99.41% | 94.94% | 90.65% |
| DN201 | CBAM Block | 99.38% | 94.98% | 90.64% |
| DN201 | Swish + Channel Attention (After every Dense Block) | 99.32% | 94.50% | 90.51% |
| DN201 | Squeeze & Excitation Attention Block | 99.46% | 94.99% | 90.48% |
| DN201 | Channel Attention + Super Latent Connections | 99.49% | 94.68% | 90.46% |
| DN201 | Deeper Last Block (48 Dense Layers in Last Block) | 99.43% | 94.89% | 90.44% |
| DN201 | Swish + Channel Attention Block + 5x5 kernel in all Blocks | 99.37% | 94.86% | 90.34% |
| DN201 | Bigger Kernel per Block (1st Block 3x3, 2nd Block 5x5, 3rd Block 7x7) | 99.35% | 94.94% | 90.21% |
| DN201 | Kernel Size 7x7 all Blocks | 99.39% | 94.76% | 90.11% |
| DN201 | Last Dense Block with growth rate 48, 48 layers and 7x7 kernel | 99.38% | 94.69% | 90.09% |
| DN201 | Growth rate 48 in 3rd Dense Block and 64 in last Dense block | 99.45% | 94.72% | 90.03% |
| DN201 | Replacing Dense Convolution Layers with ConvNeXt like blocks (and ConvNeXt depth and width) while maintaining Dense Connections | 99.34% | 94.12% | 89.97% |
| DN201 | Channel Attention (After every Dense Block) | 99.31% | 94.63% | 89.67% |
| DN201 | Light Super Latent Connection | 99.44% | 94.70% | 89.57% |
| DN201 | Light Plus Super Latent Connection | 99.46% | 94.27% | 89.54% |
| DN201 | Super Latent Connection | 99.50% | 94.60% | 89.54% |
| DN201 | Channel Attention Block + 7x7 kernel size in last Dense Block | 99.40% | 94.80% | 89.51% |
| DN201 | Bigger Kernel inside each Block (first 1/3rd of layers in Block 3x3, 2nd 1/3 5x5, 3rd 1/3 7x7) | 99.37% | 94.52% | 89.22% |
| DN201 | Replacing Dense Convolution Layers with ConvNeXt like blocks while maintaining Dense Connections | 99.15% | 92.70% | 88.03% |
| DN201 | Adding ConvNeXt block after last Dense Block | 99.09% | 93.29% | 87.89% |

The two techniques that managed to elevate the performance of the baseline model were SiLU/Swish-1 activations and Channel Attention. Both methods as well as the resulting model architecture look and work as follows:

#### 9.2.3.1 Swish Activation (SiLU)

Swish Activations, also known as Swish-1 or SiLU (Sigmoid Linear Unit) (Equations (9.1) and (9.2)) [137, 138, 139], as they are used in EfficientNet architectures (which have also performed well in the benchmarks) are an activation function that, unlike the ReLU used in DenseNet, does not set all negative inputs to 0, which helps it overcome the dying ReLU problem of vanishing gradients. SiLU is self-gated, meaning that it multiplies the result of the sigmoid function of the input by the input itself, scaling the output by the activation level. As can be seen in Figure 9.2, SiLU is a smooth activation function, especially when compared with the very sharp ReLU activation. This smoothness allows better gradient flow through the network which can improve learning. SiLU is non-monotonic, not increasing linear after 0, this allows it to achieve better generalization and can help learn more complex patterns.

$$f(x) = x * sigmoid(x) \tag{9.1}$$

$$sigmoid(x) = \frac{1}{1+e^{-x}} \tag{9.2}$$

Implementing SiLU into the DenseBlocks improved the performance of the DenseNet201 architecture when looking at the test dataset performance, which is best suited to showcase the robustness and generalization ability of the final model.

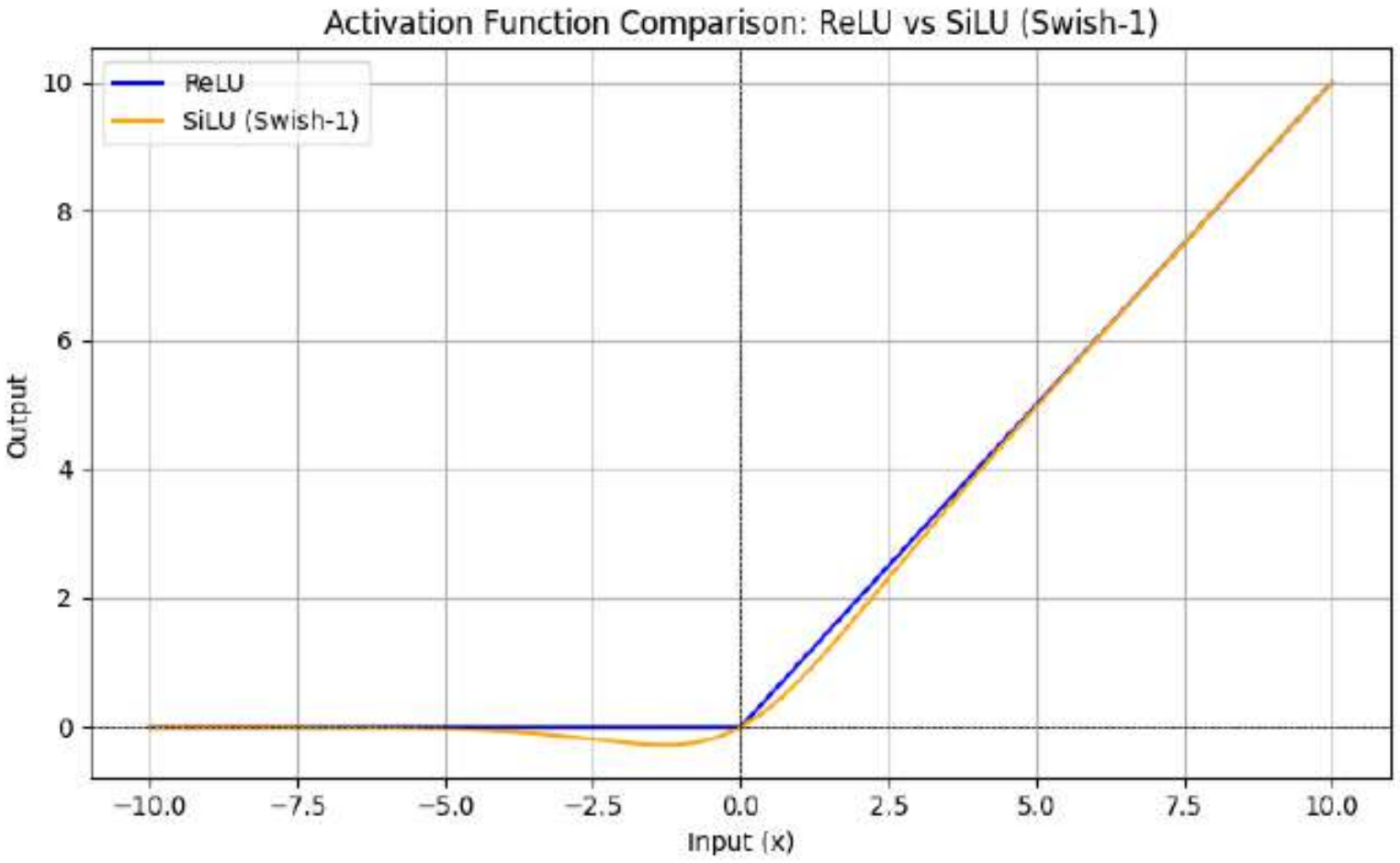

Figure 9.2: A visual comparison of the ReLU and the SiLU/Swish activation functions.

#### 9.2.3.2 Channel Attention

The second mechanism that managed to improve the performance of DenseNet201 was adding a Channel Attention block [140, 141] to the end of the DenseNet201 model. Channel Attention is a mechanism that ranks input channels based on their importance to the classification task and weights them accordingly. Channel Attention does that by first reducing each channel to a single value, once through global average pooling and global max pooling. Both of these feature vectors of size 1 x 1 x channel size are then passed into an encoder-decoder fully connected block separately, before being added. Then, after sigmoid activation, each channel is multiplied by said sigmoid value to scale them based on their importance to the task. This allows the model to focus on important feature maps that actually help the model classify images into the correct class and focus less on feature maps that are not as useful to the task at hand. See Figure 9.3 for a visual representation of that method.

#### 9.2.3.3 Combined Model

As a result of both of the above mentioned techniques managing to improve the performance of the baseline DenseNet201 architecture for the field of plant leaf disease classification, the next step is to combine them into one model to benefit from the advantages they both present. The resulting PLDC-Net architecture replaces all the ReLU connections in the Dense Preamble, the Dense Blocks and also in the Dense Postamble. The CA Block still uses ReLU, since experiments of also using Swish in the attention block yielded worse results. The CA block is positioned right after the dense post-amble block and before the global average pooling layer, where it produced the best results, when compared to other positioning and also when compared to using multiple CA blocks across the network. This resulting model will be called PLDC-Net, more details can be found in [136].

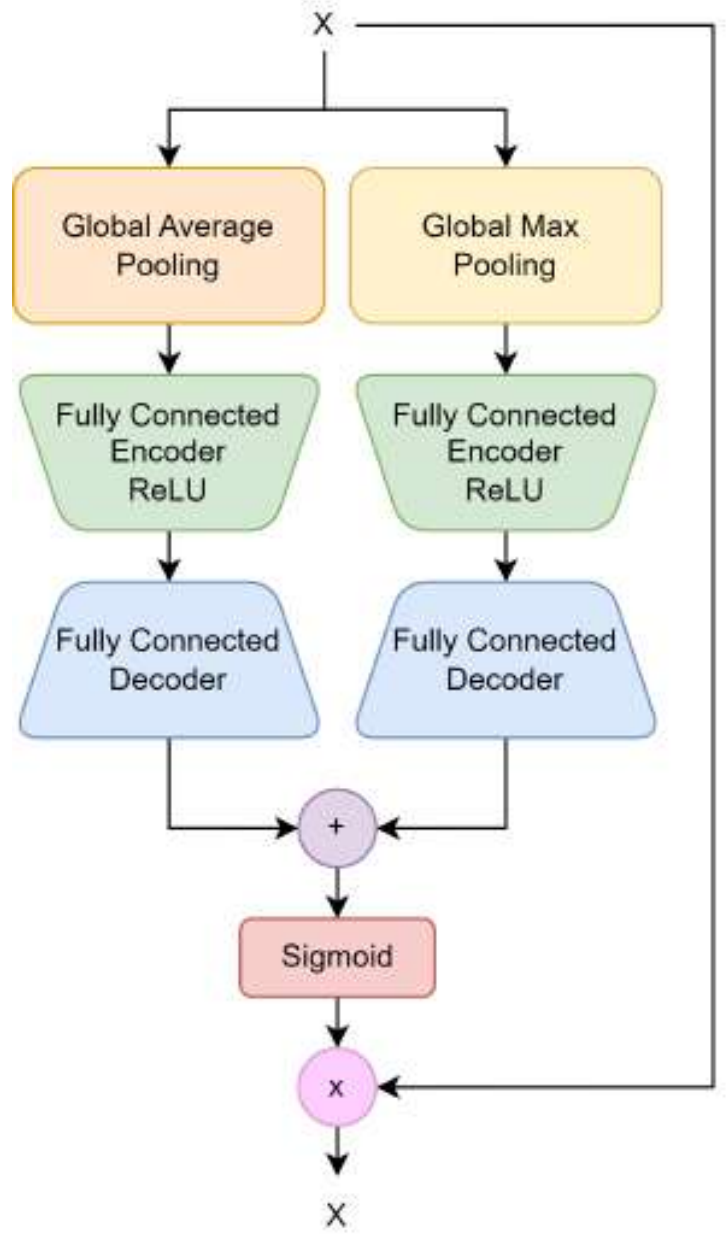

Figure 9.3: Flowchart of the Channel Attention Block.

## 9.3 Transfer-Learning

To validate that the resulting pre-trained model did not only achieve better generalization results on the PLDC-80 test dataset, one now needs to create yet another dataset that the pre-trained model can now be transfer-learned on. Ideally, this dataset does not have much overlap with the PLDC-80 dataset in terms of classes (plant-disease overlap) and it should not contain any images that were in the PLDC-80 dataset during training.

### 9.3.1 PLDC-6

To facilitate these further TL experiments, a new dataset needs to be constructed that is, as mentioned earlier, relevant to the field of plant leaf disease classification, representative of a dataset one might collect in future research, and different enough from the original PLDC-80, to actually test TL capabilities. To achieve these design goals, another 2 datasets have been obtained and combined to create the new dataset. In this case no augmentation and balancing was carried out, to more resemble a dataset that would be collected by someone that would go out into the field to simply collect and label data. The datasets chosen for this 6 class version PLDC transfer dataset for TL purposes is made up of the iBean dataset [142] and the Soybean images dataset [143, 144] (see Table 9.5).

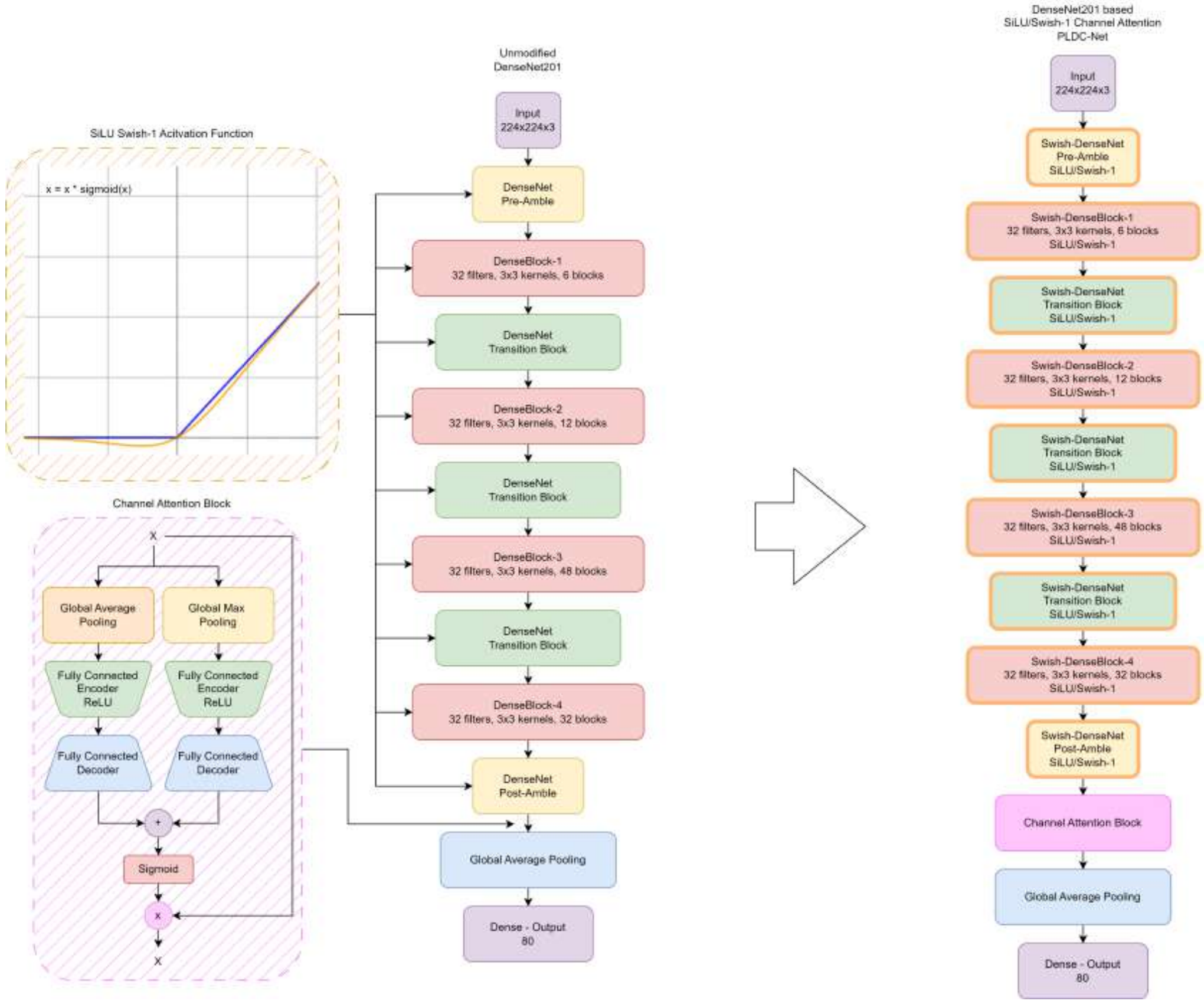

Figure 9.4: The proposed PLDC-Net architecture based on the DenseNet201 baseline model with a Channel Attention Block and Swish-1 SiLU activations.

Table 9.5: The 2 datasets used to combine into PLDC-6 for TL testing.

| Ref. | Dataset | Type | Plant | Num. Class | Total Img. |
|---|---|---|---|---|---|
| [143, 144] | mignoniSoy | Field | Soybean | 3 | 6,410 |
| [142] | iBean | Field | Beans | 3 | 1,296 |

Table 9.6: Metrics of the new PLDC-6 dataset.

| | |
|---|---|
| Number of Different Plants | 2 |
| Number of Classes | 6 |
| Number of Images Total | 7,706 |
| Image Size | 224x224x3 |
| Images Per Class | • Soybean Healthy - 896<br>• Soybean Caterpillar - 3309<br>• Soybean Diabrotica speciosa - 2205<br>• Bean Healthy - 428<br>• Bean Angular Leaf Spot - 432<br>• Bean Bean Rust - 436 |

To combine the datasets, they were simply combined to create a 6 class, 2 plant dataset with 7,706 images in total. As earlier mentioned, this data will not be augmented or adjusted for good balance. Also iBean data has been un-split (it originally comes in a pre train-validation-test split) (see Appendix F for more information about the original datasets). The resulting combined PLDC-6 dataset's information can be seen in Table 9.6.

As one can see in Table 9.6, the dataset is largely imbalanced. This can happen with data taken in field, an does as such not only pose an extra challenge, but also represents what real conditions might produce. Due to this imbalance, in the TL experiments the focus will be put on the F1-Score rather than the accuracy. Another possible limitation that real field datasets might suffer from is small image counts (a small dataset). To simulate this limitation, the dataset will be train-test split in different rations (see Table 9.7). these ratios are:

- train 10% - test 90%
- train 30% - test 70%
- train 50% - test 50%
- train 70% - test 30%
- train 90% - test 10%

These different ratios will give an insight into the effects the pre-trained PLDC-Net Base Model has on small datasets, at different levels, ending with fairly large datasets in the end.

### 9.3.2 Transfer-Learning Methodology

With the dataset prepared, the pre-trained models can now be trained on it to compare how well they adjust to the new data. The two models that will be compared in these experiments are the PLDC-80 pre-trained PLDC-Net and the baseline DenseNet201 on the ImageNet Dataset. Each model will be trained on each of the different train-test splits mentioned above, once with the classifier frozen and once with it un-frozen. Frozen training is beneficial in terms of computational costs, since only the classifier top needs to be retrained, and the huge CNN model with all its parameters remains unchanged (see Table 9.8). As a result, in total 20 models will be trained, 5 train-test splits, frozen (batch size 128) and un-frozen (batch size 32) and all that for ImageNet DenseNet201 and PLDC-80 pre-trained PLDC-Net. Each model is trained for 50 epoch. The best performing epoch is used for evaluation using the checkpoint callback.

### 9.3.3 Results

Each of the train-test split experiments, both for frozen as well as for un-frozen, will then compare the performance of the PLDC-80 pre-trained PLDC-Net and the ImageNet trained DenseNet201 baseline. Performance will be compared on speed, robustness, adaptability and final performance results. As mentioned earlier, the goal of developing a new Base Model for the domain of plant leaf disease classification is that the resulting model can train a domain-specific model faster and with less resources by utilizing the domain related knowledge obtained during the pre-training phase.

Table 9.7: Train-Test split results for TL experiments

| **Train-Test Split (0.1-0.9)** | | |
|---|---|---|
| | Train | Test |
| **Soybean Healthy** | 90 | 806 |
| **Soybean Caterpillar** | 331 | 2978 |
| **Soybean Diabrotica speciosa** | 220 | 1984 |
| **Bean Healthy** | 43 | 385 |
| **Bean Angular Leaf Spot** | 43 | 389 |
| **Bean Bean Rust** | 44 | 392 |
| **Total** | 771 | 6934 |
| **Train-Test Split (0.3-0.7)** | | |
| | Train | Test |
| **Soybean Healthy** | 269 | 627 |
| **Soybean Caterpillar** | 993 | 2316 |
| **Soybean Diabrotica speciosa** | 662 | 1544 |
| **Bean Healthy** | 128 | 300 |
| **Bean Angular Leaf Spot** | 130 | 302 |
| **Bean Bean Rust** | 131 | 305 |
| **Total** | 2313 | 5394 |
| **Train-Test Split (0.5-0.5)** | | |
| | Train | Test |
| **Soybean Healthy** | 448 | 448 |
| **Soybean Caterpillar** | 1654 | 1654 |
| **Soybean Diabrotica speciosa** | 1102 | 1102 |
| **Bean Healthy** | 214 | 214 |
| **Bean Angular Leaf Spot** | 216 | 216 |
| **Bean Bean Rust** | 218 | 218 |
| **Total** | 3852 | 3852 |
| **Train-Test Split (0.7-0.3)** | | |
| | Train | Test |
| **Soybean Healthy** | 627 | 269 |
| **Soybean Caterpillar** | 2316 | 993 |
| **Soybean Diabrotica speciosa** | 1544 | 662 |
| **Bean Healthy** | 300 | 128 |
| **Bean Angular Leaf Spot** | 302 | 130 |
| **Bean Bean Rust** | 305 | 131 |
| **Total** | 5394 | 2313 |
| **Train-Test Split (0.9-0.1)** | | |
| | Train | Test |
| **Soybean Healthy** | 806 | 90 |
| **Soybean Caterpillar** | 2978 | 331 |
| **Soybean Diabrotica speciosa** | 1984 | 220 |
| **Bean Healthy** | 385 | 43 |
| **Bean Angular Leaf Spot** | 389 | 43 |
| **Bean Bean Rust** | 392 | 44 |
| **Total** | 6934 | 771 |

Table 9.8: Model parameter comparison for frozen and un-frozen models.

| | | Model | |
|---|---|---|---|
| | | **Baseline** | **PLDC-Net** |
| **Frozen** | Total Params. | 18,333,510 | 19,255,110 |
| | Trainable Params. | 11,526 | 11,526 |
| | Non-Trainable Params. | 18,321,984 | 19,243,584 |
| **Unfrozen** | Total Params. | 18,333,510 | 19,255,110 |
| | Trainable Params. | 18,104,454 | 19,026,054 |
| | Non-Trainable Params. | 229,056 | 229,056 |

Table 9.9: The results of the frozen TL results on the frozen 10-90 split. The percentage values given are the test F-1 scores.

| Frozen 10-90 | |
|---|---|
| **Baseline** | 69.19% |
| **At Epoch** | 49 |
| **PLDC-Net** | 79.21% |
| **At Epoch** | 39 |
| **Overtake PLDC-Net** | 69.32% |
| **At Epoch** | 10 |

Less resources are used when the model is frozen and when only the classifier top is used, making this scenario the most desirable on. On top of that, as this field does lack data, good performance on small datasets is deemed the most important aspect.

#### 9.3.3.1 Frozen

This section will showcase the frozen model TL performance.

**10% Train Data - 90% Test Data** The results obtained when the PLDC-6 dataset is split into only 10% data used for training, simulating a very small and imbalanced dataset (see Table 9.7), showcase the following results that can be seen in Table 9.9 and Figure 9.5.

As can easily be seen in both the table and the figure, the results in the hardest train-test split in the frozen, and therefore more computationally efficient setting that relies and benefits more from pre-trained data, the PLDC-Net manages to outperform the baseline model by over 10% in terms of F1-score on the before un-seen test dataset. The PLDC-Net model also manages to outperform the baseline model after only 10 epochs, to reach a value higher than the baseline's max score for which it needed 49 epochs. This showcases the validity of such a domain relevant and domain related pre-trained Base Model for the field of plant leaf disease classification. The proposed PLDC-Net does not only heavily outperform the baseline on this small and imbalanced dataset, but also does so faster and more robust.

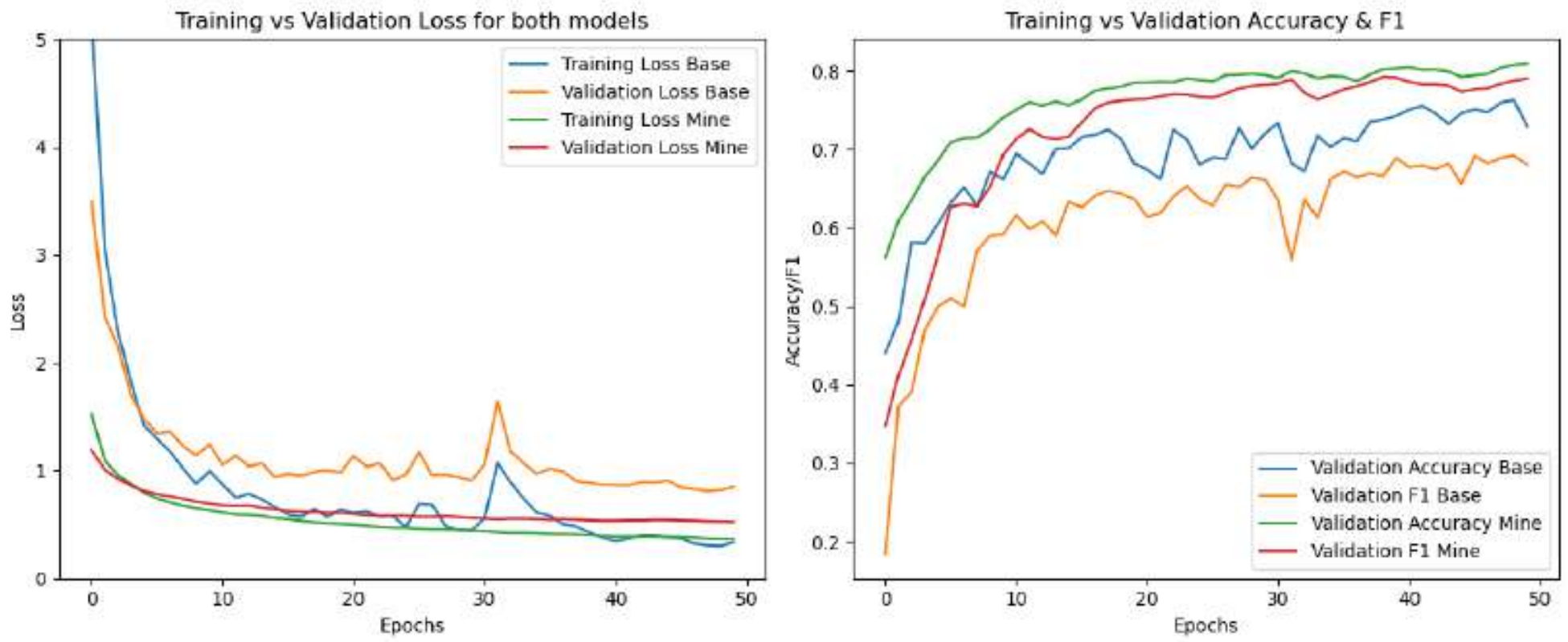

Figure 9.5: The result graph of the frozen TL results on the frozen 10-90 split.

Table 9.10: The results of the frozen TL results on the frozen 30-70 split. The percentage values given are the test F-1 scores.

| Frozen 30-70 | |
|---|---|
| **Baseline** | 77.50% |
| **At Epoch** | 50 |
| **PLDC-Net** | 86.31% |
| **At Epoch** | 47 |
| **Overtake PLDC-Net** | 77.73% |
| **At Epoch** | 5 |

**30% Train Data - 70% Test Data** With the dataset split 30-70 and frozen CNN layers, the results that can be seen in Figure 9.6 and 9.10 have been achieved.

With this somewhat bigger but still rather small (and imbalanced) train set, the baseline model managed 77.50% F1-Score. The PLDC-Net managed to outperform that score after only 5 epochs with 77.63% and with a max score of 86.31% PLDC-Net managed to outperform the baseline model nearly 9%. Figure 9.6 also showcases the difference in biasing between PLDC-Net and the baseline model. Whereas the baseline model shows significant differences between accuracy and F1-Score, indicating a bias towards some classes, PLDC-Net actually showcases a (slightly) better F1-Score when compared with the accuracy, showing no bias, signifying that the model handles the imbalanced data well.

**50% Train Data - 50% Test Data** With a 50-50 split, the train set starts to become somewhat large. It is still just as imbalanced as it was in the previous experiments, but with over 3,800 images in the training set, the dataset now enters an image count that one might no longer consider to be ”too small”, however due to said imbalance, the bean classes are still only just over 200 images per class. Results are listed and showcased in Table 9.11 and Figure 9.7.

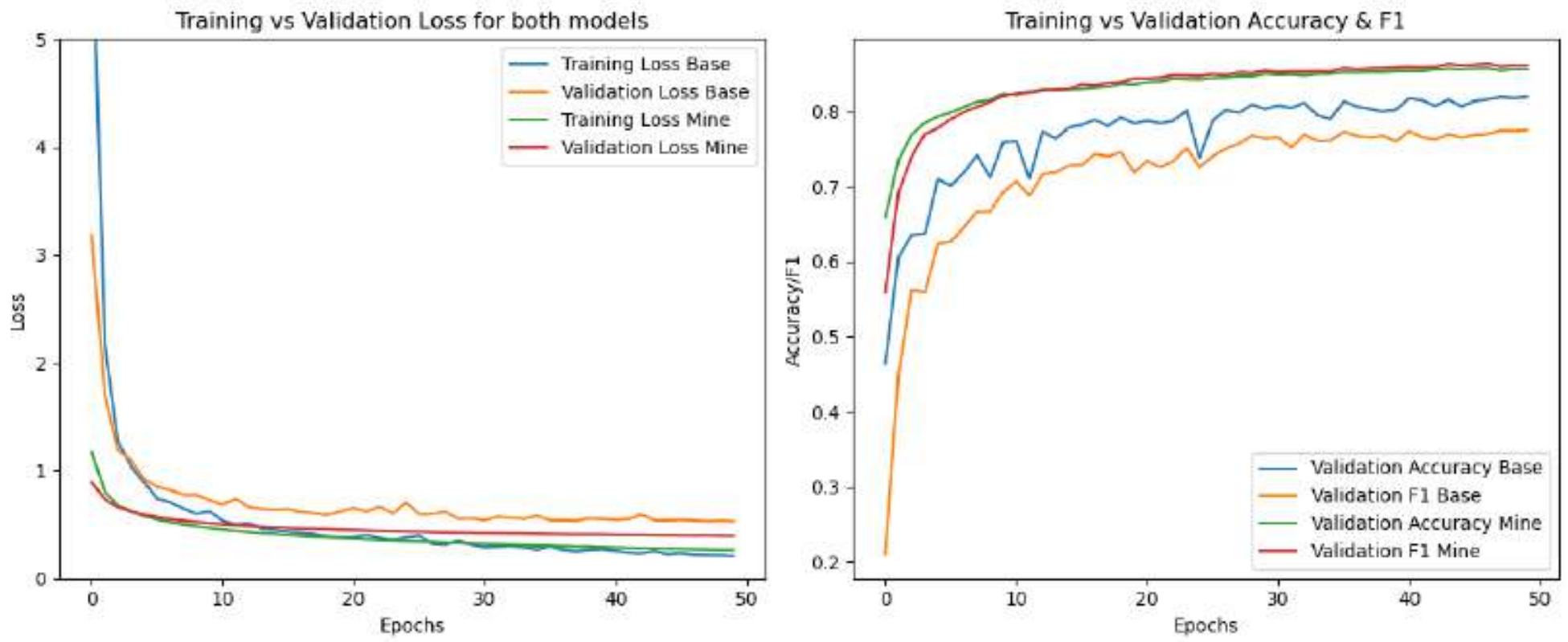

Figure 9.6: The result graph of the frozen TL results on the frozen 30-70 split.

Table 9.11: The results of the frozen TL results on the frozen 50-50 split. The percentage values given are the test F-1 scores.

| Frozen 50-50 | |
|---|---|
| **Baseline** | 81.53% |
| **At Epoch** | 45 |
| **PLDC-Net** | 88.02% |
| **At Epoch** | 45 |
| **Overtake PLDC-Net** | 82.24% |
| **At Epoch** | 7 |

The experiments on the 50-50 split dataset have, even though data is becoming increasingly bigger, produced results that are similar to the ones from the earlier and smaller dataset runs. The PLDC-Net model manages to outperform the baseline by over 6%. The proposed model manages to outperform the baseline's max score after only 5 epochs and it does again learn much more stable, faster and more robust, with the F1-Score to accuracy ratio being much more desirable.

**70% Train Data - 30% Test Data** With the 70-30 split, and a subsequent training set of over 5,300 images, the dataset is starting to resemble a larger dataset. However, due to the heavy imbalance, the smaller classes still only contain roughly 300 images. The frozen model's performance can be seen in Table 9.12 and Figure 9.8.

When comparing the performance of the frozen PLDC-Net and the baseline model on the 70-30 split of PLDC-6, one can see that the PLDC-Net model again managed to catch up and overtake the baselines max score of 82.25% in only 5 epochs. PLDC-Net managed a score with an over 6% better F1-Score of 88.63% in the end. The graph also showcases similar findings to before, where the PLDC-Net's training scores showcase more robust and faster learning, while the baseline's show a rift between the F1-Score and the accuracy, suggesting some biasing in the baselines weights.

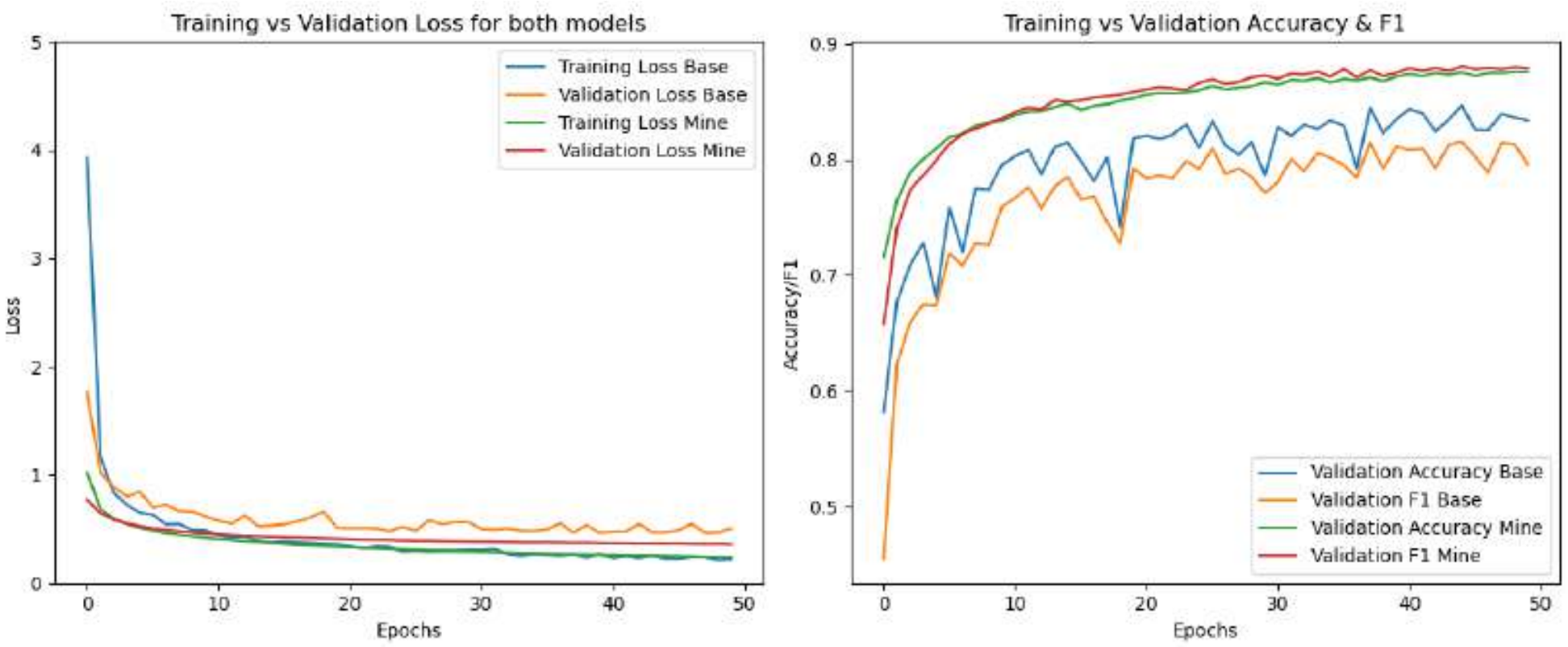

Figure 9.7: The result graph of the frozen TL results on the frozen 50-50 split.

Table 9.12: The results of the frozen TL results on the frozen 70-30 split. The percentage values given are the test F-1 scores.

| Frozen 70-30 | |
|---|---|
| **Baseline** | 82.25% |
| **At Epoch** | 43 |
| **PLDC-Net** | 88.63% |
| **At Epoch** | 47 |
| **Overtake PLDC-Net** | 83.10% |
| **At Epoch** | 5 |

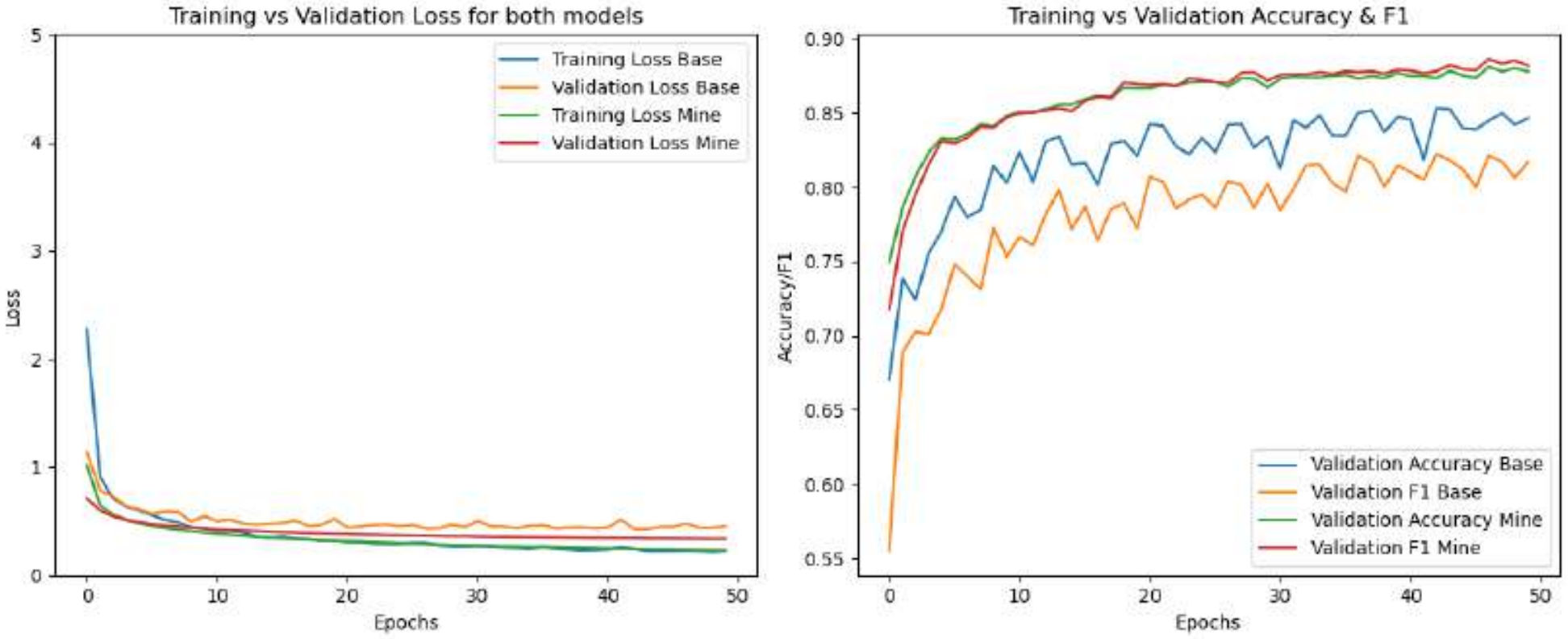

Figure 9.8: The result graph of the frozen TL results on the frozen 70-30 split.

Table 9.13: The results of the frozen TL results on the frozen 90-10 split. The percentage values given are the test F-1 scores.

| **Frozen 90-10** | |
|---|---|
| **Baseline** | 85.06% |
| **At Epoch** | 30 |
| **PLDC-Net** | 90.13% |
| **At Epoch** | 40 |
| **Overtake PLDC-Net** | 85.23% |
| **At Epoch** | 7 |

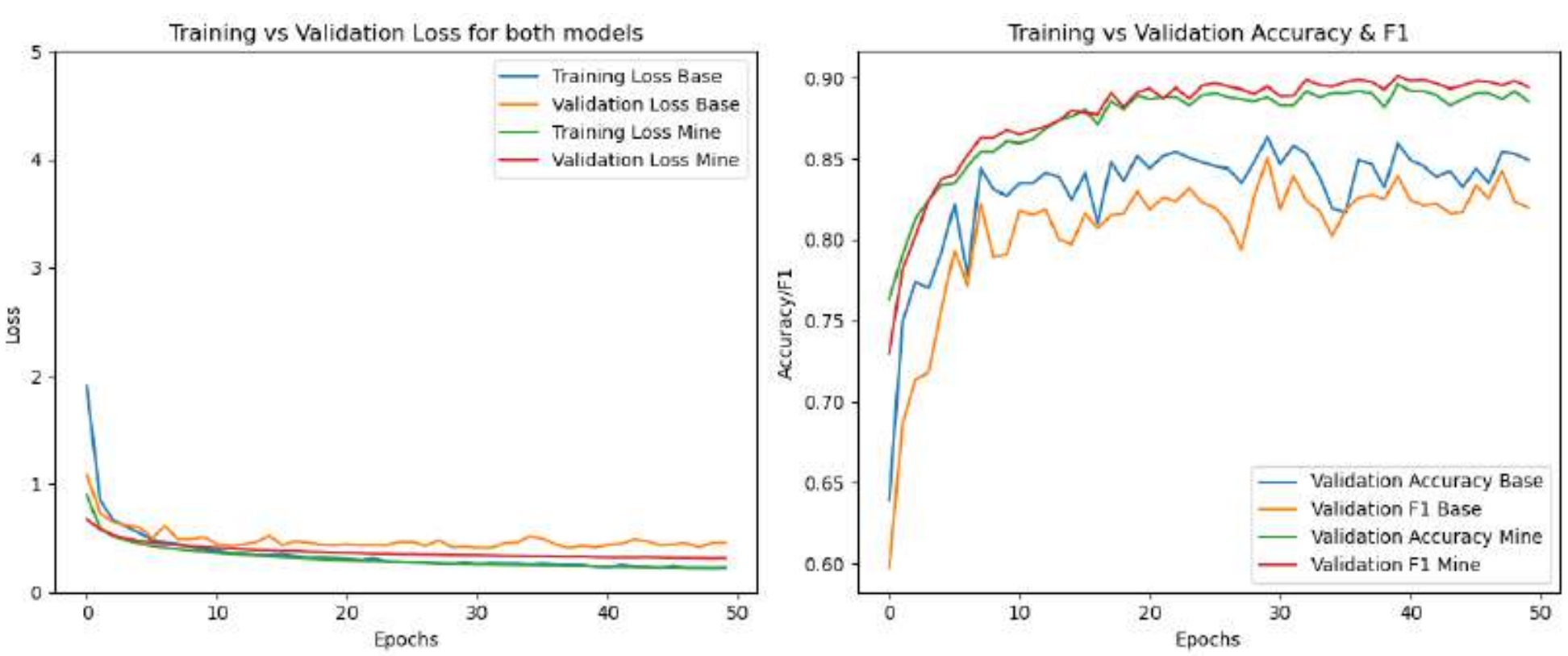

Figure 9.9: The result graph of the frozen TL results on the frozen 90-10 split.

**90% Train Data - 10% Test Data** The largest test data split - with 90% of the PLDC-6 data being used for training, allows both models to now utilize a fairly decently large dataset for training. While still imbalanced, this dataset of now nearly 7,000 images with all classes almost topping 400 images, is now of a size where many researchers or farmers would need to spend significant amounts of time into collecting and annotating the data, if they can even muster this many images per class from the plant available in their fields. Results can be seen in Figure 9.9 and Table 9.13.

Results obtained from these runs again showcase that PLDC-Net manages to outperform the baseline model when comparing max performance, with PLDC-Net reaching 90.13% on the F1-Score, while the baseline model only managed 85.06%. PLDC-Net managed to outperform these 85.06% after only 7 epochs, where PLDC-Net was already up to 85.23%. The graph also again shows the same story, where the PLDC-Net's accuracy and F-1 scores are almost identical, with F1 actually being higher, while there is a gap in the performance of the baseline model.

**Frozen Results Summary** The results obtained by training the frozen PLDC-Net model and the baseline model on the PLDC-6 dataset show how significant the advantage a domain-related and domain-relevant Base Model can have over a general Foundational Model. PLDC-Net managed to outperform the baseline consistently on each of the different dataset splits, with the model managing a higher performance with 10% training data than the baseline did with 30%, meaning it did better with only 1/3 of the data, while the baseline with 3x the data still fell behind. PLDC-Net with only 30% of the data also managed to beat the baseline that was trained on 90% of the data, again showing that it can be 3x as data efficient and still more powerful. It does all that in a setting with limited data, with a frozen classifier, meaning it does that in a setting where it is computationally lighter than it would be to re-train the entire model. It is also, across the board, much faster, much more stable, much more robust, and performs better without exception when using the computationally more efficient and therefore more desirable method to locations without high end computers and countless training data (such as many farms). These results more than validate PLDC-Net and the idea of domain-specific BM.

#### 9.3.3.2 Unfrozen

Here the results of the unfrozen experiments will be presented and analyzed. Unfrozen training runs are much more computationally expensive (see Table 9.8, since each parameter needs to be trained during these runs. This is also why the batch size had to be reduced from 128 to only 32, since the machine used for testing can not handle the full 128 batch size in un-frozen mode. As such, they are less applicable to use-cases where computational power is at an impasse, which can include farms. Unfrozen Transfer-Learning also is much more prone to overfit on small data, since the TL domain specific dataset influences every weight in the model. However, unfrozen training can lead to stronger overall results, baring all the downsides listed.

Table 9.14: The results of the unfrozen TL results on the unfrozen 10-90 split. The percentage values given are the test F-1 scores.

| Unfrozen 10-90 | |
|---|---|
| **Baseline** | 79.50% |
| **At Epoch** | 43 |
| **PLDC-Net** | 83.88% |
| **At Epoch** | 33 |
| **Overtake PLDC-Net** | 80.57% |
| **At Epoch** | 6 |

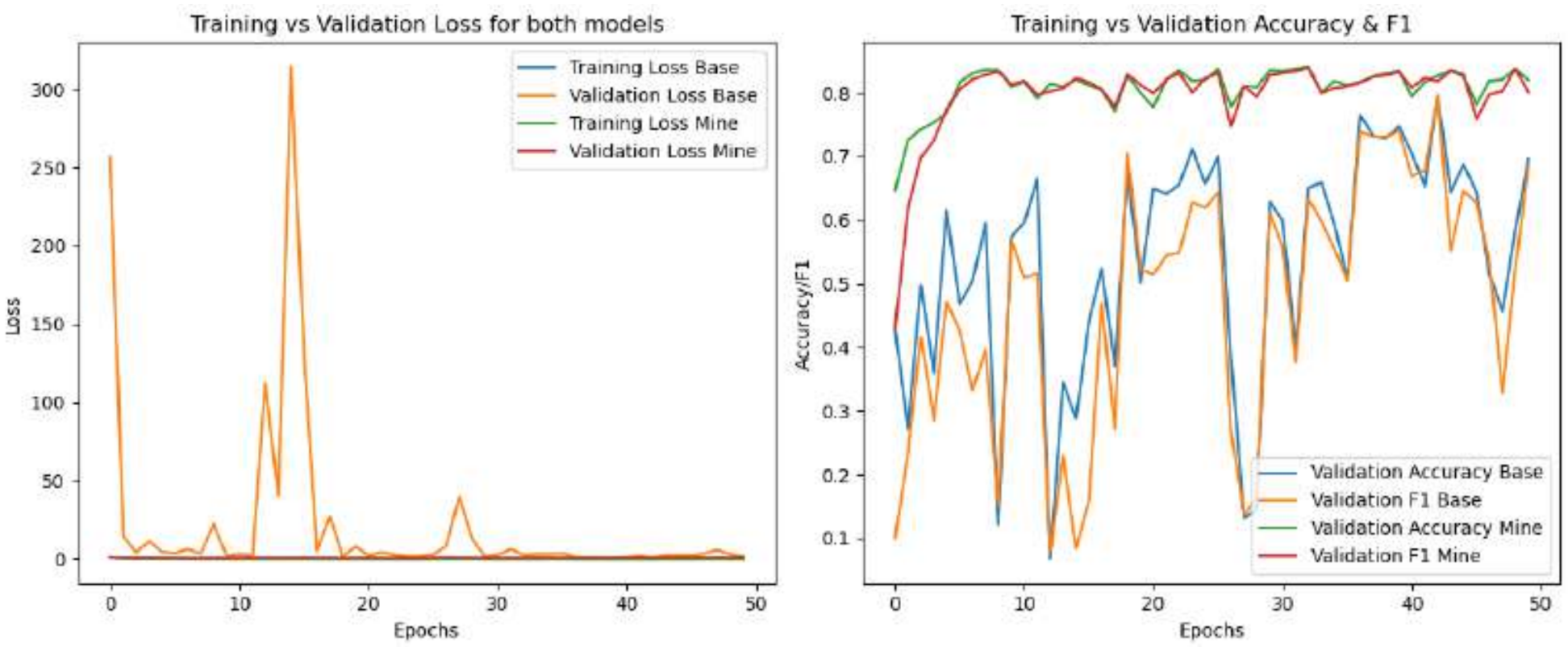

Figure 9.10: The result graph of the unfrozen TL results on the unfrozen 10-90 split.

**10% Train Data - 90% Test Data** The smallest train split (see Table 9.7) with less than 800 total images across 6 classes and with only around 40 images per bean class, used for training on the un-frozen models produced the results in Table 9.14 and in Figure 9.10, with Figure 9.11 providing a zoomed in view to the loss, as the validation loss of the baseline model is too high to see the more subtle trends of PLDC-Net.

Results showcase that PLDC-Net managed to outperform the baseline model by over 5% in overall performance, and it managed to overtake the max performance posted by the baseline after only 6 epochs. This shows that PLDC-Net manages to perform better on the most difficult train dataset split and it does also learn much faster. Even more obvious are the findings observed in the Figures (Figures 9.10 and 9.11). Here it becomes obvious how much the baseline model fluctuates, with performance being incredibly instable during the entire training period. PLDC-Net, on the other hand, performs much more stable and robust. PLDC-Net is, based on the results observed, much more suited to handle this task.

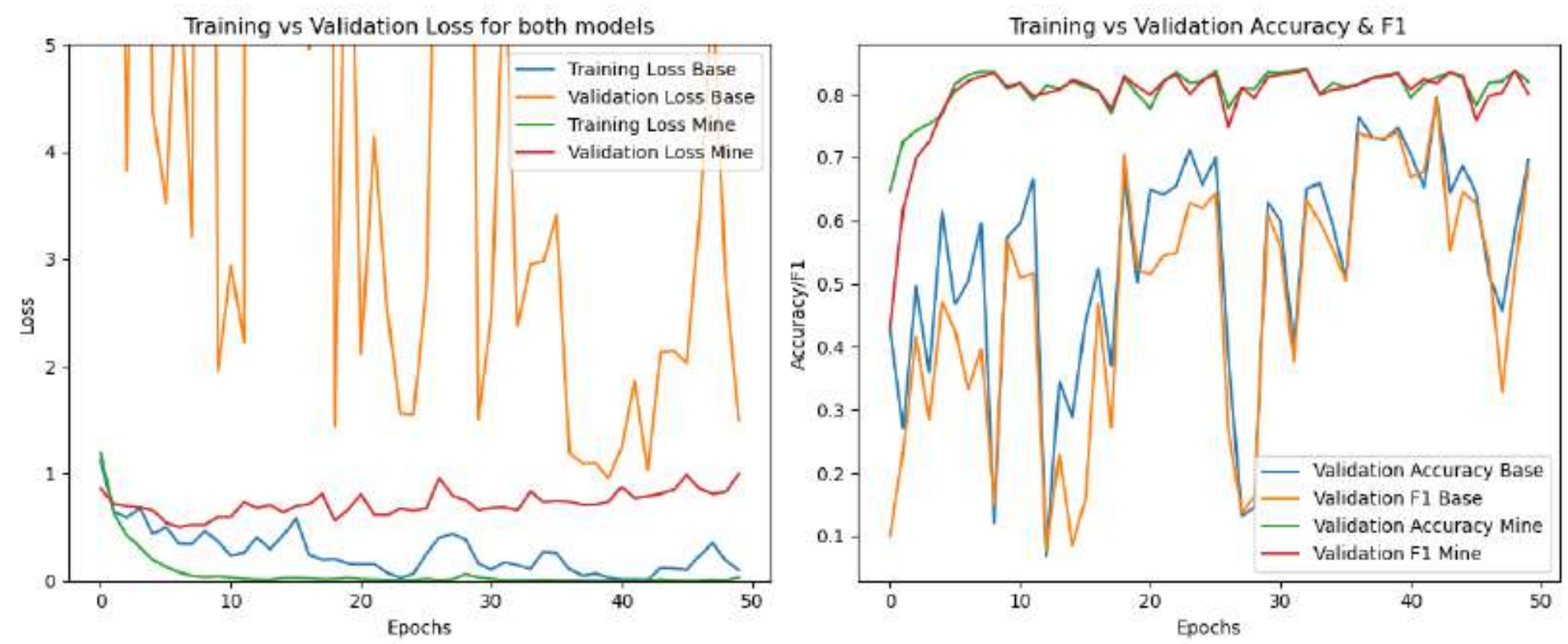

Figure 9.11: The zoomed in result graph of the unfrozen TL results on the unfrozen 10-90 split.

Table 9.15: The results of the unfrozen TL results on the unfrozen 30-70 split. The percentage values given are the test F-1 scores.

| Unfrozen 30-70 | |
|---|---|
| **Baseline** | 90.75% |
| **At Epoch** | 46 |
| **PLDC-Net** | 89.94% |
| **At Epoch** | 7 |
| **Overtake PLDC-Net** | None |
| **At Epoch** | None |

**30% Train Data - 70% Test Data** Increasing the training dataset size by a factor of 3 to a 30-70 split shows a sharp increase in overall performance for both models, as they can now make use of significantly more data. Results are available in Table 9.15 and Figure 9.12.

This is the first test scenario where the baseline model managed to achieve a higher overall F1-Score than PLDC-Net. However, one can also see that PLDC-Net manages it's 89.95% F1 max score after only 7 epochs, while the baseline took 46 to reach 90.75%, only less than 1% more. The max score achieved by PLDC-Net after 7 epochs is only outperformed by the baseline after 46 epochs of training. The graphs also indicate that the baseline model is much more vulnerable to overfitting and trains much less table and much more erratic with high variance in performance epoch to epoch.

**50% Train Data - 50% Test Data** The 50-50 split dataset with the un-frozen model again shows improved performance for both models. Results are shown in Table 9.16 and Figure 9.13.

Here too the baseline did manage to score higher in overall performance with an around

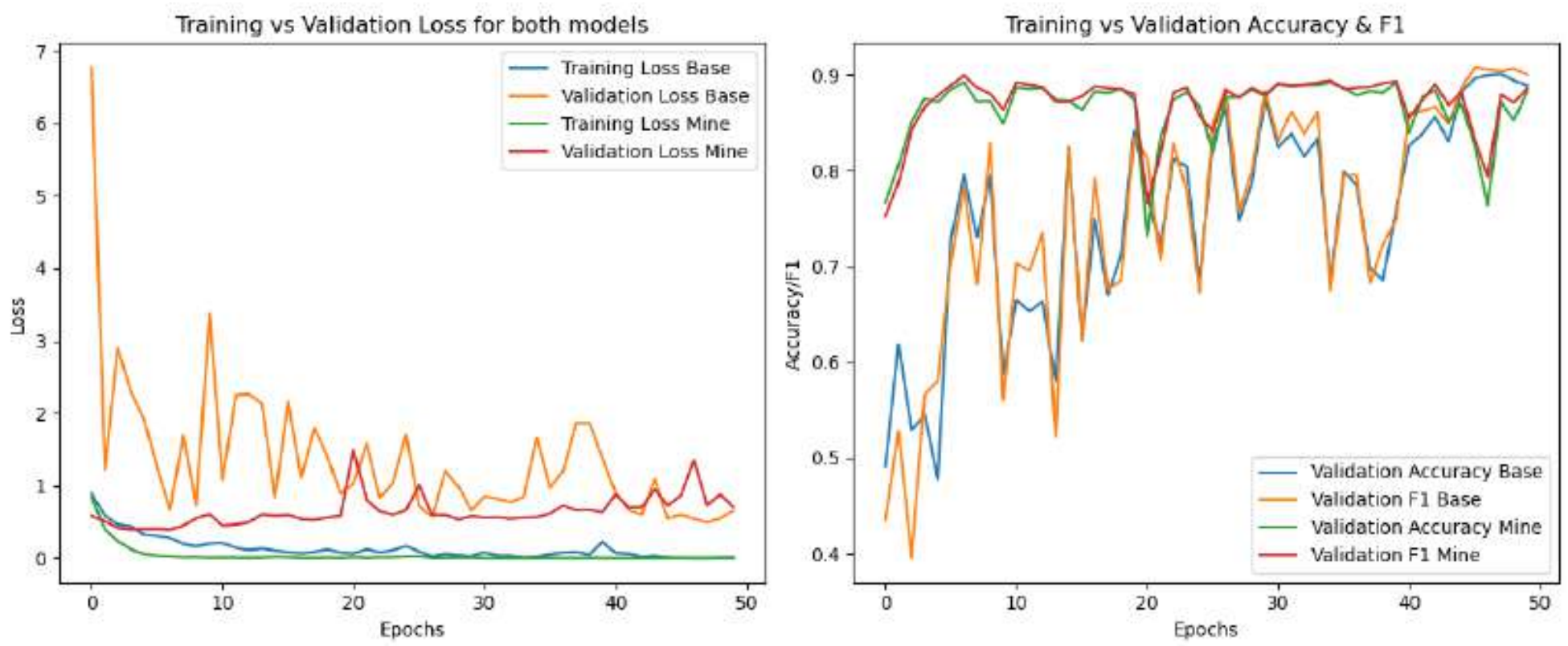

Figure 9.12: The result graph of the unfrozen TL results on the unfrozen 30-70 split.

Table 9.16: The results of the unfrozen TL results on the unfrozen 50-50 split. The percentage values given are the test F-1 scores.

| Unfrozen 50-50 | |
|---|---|
| **Baseline** | 92.63% |
| **At Epoch** | 44 |
| **PLDC-Net** | 91.99% |
| **At Epoch** | 11 |
| **Overtake PLDC-Net** | None |
| **At Epoch** | None |

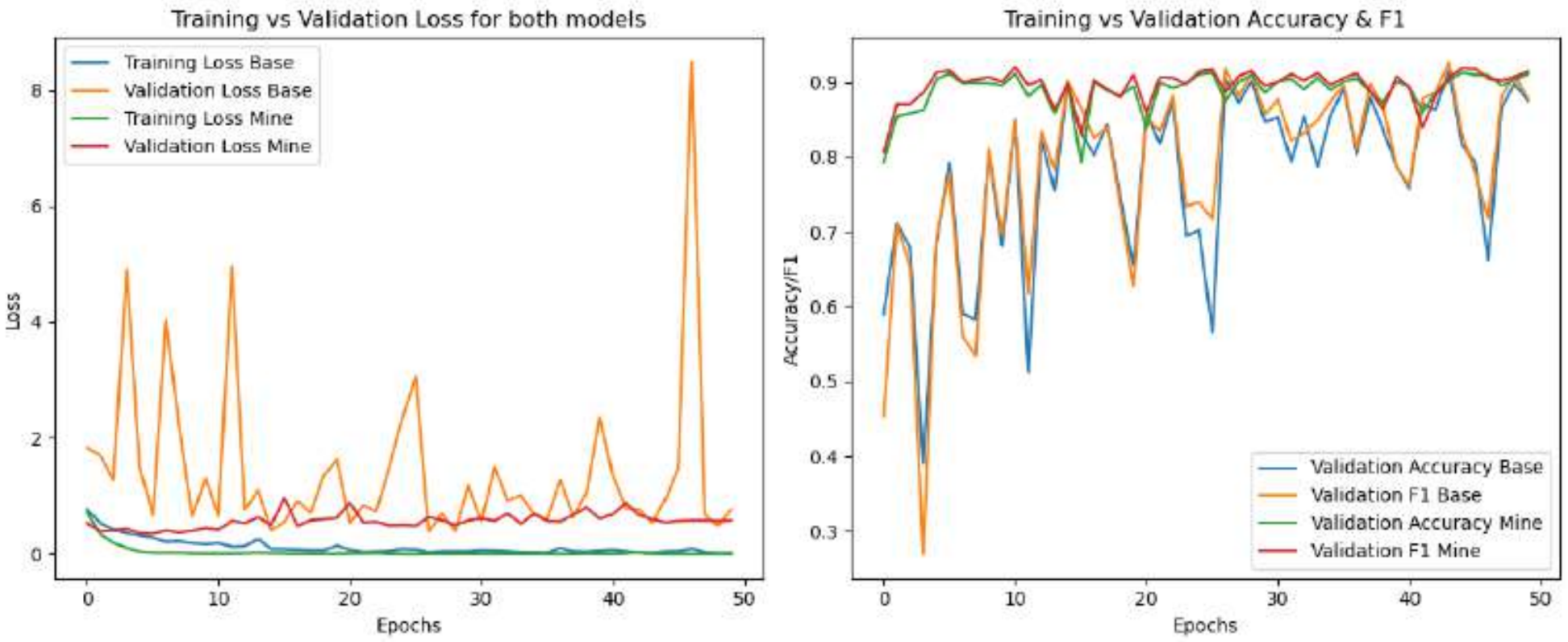

Figure 9.13: The result graph of the unfrozen TL results on the unfrozen 50-50 split.

Table 9.17: The results of the unfrozen TL results on the unfrozen 70-30 split. The percentage values given are the test F-1 scores.

| Unfrozen 70-30 | |
|---|---|
| **Baseline** | 91.74% |
| **At Epoch** | 39 |
| **PLDC-Net** | 93.08% |
| **At Epoch** | 24 |
| **Overtake PLDC-Net** | 92.50% |
| **At Epoch** | 8 |

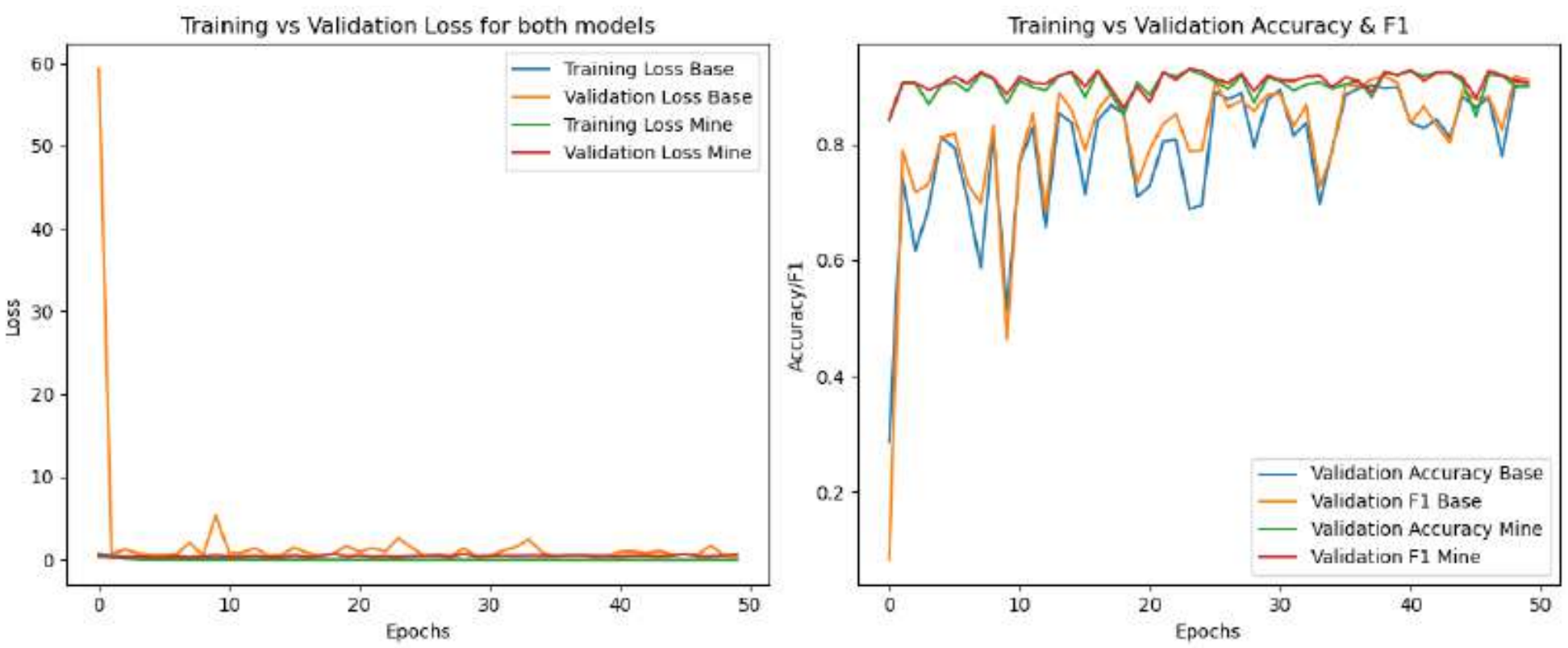

Figure 9.14: The result graph of the unfrozen TL results on the unfrozen 70-30 split.

0.6% better F1-Score, but as was the case with the 30-70 split, PLDC-Net managed its high score after only 11 epochs, while it took the baseline model 44 epochs to outscore that. The graphs also paint a similar picture to the graphs obtained by the previous un-frozen experiments, with the baseline model being very unstable and erratic, while the PLDC-Net model trains much more smoothly and stable, with no sudden variance and no sudden large drop-offs.

**70% Train Data - 30% Test Data** Again, with the 70% split used for training, the dataset starts to resemble a somewhat large scale dataset, one that might not easily be collected without significant time and financial investments, and only if enough plants per disease are available in field. Results of the un-frozen models on this larger dataset are presented in Table 9.17 and Figures 9.14 and the zoomed in Figure 9.15.

Results of this 70-30 split un-frozen run showcase that PLDC-Net managed to outperform the baseline again, reaching a max score of 93.08% compared to the 91.74% achieved by the baseline. This is also a step back in performance of the baseline when compared to the 50-50 split. One can also see that the PLDC-Net model managed to outscore the baseline's 91.74% after only 8 epochs, at which PLDC-Net already managed to score 92.50%. Again, the plots showcase a much more erratic performance by the baseline, with the PLDC-Net model performing much more stable. Although one has to point out that the baseline is less variant on the larger dataset than it was on previous splits.

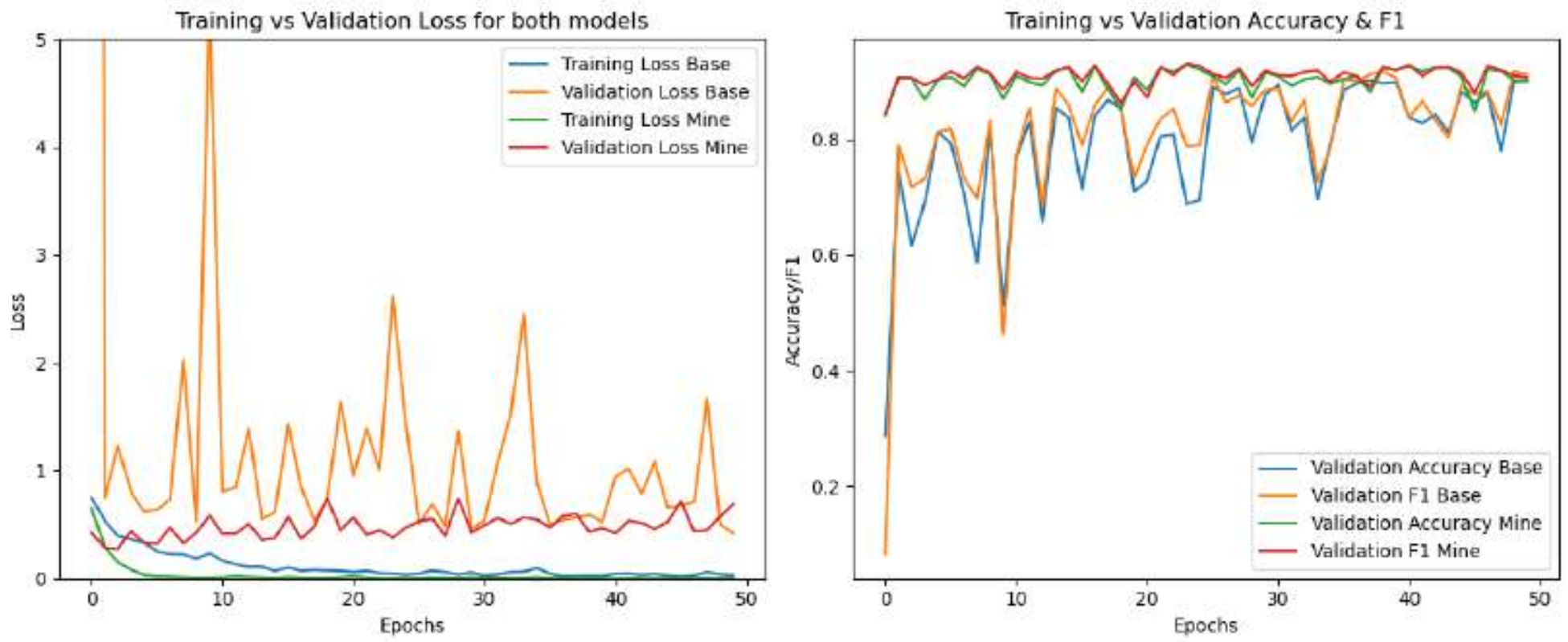

Figure 9.15: The zoomed in result graph of the unfrozen TL results on the unfrozen 70-30 split.

Table 9.18: The results of the unfrozen TL results on the unfrozen 90-10 split. The percentage values given are the test F-1 scores.

| **Unfrozen 90-10** | |
|---|---|
| **Baseline** | 94.53% |
| **At Epoch** | 36 |
| **PLDC-Net** | 95.09% |
| **At Epoch** | 20 |
| **Overtake PLDC-Net** | 94.82% |
| **At Epoch** | 19 |

**90% Train Data - 10% Test Data** The largest train split, with 90% of the data in PLDC-6, does come with a size that is not easy to collect, but does come with the best setup for un-frozen training. Results are shown in Table 9.18 and Figures 9.16 and 9.17 with the latter being the zoomed in representation.

Results showcase both models achieving their overall highest scores across all experiments, which is to be expected with the un-frozen (but heavy and time consuming) training on a rather large (but hard to collect) dataset. However, even here the PLDC-Net model managed a higher score than the baseline model did. It also managed to achieve said higher score after only 19 epochs of training. The figures also, again, showcase that the baseline model is much less stable and much more erratic than the PLDC-Net.

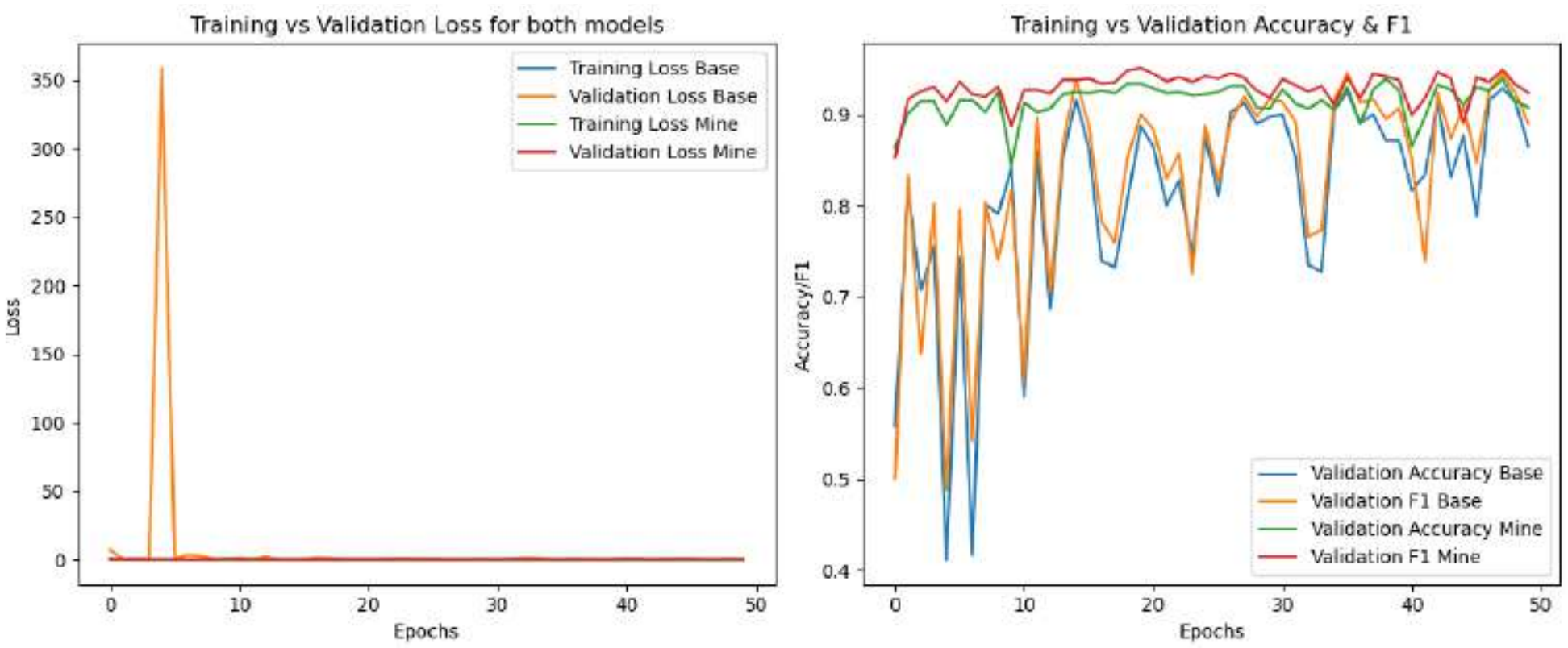

Figure 9.16: The result graph of the unfrozen TL results on the unfrozen 90-10 split.

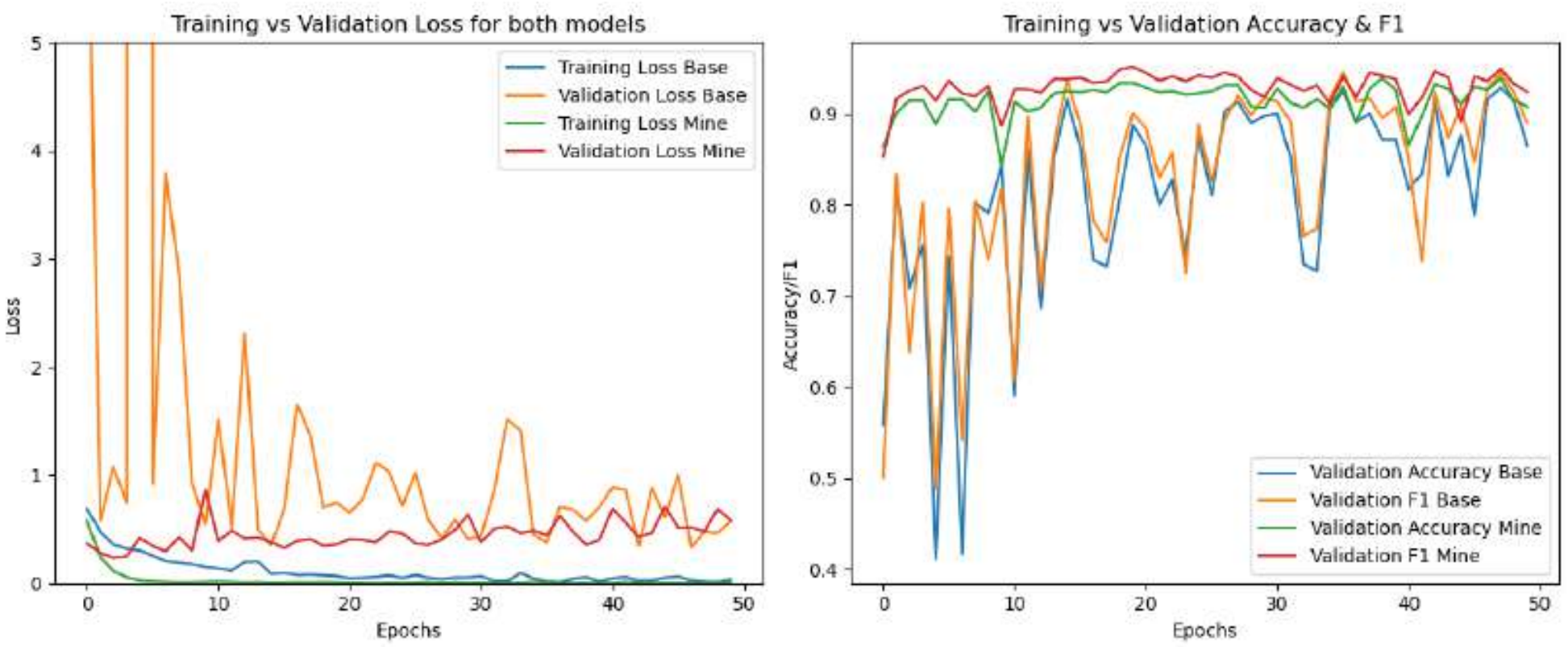

Figure 9.17: The zoomed in result graph of the unfrozen TL results on the unfrozen 90-10 split.

**Frozen Results Summary** In the un-frozen set of experiments, which are more computationally expensive and more time consuming, the PLDC-Net model managed to outperform the baseline model in 3 of 5 cases. In the 2 cases where the baseline model did manage higher max scores, the PLDC-Net model did manage results that are not too far of the baseline model's, while achieving said comparable scores much faster. Maybe more noticeable is the volatility that the baseline model operates with. All figures show that the baseline's results are very erratic, volatile, variant, and unstable. Loss, accuracy and F1-Scores of the PLDC-Net, however, are much more stable, with no large drop-offs in performance from one epoch to the other. PLDC-Net also manages to learn much faster than the baseline did, on top of being much more stable, as just mentioned.

#### 9.3.3.3 Transfer-Learning Results Summary

Through these TL experiments and the results obtained from them, once can see that PLDC-Net managed to outperform the baseline model in each case when using the preferred frozen training method, with large improvements in each case, especially so in cases where the dataset was more limited. With data being sparse, these frozen and low data count cases are the most relevant and most interesting results obtained. With PLDC-Net being this much more powerful in these test-cases, its applicability in real world scenarios, where large data and advanced computational power and time might not be guaranteed, it sets itself apart as the better, faster, more powerful, more stable, and cheaper alternative to the baseline approach. When using the un-frozen approach, which is heavier in terms of time and computation, the baseline model did manage to score a higher overall score in 2 out of 5 cases. The difference to PLDC-Net was, however, not very large and PLDC-Net still managed to learn much faster. One thing that was even more noticeable in the un-frozen experiments was the inconsistency and erraticness that the baseline model trained with. While PLDC-Net still trained stable and fast, the baseline model was extremely volatile and unstable during training, especially so in the use-cases where data was limited.

Overall these results showcase the validity of PLDC-Net and the improvements it brings to the field of plant leaf disease classification. This domain-related and domain-relevant BM trained on the PLDC-80 dataset uses the related feature maps it obtained from the PLDC-80 dataset and transfers them to the new unseen PLDC-6 dataset much more effectively than the baseline DenseNet201 model that was pre-trained on the general ImageNet dataset.

## 9.4 One-Shot and Few-Shot Learning

To further test the robustness and adaptability of PLDC-Net, especially in low data environments, one-shot and few-shot learning was employed. The above mentioned bean dataset was used, but without merging it with the soybean dataset to have more consistent data, which these techniques rely upon. Only a very small number of images (1 and 5) per class was used during ”training” of the classifier. The models, PLDC-Net and ImageNet pre-trained DenseNet201 as the baseline, are not re-trained, or fine-tuned, nor is the dense layer at the end, at all during this. Instead they are simply used as feature extractors to provide embeddings of the one-shot images and then compare future images based on those embeddings.

### 9.4.1 Dataset

The dataset used here, as mentioned above, is the iBean dataset [142] with 3 classes taken in somewhat similar conditions. Depending on the experiments, one-shot or few-shot, a different amount of images will be used for ”training”, with the rest being used for testing. In one-shot learning, only 1 image per class will be used during ”training”, 3 in total, while in few-shot learning 5 per class (15 in total) are used.

Table 9.19: iBean dataset used for few-shot learning.

| iBean Dataset | |
|---|---|
| Images | 1,296 |
| Classes | 3 |
| Images per Class | • Healthy - 428<br>• Angular Leaf Spot - 432<br>• Bean Rust - 436 |

### 9.4.2 Methodology

In the few-shot learning experiments here, the models (PLDC-Net and ImageNet pre-trained DenseNet201) are not trained at all. The models are only used in inference mode. Each model is cut off at the max pooling layer, deleting the classifier top. This feature extractor output of the max pooling layer is used for few shot learning as embeddings for the classes.

Each of the image support samples (training images), so 1 per class for one-shot and 5 per class for few-shot, are fed to the network and the predicted feature extractor embeddings are saved (and averaged in the case of few-shot). These embeddings (which represent the class of the support image(s) that generated it), which are a vector of 1920, are then used to predict classes for future images of the query (testing data). Each testing query image is also fed into the same model, and its embeddings are compared to the set of 3 prototype embeddings (1 per class) of the training support embeddings saved earlier using Euclidean distance (see (9.3)).

$$d(\mathbf{x},\mathbf{y}) = \sqrt{\sum_{i=1}^{n}(x_i - y_i)^2} \tag{9.3}$$

Here $d$ is the Euclidean distance between the predicted query image and the saved prototype support image embedding, $x_i$ is the $i$-th dimensional position value of the prototype embedding, $y_i$ is the $i$-th dimensional embedding of the new query test image, and $n$ is the number of dimensions, which is 1920 in this case for both models.

Then the the prototype embedding (and with that the class it is representing) with the smallest distance to the new test image will be used as the predicted class and compared to the images true value, to obtain accuracy scores. Like this, the model does not need to be retrained, which would be near impossible with a single image per class, saves computational cost since it does not need to train, only inference, and even makes the model more light weight, since the dense layer at the end can be omitted. All this allows the model to be adjusted to classify new image classes of before unseen plants, diseases and classes (cross-domain), with incredibly limited data.

### 9.4.3 Results

When looking at the results of both models (PLDC-Net and ImageNet trained DenseNet201), we will differentiate 2 different approaches and their results. Each model was once used for one-shot learning, and once for 5-shot few-shot learning.

#### 9.4.3.1 Cross-Domain One-Shot Classification

In one-shot learning, where each class was only embedded with a single image (train image set of 1 image per class and therefore 3 images in total), we can see the models performing decently well considering the limitations (see Table 9.20 for results). The baseline model manages almost 54% accuracy and PLDC-Net reaches almost 61%. Both of these outperform randomness (which would be around 33% for 3 classes) significantly, proving they are indeed properly predicting. The PLDC-Net, however, manages to outperform the baseline by nearly 7%, showing significant improvement over the baseline in highly limited data.

Table 9.20: The results of both models One-Shot Learning on the iBean dataset.

| 1-Shot Learning | |
|---|---|
| **Baseline** | 53.95% |
| **PLDC-Net** | 60.91% |

#### 9.4.3.2 Cross Domain Few-Shot Classification

In the test case where each class is represented by 5 images during training, with average embeddings, the PLDC-Net manages to outperform the baseline by an even bigger margin, achieving a nearly 14% better accuracy score (PLDC-Net 75.70%, Baseline 61.89% - see Table 9.21). As a result, here too the PLDC-Net manages to provide significantly better and therefore more valuable predictions, in conditions that are still highly limited in terms of image count for training.

Table 9.21: The results of both models Five-Shot Learning on the iBean dataset.

| 5-Shot Learning | |
|---|---|
| **Baseline** | 61.80% |
| **PLDC-Net** | 75.70% |

#### 9.4.3.3 One and Few-Shot Result Summary

Overall, just like it was the case with TL, one-shot and few-shot learning results show that the PLDC-Net model pre-trained on domain relevant and domain related data manages to produce features that are better and more helpful to predict new plant, diseases and classes in cross-domain settings. The five-shot baseline model only managed to slightly outperform the one-shot PLDC-Net model by less than 1% in a case where the baseline had 5x the images. The PLDC-Net model was significantly strong in both scenarios, proving it can produce stronger plant leaf disease classifiers in data and resource constraint scenarios, which are real limitations that the field is faced with, making the PLDC-80 pre-trained PLDC-Net the stronger and better choice for domain adaptation when compared to the baseline.

# 10. Discussion

Through thorough, step-by-step research, PLDC-80, PLDC-Net and PLDC-6 were created in an effort to create a Base Model that is specifically tailored to the field of plant leaf disease classification. Whereas before, as pointed out by prior work [14, 11], datasets are an issue, with large scale diverse high quality field or hybrid datasets being sparse. After having conducted a detailed and extensive benchmark, suitable datasets and models have been identified that could be utilized for future steps such as creating a large scale benchmarking dataset that can be used for large scale models and a set of models that could be used as baseline comparisons as well as the basis for improvements. While collecting a totally new dataset is out of scope of this work, and as such would still be a worthwhile and desirable addition to the field, PLDC-80 combines, merges, augments (using proven methods), and filters 9 pre-existing benchmarked datasets (hybrid, lab, and field). Through this process PLDC-80 manages to reach over 300,000 images across 80 classes, representing 25 different plant species. This size, diversity (both in terms of plant types, diseases, and capturing conditions), and quality make for a uniquely capable dataset for the field of plant leaf disease classification. This dataset allows the opportunity to properly train a large scale CNN classifying model in a way in which it would not be possible with other openly available datasets. Using this dataset, a large number of models were trained to compare their robustness and generalization ability on PLDC-80's test set. DenseNet201 was the best performing baseline model, so it was chosen to be the basis of improvements. The improvements chosen were Channel Attention and Swish-1 SiLU activation. This new model, called PLDC-Net, managed to outperform the other models on the PLDC-80 dataset in terms of test performance. As such, it was used for another set of experiments, by using it as a foundational pre-trained model for TL, comparing it to the commonly used ImageNet pre-trained DenseNet201 baseline. In these experiments, on the PLDC-6 dataset, the model managed to outperform the baseline in most cases, learning faster, more robust, less volatile weights and features. These experiments prove that the domain-specific Base Model, specific to the field of plant leaf disease classification, provides better training conditions for scenarios with small dataset and lower computational capacity. All in all, this proves the need for and the validity of PLDC-Net and the PLDC-80 dataset, as it improves the TL abilities for this field, allowing future research to train more robust models with less data in a shorter time frame.

# 11. Future Work

With PLDC-80, PLDC-6, and PLDC-Net now having been developed, future work can be built upon them. Future directions following these developments could go in a few different directions. For one, future work could try to improve PLDC-Net by trying even more different methods to change the architecture in hopes of managing better scores on the PLDC-80 benchmarking set. Alternatively, one could try to build a completely new model architecture for this field, although this task would be very difficult, since existing architectures like DenseNet201 already achieve great results

One could also make use of the benchmarking results to better select models for future experiments. The results showcase a clear difference between different model families and architectures, which should help future works to make a better informed pick for as to which model to pick.

The same benchmark, in combination with the results of the augmentation assessment, can also be used to enhance already existing datasets or even during the creation of new datasets, that are yet to be collected. Collecting new datasets is laborious, time consuming, and also potentially costly, so utilizing the benchmarking results to see what dataset size, diversity, and type are best suited for the field, as well as understanding which augmentations are most beneficial, can make creating more new datasets less of a problem.

Another option, one that could have large impact, but one that comes with a hefty cost, would be to create a completely new, large scale, diverse, high quality benchmarking dataset, completely collected in field (and maybe also in lab) conditions specifically for this field. One could then either use that dataset to train future BM, or use that new dataset in combination with PLDC-80 to create a even larger and potentially better dataset to train future models.

One could also try observe and asses different methods for training models for plant leaf disease classification. One option would be to use distributed methods such as the decentralized federated learning method, to train models not on one dataset and one machine alone, but on a cluster or a network of multiple machines at the same time, which could reduce dataset requirements and computational costs per node.

# 12. Conclusion

This thesis first presents an extensive review of current literature in the field of plant leaf disease classification. This review provides a good look at the current state of the field and identifies current trends and best practices in the field. Following this literature review, this thesis explains a lot of the background information related to the field as well as relevant and essential information regarding CNN Deep Learning methods, upon which this thesis builds in later chapters. The background chapter also explains methods and ideas specific to plant leaf diseases and classification of such. Following that is a deep look into a large number of open datasets for plant leaf disease classification, giving an overview of possible options as well as giving a good overview of each datasets characteristics such as collection conditions, size, number of classes, class balance, etc. Next, relevant and popular large scale CNN classifier models are introduced and explained. These models make up a large number of the models used in current literature and also introduce a large number of currently popular methods. As such, this chapter gives good insight into the current state of CNN classifiers and these findings are later utilized in model construction. The next chapter contains the benchmarking experiments and results conducted as part of this research. This benchmark offers very insightful and important information into the field of plant leaf disease classification, identifying which datasets are better suited to be used for classification tasks as well as pointing out which models are more powerful when applied to plant leaf classification datasets. The benchmark shows significant differences between different datasets and models in terms of performance and applicability to this field and shows useful trends that can be utilized to chose the right combination for future work. The augmentation experiments offer further insights that can be implemented when enriching or even when creating datasets for plant leaf disease classification experiments. To solve the problem of insufficient data, PLDC-80 was introduced. A 80 class, 25 plant species and 300,000 images large hybrid dataset that can be use to benchmark models, as well as to train a domain-related and domain-relevant Base Model, due to its size, diversity, conditions and quality. Such a Base Model was then conceptualized, constructed, and tested on the PLDC-80 dataset where it was also compared with baseline models to verify its legitimacy and relevance, as well as to point out its performance gains. This PLDC-Net model was, after pre-training on the large scale PLDC-80 dataset, used for TL on another, smaller, related dataset that has no overlap with PLDC-80, called PLDC-6, since it contains 6 classes (unseen to the pre-trained PLDC-Net). Transfer-Learning experiments showcase the power of PLDC-Net, when compared to the baseline DenseNet201 model pre-trained on ImageNet. PLDC-Net was faster, more robust, more stable, and better performing, especially in low data scenarios when using the computationally lighter frozen TL approach. PLDC-Net trained on PLDC-80 managed to outperform the baseline model by over 10% F1-Score on 10-90 train-test split PLDC-6 using frozen CNN layers, signifying the substantial improvement made in terms of low data and computational availability, which is heavily beneficial in the field of plant leaf disease classification. The model also showcased better learning characteristics on top of the afore mentioned numbers, with it learning much faster, it being less erratic, volatile, and variant during training, while the baseline model suffers from all these problems. This positions PLDC-Net as a powerful domain-informed Base Model for plant leaf disease classification with many beneficial and desirable characteristics.

# 식물 잎 질병 분류를 위한 강력한 사전 학습 기본 모델 개발

**David Jona Richter**

전남대학교대학원 인공지능융합학과

(지도교수: 김경백)

(국문초록)

식물, 작물, 그리고 그 수확물은 우리 생존에 필수적입니다. 우리 사회는 식량을 농업과 농부에게 크게 의존하지만, 질병과 해충은 매년 큰 손실을 초래합니다. 따라서 질병을 조기에 발견하고 적절히 치료하여 가능한 한 확산을 막는 것이 매우 중요합니다. 수작업 및 전통적인 방법은 전문 지식을 갖춘 인력이 밭을 직접 돌아다니며 증상을 ”손으로” 확인하거나 현미경과 같은 실험실 장비를 사용하여 식물의 질병을 분석해야 합니다. 이는 당연히 매우 힘들고 시간이 많이 소요되며 숙련된 인력도 필요합니다. 따라서 이 과정을 자동화해야 할 필요성이 대두되었습니다. 그 결과 머신러닝(ML) 기법이 적용되어 유망한 결과를 얻었지만, 머신러닝에는 단점이 있습니다. 특히 이미지 데이터(일반 이미지, 다중 스펙트럼 이미지, 초분광 이미지 모두)에서 연구자가 특징을 추출해야 합니다. 딥러닝(DL) 기법은 이러한 문제를 해결할 뿐만 아니라 기존 머신러닝보다 성능이 더 우수한 경향이 있습니다. CNN 모델은 수동 특징 생성 없이 이미지에서 특징을 자동으로 추출한 후 분류기(ML 또는 DL)에 입력할 수 있으므로 특히 효율적입니다. 그 이후로 다양한 DL 모델이 제안되어 식물 질병을 식별하는 데 사용되었으며, 그 성공률은 다양했습니다. 그중 대부분은 이미지 데이터를 입력으로 사용하는 CNN 아키텍처였습니다. CNN을 비롯한 모든 DL 방법은 제대로 학습하려면 방대한 양의 데이터가 필요하기 때문에, 데이터 세트는 모델의 최종 성능에 큰 영향을 미칩니다. 데이터 세트가 분야에 미치는 중요성에도 불구하고, 공개적으로 사용 가능한 데이터 세트와 완전한 기능을 갖춘 모델을 충분히 학습하는 데 필요한 데이터 세트 간에는 여전히 약간의 차이가 있는 것으로 보입니다. 이러한 단점을 극복하기 위해 본 논문

에서는 식물 잎 질병 분류 분야의 공개 데이터 세트를 파악하고, 이를 기반으로 학습 가능한 모델을 선정했으며, 적합성을 확인하기 위한 광범위한 벤치마크를 수행했습니다. 그런 다음 이러한 결과와 증강 적용성 연구 결과를 바탕으로 새로운 데이터 세트를 구성했습니다. 이 데이터 세트는 DenseNet201 아키텍처를 기반으로 하는 새로운 모델을 훈련하는 데 사용될 예정이며, 해당 새로운 데이터 세트에서 기준 모델보다 우수한 성능을 보였을 뿐만 아니라 다른 새로운 데이터 세트에서 식물 잎 질병 분류 도메인 특정 전이 학습 실험에서도 기준 모델보다 우수한 성능을 보였습니다.

# A. Appendix Datasets

## A.1 Lab Datasets

### A.1.1 Plant Village

Table A.1: Information about the Plant Village Dataset.

| | |
|---|---|
| Plants | 1. Corn (Maize)<br>2. Potato<br>3. Pepper (Bell)<br>4. Tomato<br>5. Blueberry<br>6. Cherry<br>7. Squash<br>8. Orange<br>9. Raspberry<br>10. Apple<br>11. Grape<br>12. Peach<br>13. Strawberry<br>14. Soybean |
| Diseases | 1. Apple<br>• Healthy<br>• Cedar Apple Rust<br>• Apple Scab<br>• Apple Black Rot<br>2. Blueberry<br>• Healthy<br>3. Cherry<br>• Healthy<br>• Powdery Mildew<br>4. Corn<br>• Healthy<br>• Common Rust<br>• Northern Leaf Blight<br>• Gray Leaf Spot<br>5. Grape<br>• Healthy<br>• Leaf Blight<br>• Black Measles<br>• Black Rot<br>6. Orange<br>• Citrus Greening<br>7. Peach<br>• Healthy<br>• Bacterial Spot<br>8. Bell Pepper<br>• Healthy<br>• Bacterial Spot<br>9. Potato<br>• Healthy<br>• Early Blight<br>• Late Blight<br>10. Raspberry<br>• Healthy<br>11. Soybean<br>• Healthy<br>12. Squash<br>• Powdery Mildew<br>13. Strawberry<br>• Healthy<br>• Leaf Scorch<br>14. Tomato<br>• Healthy<br>• Leaf Mold<br>• Leaf Spot<br>• Early Blight<br>• Late Blight<br>• Spider Mites<br>• Bacterial Spot<br>• Mosaic Virus<br>• Yellow Leaf Curl Virus<br>• Target Spot |
| Number of Classes | 38 |
| Number of Images | 54,304 |

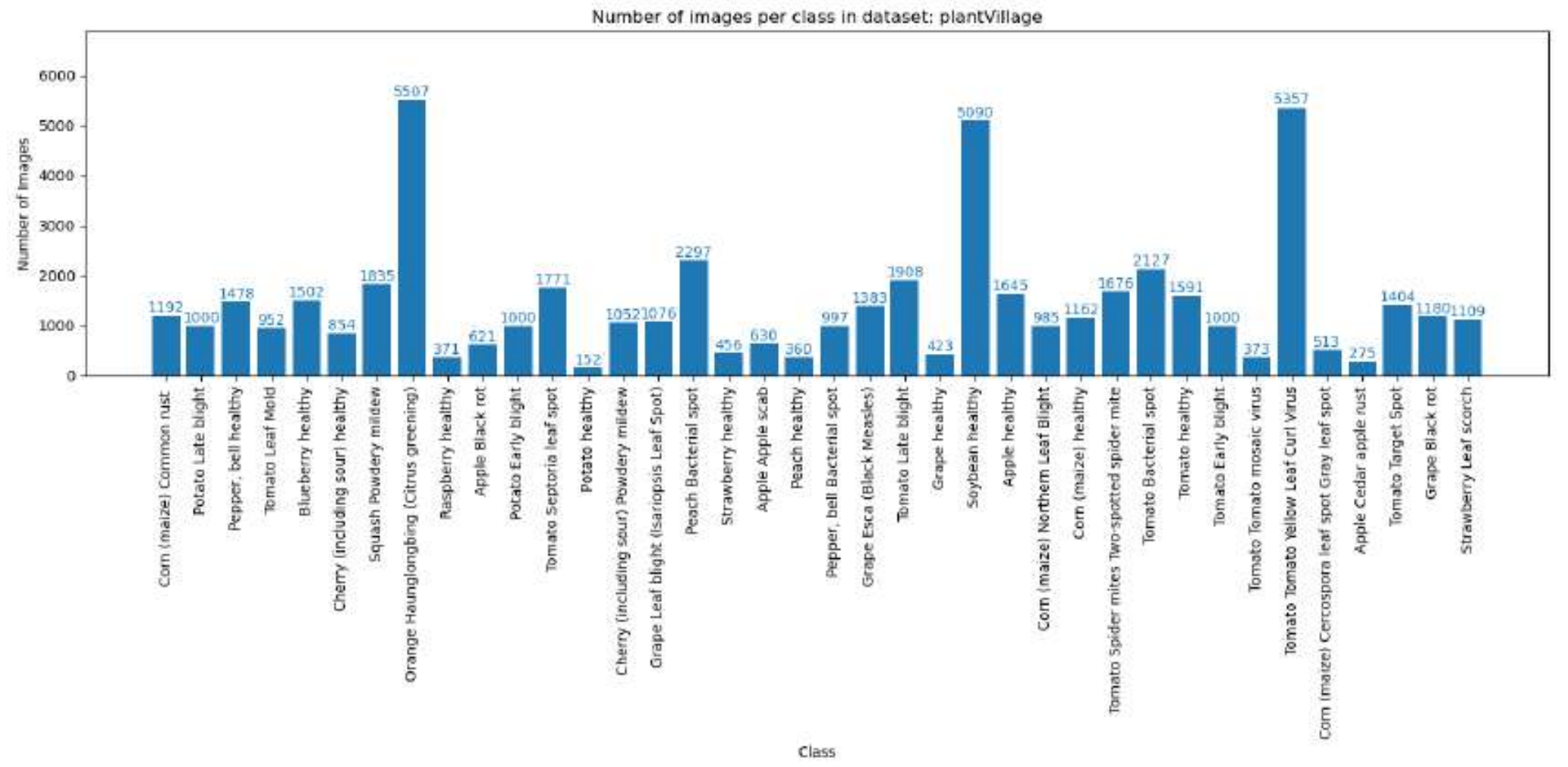

Figure A.1: Class Distribution of the Plant Village Dataset.

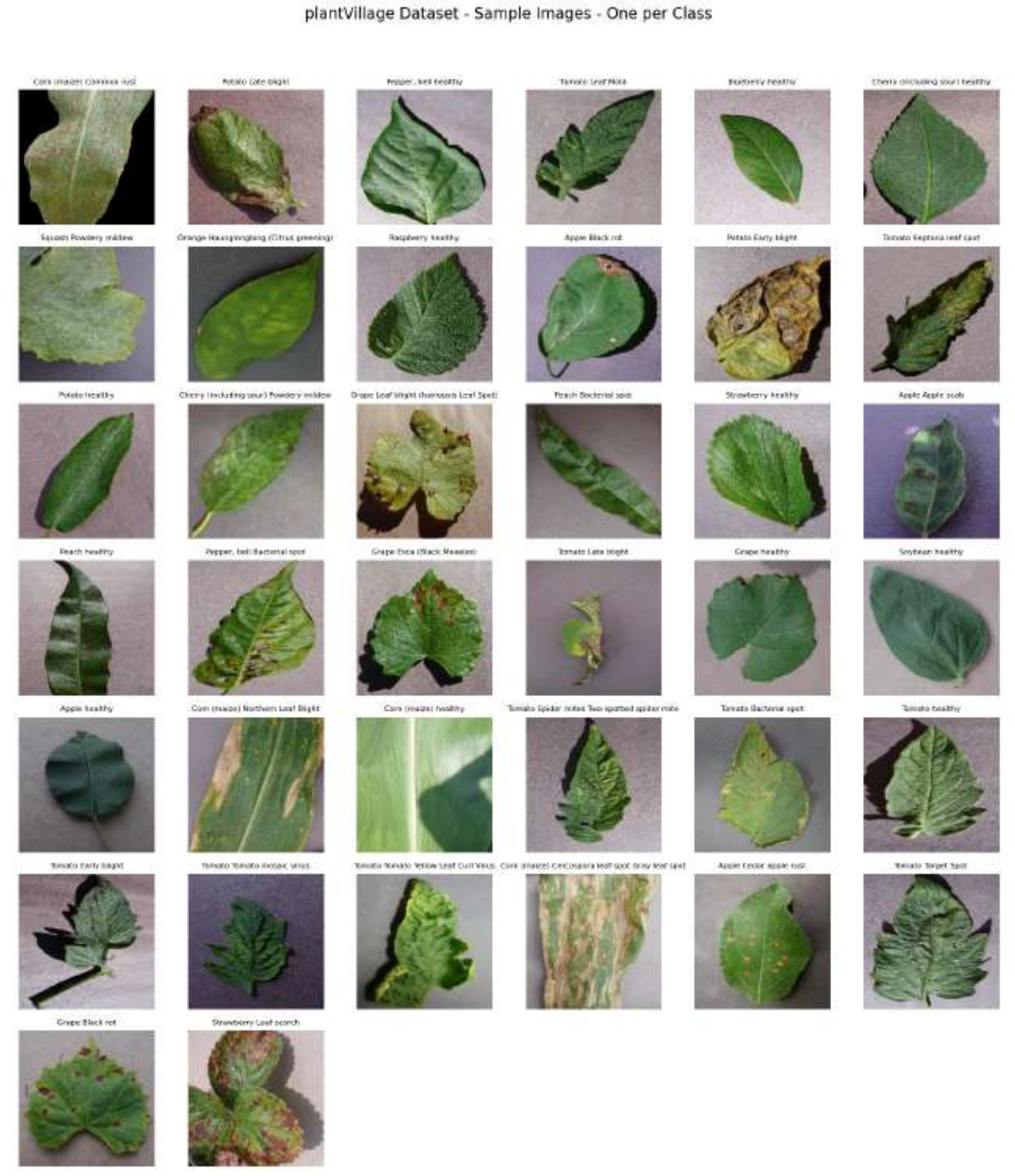

Figure A.2: A sample image of each class present in the Plant Village Dataset.

### A.1.2 Tomato Village

Table A.2: Information about the Tomato Village Dataset.

| Plants | Tomato |
|---|---|
| Diseases | 1. Healthy<br>2. Leaf Miner<br>3. Pottasium Deficiency<br>4. Spotted with Virus<br>5. Magnesium Deficiency<br>6. Early Blight<br>7. Late Blight<br>8. Nitrogen Deficiency |
| Number of Classes | 8 |
| Number of Images | 4,526 |

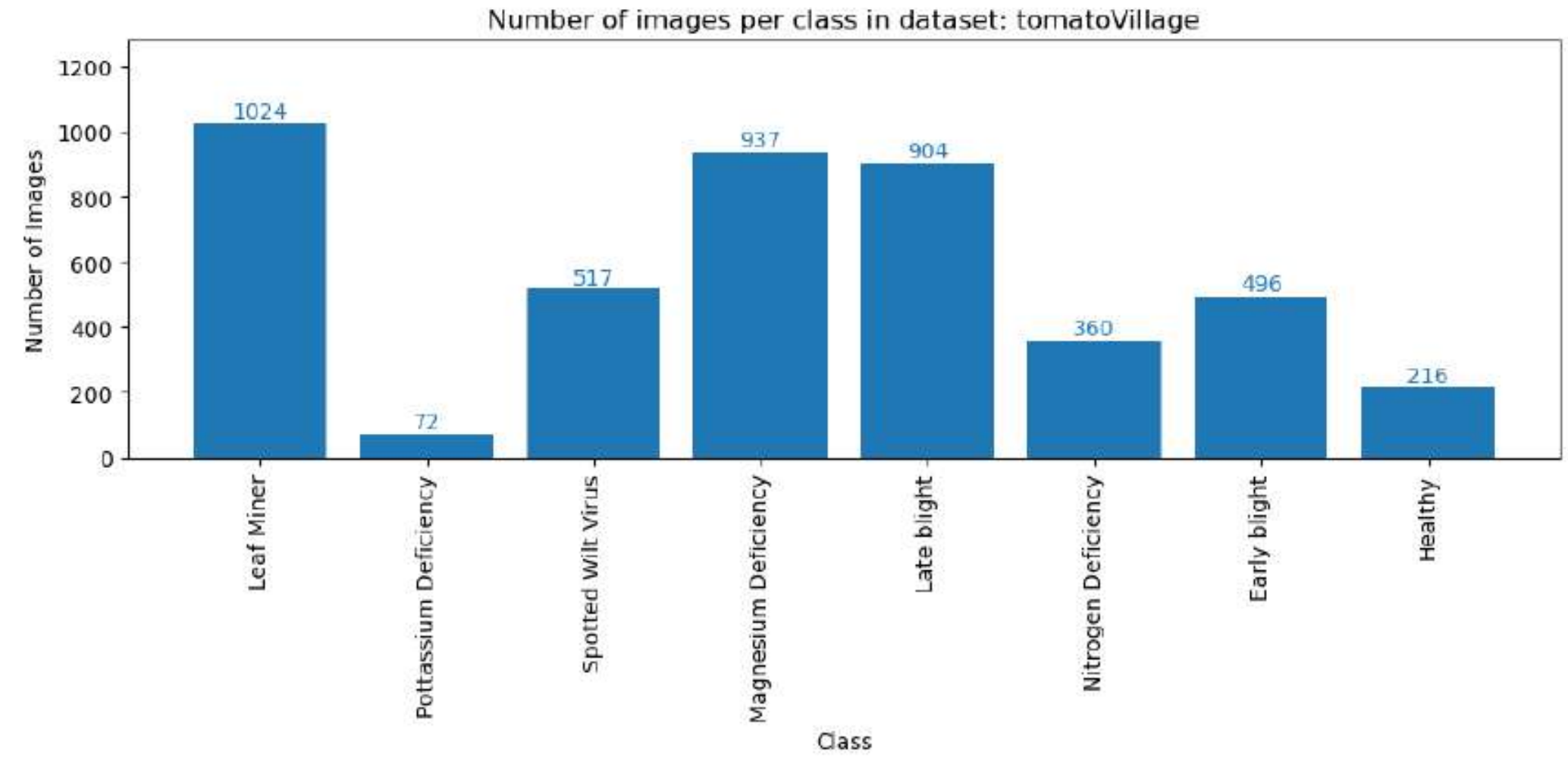

Figure A.3: Class Distribution of the Tomato Village Dataset.

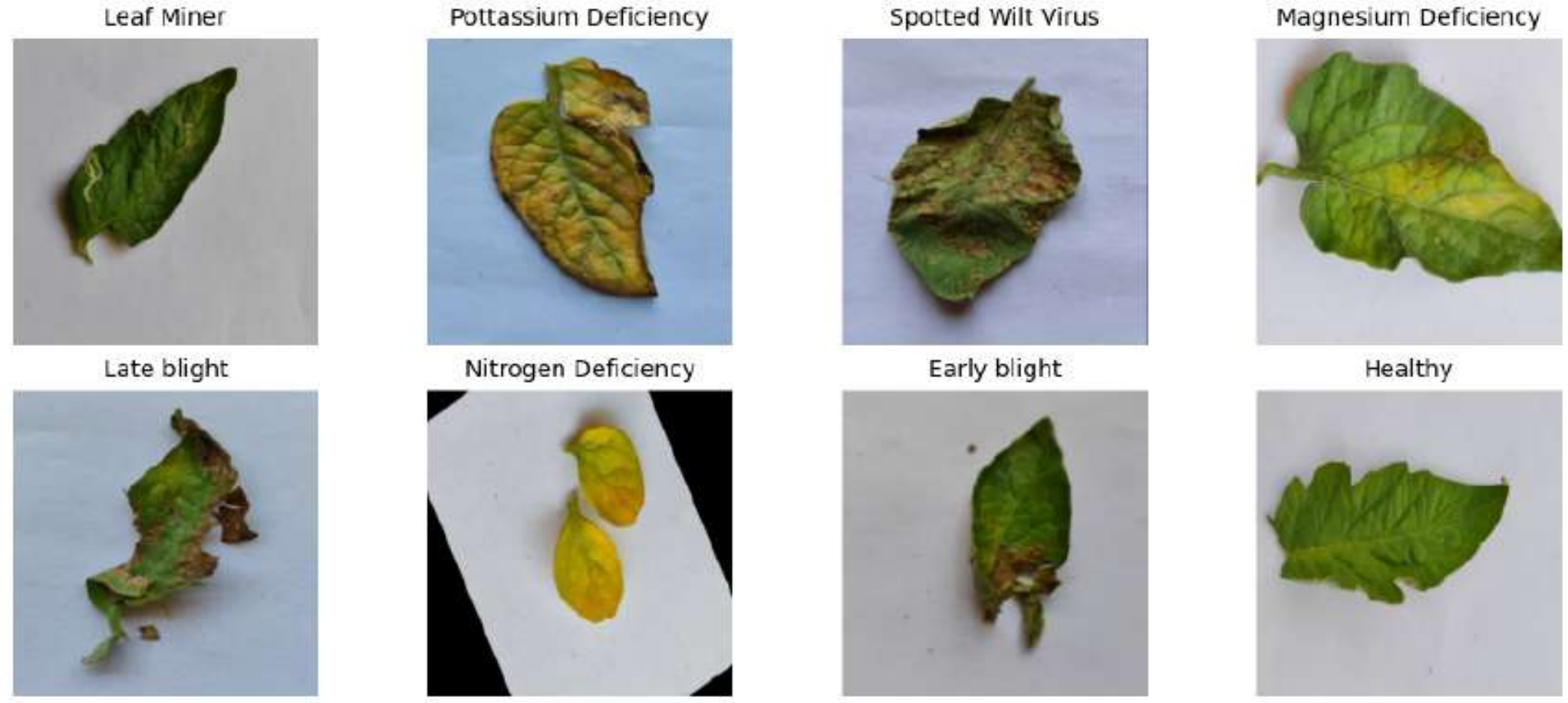

Figure A.4: A sample image of each class present in the Tomato Village Dataset.

### A.1.3 Potato Disease Leaf Dataset (PLD)

Table A.3: Information about the PLD Dataset.

| Plants | Potato |
|---|---|
| Diseases | 1. Healthy<br>2. Early Blight<br>3. Late Blight |
| Number of Classes | 3 |
| Number of Images | 4,062 |

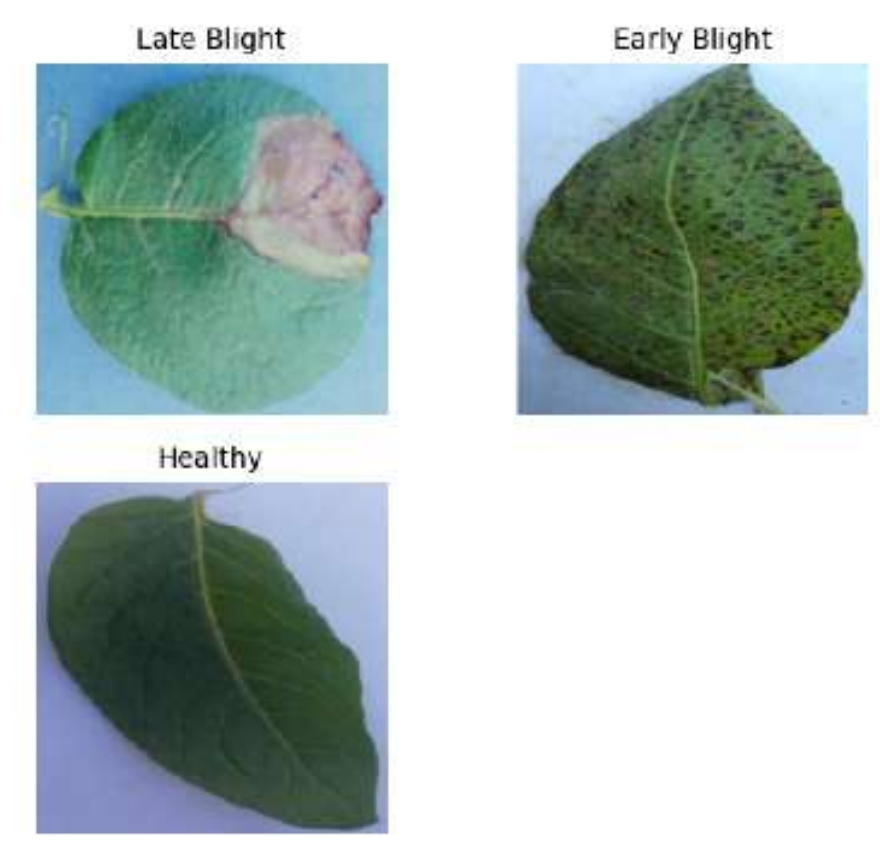

Figure A.6: A sample image of each class present in the PLD Dataset.

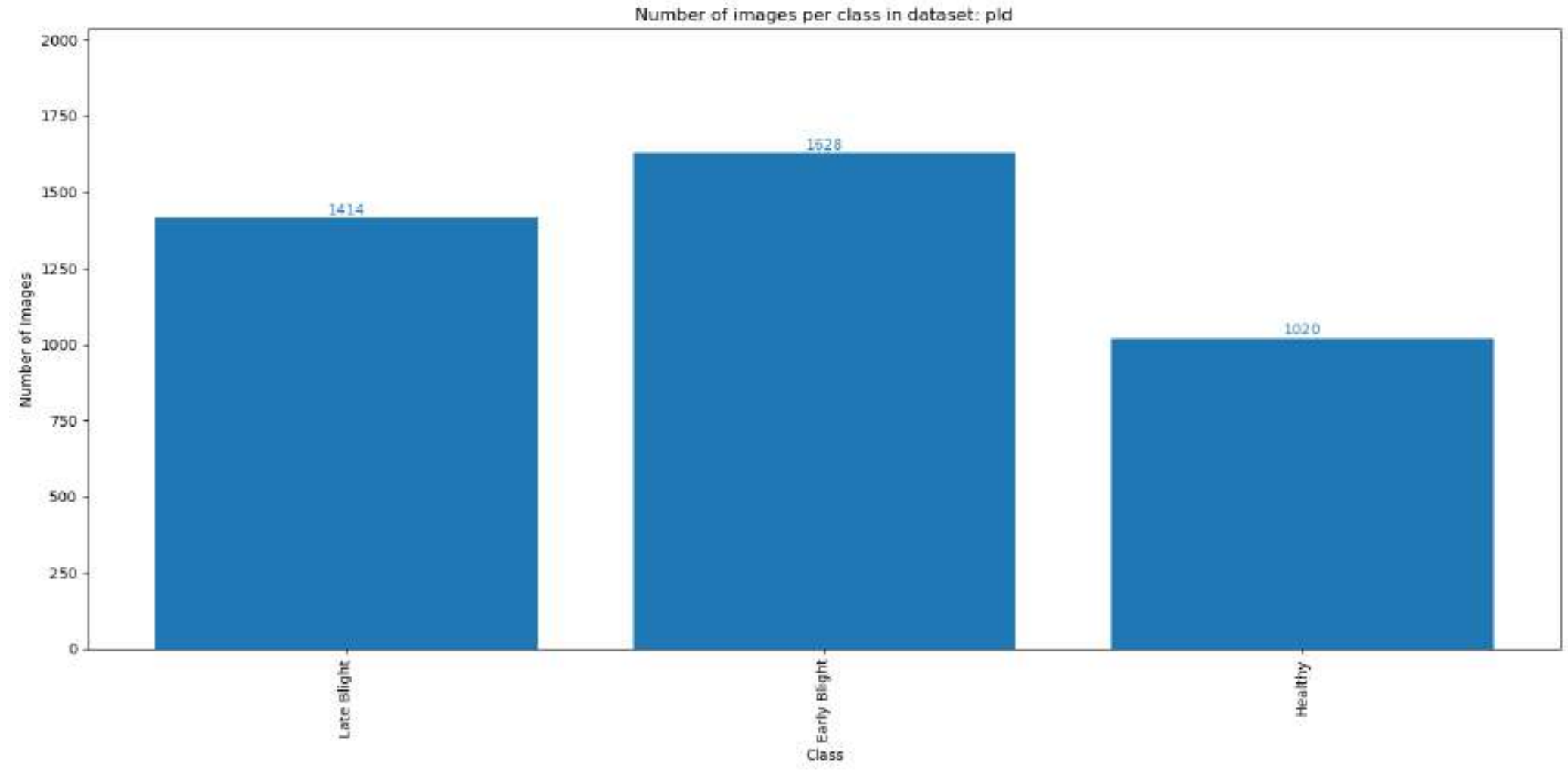

Figure A.5: Class Distribution of the PLD Dataset.

### A.1.4 Tea Sickness Dataset

Table A.4: Information about the Tea Leaf Dataset.

| Plants | Tea |
|---|---|
| Diseases | 1. Healthy<br>2. Red Leaf Spot<br>3. Bird Eye Spot<br>4. Anthracnose<br>5. Gray Light<br>6. Brown Blight<br>7. Algal Leaf<br>8. White Spot |
| Number of Classes | 8 |
| Number of Images | 885 |

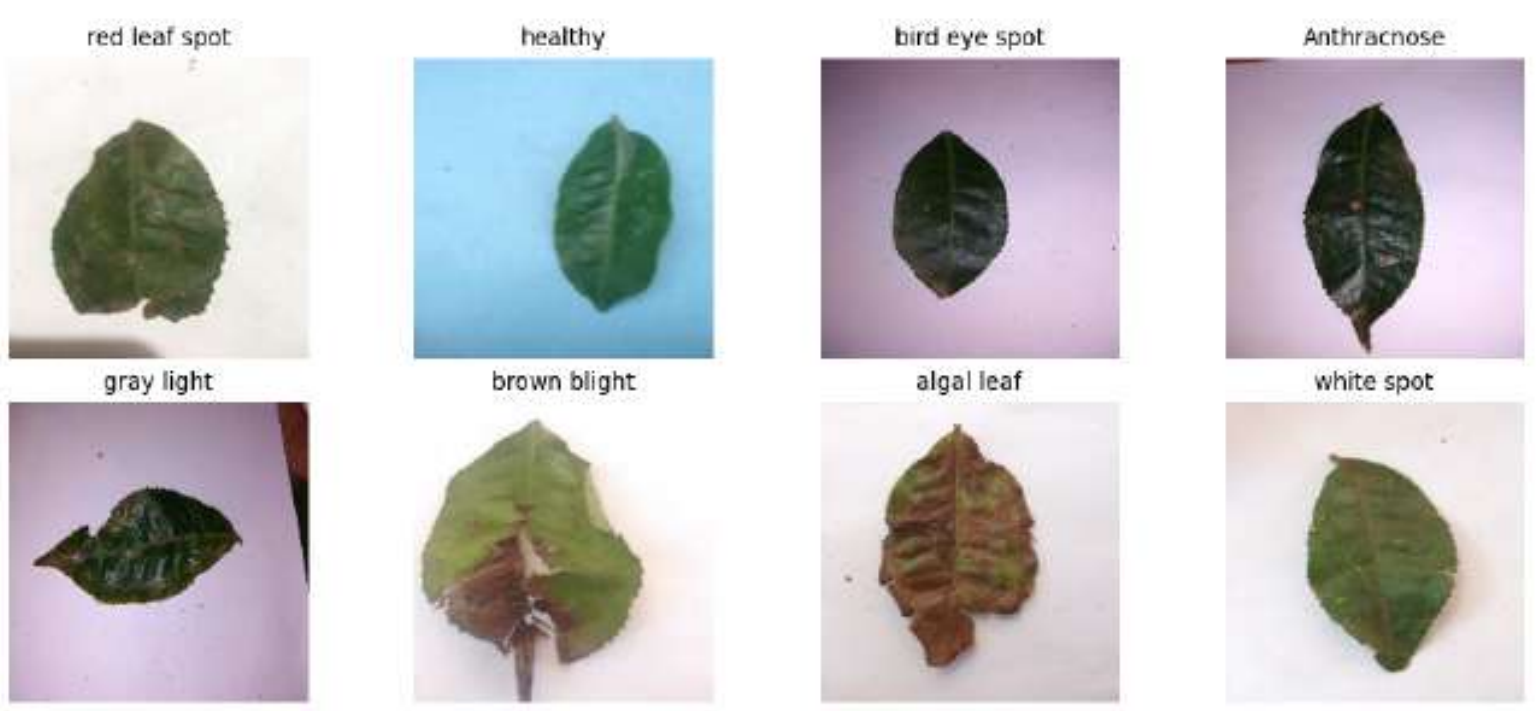

Figure A.8: A sample image of each class present in the Tea Dataset.

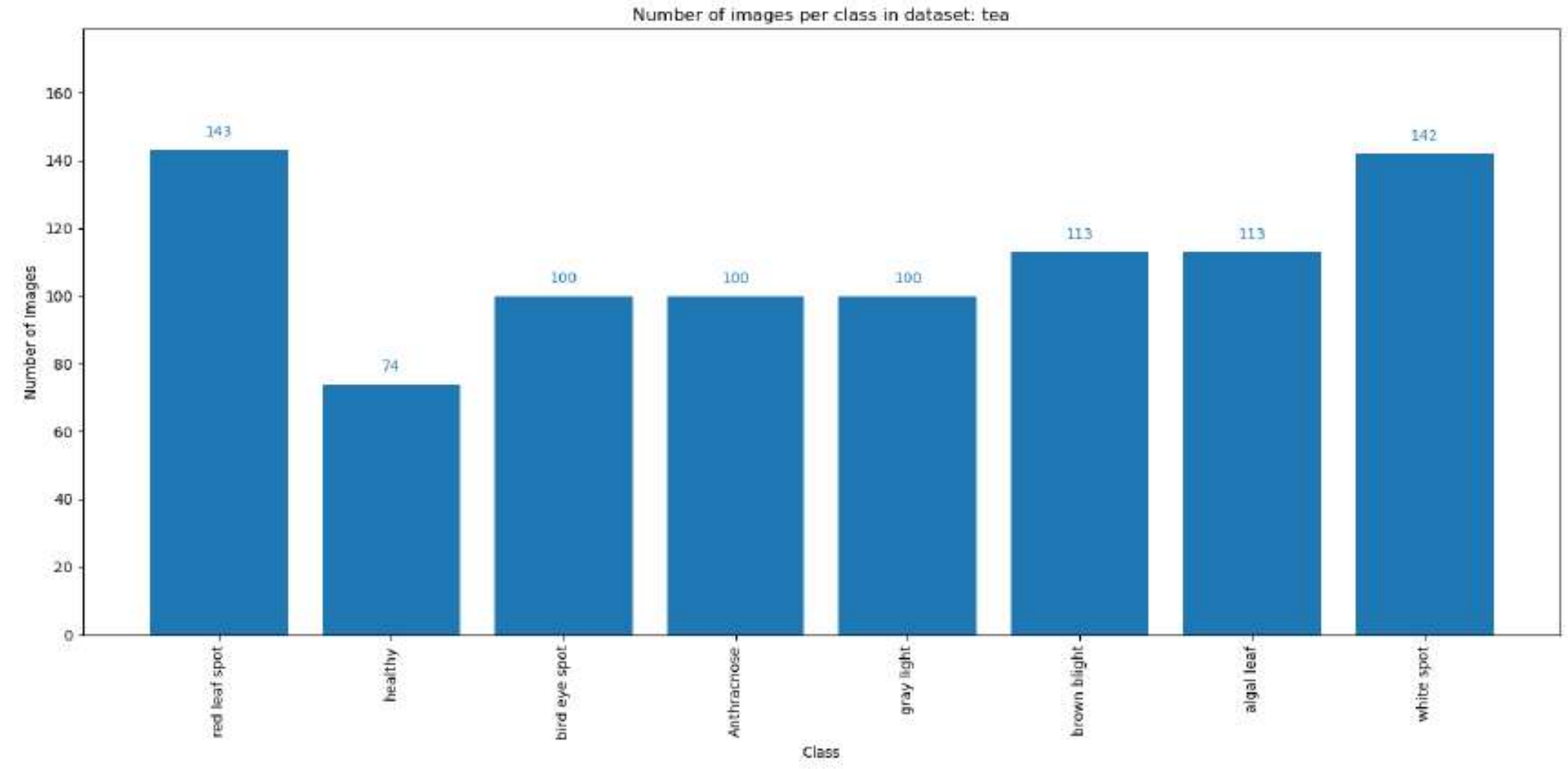

Figure A.7: Class Distribution of the Tea Dataset.

## A.2 Field Datasets

### A.2.1 Cassava Leaf Disease Classification

Table A.5: Information about the Cassava Dataset.

| Plants | Cassava |
|---|---|
| Diseases | 1. Healthy<br>2. Cassava Bacterial Blight<br>3. Cassava Mosaic Disease<br>4. Cassava Green Mottle<br>5. Cassava Brown Streak Disease |
| Number of Classes | 5 |
| Number of Images | 21,397 |

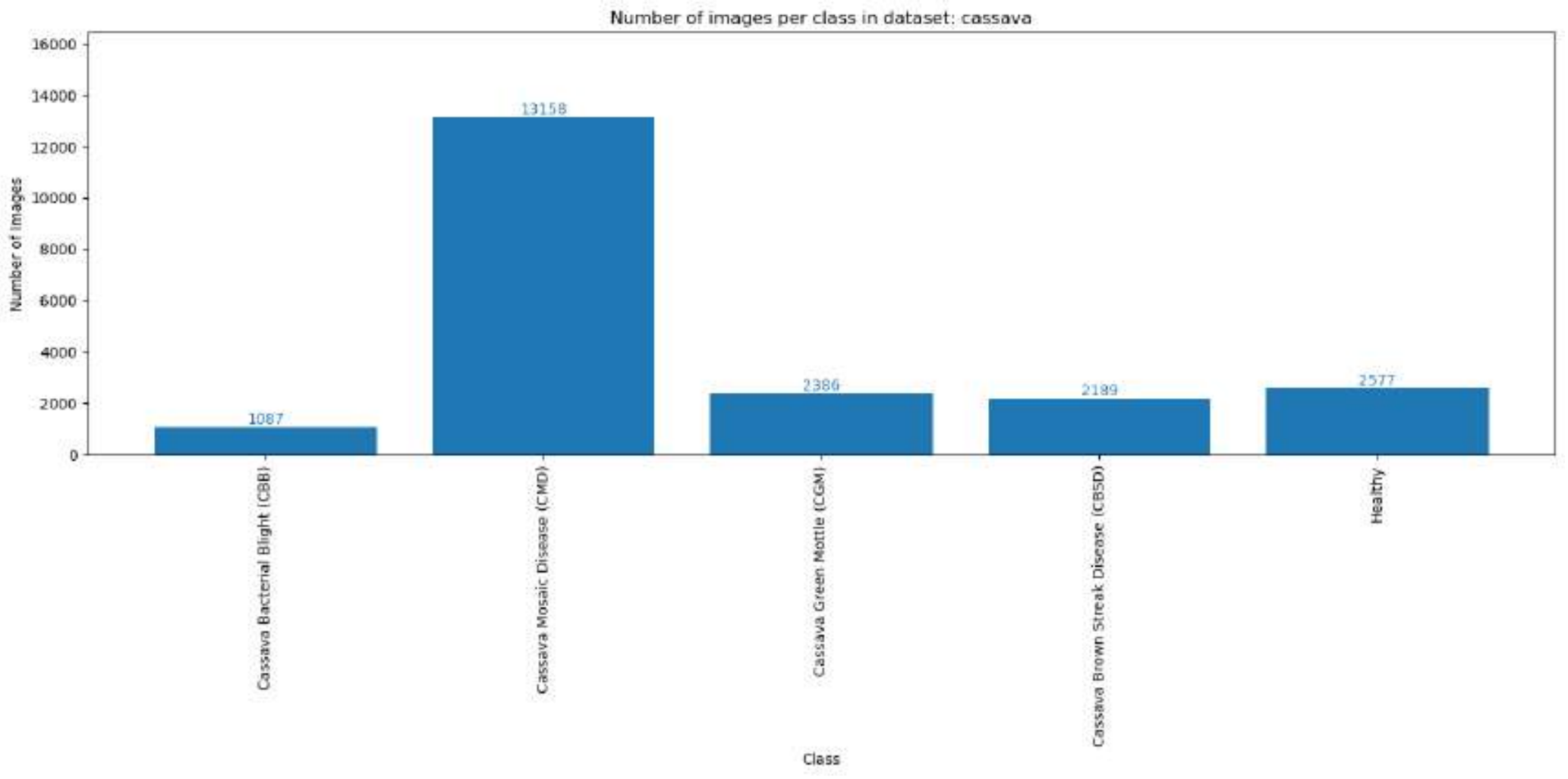

Figure A.9: Class Distribution of the Cassava Dataset.

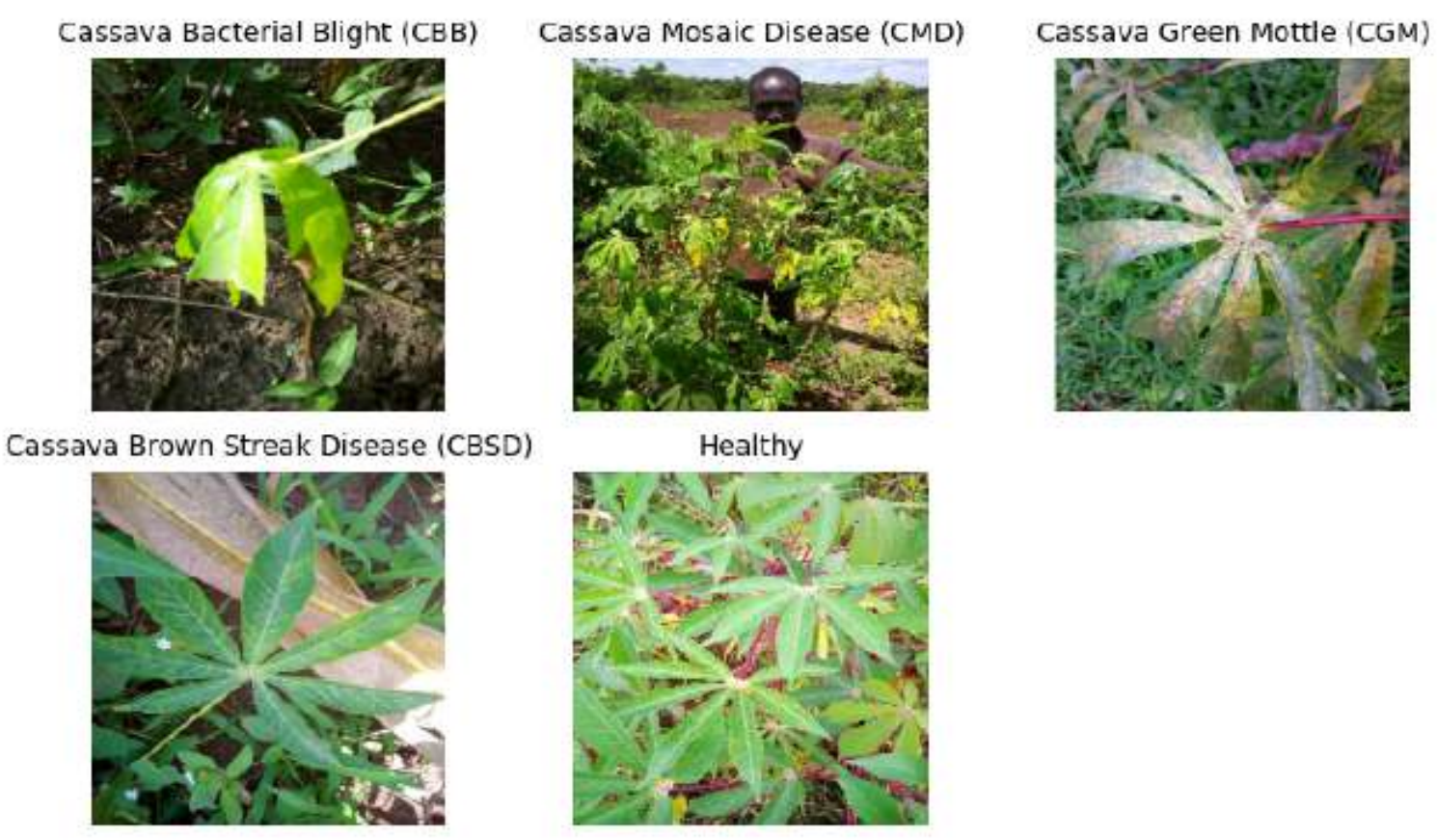

Figure A.10: A sample image of each class present in the Cassava Dataset.

### A.2.2 CD&S Dataset

Table A.6: Information about the CD&S Dataset.

| Plants | Corn |
|---|---|
| Diseases | 1. Northern Leaf Blight<br>2. Gray Leaf Spot<br>3. Northern Leaf Spot |
| Number of Classes | 3 |
| Number of Images | 1,571 |

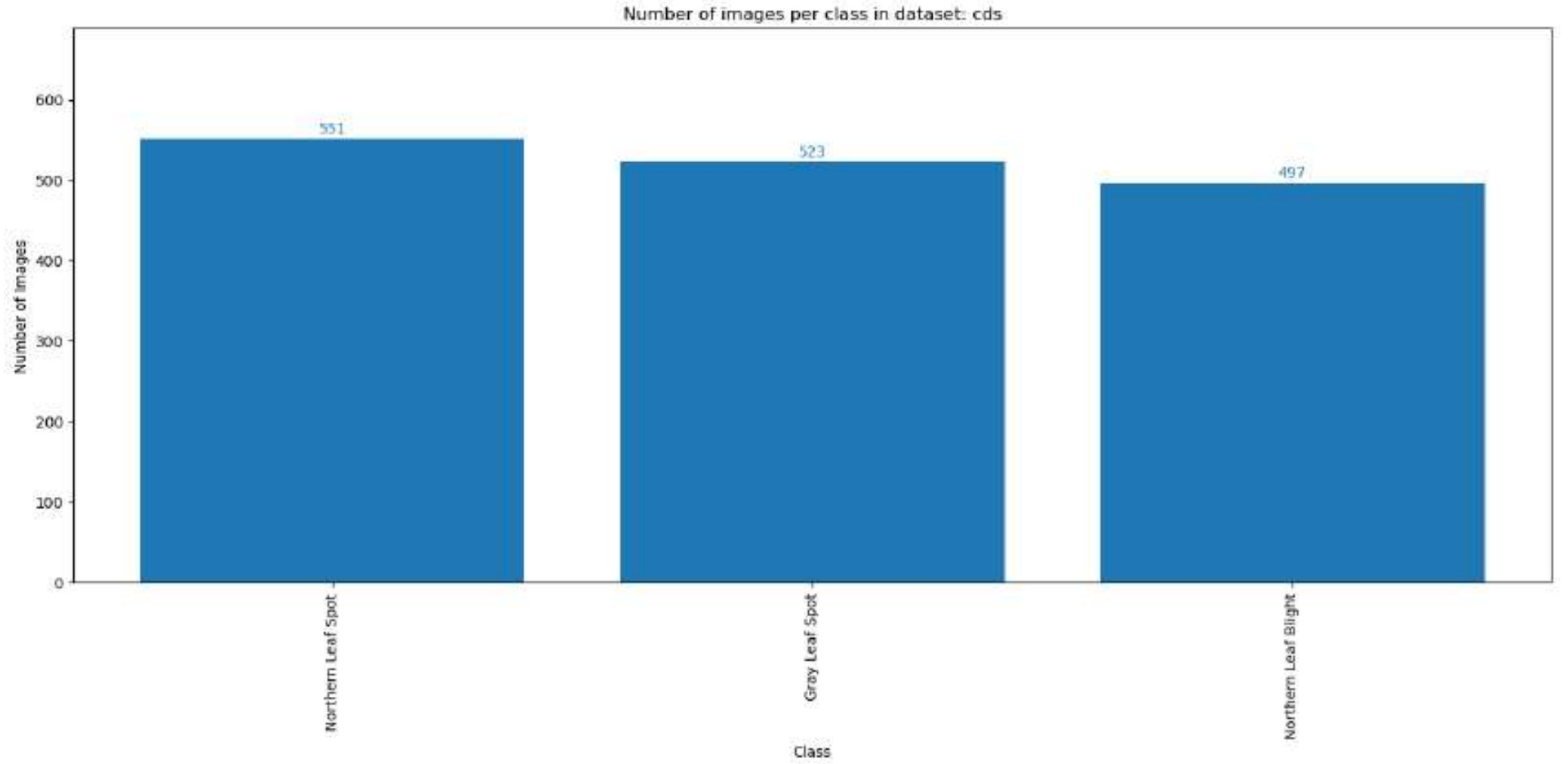

Figure A.11: Class Distribution of the CD&S Dataset.

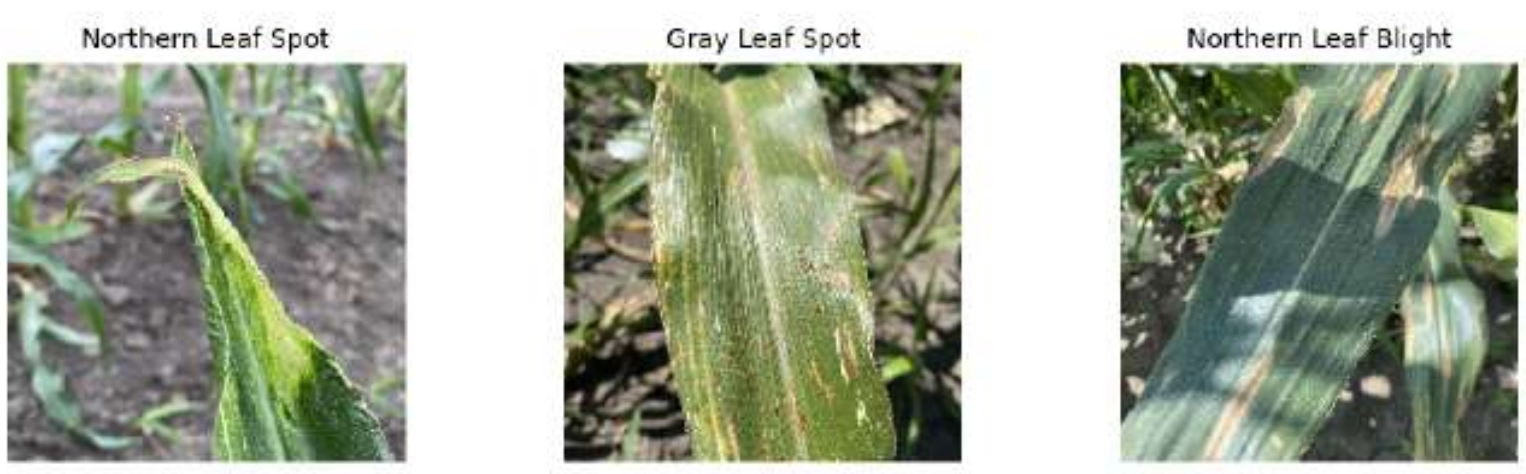

Figure A.12: A sample image of each class present in the CD&S Dataset.

### A.2.3 Cucumber Disease Recognition Dataset

Table A.7: Information about the Cucumber Dataset.

| Plants | Cucumber |
|---|---|
| Diseases | 1. Healthy<br>2. Downy Mildew<br>3. Gummy Stem Blight<br>4. Anthracnose<br>5. Bacterial Wilt |
| Number of Classes | 5 |
| Number of Images | 800 |

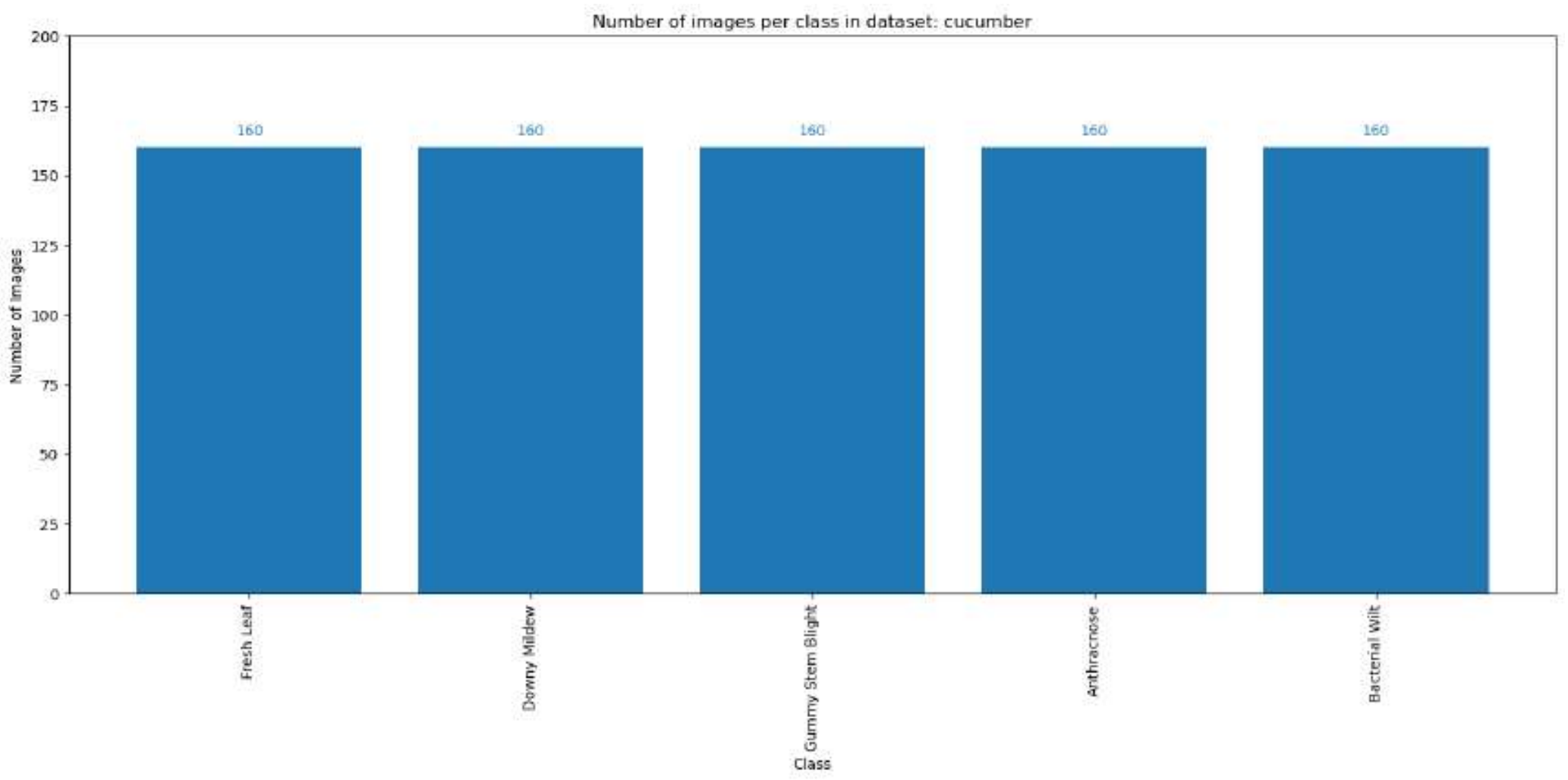

Figure A.13: Class Distribution of the Cucumber Dataset.

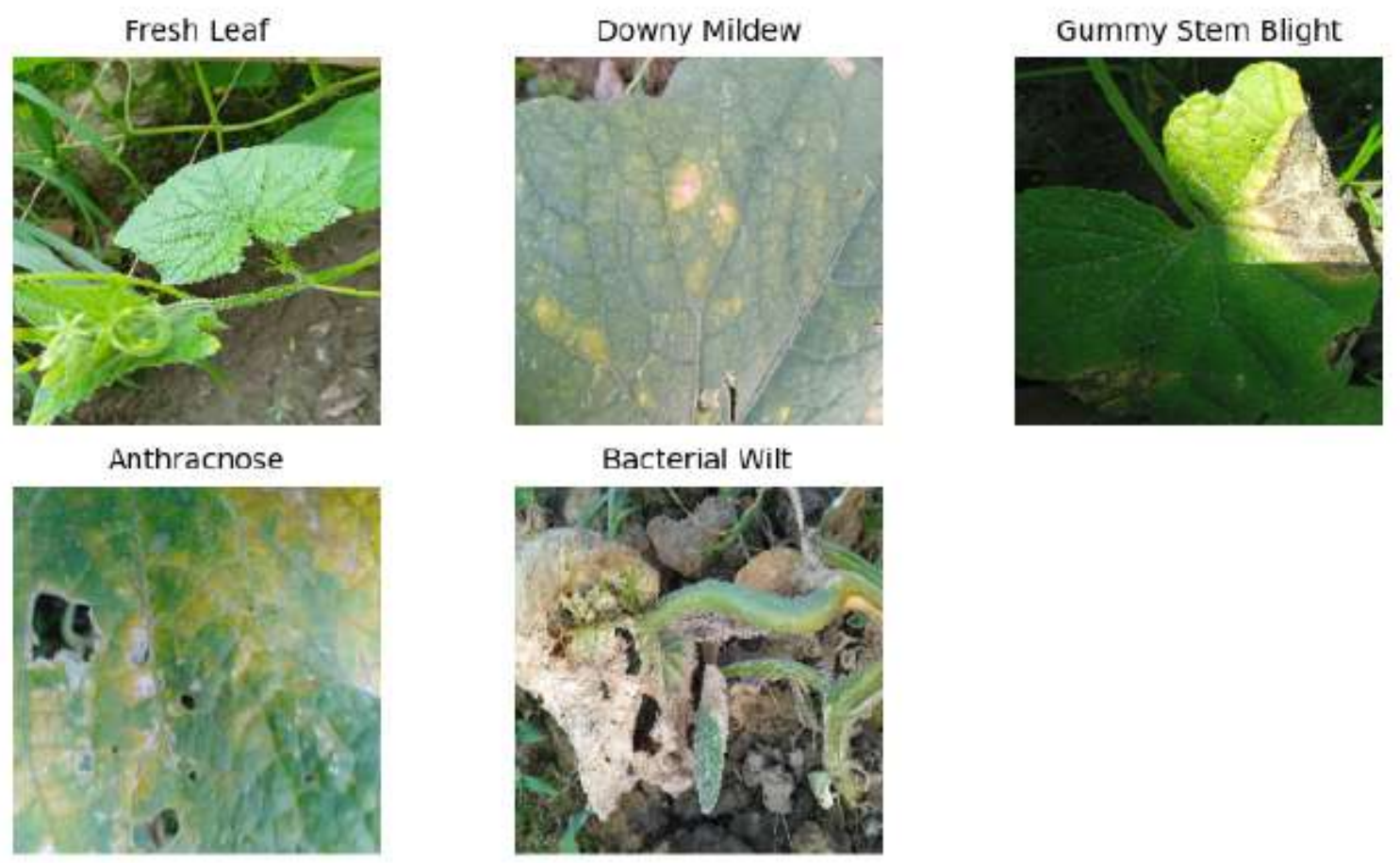

Figure A.14: A sample image of each class present in the Cucumber Dataset.

### A.2.4 DiaMOS Plant

Table A.8: Information about the diaMOS Dataset.

| Plants | Pear |
|---|---|
| Diseases | 1. Healthy<br>2. Slug<br>3. Spot<br>4. Curl |
| Number of Classes | 4 |
| Number of Images | 3,006 |

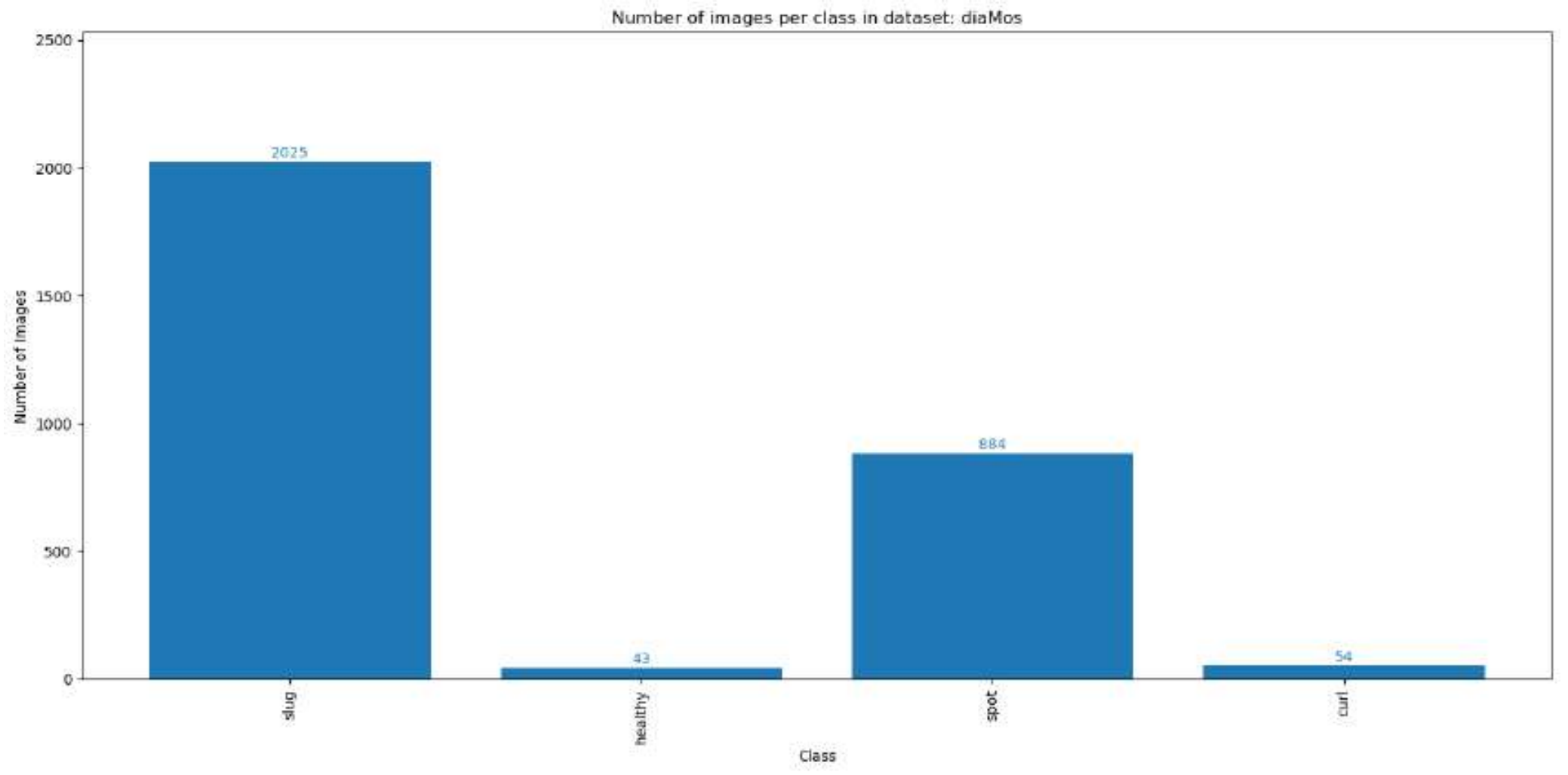

Figure A.15: Class Distribution of the diaMOS Dataset.

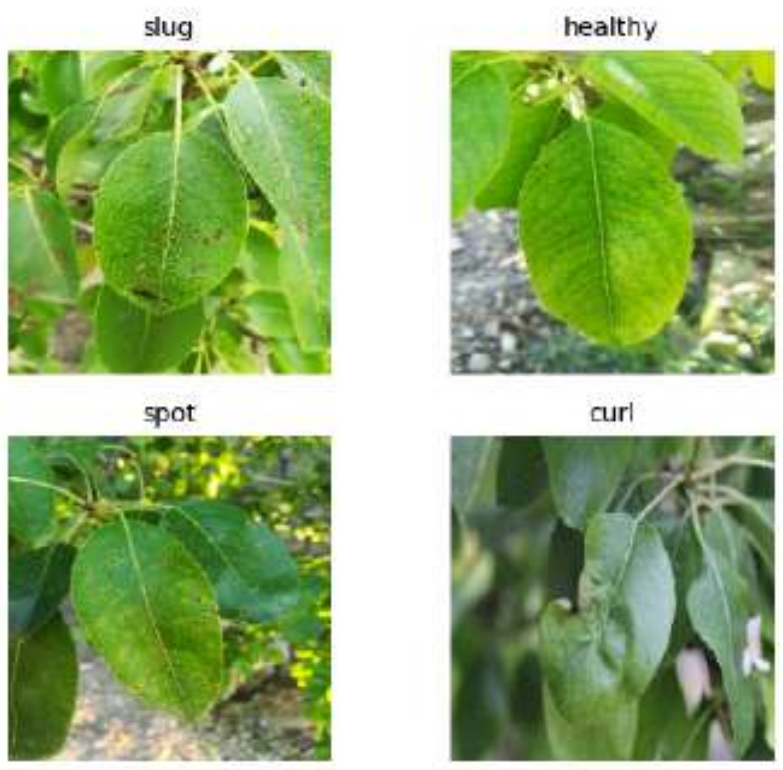

Figure A.16: A sample image of each class present in the diaMOS Dataset.

### A.2.5 Paddy Doctor

Table A.9: Information about the Paddy Doctor Dataset.

| Plants | Rice |
|---|---|
| Diseases | 1. Healthy<br>2. Bacterial Leaf Streak<br>3. Tungro<br>4. Brown Spot<br>5. Dead Heart<br>6. Hispa<br>7. Downy Mildew<br>8. Bacterial Panicle Blight<br>9. Blast<br>10. Bacterial Leaf Blight |
| Number of Classes | 10 |
| Number of Images | 10,407 |

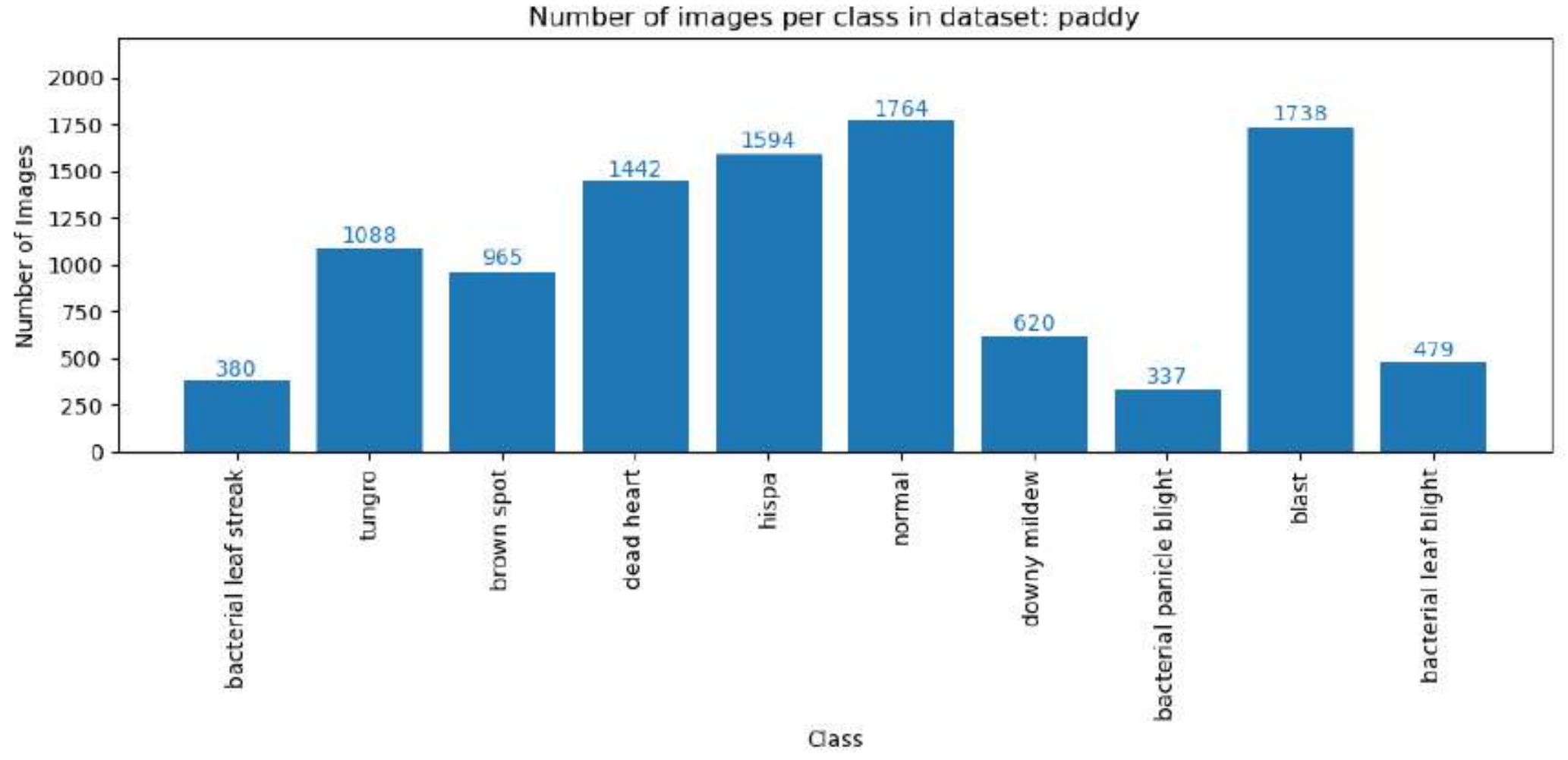

Figure A.17: Class Distribution of the Paddy Doctor Dataset.

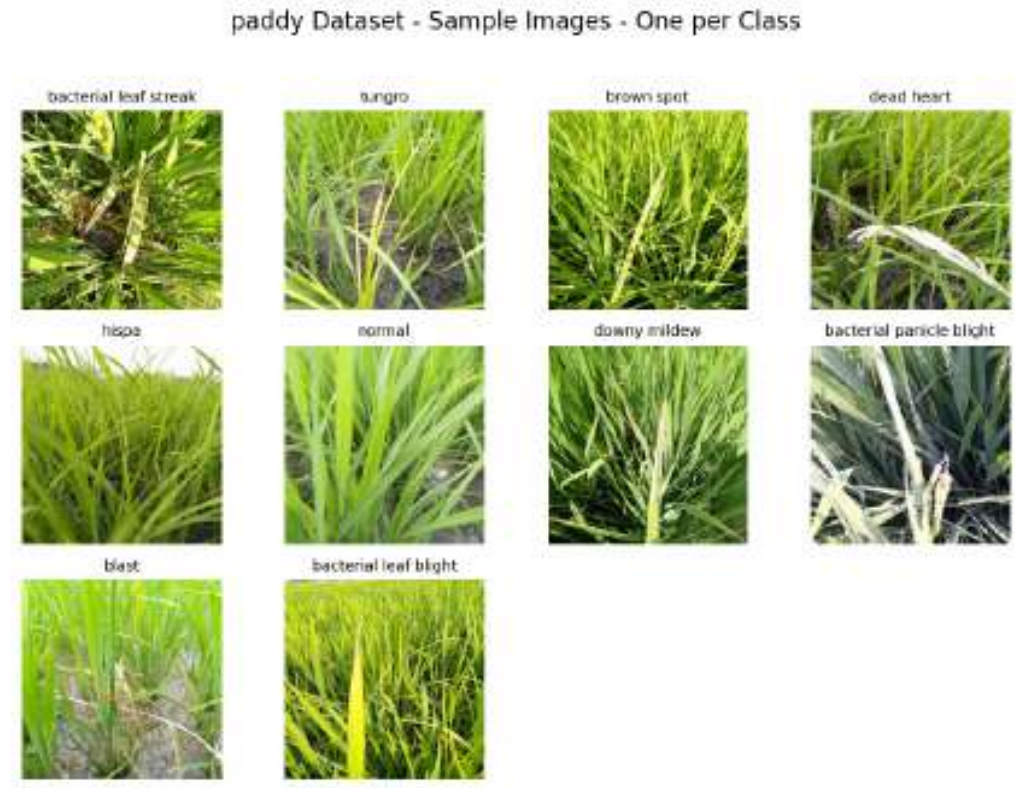

Figure A.18: A sample image of each class present in the Paddy Doctor Dataset.

### A.2.6 Plant Pathology 2020 - FGVC7

Table A.10: Information about the FGVC7 Dataset.

| | |
|---|---|
| Plants | Apple |
| Diseases | 1. Healthy<br>2. Rust<br>3. Scab<br>4. Multi-Disease |
| Number of Classes | 4 |
| Number of Images | 1,821 |

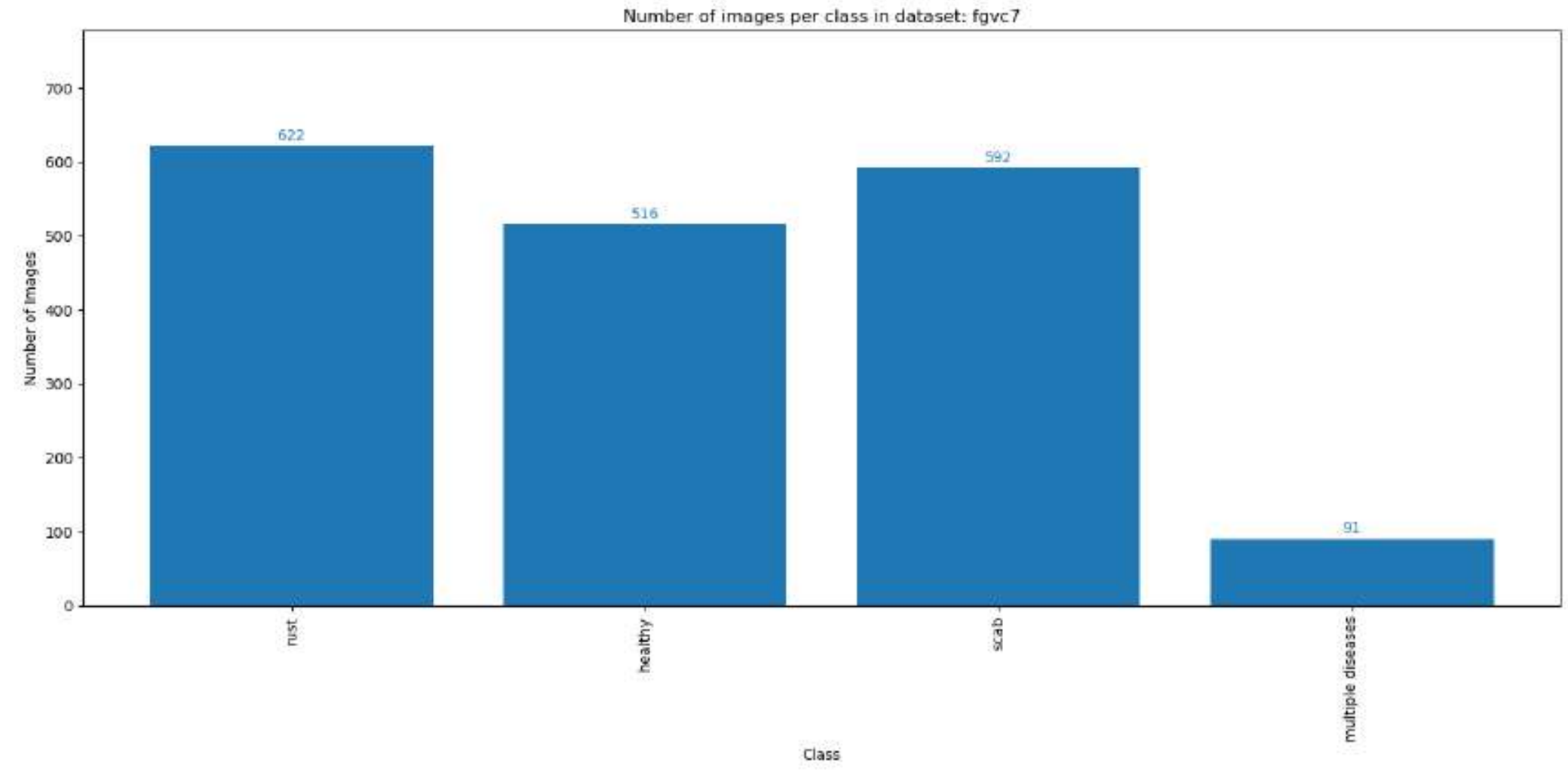

Figure A.19: Class Distribution of the FGVC7 Dataset.

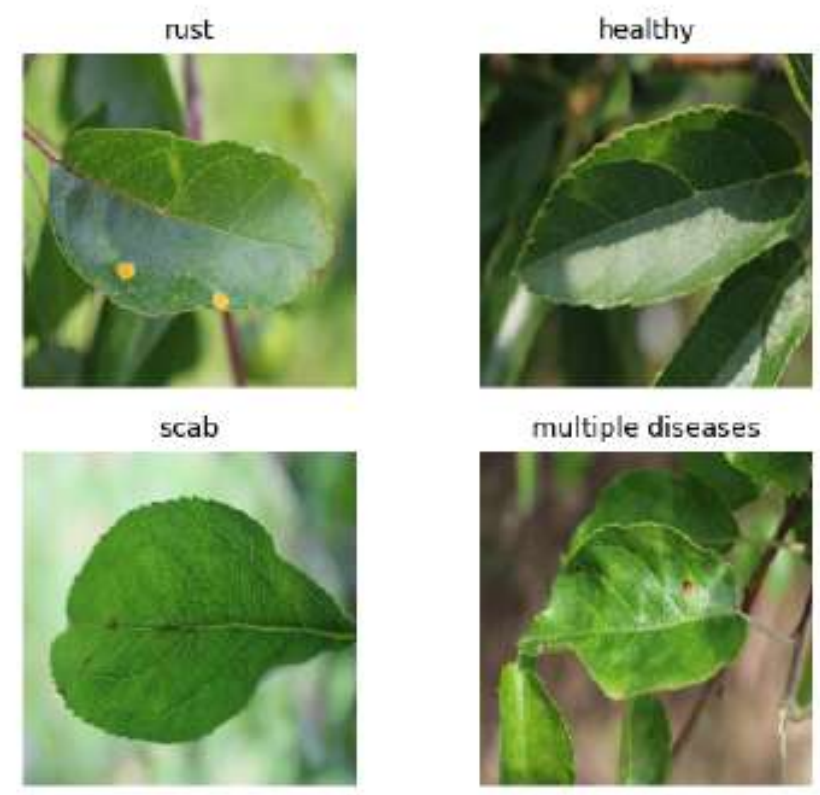

Figure A.20: A sample image of each class present in the FGVC7 Dataset.

Table A.11: Information about the FGVC8 Dataset.

| Plants | Apple |
|---|---|
| Diseases | 1. Healthy<br>2. Rust<br>3. Rust (complex)<br>4. Rust Frog Eye Leaf Spot<br>5. Scab<br>6. Scab (complex)<br>7. Scab Frog Eye Leaf Spot<br>8. Frog Eye Leaf Spot<br>9. Frog Eye Leaf Spot (complex)<br>10. Complex<br>11. Powdery Mildew<br>12. Powdery Mildew (complex) |
| Number of Classes | 12 |
| Number of Images | 18,632 |

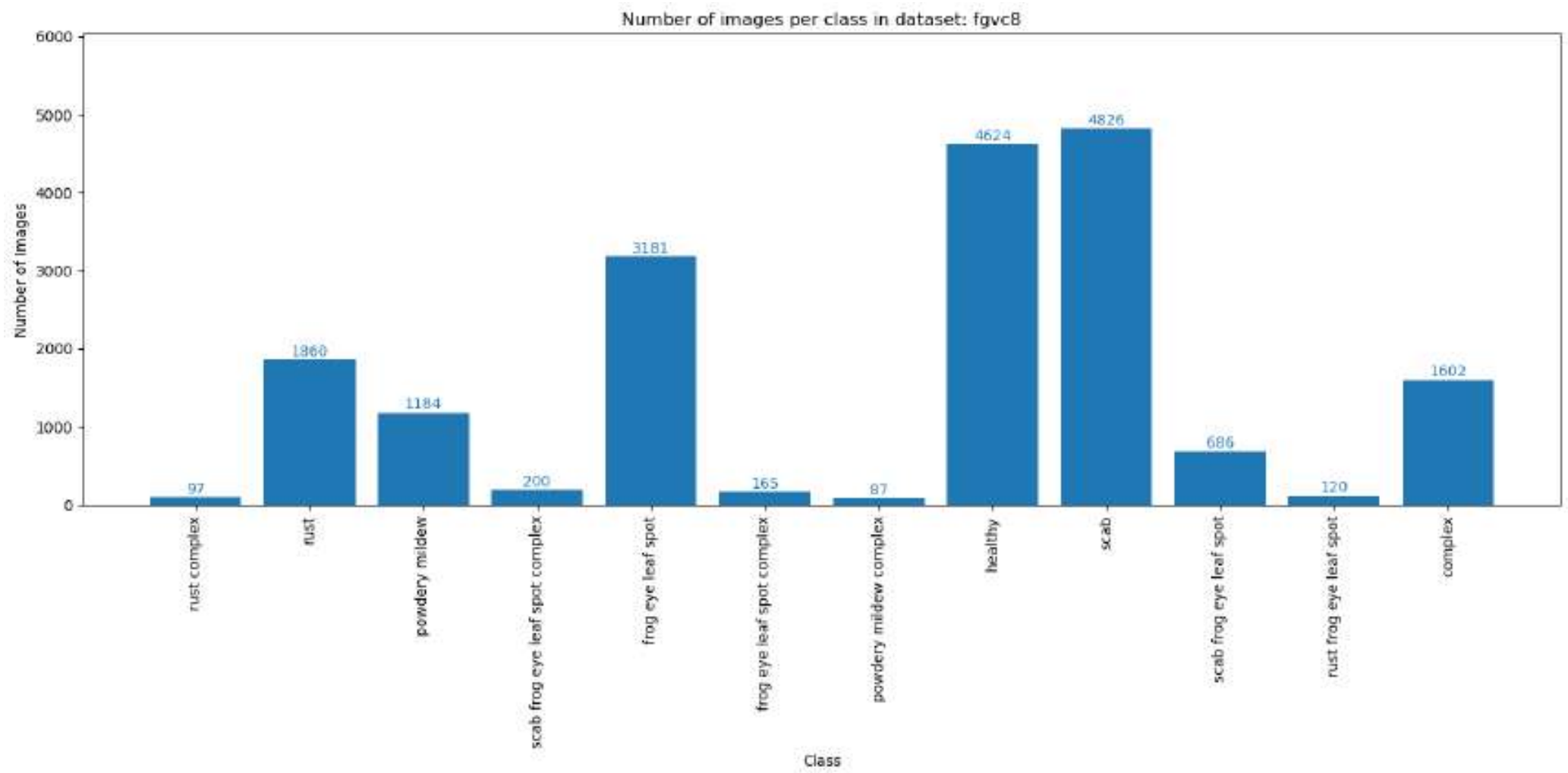

Figure A.21: Class Distribution of the FGVC8 Dataset.

### A.2.7 Plant Pathology 2021 - FGVC8

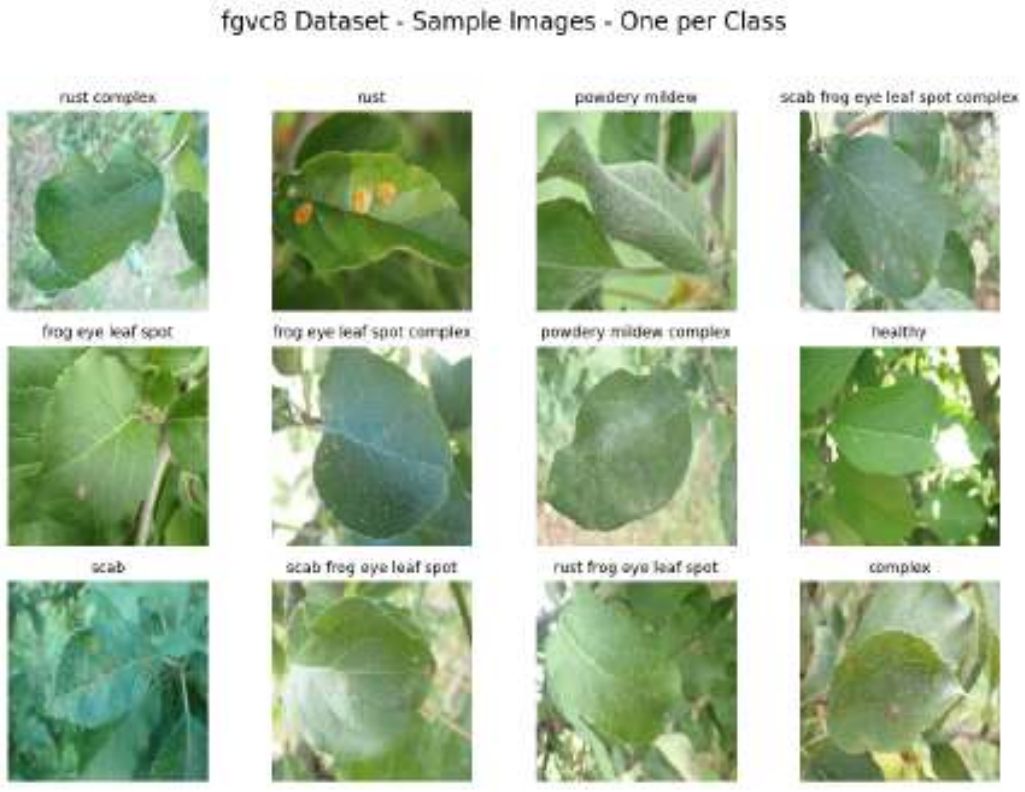

Figure A.22: A sample image of each class present in the FGVC8 Dataset.

### A.2.8 PDD271: Plant Disease Recognition Dataset

Table A.12: Information about the PDD271 Dataset.

| | |
|---|---|
| Plants | 1. Leek<br>2. Sweet Potato<br>3. Mung Bean<br>4. Radish<br>5. Ginger |
| Diseases | 1. Leek<br>• Gray Mold<br>• Hail Damage<br>2. Sweet Potato<br>• Healthy<br>• Sooty Mold<br>• Magnesium Deficiency<br>3. Mung Bean<br>• Brown Spot<br>4. Radish<br>• Black Spot<br>• Mosaic Virus<br>• Wrinkle Virus<br>5. Ginger<br>• Anthracnose |
| Number of Classes | 10 |
| Number of Images | 7,555 |

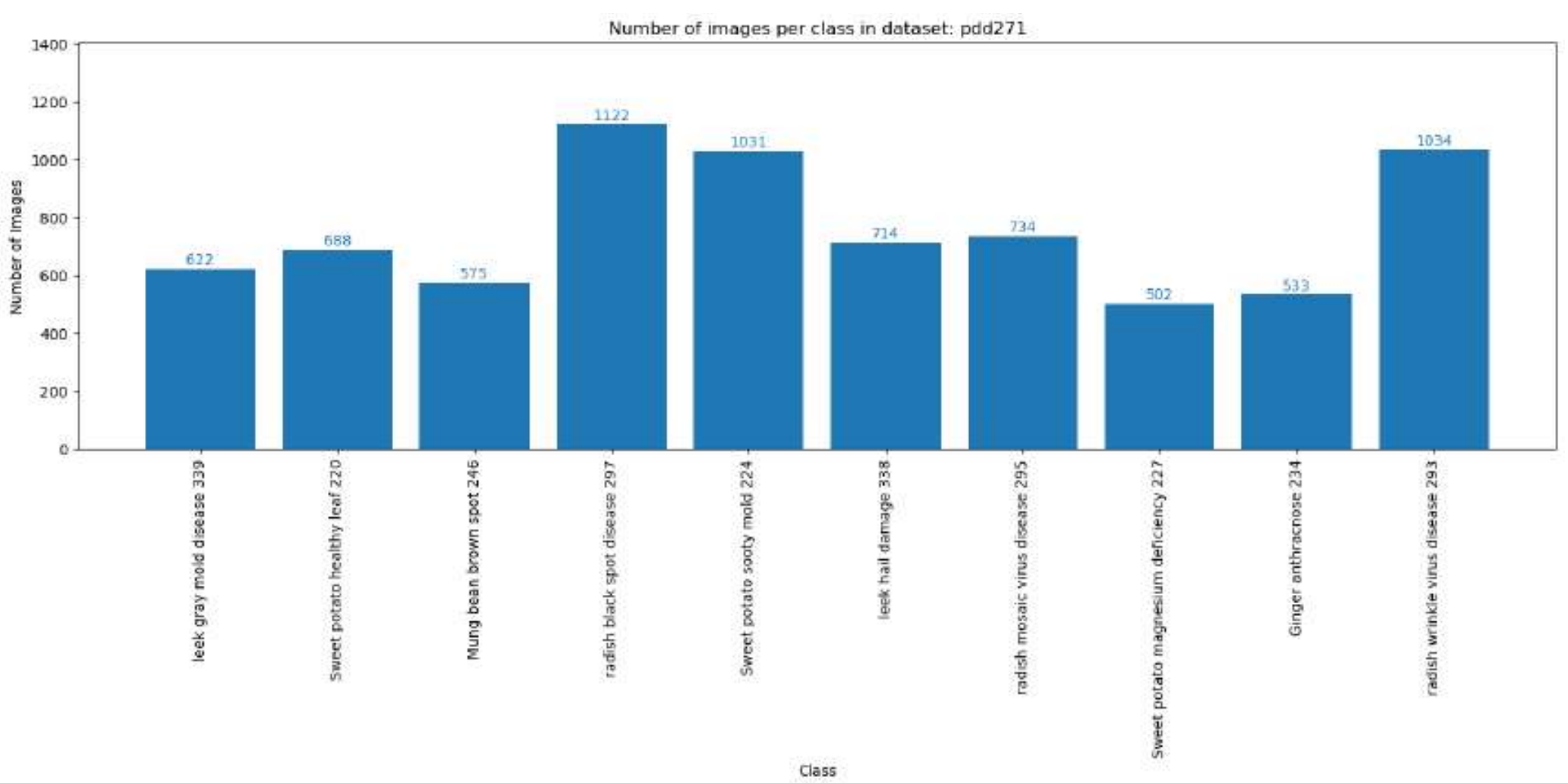

Figure A.23: Class Distribution of the PDD271 Dataset.

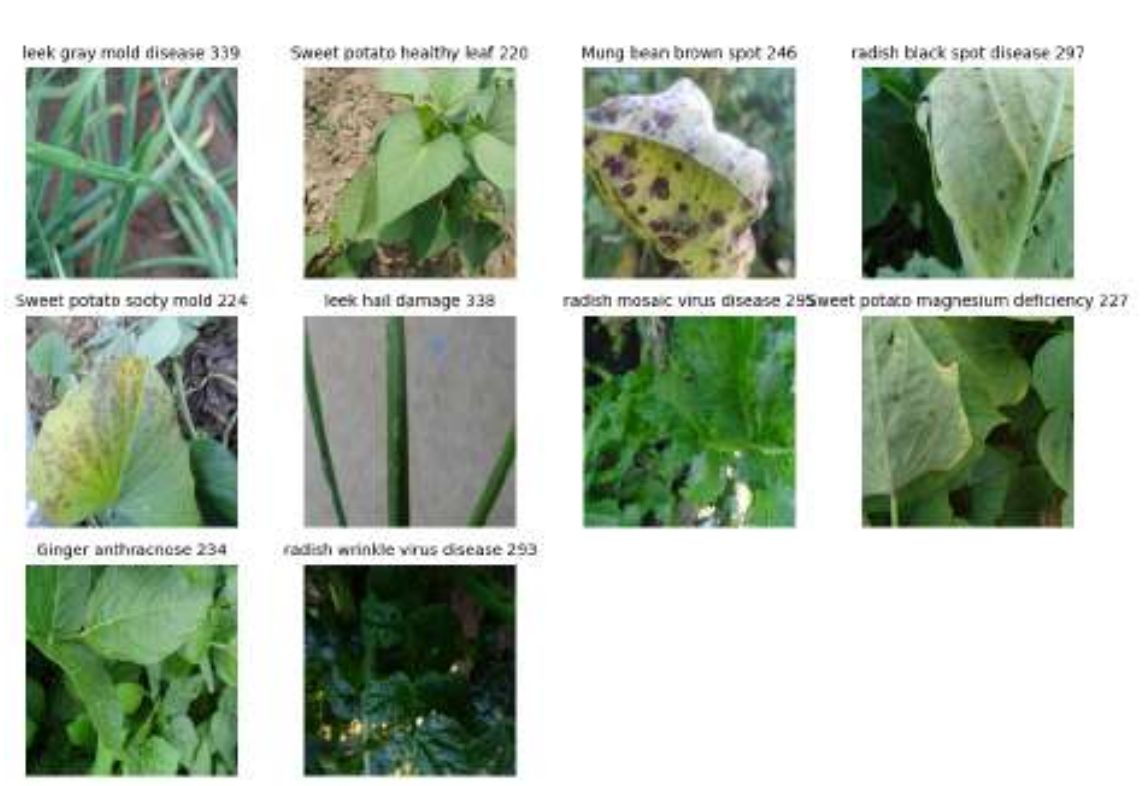

Figure A.24: A sample image of each class present in the PDD271 Dataset.

### A.2.9 Strawberry Disease Detection Dataset

Table A.13: Information about the Strawberry Dataset.

| Plants | Strawberry |
|---|---|
| Diseases | 1. Leaf Smut<br>2. Brown Spot<br>3. Powdery Mildew Bacterial Blight |
| Number of Classes | 3 |
| Number of Images | 1583 |

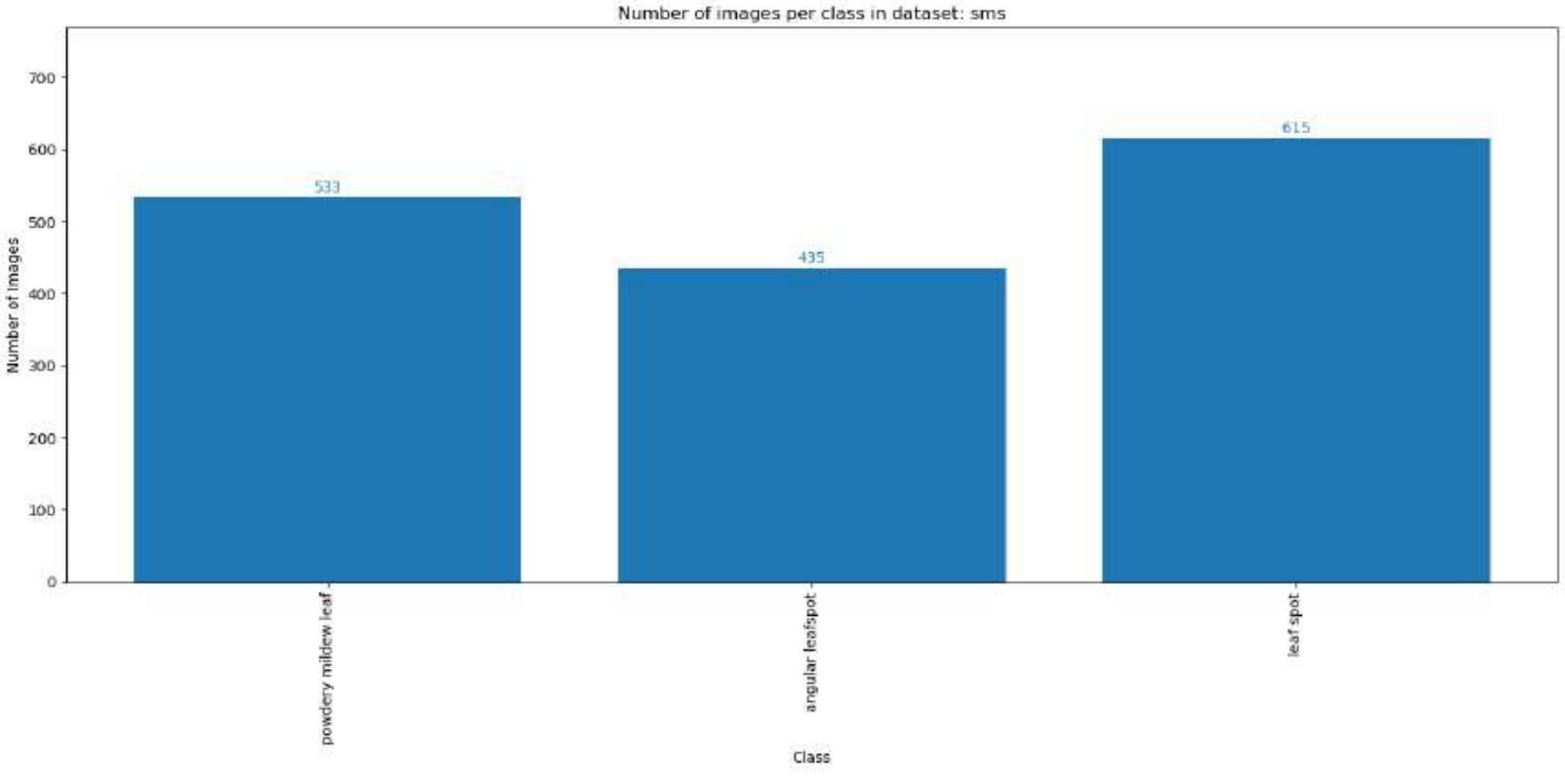

Figure A.25: Class Distribution of the Strawberry Dataset.

sms Dataset - Sample Images - One per Class

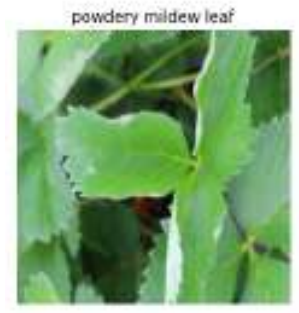

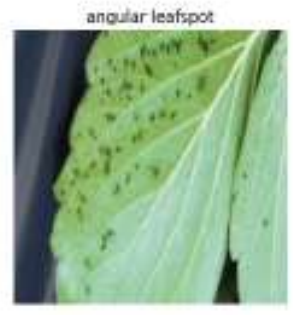

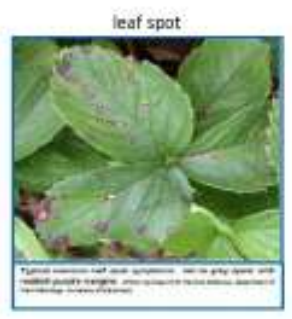

Figure A.26: A sample image of each class present in the Strawberry Dataset.

## A.3 Hybrid Datasets

### A.3.1 plantDoc

Table A.14: Information about the plantDoc Dataset.

| | |
|---|---|
| Plants | 1. Squash<br>2. Bell Pepper<br>3. Tomato<br>4. Strawberry<br>5. Grape<br>6. Blueberry<br>7. Corn<br>8. Cherry<br>9. Apple<br>10. Potato<br>11. Soybean<br>12. Raspberry<br>13. Peach |
| Diseases | 1. **Squash**<br>• Powdery Mildew<br>2. **Bell Pepper**<br>• Healthy<br>• Leaf Spot<br>3. **Tomato**<br>• Healthy<br>• Septoria Leaf Spot<br>• Yellow Virus<br>• Bacterial Spot<br>• Mosaic Virus<br>• Two Spotted Spider Mites<br>• Mold<br>• Early Blight<br>• Late Blight<br>4. **Strawberry**<br>• Healthy<br>5. **Grape**<br>• Healthy<br>• Black Rot<br>6. **Potato**<br>• Late Blight<br>• Early Blight<br>7. **Blueberry**<br>• Healthy<br>8. **Corn**<br>• Rust Leaf<br>• Gray Leaf Spot<br>• Leaf Blight<br>9. **Cherry**<br>• Cherry leaf Healthy<br>10. **Apple**<br>• Healthy<br>• Scab<br>• Rust Leaf<br>11. **Soybean**<br>• Healthy<br>12. **Raspberry**<br>• Healthy<br>13. **Peach**<br>• Healthy |
| Number of Classes | 28 |
| Number of Images | 2,577 |

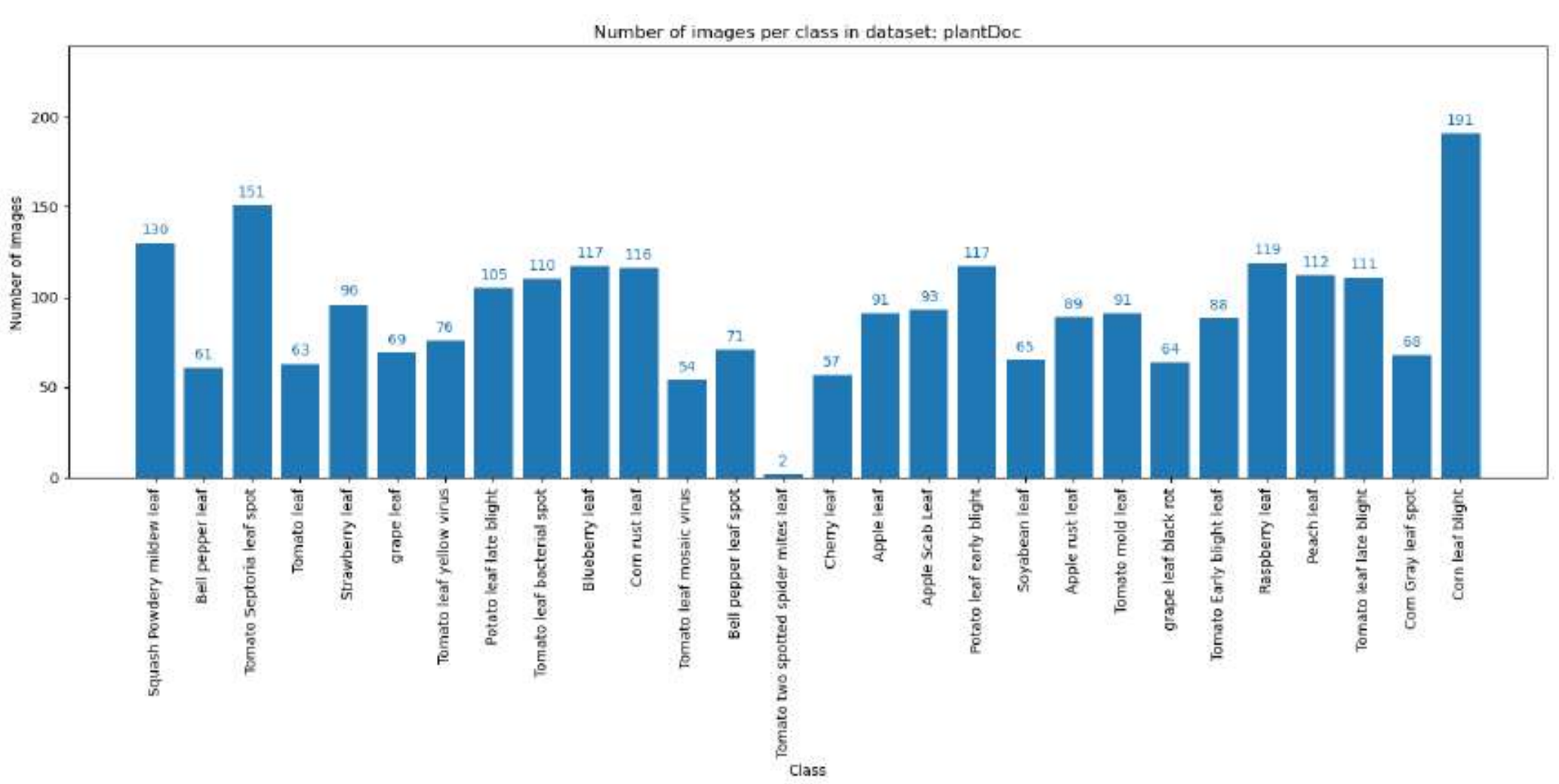

Figure A.27: Class Distribution of the plantDoc Dataset.

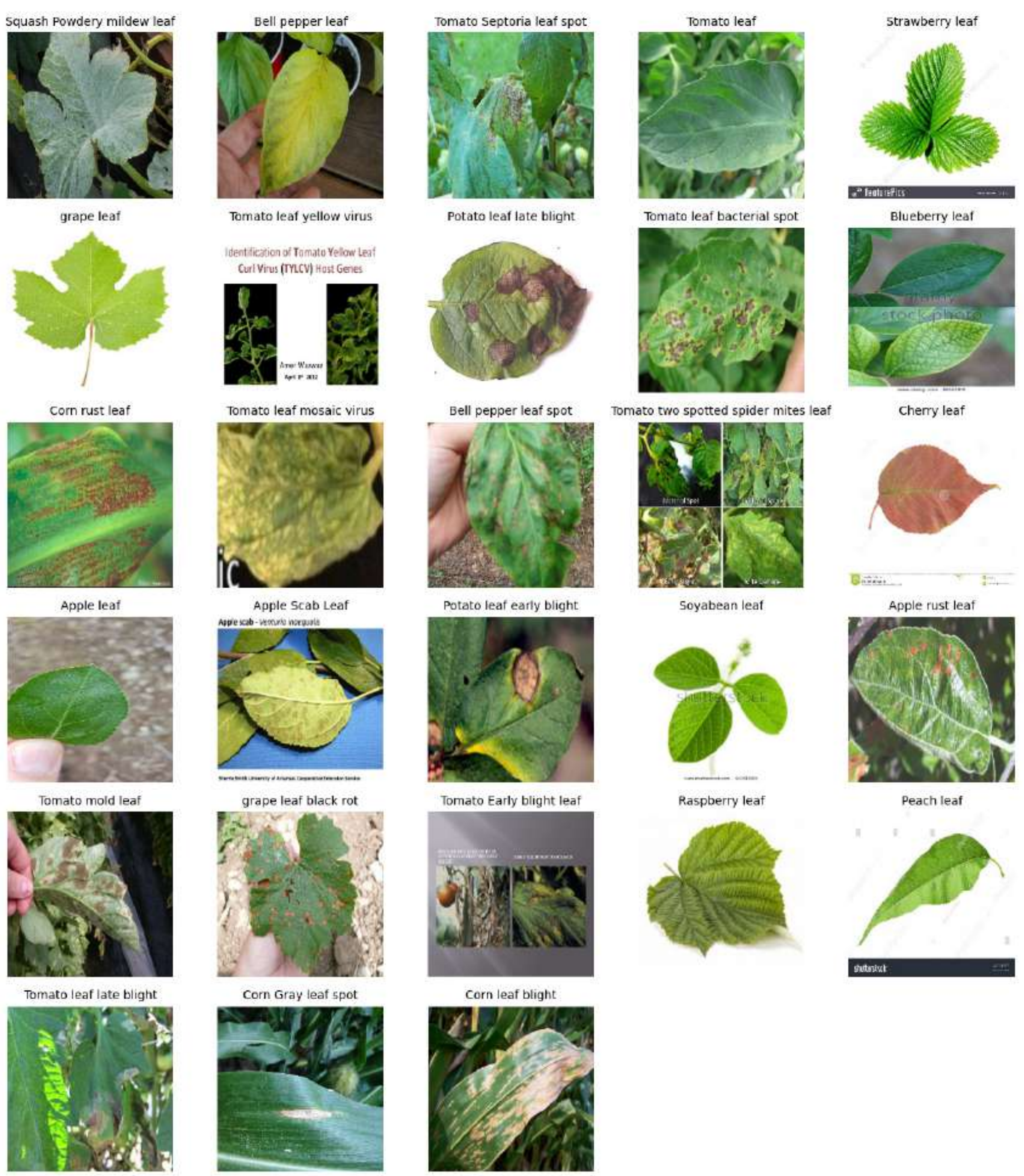

Figure A.28: A sample image of each class present in the plantDoc Dataset.

### A.3.2 Taiwan Tomato

Table A.15: Information about the Taiwan Tomato Dataset.

| Plants | Tomato |
|---|---|
| Diseases | 1. Healthy<br>2. Bacterial Spot<br>3. Gray Spot<br>4. Late Blight<br>5. Powdery Mildew<br>6. Black Mold |
| Number of Classes | 6 |
| Number of Images | 622 |

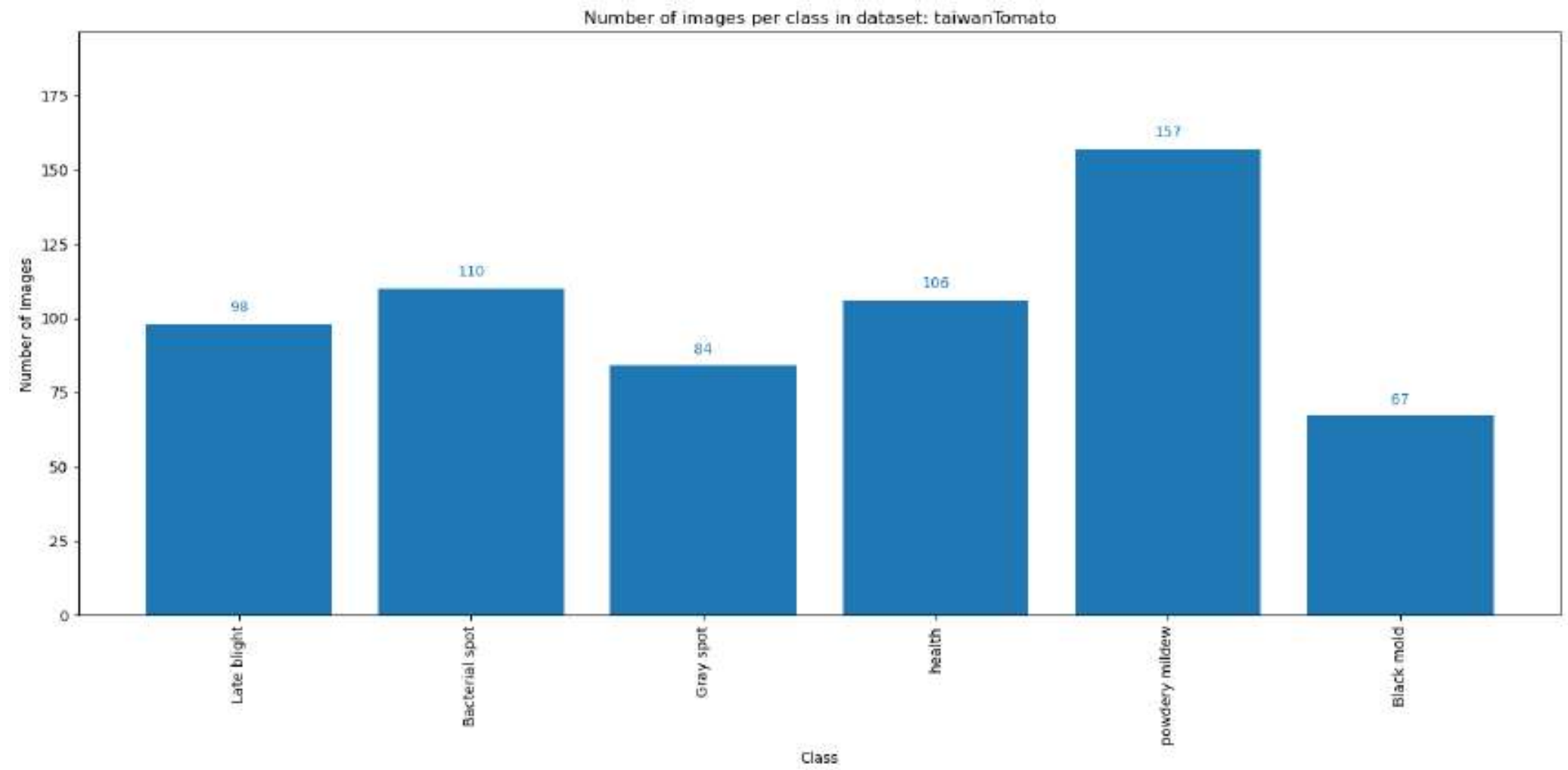

Figure A.29: Class Distribution of the Taiwan Tomato Dataset.

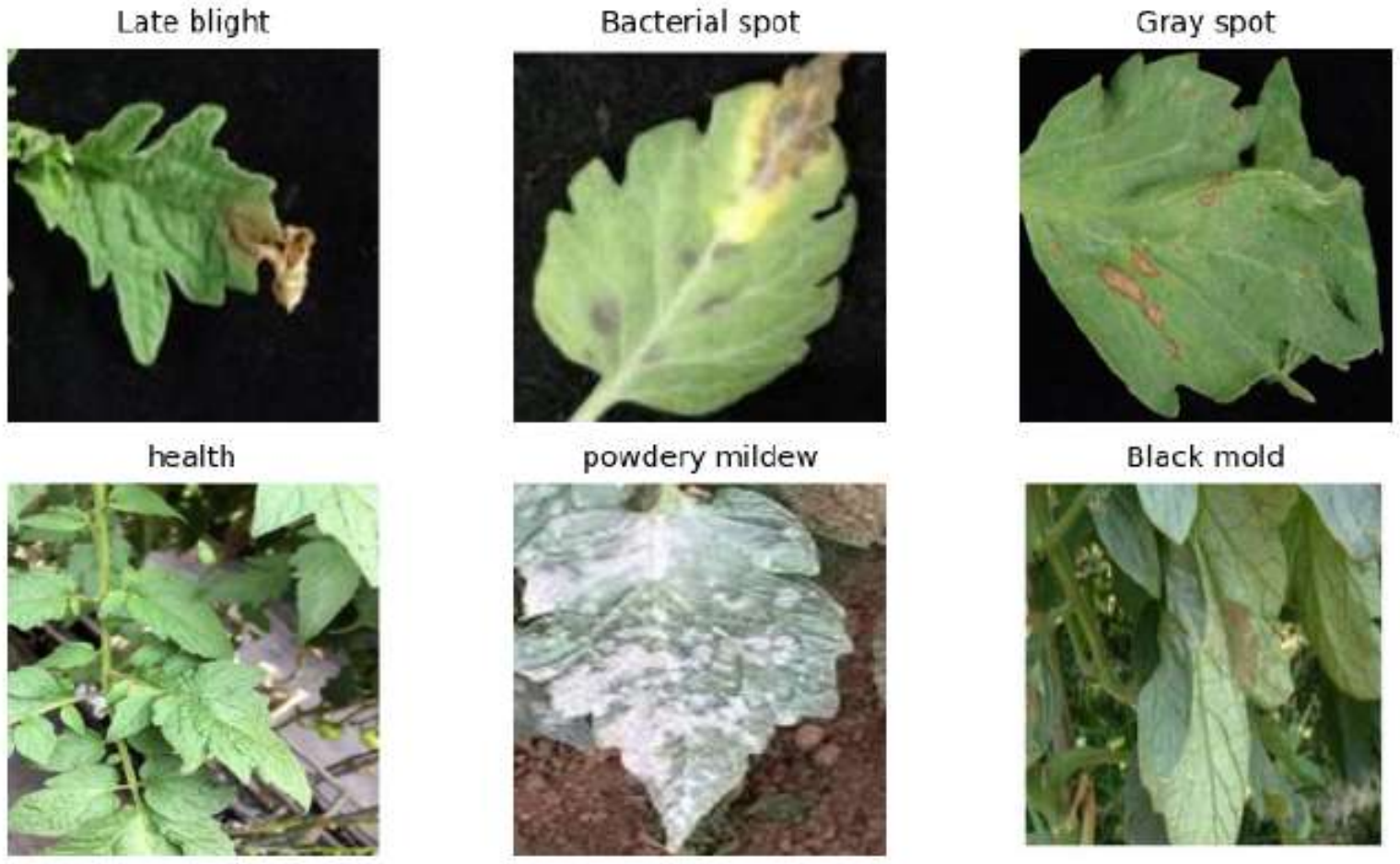

Figure A.30: A sample image of each class present in the Taiwan Tomato Dataset.

### A.3.3 Potato Leaf Disease Dataset in Uncontrolled Environment

Table A.16: Information about the Novel Potato Dataset.

| Plants | Potato |
|---|---|
| Diseases | 1. Healthy<br>2. Bacteria<br>3. Phytopthora<br>4. Nematode<br>5. Pest<br>6. Virus<br>7. Fungi |
| Number of Classes | 7 |
| Number of Images | 3,076 |

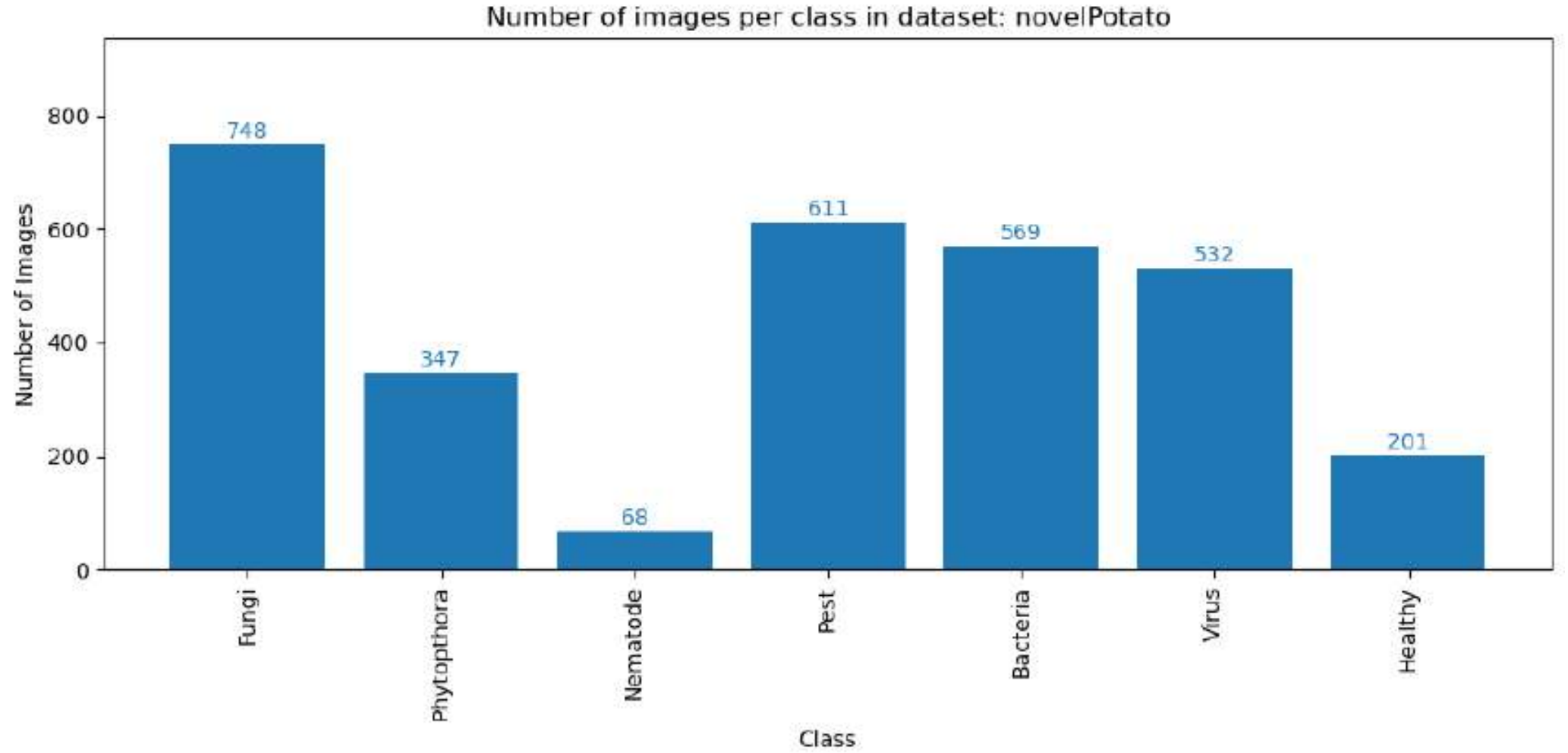

Figure A.31: Class Distribution of the Novel Potato Dataset.

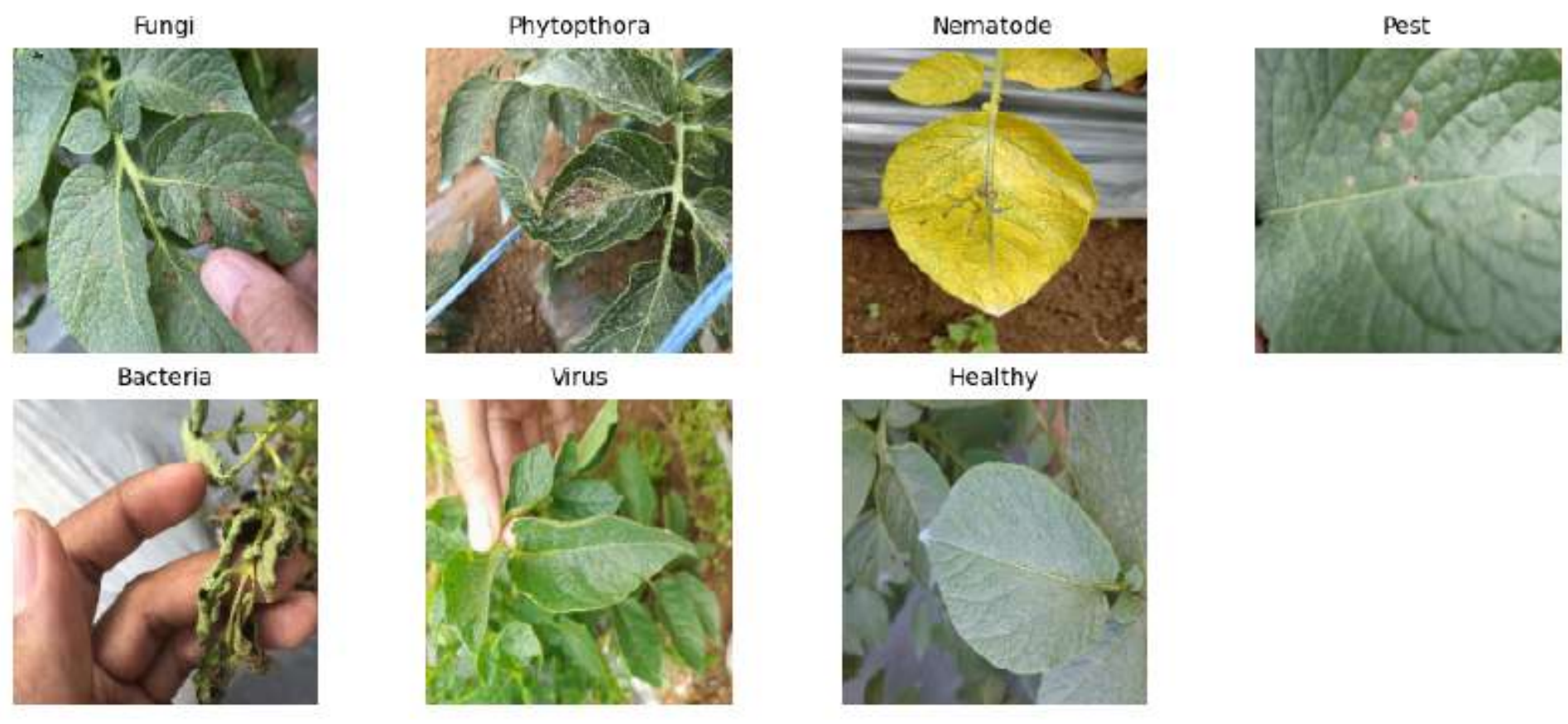

Figure A.32: A sample image of each class present in the Novel Potato Dataset.

### A.3.4 Rice Leaf Diseases Dataset

Table A.17: Information about the RLDD Dataset.

| Plants | Rice |
|---|---|
| Diseases | 1. Bacterial Blight<br>2. Brown Spot<br>3. Leaf Smut |
| Number of Classes | 3 |
| Number of Images | 524 |

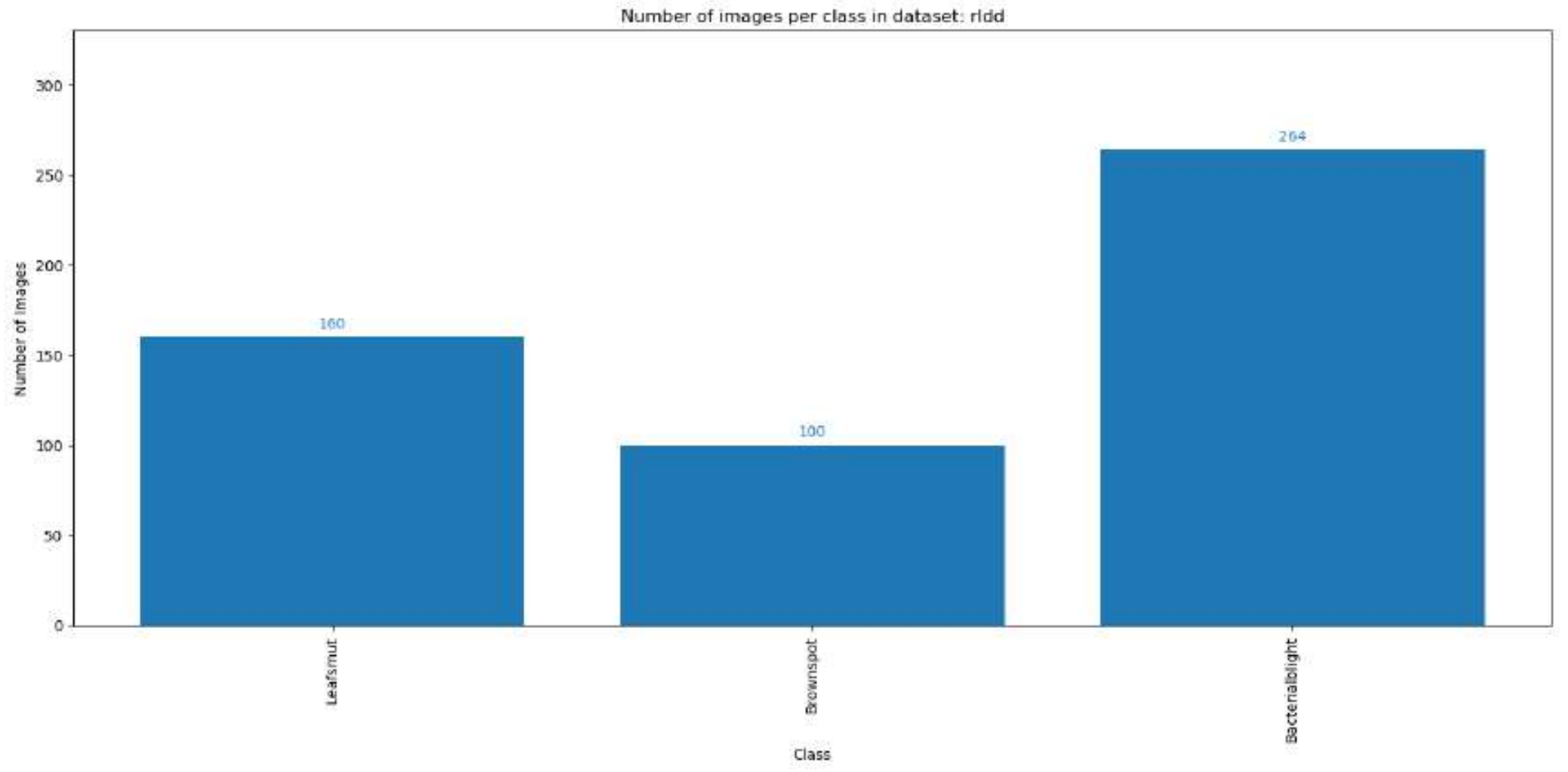

Figure A.33: Class Distribution of the RLDD Dataset.

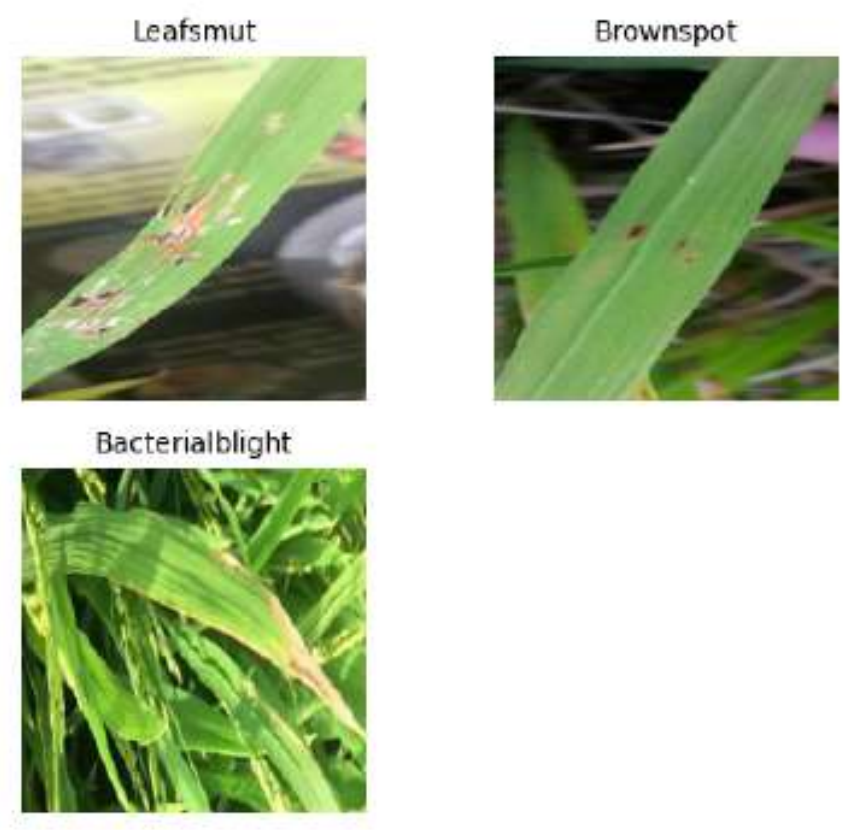

Figure A.34: A sample image of each class present in the RLDD Dataset.

Table A.18: Information about the Sugarcane Dataset.

| Plants | Sugarcane |
|---|---|
| Diseases | 1. Healthy<br>2. Viral Disease<br>3. Banded Chlorosis<br>4. Grassy shoot<br>5. Yellow Leaf<br>6. Pokkah Boeng<br>7. Brown Spot<br>8. Brown Rust<br>9. Smut<br>10. Sett Rot |
| Number of Classes | 10 |
| Number of Images | 6,405 |

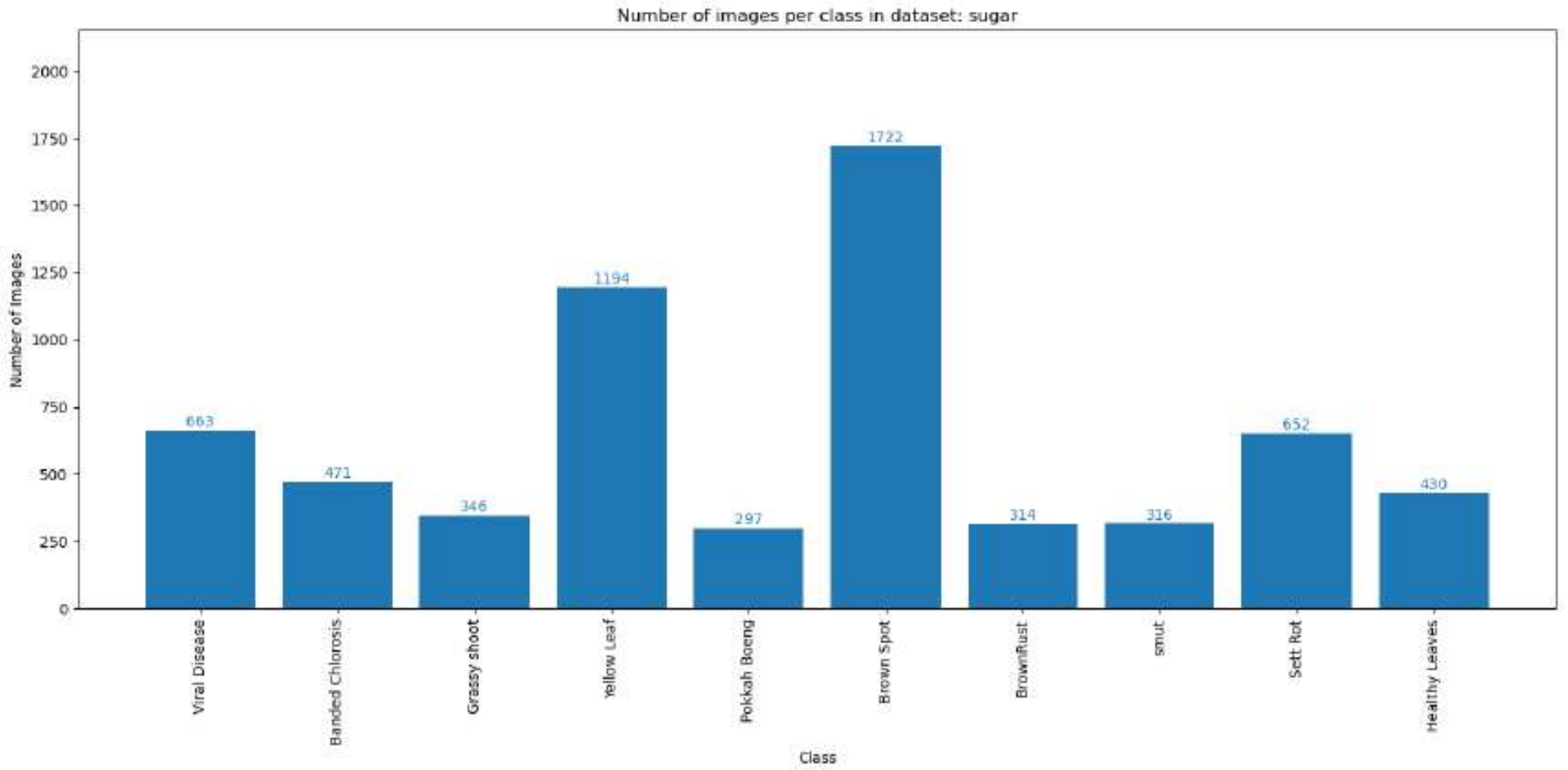

Figure A.35: Class Distribution of the Sugarcane Dataset.

### A.3.5 Sugarcane Leaf Dataset

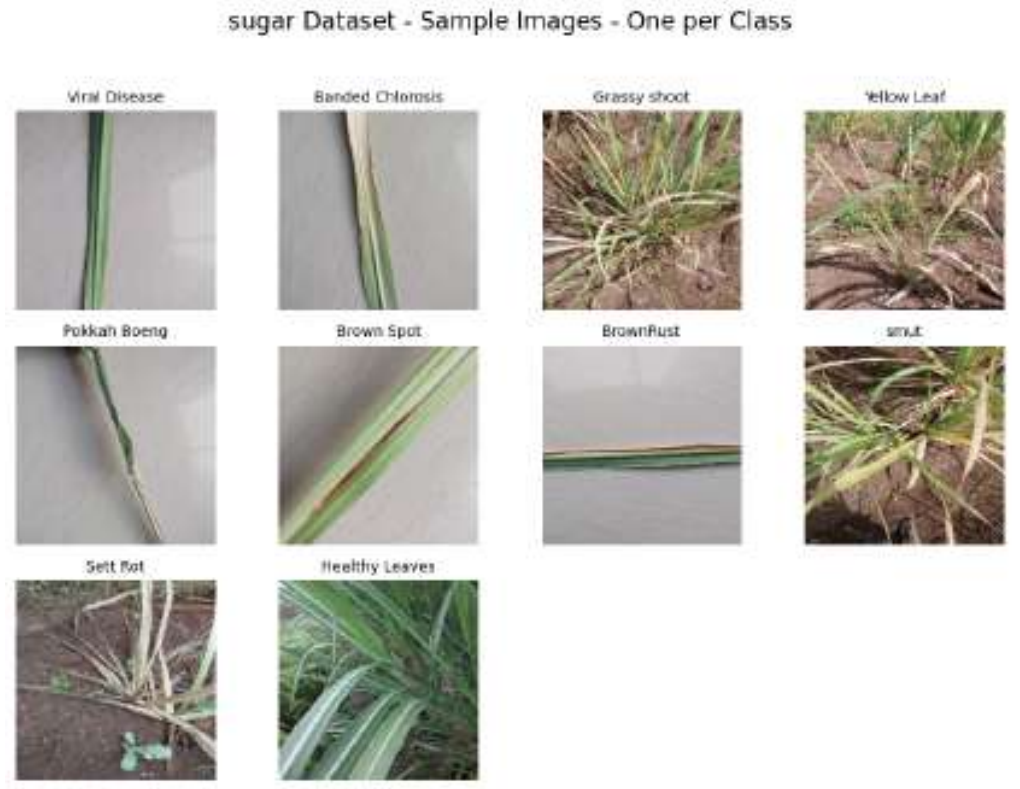

Figure A.36: A sample image of each class present in the Sugarcane Dataset.

# B. Appendix Datasets

## B.1 ResNet

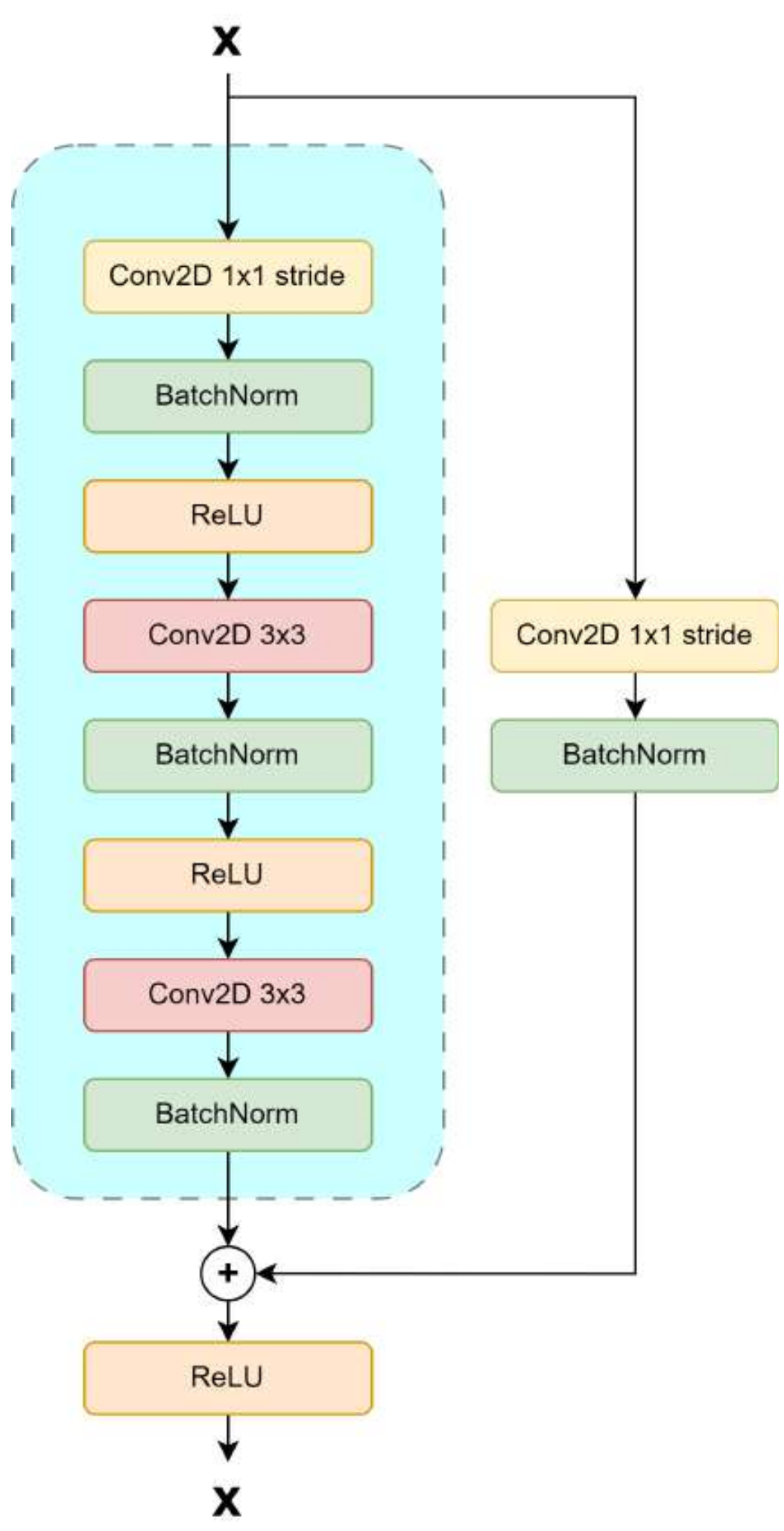

Figure B.1: A ResNet block with the residual connection. The stride in is defaulted to 1 but set to 2 if this is the first block in the layer-block. Based on TensorFlow implementation [127].

## B.2 VGG

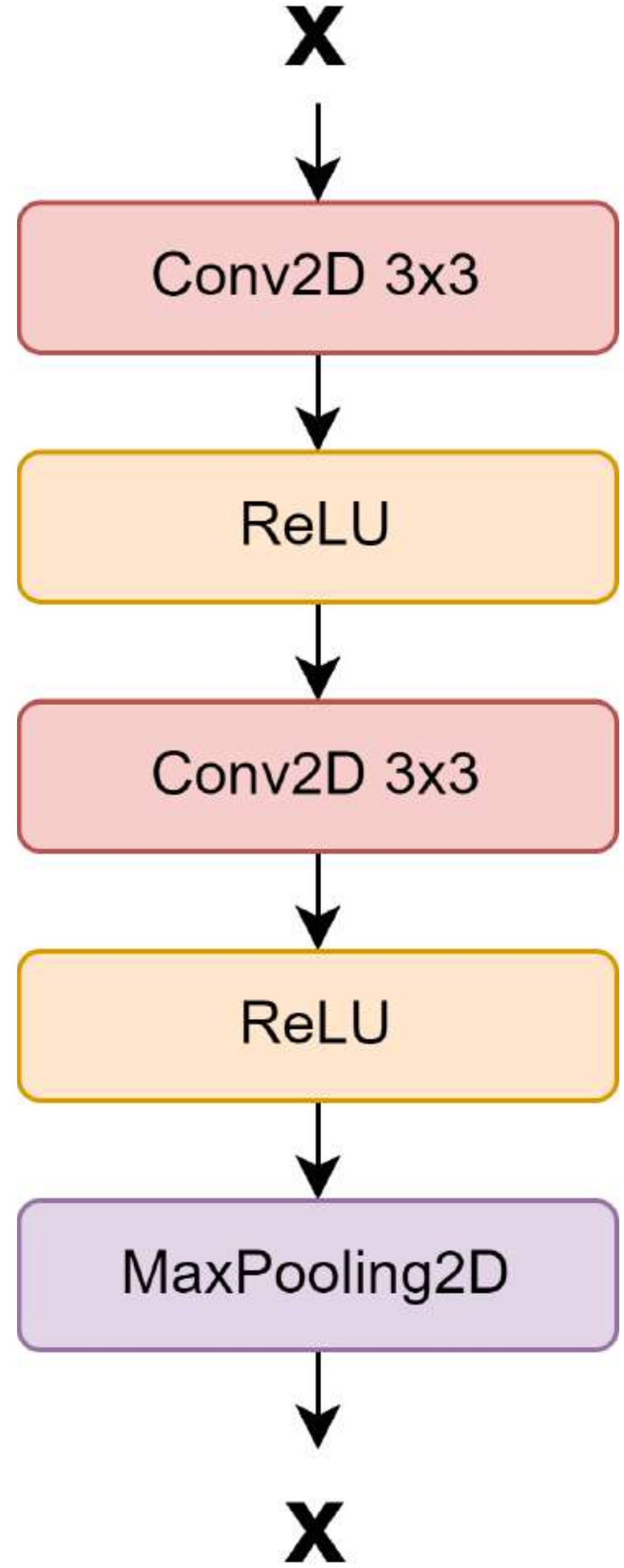

Figure B.2: VGG block of convolutional layers with pooling. Based on TensorFlow implementation [127].

## B.3 ConvNeXt

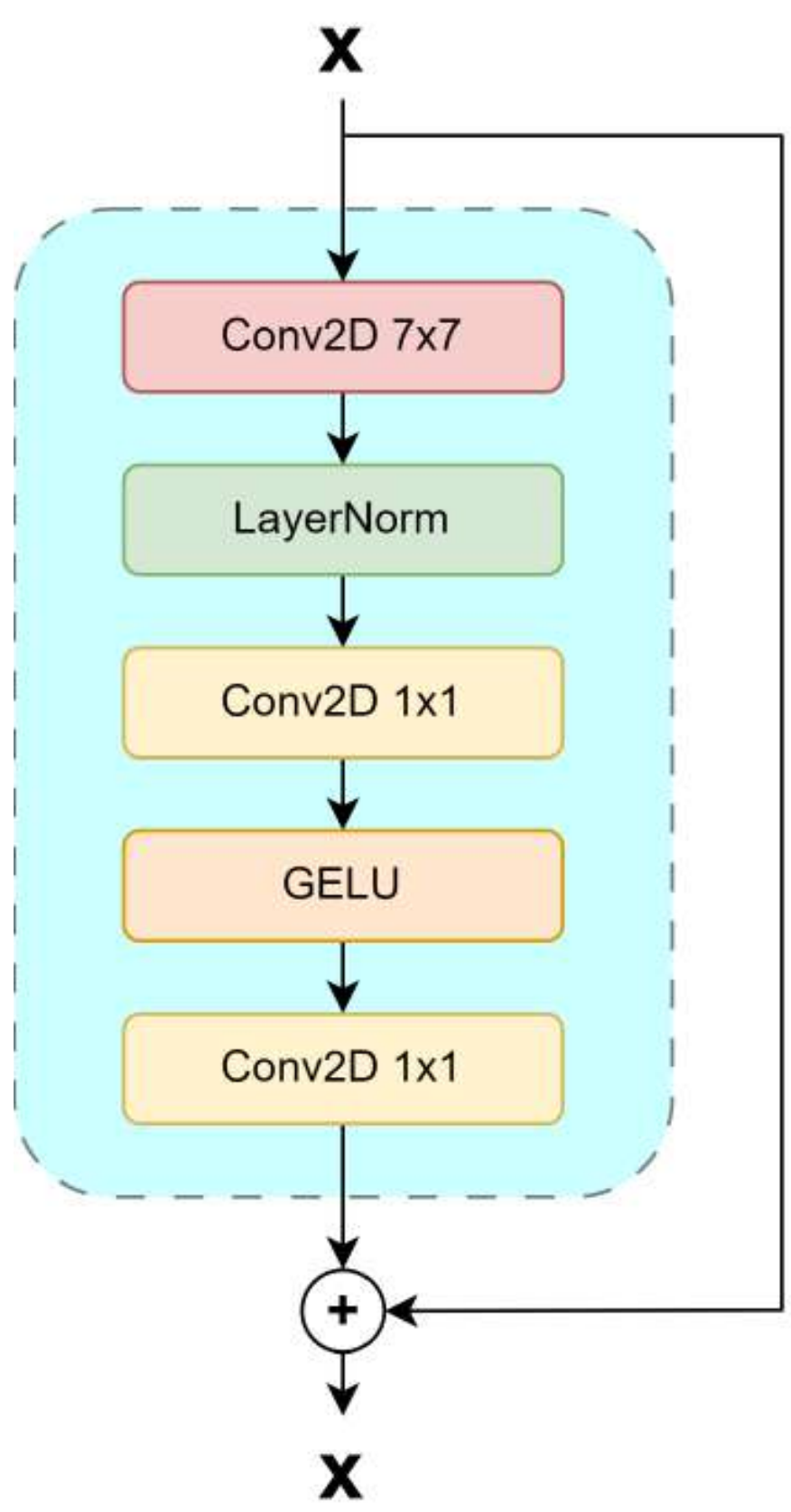

Figure B.3: ConvNeXt block with large kernel, pointwise convolutions, GELU and residual connections. Based on TensorFlow implementation [127].

## B.4 DenseNet

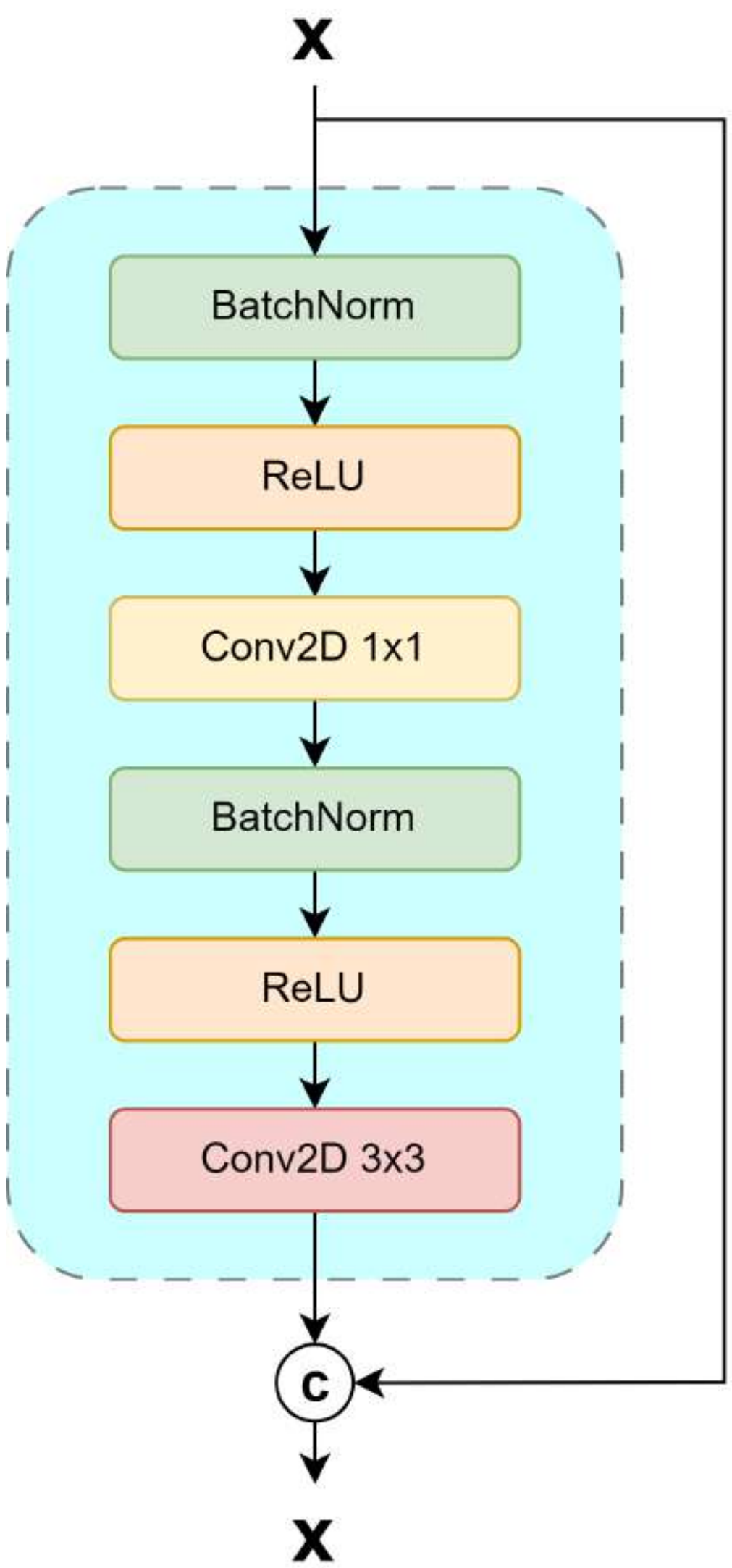

Figure B.4: DenseNet block with bottleneck and dense connection connections. Based on TensorFlow implementation [127].

# C. Appendix Benchmark

In this appendix more detailed results will be given for each combination of all 23 models across all 18 datasets, resulting in 414 different combinations, all trained first using only transfer-learning and then again with fine-tuning in addition. This is done 5 times for each combination, resulting in a total of 4,140 models having been trained. Here, in Table C.2 all 414 combinations will be listed with the corresponding results for transfer-learning only and including fine-tuning. All results are averaged results over the 5 repetitions.

Table C.1: The models rank per dataset for each combination. Column name legend: 1 = cassava, 2 = cds, 3 = cucumber, 4 = diaMos, 5 = fgvc7, 6 = fgvc8, 7 = novelPotato, 8 = paddy, 9 = pdd271, 10 = plantDoc, 11 = plantVillage, 12 = pld, 13 = rldd, 14 = sms, 15 = sugar, 16 = taiwanTomato, 17 = tea, 18 = tomatoVillage, 19 = Avg. Rank, 20 = Worst Rank, 21 = Best Rank, 22 = Avg. Rank STD.

| | **1** | **2** | **3** | **4** | **5** | **6** | **7** | **8** | **9** | **10** | **11** | **12** | **13** | **14** | **15** | **16** | **17** | **18** | **19** | **20** | **21** | **22** |
|---|---|---|---|---|---|---|---|---|---|---|---|---|---|---|---|---|---|---|---|---|---|---|
| **ConvNeXtSmall** | 1 | 13 | 18 | 1 | 4 | 4 | 7 | 2 | 5 | 8 | 8 | 20 | 8 | 14 | 16 | 10 | 11 | 1 | 8.39 | 20 | 1 | 5.96 |
| **ConvNeXtTiny** | 2 | 4 | 5 | 2 | 13 | 1 | 10 | 3 | 9 | 3 | 6 | 16 | 16 | 2 | 20 | 5 | 2 | 10 | 7.17 | 20 | 1 | 5.81 |
| **DenseNet121** | 13 | 7 | 11 | 5 | 9 | 13 | 11 | 16 | 4 | 5 | 5 | 5 | 10 | 8 | 7 | 9 | 10 | 5 | 8.50 | 16 | 4 | 3.43 |
| **DenseNet169** | 14 | 8 | 15 | 13 | 11 | 12 | 14 | 9 | 1 | 10 | 4 | 13 | 11 | 3 | 10 | 6 | 5 | 13 | 9.56 | 15 | 1 | 4.19 |
| **DenseNet201** | 6 | 9 | 9 | 3 | 3 | 16 | 9 | 6 | 2 | 9 | 3 | 1 | 2 | 9 | 4 | 13 | 12 | 9 | 6.94 | 16 | 1 | 4.28 |
| **EfficientNetV2B0** | 8 | 2 | 2 | 9 | 10 | 11 | 3 | 12 | 6 | 1 | 9 | 2 | 3 | 6 | 11 | 8 | 8 | 7 | 6.56 | 12 | 1 | 3.58 |
| **EfficientNetV2B1** | 5 | 10 | 7 | 6 | 7 | 3 | 1 | 1 | 11 | 11 | 11 | 6 | 6 | 11 | 8 | 7 | 14 | 2 | 7.06 | 14 | 1 | 3.78 |
| **EfficientNetV2B2** | 7 | 1 | 6 | 4 | 6 | 5 | 4 | 10 | 3 | 2 | 12 | 9 | 4 | 20 | 2 | 2 | 13 | 6 | 6.44 | 20 | 1 | 4.84 |
| **EfficientNetV2B3** | 10 | 6 | 8 | 14 | 15 | 9 | 6 | 7 | 15 | 6 | 10 | 7 | 5 | 4 | 5 | 1 | 9 | 8 | 8.06 | 15 | 1 | 3.76 |
| **EfficientNetV2M** | 4 | 12 | 3 | 10 | 2 | 7 | 5 | 5 | 8 | 12 | 1 | 3 | 14 | 12 | 3 | 19 | 1 | 11 | 7.33 | 19 | 1 | 5.17 |
| **EfficientNetV2S** | 3 | 3 | 13 | 11 | 8 | 2 | 2 | 4 | 12 | 4 | 2 | 10 | 12 | 7 | 6 | 4 | 15 | 3 | 6.72 | 15 | 2 | 4.38 |
| **InceptionResNetV2** | 16 | 22 | 22 | 16 | 16 | 15 | 20 | 18 | 16 | 15 | 19 | 14 | 23 | 15 | 12 | 23 | 20 | 16 | 17.67 | 23 | 12 | 3.33 |
| **InceptionV3** | 22 | 18 | 23 | 23 | 19 | 17 | 23 | 21 | 17 | 18 | 20 | 23 | 19 | 13 | 21 | 21 | 22 | 23 | 20.17 | 23 | 13 | 2.79 |
| **MobileNetV3Large** | 9 | 15 | 4 | 7 | 5 | 6 | 8 | 13 | 10 | 7 | 13 | 12 | 1 | 5 | 15 | 3 | 4 | 4 | 7.83 | 15 | 1 | 4.30 |
| **MobileNetV3Small** | 15 | 11 | 1 | 19 | 14 | 14 | 15 | 20 | 18 | 13 | 16 | 19 | 9 | 19 | 19 | 11 | 7 | 12 | 14.00 | 20 | 1 | 4.99 |
| **NASNetLarge** | 18 | 23 | 21 | 17 | 17 | 18 | 19 | 17 | 21 | 21 | 17 | 17 | 20 | 22 | 17 | 15 | 19 | 21 | 18.89 | 23 | 15 | 2.22 |
| **NASNetMobile** | 23 | 17 | 14 | 20 | 18 | 23 | 21 | 22 | 20 | 20 | 22 | 18 | 7 | 23 | 23 | 16 | 21 | 22 | 19.44 | 23 | 7 | 4.08 |
| **ResNet101V2** | 17 | 16 | 16 | 22 | 21 | 19 | 16 | 14 | 19 | 14 | 18 | 15 | 13 | 1 | 14 | 20 | 16 | 19 | 16.11 | 22 | 1 | 4.57 |
| **ResNet152V2** | 19 | 19 | 17 | 15 | 20 | 20 | 17 | 8 | 13 | 16 | 15 | 11 | 15 | 18 | 13 | 17 | 18 | 14 | 15.83 | 20 | 8 | 3.20 |
| **ResNet50V2** | 20 | 21 | 20 | 18 | 23 | 21 | 22 | 19 | 22 | 19 | 21 | 8 | 18 | 10 | 18 | 18 | 17 | 18 | 18.50 | 23 | 8 | 3.87 |
| **VGG16** | 12 | 14 | 12 | 12 | 1 | 8 | 12 | 15 | 14 | 22 | 14 | 21 | 21 | 17 | 9 | 14 | 6 | 17 | 13.39 | 22 | 1 | 5.34 |
| **VGG19** | 11 | 5 | 19 | 8 | 12 | 10 | 13 | 11 | 7 | 23 | 7 | 4 | 17 | 16 | 1 | 22 | 3 | 15 | 11.33 | 23 | 1 | 6.42 |
| **Xception** | 21 | 20 | 10 | 21 | 22 | 22 | 18 | 23 | 23 | 17 | 23 | 22 | 22 | 21 | 22 | 12 | 23 | 20 | 20.11 | 23 | 10 | 3.72 |

Table C.2: All of the results obtained during benchmarking. Each of the 23 model 18 dataset combinations are given. Each row represents the 5 run average results. Results originate from [125].

| Model | Dataset | Acc | F1 | Acc SD | F-1 SD | Acc FT | F1 FT | Acc SD FT | F-1 SD FT | Best Epoch | Best Epoch FT |
|---|---|---|---|---|---|---|---|---|---|---|---|
| ConvNeXtSmall | cassava | 0.78001 | 0.62425 | 0.00117 | 0.14137 | 0.82496 | 0.69262 | 0.00689 | 0.12900 | 34 | 5.5 |
| ConvNeXtTiny | cassava | 0.78328 | 0.63023 | 0.00047 | 0.14031 | 0.82415 | 0.69385 | 0.00631 | 0.12505 | 35.5 | 6 |
| DenseNet121 | cassava | 0.75747 | 0.60116 | 0.00105 | 0.14537 | 0.79496 | 0.66093 | 0.00841 | 0.13004 | 30 | 3.5 |
| DenseNet169 | cassava | 0.76845 | 0.60971 | 0.00058 | 0.14469 | 0.79426 | 0.65234 | 0.01401 | 0.13566 | 21.5 | 1.5 |
| DenseNet201 | cassava | 0.77265 | 0.62075 | 0.00245 | 0.13852 | 0.81282 | 0.68416 | 0.00829 | 0.12165 | 19.5 | 2.5 |
| EfficientNetV2B0 | cassava | 0.78643 | 0.63889 | 0.00222 | 0.13538 | 0.80862 | 0.67394 | 0.00736 | 0.13252 | 17.5 | 6 |
| EfficientNetV2B1 | cassava | 0.78211 | 0.63017 | 0.00093 | 0.13845 | 0.81364 | 0.67905 | 0.00187 | 0.12862 | 17 | 5 |
| EfficientNetV2B2 | cassava | 0.78515 | 0.63138 | 0.00047 | 0.14088 | 0.81119 | 0.67578 | 0.00082 | 0.13248 | 19 | 3 |
| EfficientNetV2B3 | cassava | 0.77802 | 0.62022 | 0.00199 | 0.14717 | 0.80184 | 0.65205 | 0.00362 | 0.14849 | 18.5 | 3 |
| EfficientNetV2M | cassava | 0.75829 | 0.59184 | 0.00047 | 0.15247 | 0.81773 | 0.69511 | 0.00666 | 0.11830 | 21.5 | 2 |
| EfficientNetV2S | cassava | 0.78164 | 0.62902 | 0.00280 | 0.14037 | 0.81831 | 0.68998 | 0.00140 | 0.12639 | 20.5 | 3 |
| InceptionResNetV2 | cassava | 0.73552 | 0.55229 | 0.00245 | 0.16508 | 0.77861 | 0.62860 | 0.00187 | 0.13799 | 27.5 | 2.5 |
| InceptionV3 | cassava | 0.72069 | 0.51887 | 0.00327 | 0.17793 | 0.75479 | 0.59427 | 0.00444 | 0.15118 | 14 | 1.5 |
| MobileNetV3Large | cassava | 0.77721 | 0.62504 | 0.00070 | 0.13600 | 0.80220 | 0.66483 | 0.00677 | 0.12958 | 19 | 7.5 |
| MobileNetV3Small | cassava | 0.75315 | 0.58684 | 0.00234 | 0.14969 | 0.78234 | 0.63672 | 0.00257 | 0.13387 | 27 | 14 |

Continued on next page

**Table C.2 - continued from previous page**

| Model | Dataset | Acc | F1 | Acc SD | F-1 SD | Acc FT | F1 FT | Acc SD FT | F-1 SD FT | Best Epoch | Best Epoch FT |
|---|---|---|---|---|---|---|---|---|---|---|---|
| NASNetLarge | cassava | 0.70213 | 0.48395 | 0.00175 | 0.19411 | 0.76273 | 0.59598 | 0.01051 | 0.15526 | 10 | 2 |
| NASNetMobile | cassava | 0.71976 | 0.52050 | 0.00304 | 0.17424 | 0.75058 | 0.58012 | 0.00560 | 0.15615 | 20 | 5 |
| ResNet101V2 | cassava | 0.73027 | 0.53108 | 0.00117 | 0.17458 | 0.76436 | 0.59547 | 0.01028 | 0.15580 | 13 | 1.5 |
| ResNet152V2 | cassava | 0.72279 | 0.52208 | 0.00257 | 0.17665 | 0.76133 | 0.58455 | 0.01775 | 0.16701 | 12.5 | 1.5 |
| ResNet50V2 | cassava | 0.72980 | 0.53629 | 0.00140 | 0.17492 | 0.76039 | 0.59074 | 0.01355 | 0.15854 | 12.5 | 1 |
| VGG16 | cassava | 0.72875 | 0.53920 | 0.00245 | 0.16608 | 0.80044 | 0.66208 | 0.02067 | 0.13597 | 27 | 5.5 |
| VGG19 | cassava | 0.71952 | 0.51435 | 0.00187 | 0.17995 | 0.80056 | 0.65480 | 0.00794 | 0.13737 | 25 | 5.5 |
| Xception | cassava | 0.72560 | 0.53492 | 0.00304 | 0.17237 | 0.75525 | 0.59027 | 0.01308 | 0.15398 | 17 | 2 |
| ConvNeXtSmall | cds | 0.96994 | 0.96945 | 0.00791 | 0.01403 | 0.98259 | 0.98238 | 0.00158 | 0.00511 | 143 | 39.5 |
| ConvNeXtTiny | cds | 0.96835 | 0.96774 | 0.00000 | 0.01838 | 0.99051 | 0.99024 | 0.00316 | 0.00797 | 101.5 | 48 |
| DenseNet121 | cds | 0.96519 | 0.96497 | 0.00633 | 0.00882 | 0.98734 | 0.98734 | 0.00316 | 0.00461 | 172.5 | 56 |
| DenseNet169 | cds | 0.96677 | 0.96622 | 0.00158 | 0.01683 | 0.98576 | 0.98558 | 0.00158 | 0.00577 | 76 | 82.5 |
| DenseNet201 | cds | 0.97152 | 0.97117 | 0.00316 | 0.00990 | 0.98418 | 0.98394 | 0.00316 | 0.00851 | 89.5 | 24.5 |
| EfficientNetV2B0 | cds | 0.98101 | 0.98077 | 0.00316 | 0.00827 | 0.99367 | 0.99364 | 0.00000 | 0.00231 | 78.5 | 89 |
| EfficientNetV2B1 | cds | 0.97627 | 0.97582 | 0.00158 | 0.01251 | 0.98418 | 0.98386 | 0.00316 | 0.00941 | 89.5 | 44.5 |
| EfficientNetV2B2 | cds | 0.96835 | 0.96781 | 0.00000 | 0.01339 | 0.99525 | 0.99512 | 0.00158 | 0.00399 | 64.5 | 119 |

Continued on next page

**Table C.2 - continued from previous page**

| Model | Dataset | Acc | F1 | Acc SD | F-1 SD | Acc FT | F1 FT | Acc SD FT | F-1 SD FT | Best Epoch | Best Epoch FT |
|---|---|---|---|---|---|---|---|---|---|---|---|
| EfficientNetV2B3 | cds | 0.97152 | 0.97111 | 0.00316 | 0.01170 | 0.98892 | 0.98873 | 0.00158 | 0.00560 | 103.5 | 97 |
| EfficientNetV2M | cds | 0.95411 | 0.95377 | 0.00475 | 0.01189 | 0.98259 | 0.98257 | 0.00158 | 0.00362 | 156.5 | 68.5 |
| EfficientNetV2S | cds | 0.97943 | 0.97903 | 0.00158 | 0.01067 | 0.99209 | 0.99194 | 0.00158 | 0.00460 | 74.5 | 49.5 |
| InceptionResNetV2 | cds | 0.84019 | 0.83703 | 0.00158 | 0.04516 | 0.92880 | 0.92798 | 0.00158 | 0.02322 | 65.5 | 10.5 |
| InceptionV3 | cds | 0.83070 | 0.82909 | 0.00158 | 0.04169 | 0.94620 | 0.94517 | 0.00000 | 0.02506 | 50.5 | 29 |
| MobileNetV3Large | cds | 0.96361 | 0.96338 | 0.00158 | 0.00700 | 0.97785 | 0.97781 | 0.00000 | 0.00370 | 111.5 | 75.5 |
| MobileNetV3Small | cds | 0.97468 | 0.97418 | 0.00000 | 0.01513 | 0.98259 | 0.98243 | 0.00475 | 0.00864 | 116 | 125 |
| NASNetLarge | cds | 0.84968 | 0.84730 | 0.00475 | 0.05216 | 0.91930 | 0.91798 | 0.02057 | 0.04176 | 35.5 | 21 |
| NASNetMobile | cds | 0.87658 | 0.87488 | 0.00000 | 0.02972 | 0.96677 | 0.96644 | 0.00158 | 0.01037 | 94.5 | 73.5 |
| ResNet101V2 | cds | 0.91772 | 0.91733 | 0.00000 | 0.02671 | 0.97152 | 0.97128 | 0.01582 | 0.01955 | 47 | 18 |
| ResNet152V2 | cds | 0.90032 | 0.89906 | 0.00475 | 0.02857 | 0.94620 | 0.94583 | 0.00633 | 0.01566 | 61.5 | 71 |
| ResNet50V2 | cds | 0.89557 | 0.89414 | 0.00000 | 0.04335 | 0.94146 | 0.94079 | 0.02057 | 0.03197 | 54.5 | 8.5 |
| VGG16 | cds | 0.94304 | 0.94268 | 0.00633 | 0.01272 | 0.98101 | 0.98089 | 0.00633 | 0.01022 | 131 | 144.5 |
| VGG19 | cds | 0.94146 | 0.94072 | 0.00791 | 0.01817 | 0.99051 | 0.99044 | 0.00316 | 0.00406 | 105.5 | 119 |
| Xception | cds | 875.00000 | 0.87348 | 0.00475 | 0.03620 | 0.94304 | 0.94213 | 0.00000 | 0.01913 | 49 | 71 |
| ConvNeXtSmall | cucumber | 0.81875 | 0.81721 | 0.01875 | 0.11032 | 0.84375 | 0.84344 | 0.00000 | 0.09871 | 115.5 | 12 |

Continued on next page

**Table C.2 - continued from previous page**

| Model | Dataset | Acc | F1 | Acc SD | F-1 SD | Acc FT | F1 FT | Acc SD FT | F-1 SD FT | Best Epoch | Best Epoch FT |
|---|---|---|---|---|---|---|---|---|---|---|---|
| ConvNeXtTiny | cucumber | 0.86563 | 0.86652 | 0.01563 | 0.07333 | 0.89375 | 0.89313 | 0.00625 | 0.05209 | 115.5 | 29 |
| DenseNet121 | cucumber | 0.84062 | 0.84082 | 0.00313 | 0.10035 | 0.86875 | 0.86702 | 0.01250 | 0.07383 | 75 | 2 |
| DenseNet169 | cucumber | 0.82813 | 0.82875 | 0.01563 | 0.08853 | 0.85313 | 0.85208 | 0.00312 | 0.07025 | 56 | 4 |
| DenseNet201 | cucumber | 0.86250 | 0.86106 | 0.00000 | 0.06093 | 0.87813 | 0.87518 | 0.00313 | 0.07824 | 53.5 | 3 |
| EfficientNetV2B0 | cucumber | 0.88125 | 0.88196 | 0.00625 | 0.07104 | 0.89687 | 0.89769 | 0.00312 | 0.05553 | 58.5 | 11.5 |
| EfficientNetV2B1 | cucumber | 0.86562 | 0.86351 | 0.00938 | 0.08855 | 0.88125 | 0.87999 | 0.00625 | 0.08022 | 56.5 | 15 |
| EfficientNetV2B2 | cucumber | 0.86250 | 0.86349 | 0.01250 | 0.08288 | 0.88438 | 0.88592 | 0.00938 | 0.07105 | 50 | 10 |
| EfficientNetV2B3 | cucumber | 0.85000 | 0.85159 | 0.01250 | 0.07936 | 0.88125 | 0.88188 | 0.00625 | 0.06592 | 71 | 14 |
| EfficientNetV2M | cucumber | 0.85313 | 0.85315 | 0.02813 | 0.08377 | 0.89687 | 0.89825 | 0.00938 | 0.06460 | 85 | 7.5 |
| EfficientNetV2S | cucumber | 0.86250 | 0.86378 | 0.00625 | 0.07255 | 0.86875 | 0.86866 | 0.00000 | 0.07907 | 73.5 | 4.5 |
| InceptionResNetV2 | cucumber | 0.73750 | 0.73738 | 0.01875 | 0.12725 | 0.79063 | 0.78799 | 0.04063 | 0.13374 | 118 | 1 |
| InceptionV3 | cucumber | 0.74375 | 0.74095 | 0.00625 | 0.13858 | 0.77500 | 0.77154 | 0.03125 | 0.14104 | 50 | 1.5 |
| MobileNetV3Large | cucumber | 0.89687 | 0.89606 | 0.00938 | 0.05367 | 0.89375 | 0.89341 | 0.00625 | 0.06767 | 67.5 | 1.5 |
| MobileNetV3Small | cucumber | 0.91250 | 0.91108 | 0.00625 | 0.05778 | 0.89687 | 0.89592 | 0.00312 | 0.05804 | 110 | 7.5 |
| NASNetLarge | cucumber | 0.75000 | 0.75064 | 0.01875 | 0.11984 | 0.79375 | 0.79340 | 0.03750 | 0.11990 | 45 | 2 |
| NASNetMobile | cucumber | 0.80313 | 0.79795 | 0.01563 | 0.09004 | 0.86250 | 0.85916 | 0.03125 | 0.07576 | 71.5 | 5.5 |

Continued on next page

**Table C.2 - continued from previous page**

| Model | Dataset | Acc | F1 | Acc SD | F-1 SD | Acc FT | F1 FT | Acc SD FT | F-1 SD FT | Best Epoch | Best Epoch FT |
|---|---|---|---|---|---|---|---|---|---|---|---|
| ResNet101V2 | cucumber | 0.79688 | 0.79432 | 0.00313 | 0.09929 | 0.84375 | 0.84105 | 0.01875 | 0.07297 | 36.5 | 2.5 |
| ResNet152V2 | cucumber | 0.82500 | 0.82416 | 0.01250 | 0.08587 | 0.84375 | 0.84312 | 0.03125 | 0.08643 | 45 | 1 |
| ResNet50V2 | cucumber | 0.81562 | 0.81491 | 0.00938 | 0.09511 | 0.82188 | 0.82169 | 0.00312 | 0.07182 | 39.5 | 1.5 |
| VGG16 | cucumber | 0.82188 | 0.82181 | 0.03437 | 0.09345 | 0.86875 | 0.86823 | 0.02500 | 0.06336 | 100 | 5.5 |
| VGG19 | cucumber | 0.80937 | 0.80940 | 0.01563 | 0.07588 | 0.83750 | 0.83840 | 0.04375 | 0.09318 | 85 | 4 |
| Xception | cucumber | 0.83750 | 0.83852 | 0.00000 | 0.08981 | 0.87188 | 0.87223 | 0.00938 | 0.07949 | 78.5 | 2.5 |
| ConvNeXtSmall | diaMos | 0.86213 | 0.84897 | 0.00498 | 0.08569 | 0.88040 | 0.85283 | 0.00498 | 0.06588 | 58.5 | 5 |
| ConvNeXtTiny | diaMos | 0.85880 | 0.83793 | 0.00664 | 0.11707 | 0.87708 | 0.84893 | 0.00332 | 0.11475 | 57.5 | 3.5 |
| DenseNet121 | diaMos | 0.84718 | 0.83566 | 0.00332 | 0.12595 | 0.87043 | 0.88052 | 0.00166 | 0.07445 | 34.5 | 3 |
| DenseNet169 | diaMos | 0.84718 | 0.73766 | 0.00000 | 0.18964 | 0.85797 | 0.83350 | 0.01080 | 0.11000 | 33 | 2 |
| DenseNet201 | diaMos | 0.86545 | 0.86713 | 0.00332 | 0.05530 | 0.87458 | 0.88296 | 0.00581 | 0.05537 | 29.5 | 4 |
| EfficientNetV2B0 | diaMos | 0.85382 | 0.84989 | 0.00332 | 0.06426 | 0.86296 | 0.87882 | 0.01412 | 0.07092 | 36.5 | 3 |
| EfficientNetV2B1 | diaMos | 0.85963 | 0.84283 | 0.01080 | 0.10764 | 0.86960 | 0.87193 | 0.00083 | 0.06647 | 45 | 3 |
| EfficientNetV2B2 | diaMos | 0.85299 | 0.80524 | 0.00748 | 0.10470 | 0.87375 | 0.86767 | 0.00166 | 0.06332 | 32.5 | 4 |
| EfficientNetV2B3 | diaMos | 0.83638 | 0.80016 | 0.00083 | 0.11112 | 0.85216 | 0.84935 | 0.01329 | 0.08003 | 39 | 1 |
| EfficientNetV2M | diaMos | 0.82226 | 0.77528 | 0.00166 | 0.06816 | 0.86130 | 0.83250 | 0.01412 | 0.10182 | 48.5 | 3.5 |

Continued on next page

**Table C.2 - continued from previous page**

| Model | Dataset | Acc | F1 | Acc SD | F-1 SD | Acc FT | F1 FT | Acc SD FT | F-1 SD FT | Best Epoch | Best Epoch FT |
|---|---|---|---|---|---|---|---|---|---|---|---|
| EfficientNetV2S | diaMos | 0.84551 | 0.76510 | 0.00332 | 0.18956 | 0.86130 | 0.81664 | 0.00914 | 0.12668 | 31.5 | 3.5 |
| InceptionResNetV2 | diaMos | 0.79485 | 0.77086 | 0.00083 | 0.08811 | 0.83970 | 0.83068 | 0.01744 | 0.07689 | 44.5 | 2 |
| InceptionV3 | diaMos | 0.78821 | 0.66302 | 0.01246 | 0.28057 | 0.82641 | 0.79891 | 0.02575 | 0.11158 | 30.5 | 2 |
| MobileNetV3Large | diaMos | 0.84967 | 0.75074 | 0.00249 | 0.16749 | 0.86877 | 0.79644 | 0.00332 | 0.13719 | 30 | 2.5 |
| MobileNetV3Small | diaMos | 0.82392 | 0.73654 | 0.00000 | 0.09735 | 0.83555 | 0.76536 | 0.00498 | 0.08361 | 59 | 3 |
| NASNetLarge | diaMos | 0.75083 | 0.58548 | 0.00831 | 0.15768 | 0.83804 | 0.79104 | 0.02907 | 0.12968 | 17 | 2.5 |
| NASNetMobile | diaMos | 0.79319 | 0.73738 | 0.00415 | 0.14046 | 0.83472 | 0.77049 | 0.02409 | 0.15828 | 43 | 4.5 |
| ResNet101V2 | diaMos | 0.80066 | 0.66903 | 0.00498 | 0.19162 | 0.82973 | 0.79110 | 0.01080 | 0.12124 | 24 | 1.5 |
| ResNet152V2 | diaMos | 0.80731 | 0.63650 | 0.00498 | 0.26020 | 0.84053 | 0.81604 | 0.02159 | 0.12358 | 26 | 1 |
| ResNet50V2 | diaMos | 0.80150 | 0.73303 | 0.00914 | 0.16123 | 0.83721 | 0.82918 | 0.01163 | 0.08373 | 27.5 | 1.5 |
| VGG16 | diaMos | 0.81561 | 0.76360 | 0.01329 | 0.12034 | 0.86130 | 0.78359 | 0.02575 | 0.17792 | 53.5 | 4.5 |
| VGG19 | diaMos | 0.81146 | 0.73417 | 0.01080 | 0.12653 | 0.86794 | 0.79256 | 0.00914 | 0.12899 | 62 | 4.5 |
| Xception | diaMos | 0.79153 | 0.59321 | 0.00415 | 0.29989 | 0.83056 | 0.74193 | 0.02159 | 0.21693 | 28 | 2.5 |
| ConvNeXtSmall | fgvc7 | 0.85967 | 0.71583 | 0.00681 | 0.28318 | 0.90599 | 0.77388 | 0.00681 | 0.27519 | 61 | 13 |
| ConvNeXtTiny | fgvc7 | 0.85014 | 0.75550 | 0.00272 | 0.20168 | 0.89101 | 0.77237 | 0.01907 | 0.24835 | 95.5 | 10.5 |
| DenseNet121 | fgvc7 | 0.83243 | 0.68355 | 0.00681 | 0.29268 | 0.89782 | 0.79357 | 0.00681 | 0.21757 | 69.5 | 7 |

Continued on next page

**Table C.2 - continued from previous page**

| Model | Dataset | Acc | F1 | Acc SD | F-1 SD | Acc FT | F1 FT | Acc SD FT | F-1 SD FT | Best Epoch | Best Epoch FT |
|---|---|---|---|---|---|---|---|---|---|---|---|
| DenseNet169 | fgvc7 | 0.83924 | 0.69955 | 0.00000 | 0.27186 | 0.89510 | 0.78923 | 0.02316 | 0.22254 | 44 | 2.5 |
| DenseNet201 | fgvc7 | 0.84332 | 0.64892 | 0.00409 | 0.37519 | 0.90599 | 0.80389 | 0.00954 | 0.21137 | 41.5 | 5.5 |
| EfficientNetV2B0 | fgvc7 | 0.86921 | 0.69105 | 0.00000 | 0.35367 | 0.89510 | 0.80029 | 0.00681 | 0.20116 | 57.5 | 8.5 |
| EfficientNetV2B1 | fgvc7 | 0.84332 | 0.70901 | 0.00681 | 0.26612 | 0.90054 | 0.79336 | 0.01499 | 0.22689 | 44.5 | 10.5 |
| EfficientNetV2B2 | fgvc7 | 0.86785 | 0.75891 | 0.00954 | 0.22070 | 0.90191 | 0.80841 | 0.01362 | 0.20808 | 41.5 | 5 |
| EfficientNetV2B3 | fgvc7 | 0.83787 | 0.66849 | 0.00409 | 0.33284 | 0.87738 | 0.72236 | 0.01635 | 0.32486 | 52.5 | 6.5 |
| EfficientNetV2M | fgvc7 | 0.84469 | 0.70063 | 0.00272 | 0.28474 | 0.90872 | 0.78980 | 0.00954 | 0.24659 | 55.5 | 4.5 |
| EfficientNetV2S | fgvc7 | 0.88556 | 0.72589 | 0.00545 | 0.31462 | 0.90054 | 0.79349 | 0.00409 | 0.21983 | 56 | 3.5 |
| InceptionResNetV2 | fgvc7 | 0.84196 | 0.68733 | 0.01090 | 0.31421 | 0.87057 | 0.75214 | 0.00954 | 0.26032 | 54 | 2 |
| InceptionV3 | fgvc7 | 0.76158 | 0.59557 | 0.00681 | 0.32074 | 0.84605 | 0.69933 | 0.03134 | 0.31039 | 25 | 1 |
| MobileNetV3Large | fgvc7 | 0.87466 | 0.76169 | 0.01362 | 0.23266 | 0.90463 | 0.80751 | 0.00817 | 0.20164 | 40 | 11 |
| MobileNetV3Small | fgvc7 | 0.82834 | 0.63922 | 0.00272 | 0.36927 | 0.87738 | 0.71675 | 0.00000 | 0.32358 | 59 | 21.5 |
| NASNetLarge | fgvc7 | 0.76567 | 0.59957 | 0.00000 | 0.31924 | 0.85967 | 0.72154 | 0.02589 | 0.29223 | 24.5 | 2.5 |
| NASNetMobile | fgvc7 | 0.76703 | 0.59028 | 0.00409 | 0.34115 | 0.85559 | 0.73945 | 0.01362 | 0.24708 | 41.5 | 6.5 |
| ResNet101V2 | fgvc7 | 0.79019 | 0.62003 | 0.00272 | 0.33214 | 0.83379 | 0.71219 | 0.01907 | 0.24976 | 42.5 | 2.5 |
| ResNet152V2 | fgvc7 | 0.78883 | 0.61801 | 0.00681 | 0.32975 | 0.83787 | 0.74892 | 0.02316 | 0.19150 | 32.5 | 2 |

Continued on next page

**Table C.2 - continued from previous page**

| Model | Dataset | Acc | F1 | Acc SD | F-1 SD | Acc FT | F1 FT | Acc SD FT | F-1 SD FT | Best Epoch | Best Epoch FT |
|---|---|---|---|---|---|---|---|---|---|---|---|
| ResNet50V2 | fgvc7 | 0.79019 | 0.65148 | 0.00272 | 0.27253 | 0.82561 | 0.70557 | 0.01362 | 0.24273 | 38.5 | 2 |
| VGG16 | fgvc7 | 0.79292 | 0.63261 | 0.00817 | 0.31216 | 0.91417 | 0.80363 | 0.00681 | 0.22799 | 68.5 | 9.5 |
| VGG19 | fgvc7 | 0.79428 | 0.62217 | 0.00409 | 0.33410 | 0.89237 | 0.78148 | 0.01771 | 0.23919 | 57 | 7 |
| Xception | fgvc7 | 0.76839 | 0.59111 | 0.00545 | 0.34203 | 0.82834 | 0.71290 | 0.02725 | 0.24180 | 31.5 | 2 |
| ConvNeXtSmall | fgvc8 | 0.80287 | 0.44037 | 0.00067 | 0.38906 | 0.83892 | 0.46703 | 0.00616 | 0.39645 | 86 | 5.5 |
| ConvNeXtTiny | fgvc8 | 0.80193 | 0.46220 | 0.00027 | 0.36778 | 0.84723 | 0.51126 | 0.00322 | 0.36425 | 78.5 | 7 |
| DenseNet121 | fgvc8 | 0.77057 | 0.42374 | 0.00161 | 0.36817 | 0.81694 | 0.45988 | 0.00027 | 0.37844 | 56 | 4.5 |
| DenseNet169 | fgvc8 | 0.76441 | 0.41911 | 0.00027 | 0.37086 | 0.82109 | 0.48338 | 0.00389 | 0.37173 | 38 | 7.5 |
| DenseNet201 | fgvc8 | 0.75905 | 0.41633 | 0.00456 | 0.36461 | 0.80434 | 0.49161 | 0.01555 | 0.35034 | 44.5 | 3 |
| EfficientNetV2B0 | fgvc8 | 0.80032 | 0.43265 | 0.00107 | 0.38954 | 0.82337 | 0.47979 | 0.00322 | 0.37342 | 47 | 7.5 |
| EfficientNetV2B1 | fgvc8 | 0.80582 | 0.44370 | 0.00067 | 0.38564 | 0.84093 | 0.48930 | 0.00121 | 0.37715 | 45.5 | 6.5 |
| EfficientNetV2B2 | fgvc8 | 0.79831 | 0.43618 | 0.00174 | 0.38533 | 0.83516 | 0.47725 | 0.00241 | 0.38311 | 41 | 6 |
| EfficientNetV2B3 | fgvc8 | 0.79241 | 0.43647 | 0.00067 | 0.37907 | 0.82726 | 0.48462 | 0.00737 | 0.37149 | 43 | 2.5 |
| EfficientNetV2M | fgvc8 | 0.76159 | 0.39810 | 0.00121 | 0.38204 | 0.83128 | 0.49593 | 0.00201 | 0.36433 | 51 | 3 |
| EfficientNetV2S | fgvc8 | 0.80099 | 0.44135 | 0.00147 | 0.38013 | 0.84361 | 0.51394 | 0.00791 | 0.35942 | 51 | 2.5 |
| InceptionResNetV2 | fgvc8 | 0.74765 | 0.39817 | 0.00281 | 0.36445 | 0.80541 | 0.47069 | 0.00750 | 0.35894 | 60 | 2 |

Continued on next page

**Table C.2 - continued from previous page**

| Model | Dataset | Acc | F1 | Acc SD | F-1 SD | Acc FT | F1 FT | Acc SD FT | F-1 SD FT | Best Epoch | Best Epoch FT |
|---|---|---|---|---|---|---|---|---|---|---|---|
| InceptionV3 | fgvc8 | 0.68681 | 0.35759 | 0.00094 | 0.34461 | 0.79804 | 0.44809 | 0.01220 | 0.36924 | 25 | 4 |
| MobileNetV3Large | fgvc8 | 0.78196 | 0.41994 | 0.00255 | 0.38186 | 0.83463 | 0.48118 | 0.00724 | 0.37869 | 32 | 9 |
| MobileNetV3Small | fgvc8 | 0.76374 | 0.40846 | 0.00281 | 0.37619 | 0.81238 | 0.45943 | 0.00295 | 0.37302 | 51.5 | 25.5 |
| NASNetLarge | fgvc8 | 0.67649 | 0.35903 | 0.00456 | 0.33558 | 0.78397 | 0.44383 | 0.03189 | 0.36298 | 25 | 2 |
| NASNetMobile | fgvc8 | 0.66671 | 0.34277 | 0.00228 | 0.34291 | 0.75771 | 0.43410 | 0.00884 | 0.35203 | 45.5 | 6.5 |
| ResNet101V2 | fgvc8 | 0.70866 | 0.38230 | 0.00456 | 0.34800 | 0.78169 | 0.45845 | 0.01756 | 0.35464 | 24.5 | 3.5 |
| ResNet152V2 | fgvc8 | 0.70209 | 0.37233 | 0.00094 | 0.35390 | 0.78009 | 0.43756 | 0.02024 | 0.36509 | 25 | 2.5 |
| ResNet50V2 | fgvc8 | 0.72755 | 0.39012 | 0.00013 | 0.35888 | 0.77191 | 0.45005 | 0.01689 | 0.35451 | 24.5 | 2.5 |
| VGG16 | fgvc8 | 0.71000 | 0.39447 | 0.00241 | 0.33716 | 0.82806 | 0.49054 | 0.00925 | 0.36455 | 46.5 | 7.5 |
| VGG19 | fgvc8 | 0.68655 | 0.36973 | 0.00255 | 0.33881 | 0.82337 | 0.47576 | 0.01474 | 0.37343 | 40 | 6 |
| Xception | fgvc8 | 0.68681 | 0.35185 | 0.00228 | 0.35145 | 0.77137 | 0.44130 | 0.02010 | 0.35655 | 34 | 3 |
| ConvNeXtSmall | novelPotato | 0.75606 | 0.73006 | 0.00485 | 0.10755 | 0.79725 | 0.77121 | 0.00242 | 0.11059 | 71 | 19.5 |
| ConvNeXtTiny | novelPotato | 0.72132 | 0.69857 | 0.00727 | 0.10969 | 0.78675 | 0.75902 | 0.00323 | 0.10402 | 93.5 | 22.5 |
| DenseNet121 | novelPotato | 0.74637 | 0.72272 | 0.00323 | 0.10347 | 0.78271 | 0.76615 | 0.00081 | 0.09476 | 64 | 5 |
| DenseNet169 | novelPotato | 0.74071 | 0.72908 | 0.00889 | 0.08927 | 0.76979 | 0.74928 | 0.00889 | 0.10239 | 50 | 4 |
| DenseNet201 | novelPotato | 0.74960 | 0.73595 | 0.01454 | 0.08407 | 0.78918 | 0.76043 | 0.03312 | 0.11420 | 54.5 | 2 |

Continued on next page

**Table C.2 - continued from previous page**

| Model | Dataset | Acc | F1 | Acc SD | F-1 SD | Acc FT | F1 FT | Acc SD FT | F-1 SD FT | Best Epoch | Best Epoch FT |
|---|---|---|---|---|---|---|---|---|---|---|---|
| EfficientNetV2B0 | novelPotato | 0.77868 | 0.74319 | 0.00485 | 0.13013 | 0.80775 | 0.78651 | 0.01131 | 0.09571 | 38 | 6 |
| EfficientNetV2B1 | novelPotato | 0.77302 | 0.74951 | 0.00404 | 0.11690 | 0.81987 | 0.80592 | 0.00242 | 0.09205 | 51.5 | 9.5 |
| EfficientNetV2B2 | novelPotato | 0.78595 | 0.75768 | 0.00242 | 0.10515 | 0.80210 | 0.76551 | 0.00404 | 0.12744 | 54 | 5 |
| EfficientNetV2B3 | novelPotato | 0.75848 | 0.71698 | 0.00565 | 0.14877 | 0.79725 | 0.75150 | 0.01212 | 0.14230 | 43.5 | 3 |
| EfficientNetV2M | novelPotato | 0.72294 | 0.68139 | 0.00404 | 0.13412 | 0.79806 | 0.78217 | 0.02262 | 0.08672 | 57 | 4.5 |
| EfficientNetV2S | novelPotato | 0.77868 | 0.72141 | 0.00162 | 0.14792 | 0.80937 | 0.78871 | 0.01292 | 0.09410 | 57.5 | 2.5 |
| InceptionResNetV2 | novelPotato | 0.67124 | 0.62520 | 0.00404 | 0.13814 | 0.70840 | 0.68100 | 0.01696 | 0.11317 | 73.5 | 2 |
| InceptionV3 | novelPotato | 0.61551 | 0.54641 | 0.01292 | 0.18519 | 0.68821 | 0.62337 | 0.02746 | 0.20032 | 31.5 | 3 |
| MobileNetV3Large | novelPotato | 0.77464 | 0.73683 | 0.00081 | 0.12030 | 0.78998 | 0.74740 | 0.00323 | 0.13979 | 49 | 4.5 |
| MobileNetV3Small | novelPotato | 0.74798 | 0.73297 | 0.01131 | 0.10175 | 0.76737 | 0.74989 | 0.01777 | 0.10368 | 67 | 7 |
| NASNetLarge | novelPotato | 0.65105 | 0.60857 | 0.00969 | 0.13528 | 0.71325 | 0.67753 | 0.02989 | 0.15628 | 25 | 1.5 |
| NASNetMobile | novelPotato | 0.65428 | 0.57941 | 0.00000 | 0.20936 | 0.70679 | 0.64430 | 0.01696 | 0.17954 | 58 | 3.5 |
| ResNet101V2 | novelPotato | 0.72294 | 0.66344 | 0.00565 | 0.17463 | 0.73829 | 0.71156 | 0.01292 | 0.11916 | 37.5 | 1.5 |
| ResNet152V2 | novelPotato | 0.67205 | 0.60692 | 0.00808 | 0.18439 | 0.73506 | 0.71535 | 0.03231 | 0.11212 | 29.5 | 2 |
| ResNet50V2 | novelPotato | 0.67528 | 0.61586 | 0.00323 | 0.20776 | 0.70113 | 0.66073 | 0.01292 | 0.16085 | 34.5 | 1 |
| VGG16 | novelPotato | 0.68821 | 0.67304 | 0.00323 | 0.09807 | 0.78110 | 0.75588 | 0.02342 | 0.09780 | 79.5 | 9.5 |

Continued on next page

**Table C.2 - continued from previous page**

| Model | Dataset | Acc | F1 | Acc SD | F-1 SD | Acc FT | F1 FT | Acc SD FT | F-1 SD FT | Best Epoch | Best Epoch FT |
|---|---|---|---|---|---|---|---|---|---|---|---|
| VGG19 | novelPotato | 0.68255 | 0.65140 | 0.01696 | 0.11670 | 0.77948 | 0.76140 | 0.02666 | 0.08908 | 76 | 10 |
| Xception | novelPotato | 0.68336 | 0.63319 | 0.00000 | 0.16357 | 0.72698 | 0.68976 | 0.01292 | 0.14069 | 45.5 | 1.5 |
| ConvNeXtSmall | paddy | 0.92107 | 0.91175 | 0.00408 | 0.03100 | 0.93930 | 0.93103 | 0.00264 | 0.02986 | 180.5 | 19 |
| ConvNeXtTiny | paddy | 0.92035 | 0.91148 | 0.00096 | 0.03798 | 0.93930 | 0.93078 | 0.00648 | 0.03381 | 188 | 26.5 |
| DenseNet121 | paddy | 0.87908 | 0.86925 | 0.00384 | 0.04213 | 0.91579 | 0.90444 | 0.01080 | 0.03598 | 143 | 7 |
| DenseNet169 | paddy | 0.89923 | 0.88820 | 0.00336 | 0.04266 | 0.92922 | 0.91750 | 0.00984 | 0.04038 | 111.5 | 12 |
| DenseNet201 | paddy | 0.90787 | 0.90417 | 0.00192 | 0.03148 | 0.93666 | 0.92970 | 0.00528 | 0.03129 | 99 | 12.5 |
| EfficientNetV2B0 | paddy | 0.91963 | 0.91048 | 0.00264 | 0.03148 | 0.92370 | 0.91587 | 0.00672 | 0.03184 | 137.5 | 1.5 |
| EfficientNetV2B1 | paddy | 0.92850 | 0.91755 | 0.00336 | 0.03718 | 0.94218 | 0.93288 | 0.00120 | 0.02855 | 120 | 23 |
| EfficientNetV2B2 | paddy | 0.91939 | 0.91090 | 0.00000 | 0.03491 | 0.92922 | 0.92256 | 0.00504 | 0.03593 | 104 | 17 |
| EfficientNetV2B3 | paddy | 0.92202 | 0.91524 | 0.00312 | 0.02812 | 0.93330 | 0.92464 | 0.00816 | 0.02978 | 107 | 3.5 |
| EfficientNetV2M | paddy | 0.89803 | 0.88608 | 0.00024 | 0.03544 | 0.93690 | 0.92823 | 0.00600 | 0.02945 | 142.5 | 15 |
| EfficientNetV2S | paddy | 0.91483 | 0.90696 | 0.00744 | 0.03267 | 0.93738 | 0.92886 | 0.00984 | 0.02772 | 115.5 | 14.5 |
| InceptionResNetV2 | paddy | 0.84405 | 0.82390 | 0.00480 | 0.05722 | 0.91099 | 0.89987 | 0.00840 | 0.03844 | 104 | 13 |
| InceptionV3 | paddy | 0.80686 | 0.78986 | 0.00792 | 0.06084 | 0.90115 | 0.89162 | 0.00960 | 0.04031 | 60.5 | 25.5 |
| MobileNetV3Large | paddy | 0.90499 | 0.89595 | 0.00096 | 0.04295 | 0.92298 | 0.91352 | 0.00312 | 0.03440 | 112.5 | 15 |

Continued on next page

**Table C.2 - continued from previous page**

| Model | Dataset | Acc | F1 | Acc SD | F-1 SD | Acc FT | F1 FT | Acc SD FT | F-1 SD FT | Best Epoch | Best Epoch FT |
|---|---|---|---|---|---|---|---|---|---|---|---|
| MobileNetV3Small | paddy | 0.88580 | 0.87581 | 0.00144 | 0.03723 | 0.90571 | 0.89631 | 0.00504 | 0.03444 | 140.5 | 31.5 |
| NASNetLarge | paddy | 0.81958 | 0.80516 | 0.00576 | 0.05489 | 0.91267 | 0.89951 | 0.00480 | 0.03445 | 67 | 35.5 |
| NASNetMobile | paddy | 0.80902 | 0.78593 | 0.00240 | 0.06561 | 0.88220 | 0.86636 | 0.01272 | 0.05516 | 87 | 12.5 |
| ResNet101V2 | paddy | 0.86204 | 0.85447 | 0.00312 | 0.03967 | 0.91963 | 0.90913 | 0.00696 | 0.04142 | 71 | 10.5 |
| ResNet152V2 | paddy | 0.85581 | 0.83416 | 0.00408 | 0.05529 | 0.93018 | 0.92295 | 0.00120 | 0.03177 | 73 | 51 |
| ResNet50V2 | paddy | 0.84789 | 0.83700 | 0.00048 | 0.05411 | 0.90931 | 0.89647 | 0.00528 | 0.04132 | 66 | 9.5 |
| VGG16 | paddy | 0.84717 | 0.83798 | 0.00360 | 0.04570 | 0.91915 | 0.90873 | 0.00888 | 0.03416 | 132.5 | 24.5 |
| VGG19 | paddy | 0.84549 | 0.83944 | 0.00240 | 0.03999 | 0.92562 | 0.91938 | 0.01200 | 0.02698 | 152 | 25.5 |
| Xception | paddy | 0.83373 | 0.81732 | 0.00024 | 0.04852 | 0.87812 | 0.86039 | 0.01583 | 0.05227 | 88 | 4 |
| ConvNeXtSmall | pdd271 | 0.99076 | 0.99164 | 0.00066 | 0.00817 | 0.99406 | 0.99448 | 0.00000 | 0.00474 | 151.5 | 172.5 |
| ConvNeXtTiny | pdd271 | 0.99109 | 0.99133 | 0.00033 | 0.00769 | 0.99274 | 0.99288 | 0.00066 | 0.00611 | 180.5 | 120 |
| DenseNet121 | pdd271 | 0.98713 | 0.98775 | 0.00099 | 0.01201 | 0.99406 | 0.99403 | 0.00132 | 0.00531 | 85 | 25 |
| DenseNet169 | pdd271 | 0.98977 | 0.98904 | 0.00099 | 0.00889 | 0.99670 | 0.99657 | 0.00066 | 0.00428 | 56.5 | 95 |
| DenseNet201 | pdd271 | 0.98581 | 0.98546 | 0.00033 | 0.01059 | 0.99472 | 0.99461 | 0.00198 | 0.00517 | 82.5 | 52 |
| EfficientNetV2B0 | pdd271 | 0.98779 | 0.98817 | 0.00165 | 0.00775 | 0.99373 | 0.99395 | 0.00099 | 0.00359 | 80.5 | 177 |
| EfficientNetV2B1 | pdd271 | 0.98548 | 0.98496 | 0.00066 | 0.00878 | 0.99208 | 0.99200 | 0.00132 | 0.00564 | 68 | 67.5 |

Continued on next page

**Table C.2 - continued from previous page**

| Model | Dataset | Acc | F1 | Acc SD | F-1 SD | Acc FT | F1 FT | Acc SD FT | F-1 SD FT | Best Epoch | Best Epoch FT |
|---|---|---|---|---|---|---|---|---|---|---|---|
| EfficientNetV2B2 | pdd271 | 0.99109 | 0.99130 | 0.00099 | 0.00705 | 0.99439 | 0.99466 | 0.00033 | 0.00465 | 113.5 | 49 |
| EfficientNetV2B3 | pdd271 | 0.98779 | 0.98766 | 0.00099 | 0.00829 | 0.98911 | 0.98912 | 0.00297 | 0.00817 | 139 | 110.5 |
| EfficientNetV2M | pdd271 | 0.98548 | 0.98577 | 0.00066 | 0.00971 | 0.99274 | 0.99294 | 0.00000 | 0.00616 | 111 | 91.5 |
| EfficientNetV2S | pdd271 | 0.98548 | 0.98444 | 0.00066 | 0.00990 | 0.99109 | 0.99121 | 0.00033 | 0.00636 | 92 | 38.5 |
| InceptionResNetV2 | pdd271 | 0.96139 | 0.95998 | 0.00033 | 0.02858 | 0.98845 | 0.98818 | 0.00099 | 0.00985 | 80 | 126.5 |
| InceptionV3 | pdd271 | 0.95083 | 0.94697 | 0.00033 | 0.03588 | 0.98779 | 0.98679 | 0.00165 | 0.01055 | 47.5 | 47.5 |
| MobileNetV3Large | pdd271 | 0.98350 | 0.98357 | 0.00000 | 0.00995 | 0.99208 | 0.99270 | 0.00132 | 0.00663 | 76.5 | 120 |
| MobileNetV3Small | pdd271 | 0.97525 | 0.97387 | 0.00033 | 0.01588 | 0.98647 | 0.98627 | 0.00099 | 0.00731 | 68.5 | 122 |
| NASNetLarge | pdd271 | 0.94752 | 0.94423 | 0.00165 | 0.03676 | 0.98548 | 0.98495 | 0.00066 | 0.01113 | 52 | 53.5 |
| NASNetMobile | pdd271 | 0.94389 | 0.93925 | 0.00198 | 0.03974 | 0.98614 | 0.98622 | 0.00066 | 0.01216 | 71 | 129.5 |
| ResNet101V2 | pdd271 | 0.96535 | 0.96302 | 0.00231 | 0.02391 | 0.98614 | 0.98573 | 0.00000 | 0.01039 | 41 | 49.5 |
| ResNet152V2 | pdd271 | 0.96073 | 0.95941 | 0.00231 | 0.02845 | 0.99109 | 0.99079 | 0.00033 | 0.00719 | 47 | 56.5 |
| ResNet50V2 | pdd271 | 0.96205 | 0.96058 | 0.00231 | 0.02625 | 0.98515 | 0.98509 | 0.00033 | 0.00921 | 49.5 | 139.5 |
| VGG16 | pdd271 | 0.97426 | 0.97441 | 0.00066 | 0.01602 | 0.98977 | 0.98964 | 0.00033 | 0.00476 | 95 | 29.5 |
| VGG19 | pdd271 | 0.97129 | 0.97073 | 0.00033 | 0.01954 | 0.99340 | 0.99330 | 0.00000 | 0.00497 | 85 | 60.5 |
| Xception | pdd271 | 0.95743 | 0.95626 | 0.00231 | 0.02898 | 0.98185 | 0.98138 | 0.00099 | 0.01389 | 65 | 106 |

Continued on next page

**Table C.2 - continued from previous page**

| Model | Dataset | Acc | F1 | Acc SD | F-1 SD | Acc FT | F1 FT | Acc SD FT | F-1 SD FT | Best Epoch | Best Epoch FT |
|---|---|---|---|---|---|---|---|---|---|---|---|
| ConvNeXtSmall | plantDoc | 0.50633 | 0.47354 | 0.01266 | 0.24134 | 0.54219 | 0.51282 | 0.01477 | 0.24550 | 120 | 10.5 |
| ConvNeXtTiny | plantDoc | 0.51055 | 0.48895 | 0.00422 | 0.24082 | 0.56118 | 0.53864 | 0.00000 | 0.23024 | 106 | 26 |
| DenseNet121 | plantDoc | 0.52743 | 0.49128 | 0.00844 | 0.24912 | 0.55485 | 0.52134 | 0.00633 | 0.26440 | 100.5 | 3 |
| DenseNet169 | plantDoc | 0.54430 | 0.51867 | 0.00844 | 0.23776 | 0.53797 | 0.53326 | 0.01055 | 0.25685 | 74 | 4.5 |
| DenseNet201 | plantDoc | 0.54219 | 0.50797 | 0.01055 | 0.22551 | 0.54008 | 0.50536 | 0.01266 | 0.24440 | 74 | 2 |
| EfficientNetV2B0 | plantDoc | 0.55063 | 0.52007 | 0.00633 | 0.23558 | 0.58017 | 0.54747 | 0.02743 | 0.23682 | 82.5 | 1 |
| EfficientNetV2B1 | plantDoc | 0.52110 | 0.49484 | 0.00211 | 0.22620 | 0.53586 | 0.52806 | 0.00422 | 0.23081 | 82.5 | 1.5 |
| EfficientNetV2B2 | plantDoc | 0.54219 | 0.51480 | 0.01055 | 0.24146 | 0.56962 | 0.55690 | 0.00844 | 0.23671 | 69.5 | 2.5 |
| EfficientNetV2B3 | plantDoc | 0.54008 | 0.51097 | 0.00000 | 0.24688 | 0.55063 | 0.53645 | 0.01055 | 0.22282 | 66 | 1 |
| EfficientNetV2M | plantDoc | 0.49578 | 0.46099 | 0.01055 | 0.21192 | 0.52532 | 0.49595 | 0.02743 | 0.22438 | 74.5 | 1 |
| EfficientNetV2S | plantDoc | 0.55696 | 0.52960 | 0.00422 | 0.21822 | 0.55907 | 0.53078 | 0.00633 | 0.20914 | 85 | 2.5 |
| InceptionResNetV2 | plantDoc | 0.47257 | 0.44077 | 0.01266 | 0.25479 | 0.49367 | 0.48928 | 0.01688 | 0.25996 | 89 | 1.5 |
| InceptionV3 | plantDoc | 0.47468 | 0.43594 | 0.00211 | 0.21667 | 0.48101 | 0.44127 | 0.00844 | 0.25704 | 55.5 | 1 |
| MobileNetV3Large | plantDoc | 0.52954 | 0.49688 | 0.00633 | 0.23420 | 0.54641 | 0.51422 | 0.00211 | 0.23648 | 63.5 | 3.5 |
| MobileNetV3Small | plantDoc | 0.47890 | 0.44954 | 0.01055 | 0.21881 | 0.50211 | 0.47298 | 0.01266 | 0.22570 | 109 | 3.5 |
| NASNetLarge | plantDoc | 0.45781 | 0.43575 | 0.00211 | 0.20073 | 0.46414 | 0.43497 | 0.00000 | 0.20437 | 40.5 | 1 |

Continued on next page

**Table C.2 - continued from previous page**

| Model | Dataset | Acc | F1 | Acc SD | F-1 SD | Acc FT | F1 FT | Acc SD FT | F-1 SD FT | Best Epoch | Best Epoch FT |
|---|---|---|---|---|---|---|---|---|---|---|---|
| NASNetMobile | plantDoc | 0.46835 | 0.43717 | 0.00000 | 0.21926 | 0.46835 | 0.43422 | 0.01266 | 0.22485 | 70 | 1.5 |
| ResNet101V2 | plantDoc | 0.48101 | 0.45639 | 0.00422 | 0.25660 | 0.50211 | 0.48097 | 0.01688 | 0.24731 | 44 | 1 |
| ResNet152V2 | plantDoc | 0.46624 | 0.43658 | 0.00211 | 0.22992 | 0.48734 | 0.46305 | 0.00211 | 0.26638 | 42 | 1.5 |
| ResNet50V2 | plantDoc | 0.48734 | 0.46941 | 0.01477 | 0.23171 | 0.47890 | 0.48336 | 0.01477 | 0.24384 | 47 | 2 |
| VGG16 | plantDoc | 0.40506 | 0.38695 | 0.00422 | 0.21006 | 0.46414 | 0.43812 | 0.02954 | 0.23697 | 96.5 | 3 |
| VGG19 | plantDoc | 0.41772 | 0.38441 | 0.00000 | 0.24315 | 0.41772 | 0.37304 | 0.03376 | 0.24476 | 87 | 1.5 |
| Xception | plantDoc | 0.42827 | 0.39659 | 0.00633 | 0.20465 | 0.48523 | 0.47204 | 0.02532 | 0.22724 | 65 | 1 |
| ConvNeXtSmall | plantVillage | 0.98667 | 0.98182 | 0.00009 | 0.02560 | 0.99568 | 0.99320 | 0.00000 | 0.01423 | 113.5 | 62.5 |
| ConvNeXtTiny | plantVillage | 0.98667 | 0.97984 | 0.00028 | 0.03331 | 0.99600 | 0.99376 | 0.00041 | 0.01214 | 113.5 | 149.5 |
| DenseNet121 | plantVillage | 0.98299 | 0.97683 | 0.00037 | 0.03026 | 0.99632 | 0.99461 | 0.00009 | 0.00981 | 93 | 131 |
| DenseNet169 | plantVillage | 0.98271 | 0.97616 | 0.00018 | 0.03253 | 0.99637 | 0.99473 | 0.00023 | 0.01079 | 43 | 36 |
| DenseNet201 | plantVillage | 0.98750 | 0.98230 | 0.00046 | 0.02593 | 0.99646 | 0.99454 | 0.00005 | 0.01315 | 57.5 | 51 |
| EfficientNetV2B0 | plantVillage | 0.98722 | 0.98116 | 0.00028 | 0.02533 | 0.99554 | 0.99290 | 0.00051 | 0.01532 | 60 | 63 |
| EfficientNetV2B1 | plantVillage | 0.98796 | 0.98255 | 0.00037 | 0.02299 | 0.99513 | 0.99283 | 0.00064 | 0.01147 | 84 | 62.5 |
| EfficientNetV2B2 | plantVillage | 0.98750 | 0.98284 | 0.00000 | 0.02381 | 0.99481 | 0.99177 | 0.00051 | 0.01537 | 89 | 33.5 |
| EfficientNetV2B3 | plantVillage | 0.98506 | 0.97990 | 0.00032 | 0.02805 | 0.99545 | 0.99359 | 0.00032 | 0.01214 | 50.5 | 61.5 |

Continued on next page

**Table C.2 - continued from previous page**

| Model | Dataset | Acc | F1 | Acc SD | F-1 SD | Acc FT | F1 FT | Acc SD FT | F-1 SD FT | Best Epoch | Best Epoch FT |
|---|---|---|---|---|---|---|---|---|---|---|---|
| EfficientNetV2M | plantVillage | 0.98129 | 0.97569 | 0.00023 | 0.03028 | 0.99692 | 0.99524 | 0.00032 | 0.00903 | 89.5 | 85 |
| EfficientNetV2S | plantVillage | 0.98800 | 0.98405 | 0.00014 | 0.02116 | 0.99660 | 0.99496 | 0.00000 | 0.01157 | 81 | 39 |
| InceptionResNetV2 | plantVillage | 0.96506 | 0.95263 | 0.00028 | 0.05297 | 0.99099 | 0.98708 | 0.00211 | 0.01711 | 88.5 | 13 |
| InceptionV3 | plantVillage | 0.94221 | 0.92675 | 0.00106 | 0.06987 | 0.99039 | 0.98604 | 0.00106 | 0.02046 | 51.5 | 59 |
| MobileNetV3Large | plantVillage | 0.98676 | 0.98236 | 0.00083 | 0.02475 | 0.99361 | 0.99131 | 0.00023 | 0.01302 | 64.5 | 63 |
| MobileNetV3Small | plantVillage | 0.98396 | 0.97804 | 0.00023 | 0.02832 | 0.99159 | 0.98856 | 0.00051 | 0.01688 | 72 | 84 |
| NASNetLarge | plantVillage | 0.94373 | 0.92686 | 0.00129 | 0.06987 | 0.99159 | 0.98858 | 0.00069 | 0.01821 | 41 | 68 |
| NASNetMobile | plantVillage | 0.94874 | 0.93240 | 0.00005 | 0.06378 | 0.98740 | 0.98310 | 0.00092 | 0.02206 | 80.5 | 76 |
| ResNet101V2 | plantVillage | 0.96998 | 0.95885 | 0.00005 | 0.04571 | 0.99104 | 0.98707 | 0.00106 | 0.01924 | 33.5 | 37 |
| ResNet152V2 | plantVillage | 0.96865 | 0.95880 | 0.00009 | 0.04922 | 0.99223 | 0.98956 | 0.00097 | 0.01454 | 41 | 37 |
| ResNet50V2 | plantVillage | 0.96980 | 0.96002 | 0.00051 | 0.04454 | 0.99030 | 0.98686 | 0.00143 | 0.01879 | 45 | 56 |
| VGG16 | plantVillage | 0.97232 | 0.96422 | 0.00055 | 0.03996 | 0.99338 | 0.99052 | 0.00175 | 0.01545 | 54.5 | 35 |
| VGG19 | plantVillage | 0.96897 | 0.96036 | 0.00041 | 0.04203 | 0.99572 | 0.99373 | 0.00032 | 0.00960 | 69.5 | 52.5 |
| Xception | plantVillage | 0.95380 | 0.93931 | 0.00115 | 0.06101 | 0.98506 | 0.97992 | 0.00216 | 0.02332 | 52 | 33 |
| ConvNeXtSmall | pld | 0.97160 | 0.97059 | 0.00123 | 0.00622 | 0.98272 | 0.98245 | 0.00000 | 0.00155 | 62 | 23 |
| ConvNeXtTiny | pld | 0.97531 | 0.97436 | 0.00000 | 0.00582 | 0.98642 | 0.98531 | 0.00123 | 0.00687 | 84.5 | 32.5 |

Continued on next page

**Table C.2 - continued from previous page**

| Model | Dataset | Acc | F1 | Acc SD | F-1 SD | Acc FT | F1 FT | Acc SD FT | F-1 SD FT | Best Epoch | Best Epoch FT |
|---|---|---|---|---|---|---|---|---|---|---|---|
| DenseNet121 | pld | 0.98395 | 0.98298 | 0.00123 | 0.00633 | 0.99012 | 0.98969 | 0.00000 | 0.00310 | 67.5 | 10 |
| DenseNet169 | pld | 0.97778 | 0.97673 | 0.00000 | 0.00622 | 0.98765 | 0.98648 | 0.00494 | 0.00854 | 35.5 | 8.5 |
| DenseNet201 | pld | 0.97901 | 0.97840 | 0.00123 | 0.00563 | 0.99753 | 0.99719 | 0.00000 | 0.00206 | 44.5 | 20 |
| EfficientNetV2B0 | pld | 0.99259 | 0.99173 | 0.00000 | 0.00478 | 0.99383 | 0.99344 | 0.00123 | 0.00262 | 73.5 | 18.5 |
| EfficientNetV2B1 | pld | 0.98642 | 0.98521 | 0.00123 | 0.00702 | 0.99012 | 0.98894 | 0.00000 | 0.00667 | 43 | 34.5 |
| EfficientNetV2B2 | pld | 0.98765 | 0.98692 | 0.00000 | 0.00438 | 0.98889 | 0.98838 | 0.00123 | 0.00433 | 35 | 8.5 |
| EfficientNetV2B3 | pld | 0.98519 | 0.98390 | 0.00247 | 0.00787 | 0.99012 | 0.98894 | 0.00000 | 0.00667 | 31.5 | 4 |
| EfficientNetV2M | pld | 0.97778 | 0.97686 | 0.00247 | 0.00651 | 0.99136 | 0.99087 | 0.00123 | 0.00483 | 46.5 | 5.5 |
| EfficientNetV2S | pld | 0.98148 | 0.98018 | 0.00370 | 0.00986 | 0.98889 | 0.98802 | 0.00123 | 0.00850 | 41.5 | 23 |
| InceptionResNetV2 | pld | 0.96049 | 0.95791 | 0.00247 | 0.01643 | 0.98642 | 0.98556 | 0.00123 | 0.00545 | 72.5 | 4.5 |
| InceptionV3 | pld | 0.94198 | 0.94090 | 0.00370 | 0.00775 | 0.97778 | 0.97752 | 0.00247 | 0.00348 | 24 | 7.5 |
| MobileNetV3Large | pld | 0.98025 | 0.97915 | 0.00000 | 0.00812 | 0.98765 | 0.98716 | 0.00247 | 0.00396 | 45 | 5.5 |
| MobileNetV3Small | pld | 0.98148 | 0.98032 | 0.00123 | 0.00686 | 0.98272 | 0.98153 | 0.00000 | 0.00759 | 70.5 | 11 |
| NASNetLarge | pld | 0.95185 | 0.95047 | 0.00123 | 0.00937 | 0.98272 | 0.98268 | 0.00247 | 0.00464 | 25.5 | 5.5 |
| NASNetMobile | pld | 0.94321 | 0.94054 | 0.00247 | 0.01564 | 0.98272 | 0.98243 | 0.00247 | 0.00385 | 47.5 | 33 |
| ResNet101V2 | pld | 0.96790 | 0.96560 | 0.00247 | 0.01363 | 0.98642 | 0.98615 | 0.00123 | 0.00538 | 40 | 2 |

Continued on next page

**Table C.2 - continued from previous page**

| Model | Dataset | Acc | F1 | Acc SD | F-1 SD | Acc FT | F1 FT | Acc SD FT | F-1 SD FT | Best Epoch | Best Epoch FT |
|---|---|---|---|---|---|---|---|---|---|---|---|
| ResNet152V2 | pld | 0.97407 | 0.97235 | 0.00123 | 0.01041 | 0.98889 | 0.98823 | 0.00123 | 0.00413 | 35.5 | 5 |
| ResNet50V2 | pld | 0.97284 | 0.97173 | 0.00000 | 0.00672 | 0.99012 | 0.98930 | 0.00247 | 0.00523 | 36 | 16 |
| VGG16 | pld | 0.96420 | 0.96291 | 0.00370 | 0.00946 | 0.98148 | 0.97987 | 0.00370 | 0.01188 | 56 | 9 |
| VGG19 | pld | 0.96790 | 0.96604 | 0.00000 | 0.01119 | 0.99136 | 0.99060 | 0.00123 | 0.00579 | 51 | 18 |
| Xception | pld | 0.95309 | 0.95136 | 0.00247 | 0.01092 | 0.98025 | 0.97908 | 0.00000 | 0.00771 | 36 | 7 |
| ConvNeXtSmall | rldd | 0.98095 | 0.97236 | 0.00000 | 0.02157 | 0.97619 | 0.96606 | 0.00476 | 0.02694 | 190 | 48 |
| ConvNeXtTiny | rldd | 0.96667 | 0.95149 | 0.00476 | 0.03850 | 0.96667 | 0.95149 | 0.00476 | 0.03850 | 128 | 3 |
| DenseNet121 | rldd | 0.97143 | 0.95805 | 0.00000 | 0.03316 | 0.97143 | 0.95853 | 0.00952 | 0.03680 | 170 | 36.5 |
| DenseNet169 | rldd | 0.97143 | 0.95805 | 0.00000 | 0.03316 | 0.97143 | 0.95805 | 0.00000 | 0.03316 | 184.5 | 14 |
| DenseNet201 | rldd | 0.98571 | 0.97934 | 0.00476 | 0.01836 | 0.98095 | 0.97236 | 0.00000 | 0.02157 | 107.5 | 38.5 |
| EfficientNetV2B0 | rldd | 0.97143 | 0.95805 | 0.00000 | 0.03316 | 0.98095 | 0.97236 | 0.00000 | 0.02157 | 105.5 | 34.5 |
| EfficientNetV2B1 | rldd | 0.96667 | 0.95070 | 0.00476 | 0.04042 | 0.97619 | 0.96520 | 0.00476 | 0.02887 | 84.5 | 55.5 |
| EfficientNetV2B2 | rldd | 0.97143 | 0.95805 | 0.00000 | 0.03316 | 0.98095 | 0.97236 | 0.00000 | 0.02157 | 155.5 | 7.5 |
| EfficientNetV2B3 | rldd | 0.97619 | 0.96566 | 0.00476 | 0.02788 | 0.98095 | 0.97236 | 0.00000 | 0.02157 | 92 | 12.5 |
| EfficientNetV2M | rldd | 0.93333 | 0.90623 | 0.00952 | 0.06875 | 0.96667 | 0.95070 | 0.00476 | 0.04042 | 185.5 | 34.5 |
| EfficientNetV2S | rldd | 0.97143 | 0.95938 | 0.00952 | 0.03437 | 0.97143 | 0.95853 | 0.00952 | 0.03680 | 159 | 42 |

Continued on next page

Table C.2 - continued from previous page

| Model | Dataset | Acc | F1 | Acc SD | F-1 SD | Acc FT | F1 FT | Acc SD FT | F-1 SD FT | Best Epoch | Best Epoch FT |
|---|---|---|---|---|---|---|---|---|---|---|---|
| InceptionResNetV2 | rldd | 0.92857 | 0.91307 | 0.00476 | 0.03981 | 0.93810 | 0.92203 | 0.00476 | 0.04041 | 101.5 | 8.5 |
| InceptionV3 | rldd | 0.94286 | 0.92796 | 0.00952 | 0.03921 | 0.95714 | 0.94483 | 0.00476 | 0.03163 | 66.5 | 24 |
| MobileNetV3Large | rldd | 0.97619 | 0.96606 | 0.00476 | 0.02694 | 0.98095 | 0.97236 | 0.00000 | 0.02157 | 166.5 | 63 |
| MobileNetV3Small | rldd | 0.97143 | 0.96100 | 0.00000 | 0.02721 | 0.97143 | 0.95805 | 0.00000 | 0.03316 | 193.5 | 63.5 |
| NASNetLarge | rldd | 0.94286 | 0.92054 | 0.00000 | 0.05590 | 0.95714 | 0.93806 | 0.01429 | 0.05378 | 105.5 | 4.5 |
| NASNetMobile | rldd | 0.91905 | 0.89776 | 0.00476 | 0.05344 | 0.97619 | 0.96606 | 0.00476 | 0.02694 | 74 | 27.5 |
| ResNet101V2 | rldd | 0.96190 | 0.95032 | 0.00000 | 0.02946 | 0.97143 | 0.96120 | 0.00000 | 0.02712 | 79.5 | 30 |
| ResNet152V2 | rldd | 0.95238 | 0.93484 | 0.00000 | 0.04567 | 0.96667 | 0.95184 | 0.00476 | 0.03848 | 95.5 | 10.5 |
| ResNet50V2 | rldd | 0.94762 | 0.93046 | 0.00476 | 0.04311 | 0.96190 | 0.94612 | 0.00000 | 0.04039 | 72 | 15.5 |
| VGG16 | rldd | 0.94762 | 0.92825 | 0.00476 | 0.04784 | 0.95714 | 0.94312 | 0.00476 | 0.03485 | 196.5 | 33 |
| VGG19 | rldd | 0.95238 | 0.92966 | 0.00000 | 0.05511 | 0.96667 | 0.95138 | 0.00476 | 0.03905 | 187 | 24.5 |
| Xception | rldd | 0.93333 | 0.91580 | 0.00952 | 0.04531 | 0.94762 | 0.93163 | 0.00476 | 0.04307 | 92 | 7.5 |
| ConvNeXtSmall | sms | 0.92500 | 0.92522 | 0.00543 | 0.04110 | 0.92935 | 0.92962 | 0.00761 | 0.04162 | 173 | 1 |
| ConvNeXtTiny | sms | 0.94674 | 0.94681 | 0.00109 | 0.02962 | 0.95652 | 0.95666 | 0.00217 | 0.02500 | 119 | 51 |
| DenseNet121 | sms | 0.92609 | 0.92600 | 0.00000 | 0.04133 | 0.94565 | 0.94573 | 0.00000 | 0.03083 | 139.5 | 23 |
| DenseNet169 | sms | 0.93370 | 0.93388 | 0.00761 | 0.03966 | 0.95543 | 0.95551 | 0.00978 | 0.02695 | 194.5 | 46.5 |

Continued on next page

**Table C.2 - continued from previous page**

| **Model** | **Dataset** | **Acc** | **F1** | **Acc SD** | **F-1 SD** | **Acc FT** | **F1 FT** | **Acc SD FT** | **F-1 SD FT** | **Best Epoch** | **Best Epoch FT** |
|---|---|---|---|---|---|---|---|---|---|---|---|
| DenseNet201 | sms | 0.92609 | 0.92588 | 0.00870 | 0.03811 | 0.94565 | 0.94588 | 0.00217 | 0.03379 | 60 | 49 |
| EfficientNetV2B0 | sms | 0.92826 | 0.92744 | 0.02391 | 0.04596 | 0.94891 | 0.94897 | 0.00109 | 0.02582 | 152 | 17.5 |
| EfficientNetV2B1 | sms | 0.93696 | 0.93665 | 0.00435 | 0.03042 | 0.93913 | 0.93891 | 0.00217 | 0.03034 | 189.5 | 148.5 |
| EfficientNetV2B2 | sms | 0.91304 | 0.91289 | 0.00435 | 0.05656 | 0.91196 | 0.91175 | 0.00109 | 0.05721 | 195 | 167 |
| EfficientNetV2B3 | sms | 0.94457 | 0.94458 | 0.00761 | 0.03108 | 0.95000 | 0.95009 | 0.00435 | 0.02763 | 139 | 198 |
| EfficientNetV2M | sms | 0.92935 | 0.93018 | 0.00109 | 0.02587 | 0.93696 | 0.93731 | 0.00217 | 0.02768 | 77.5 | 13.5 |
| EfficientNetV2S | sms | 0.94239 | 0.94238 | 0.00109 | 0.02196 | 0.94783 | 0.94785 | 0.00217 | 0.02565 | 137 | 176 |
| InceptionResNetV2 | sms | 0.90978 | 0.91009 | 0.00109 | 0.02137 | 0.92826 | 0.92820 | 0.00000 | 0.02951 | 190.5 | 13 |
| InceptionV3 | sms | 0.89783 | 0.89691 | 0.01522 | 0.05946 | 0.93696 | 0.93703 | 0.00870 | 0.03751 | 72 | 47 |
| MobileNetV3Large | sms | 0.93913 | 0.93867 | 0.00435 | 0.02545 | 0.94891 | 0.94875 | 0.00326 | 0.02227 | 167.5 | 192 |
| MobileNetV3Small | sms | 0.87609 | 0.87155 | 0.01304 | 0.06509 | 0.91522 | 0.91418 | 0.00217 | 0.03671 | 90.5 | 77.5 |
| NASNetLarge | sms | 0.84674 | 0.84852 | 0.00326 | 0.03488 | 0.87935 | 0.88119 | 0.00326 | 0.03910 | 32.5 | 18 |
| NASNetMobile | sms | 0.84348 | 0.84106 | 0.00652 | 0.07708 | 0.87174 | 0.87032 | 0.01739 | 0.07253 | 98 | 29 |
| ResNet101V2 | sms | 0.91848 | 0.91902 | 0.00543 | 0.03889 | 0.95978 | 0.96003 | 0.00761 | 0.02301 | 69 | 57 |
| ResNet152V2 | sms | 0.89783 | 0.89784 | 0.00870 | 0.03624 | 0.91630 | 0.91612 | 0.00326 | 0.02724 | 56.5 | 1.5 |
| ResNet50V2 | sms | 0.90435 | 0.90511 | 0.00870 | 0.02985 | 0.94348 | 0.94351 | 0.01957 | 0.02923 | 46 | 25 |

Continued on next page

**Table C.2 - continued from previous page**

| Model | Dataset | Acc | F1 | Acc SD | F-1 SD | Acc FT | F1 FT | Acc SD FT | F-1 SD FT | Best Epoch | Best Epoch FT |
|---|---|---|---|---|---|---|---|---|---|---|---|
| VGG16 | sms | 0.85761 | 0.85419 | 0.00326 | 0.07896 | 0.91739 | 0.91727 | 0.01739 | 0.05392 | 104 | 28.5 |
| VGG19 | sms | 0.84022 | 0.83618 | 0.01630 | 0.07131 | 0.92283 | 0.92257 | 0.00761 | 0.03738 | 48.5 | 32.5 |
| Xception | sms | 0.90326 | 0.90254 | 0.00109 | 0.05188 | 0.90435 | 0.90363 | 0.00435 | 0.05325 | 138 | 56 |
| ConvNeXtSmall | sugar | 0.91680 | 0.87303 | 0.00078 | 0.12138 | 0.91563 | 0.86859 | 0.00039 | 0.13001 | 87 | 2.5 |
| ConvNeXtTiny | sugar | 0.91369 | 0.86696 | 0.00156 | 0.13206 | 0.91446 | 0.86707 | 0.00078 | 0.13114 | 89 | 3 |
| DenseNet121 | sugar | 0.91602 | 0.87182 | 0.00233 | 0.12841 | 0.91913 | 0.87545 | 0.00233 | 0.12572 | 91.5 | 7 |
| DenseNet169 | sugar | 0.91991 | 0.87701 | 0.00311 | 0.12339 | 0.91757 | 0.87257 | 0.00078 | 0.12650 | 50 | 2.5 |
| DenseNet201 | sugar | 0.91796 | 0.87587 | 0.00039 | 0.12035 | 0.91991 | 0.87660 | 0.00078 | 0.12224 | 56 | 2.5 |
| EfficientNetV2B0 | sugar | 0.92224 | 0.88013 | 0.00078 | 0.12115 | 0.91757 | 0.87426 | 0.00000 | 0.12395 | 81.5 | 5 |
| EfficientNetV2B1 | sugar | 0.92302 | 0.88153 | 0.00078 | 0.11917 | 0.91835 | 0.87535 | 0.00078 | 0.12160 | 75.5 | 5 |
| EfficientNetV2B2 | sugar | 0.92341 | 0.88319 | 0.00117 | 0.11863 | 0.92107 | 0.87772 | 0.00739 | 0.12384 | 62 | 2.5 |
| EfficientNetV2B3 | sugar | 0.92185 | 0.88095 | 0.00039 | 0.11831 | 0.91991 | 0.87694 | 0.00233 | 0.12256 | 60.5 | 4 |
| EfficientNetV2M | sugar | 0.91213 | 0.86541 | 0.00078 | 0.13132 | 0.92030 | 0.87747 | 0.00350 | 0.12232 | 62.5 | 4 |
| EfficientNetV2S | sugar | 0.92224 | 0.88356 | 0.00078 | 0.11273 | 0.91991 | 0.87557 | 0.00078 | 0.12178 | 54.5 | 3 |
| InceptionResNetV2 | sugar | 0.91252 | 0.86760 | 0.00039 | 0.12353 | 0.91757 | 0.87403 | 0.00233 | 0.12162 | 112 | 3 |
| InceptionV3 | sugar | 0.91174 | 0.86753 | 0.00117 | 0.12410 | 0.91330 | 0.86642 | 0.00194 | 0.13068 | 58.5 | 8 |

Continued on next page

**Table C.2 - continued from previous page**

| Model | Dataset | Acc | F1 | Acc SD | F-1 SD | Acc FT | F1 FT | Acc SD FT | F-1 SD FT | Best Epoch | Best Epoch FT |
|---|---|---|---|---|---|---|---|---|---|---|---|
| MobileNetV3Large | sugar | 0.92379 | 0.88625 | 0.00078 | 0.11093 | 0.91680 | 0.87218 | 0.00000 | 0.12640 | 56.5 | 6 |
| MobileNetV3Small | sugar | 0.91796 | 0.87403 | 0.00039 | 0.12404 | 0.91446 | 0.86999 | 0.00000 | 0.12661 | 90.5 | 13 |
| NASNetLarge | sugar | 0.91213 | 0.87200 | 0.00078 | 0.11371 | 0.91524 | 0.87131 | 0.00156 | 0.12304 | 39 | 2.5 |
| NASNetMobile | sugar | 0.90669 | 0.86298 | 0.00389 | 0.12393 | 0.90941 | 0.86393 | 0.00350 | 0.12775 | 86 | 4.5 |
| ResNet101V2 | sugar | 0.91524 | 0.87193 | 0.00078 | 0.12469 | 0.91719 | 0.87220 | 0.00272 | 0.12550 | 39 | 2 |
| ResNet152V2 | sugar | 0.91252 | 0.86704 | 0.00194 | 0.13254 | 0.91757 | 0.87436 | 0.00156 | 0.12379 | 49 | 2.5 |
| ResNet50V2 | sugar | 0.91407 | 0.86857 | 0.00039 | 0.12799 | 0.91485 | 0.86772 | 0.00194 | 0.13198 | 43 | 4.5 |
| VGG16 | sugar | 0.91835 | 0.87386 | 0.00233 | 0.12472 | 0.91796 | 0.87459 | 0.00194 | 0.12410 | 91 | 3 |
| VGG19 | sugar | 0.92030 | 0.87953 | 0.00039 | 0.11689 | 0.92302 | 0.88213 | 0.00700 | 0.11524 | 84.5 | 9.5 |
| Xception | sugar | 0.90747 | 0.86222 | 0.00233 | 0.12825 | 0.91213 | 0.86438 | 0.00156 | 0.13253 | 58 | 3 |
| ConvNeXtSmall | taiwanTomato | 0.64961 | 0.61041 | 0.00394 | 0.19342 | 0.67717 | 0.62739 | 0.00787 | 0.21015 | 74 | 8.5 |
| ConvNeXtTiny | taiwanTomato | 0.69291 | 0.65217 | 0.01575 | 0.17401 | 0.71654 | 0.68086 | 0.02362 | 0.16097 | 70 | 18 |
| DenseNet121 | taiwanTomato | 0.67323 | 0.63490 | 0.00394 | 0.18072 | 0.67717 | 0.64852 | 0.01575 | 0.16432 | 89.5 | 3 |
| DenseNet169 | taiwanTomato | 0.63780 | 0.60718 | 0.00787 | 0.19693 | 0.71260 | 0.68637 | 0.01181 | 0.14401 | 58.5 | 4 |
| DenseNet201 | taiwanTomato | 0.64173 | 0.61013 | 0.01181 | 0.18384 | 0.65748 | 0.61567 | 0.01181 | 0.18329 | 52.5 | 1.5 |
| EfficientNetV2B0 | taiwanTomato | 0.65354 | 0.60240 | 0.00787 | 0.19437 | 0.69685 | 0.66654 | 0.00394 | 0.15015 | 54.5 | 4.5 |

Continued on next page

**Table C.2 - continued from previous page**

| Model | Dataset | Acc | F1 | Acc SD | F-1 SD | Acc FT | F1 FT | Acc SD FT | F-1 SD FT | Best Epoch | Best Epoch FT |
|---|---|---|---|---|---|---|---|---|---|---|---|
| EfficientNetV2B1 | taiwanTomato | 0.68110 | 0.63758 | 0.00394 | 0.18297 | 0.70079 | 0.65509 | 0.01575 | 0.19098 | 53.5 | 5.5 |
| EfficientNetV2B2 | taiwanTomato | 0.74409 | 0.71225 | 0.00394 | 0.16147 | 0.74016 | 0.71106 | 0.01575 | 0.17241 | 60 | 5.5 |
| EfficientNetV2B3 | taiwanTomato | 0.71654 | 0.67415 | 0.00787 | 0.19439 | 0.74803 | 0.71218 | 0.00787 | 0.17379 | 58.5 | 7 |
| EfficientNetV2M | taiwanTomato | 0.59055 | 0.55584 | 0.00787 | 0.20906 | 0.63780 | 0.60465 | 0.01575 | 0.18905 | 57.5 | 2 |
| EfficientNetV2S | taiwanTomato | 0.70866 | 0.66703 | 0.00787 | 0.16376 | 0.72047 | 0.67831 | 0.01969 | 0.17900 | 63 | 3 |
| InceptionResNetV2 | taiwanTomato | 0.61024 | 0.56941 | 0.01181 | 0.19420 | 0.59449 | 0.54194 | 0.06693 | 0.24530 | 81.5 | 1 |
| InceptionV3 | taiwanTomato | 0.60630 | 0.56519 | 0.01575 | 0.19871 | 0.61417 | 0.57232 | 0.04724 | 0.21336 | 40.5 | 1.5 |
| MobileNetV3Large | taiwanTomato | 0.72047 | 0.68392 | 0.00394 | 0.17217 | 0.73228 | 0.70010 | 0.00000 | 0.15974 | 71 | 1.5 |
| MobileNetV3Small | taiwanTomato | 0.65354 | 0.60044 | 0.01575 | 0.19865 | 0.66142 | 0.60057 | 0.00787 | 0.23549 | 90.5 | 1 |
| NASNetLarge | taiwanTomato | 0.62598 | 0.59094 | 0.01181 | 0.17441 | 0.64567 | 0.61207 | 0.04724 | 0.18459 | 26.5 | 2 |
| NASNetMobile | taiwanTomato | 0.57874 | 0.52630 | 0.01181 | 0.21748 | 0.64567 | 0.61967 | 0.00787 | 0.17367 | 31 | 14 |
| ResNet101V2 | taiwanTomato | 0.63386 | 0.59668 | 0.00394 | 0.18217 | 0.63386 | 0.60876 | 0.01181 | 0.17034 | 39.5 | 2 |
| ResNet152V2 | taiwanTomato | 0.57874 | 0.53880 | 0.01969 | 0.17672 | 0.64567 | 0.59309 | 0.00787 | 0.21611 | 31.5 | 1 |
| ResNet50V2 | taiwanTomato | 0.62598 | 0.59060 | 0.00394 | 0.18307 | 0.64173 | 0.57734 | 0.03543 | 0.25968 | 37.5 | 1 |
| VGG16 | taiwanTomato | 0.59055 | 0.56380 | 0.03937 | 0.18926 | 0.65354 | 0.61630 | 0.03150 | 0.19438 | 92.5 | 5 |
| VGG19 | taiwanTomato | 0.61811 | 0.58424 | 0.02756 | 0.18938 | 0.60630 | 0.56301 | 0.04724 | 0.23638 | 84.5 | 1 |

Continued on next page

**Table C.2 - continued from previous page**

| Model | Dataset | Acc | F1 | Acc SD | F-1 SD | Acc FT | F1 FT | Acc SD FT | F-1 SD FT | Best Epoch | Best Epoch FT |
|---|---|---|---|---|---|---|---|---|---|---|---|
| Xception | taiwanTomato | 0.60236 | 0.56532 | 0.02756 | 0.15885 | 0.66142 | 0.63185 | 0.02362 | 0.19049 | 43 | 1.5 |
| ConvNeXtSmall | tea | 0.84637 | 0.84034 | 0.00838 | 0.10723 | 0.89106 | 0.88605 | 0.00279 | 0.08587 | 166.5 | 42.5 |
| ConvNeXtTiny | tea | 0.89106 | 0.88498 | 0.00279 | 0.09197 | 0.92458 | 0.91931 | 0.00279 | 0.06800 | 173 | 96.5 |
| DenseNet121 | tea | 0.87151 | 0.86606 | 0.00559 | 0.09215 | 0.89106 | 0.88905 | 0.00838 | 0.08281 | 143 | 44 |
| DenseNet169 | tea | 0.89106 | 0.89064 | 0.00838 | 0.06795 | 0.91341 | 0.91321 | 0.00279 | 0.05713 | 159.5 | 27.5 |
| DenseNet201 | tea | 0.88268 | 0.87815 | 0.01117 | 0.09176 | 0.88827 | 0.88163 | 0.01117 | 0.10691 | 119 | 23 |
| EfficientNetV2B0 | tea | 0.85475 | 0.85099 | 0.00559 | 0.09818 | 0.90782 | 0.90452 | 0.00838 | 0.07205 | 135.5 | 54 |
| EfficientNetV2B1 | tea | 0.86592 | 0.86597 | 0.00559 | 0.08003 | 0.88827 | 0.88973 | 0.00559 | 0.07433 | 121.5 | 12 |
| EfficientNetV2B2 | tea | 0.87151 | 0.86783 | 0.00000 | 0.09206 | 0.88827 | 0.88177 | 0.00000 | 0.09909 | 105.5 | 24 |
| EfficientNetV2B3 | tea | 0.88268 | 0.87568 | 0.01117 | 0.09357 | 0.90503 | 0.89872 | 0.00559 | 0.08280 | 110 | 15 |
| EfficientNetV2M | tea | 0.87989 | 0.87730 | 0.00279 | 0.07816 | 0.92737 | 0.92408 | 0.01117 | 0.06271 | 120.5 | 10.5 |
| EfficientNetV2S | tea | 0.84637 | 0.83814 | 0.01397 | 0.13286 | 0.88547 | 0.87965 | 0.01397 | 0.10119 | 138 | 35 |
| InceptionResNetV2 | tea | 0.74581 | 0.73822 | 0.01397 | 0.11962 | 0.82961 | 0.82566 | 0.01955 | 0.11707 | 81.5 | 2.5 |
| InceptionV3 | tea | 0.75419 | 0.74581 | 0.00559 | 0.09605 | 0.82123 | 0.81231 | 0.01117 | 0.10362 | 75.5 | 17 |
| MobileNetV3Large | tea | 0.91620 | 0.90992 | 0.00559 | 0.08557 | 0.92179 | 0.91417 | 0.00000 | 0.09072 | 182 | 50 |
| MobileNetV3Small | tea | 0.89665 | 0.89124 | 0.00279 | 0.08426 | 0.90782 | 0.90316 | 0.00279 | 0.07935 | 196 | 21 |

Continued on next page

**Table C.2 - continued from previous page**

| Model | Dataset | Acc | F1 | Acc SD | F-1 SD | Acc FT | F1 FT | Acc SD FT | F-1 SD FT | Best Epoch | Best Epoch FT |
|---|---|---|---|---|---|---|---|---|---|---|---|
| NASNetLarge | tea | 0.77095 | 0.76284 | 0.01676 | 0.10449 | 0.86034 | 0.85484 | 0.00559 | 0.08292 | 65.5 | 27.5 |
| NASNetMobile | tea | 0.76257 | 0.76145 | 0.00279 | 0.11468 | 0.82682 | 0.82378 | 0.01117 | 0.10563 | 146.5 | 27.5 |
| ResNet101V2 | tea | 0.81844 | 0.81320 | 0.00279 | 0.10909 | 0.88268 | 0.87957 | 0.00559 | 0.09470 | 89.5 | 13 |
| ResNet152V2 | tea | 0.80726 | 0.80332 | 0.00279 | 0.12217 | 0.86592 | 0.85995 | 0.01676 | 0.10792 | 77.5 | 16.5 |
| ResNet50V2 | tea | 0.83520 | 0.82847 | 0.00279 | 0.11025 | 0.87709 | 0.87065 | 0.02235 | 0.09898 | 83.5 | 4 |
| VGG16 | tea | 0.82961 | 0.82739 | 0.01397 | 0.09496 | 0.91061 | 0.90758 | 0.00000 | 0.06895 | 173 | 77 |
| VGG19 | tea | 0.86034 | 0.86039 | 0.00559 | 0.07989 | 0.92458 | 0.92041 | 0.01397 | 0.06838 | 132 | 51.5 |
| Xception | tea | 0.77933 | 0.77736 | 0.00279 | 0.13377 | 0.82123 | 0.82183 | 0.02235 | 0.10848 | 89.5 | 6.5 |
| ConvNeXtSmall | tomatoVillage | 0.87844 | 0.85596 | 0.00385 | 0.09042 | 0.90594 | 0.88736 | 0.00275 | 0.07742 | 136 | 15.5 |
| ConvNeXtTiny | tomatoVillage | 0.85534 | 0.83308 | 0.00495 | 0.11388 | 0.88724 | 0.86835 | 0.00495 | 0.08991 | 128.5 | 17 |
| DenseNet121 | tomatoVillage | 0.87459 | 0.84713 | 0.00000 | 0.09414 | 0.89439 | 0.87393 | 0.00330 | 0.08298 | 109 | 11.5 |
| DenseNet169 | tomatoVillage | 0.85919 | 0.83749 | 0.00110 | 0.10458 | 0.87514 | 0.85643 | 0.01155 | 0.09321 | 72.5 | 9.5 |
| DenseNet201 | tomatoVillage | 0.86359 | 0.84535 | 0.00330 | 0.10398 | 0.88834 | 0.87404 | 0.00935 | 0.07942 | 72 | 6.5 |
| EfficientNetV2B0 | tomatoVillage | 0.88669 | 0.86938 | 0.00330 | 0.09981 | 0.89164 | 0.87499 | 0.00715 | 0.09293 | 98.5 | 1 |
| EfficientNetV2B1 | tomatoVillage | 0.88944 | 0.87374 | 0.00935 | 0.06896 | 0.90374 | 0.89314 | 0.00165 | 0.06730 | 99 | 4.5 |
| EfficientNetV2B2 | tomatoVillage | 0.88614 | 0.86014 | 0.00605 | 0.11533 | 0.89329 | 0.87218 | 0.01320 | 0.09320 | 83.5 | 4 |

Continued on next page

**Table C.2 - continued from previous page**

| Model | Dataset | Acc | F1 | Acc SD | F-1 SD | Acc FT | F1 FT | Acc SD FT | F-1 SD FT | Best Epoch | Best Epoch FT |
|---|---|---|---|---|---|---|---|---|---|---|---|
| EfficientNetV2B3 | tomatoVillage | 0.88064 | 0.85709 | 0.00275 | 0.10317 | 0.89054 | 0.86953 | 0.00715 | 0.09445 | 85.5 | 2 |
| EfficientNetV2M | tomatoVillage | 0.86194 | 0.83387 | 0.00275 | 0.11037 | 0.88174 | 0.85978 | 0.02365 | 0.09754 | 98.5 | 4.5 |
| EfficientNetV2S | tomatoVillage | 0.87239 | 0.84801 | 0.00110 | 0.08970 | 0.89824 | 0.87401 | 0.00385 | 0.08204 | 81.5 | 5 |
| InceptionResNetV2 | tomatoVillage | 0.81023 | 0.77917 | 0.00715 | 0.10855 | 0.86799 | 0.84181 | 0.01760 | 0.09139 | 86.5 | 4 |
| InceptionV3 | tomatoVillage | 0.77503 | 0.74031 | 0.00165 | 0.16666 | 0.80913 | 0.77851 | 0.05226 | 0.14778 | 55 | 2.5 |
| MobileNetV3Large | tomatoVillage | 0.87734 | 0.84756 | 0.00275 | 0.11085 | 0.89769 | 0.87829 | 0.00550 | 0.09231 | 76.5 | 7.5 |
| MobileNetV3Small | tomatoVillage | 0.85809 | 0.83878 | 0.00220 | 0.08539 | 0.87624 | 0.85784 | 0.00385 | 0.08278 | 122 | 4 |
| NASNetLarge | tomatoVillage | 0.80363 | 0.76654 | 0.00495 | 0.11218 | 0.82783 | 0.79325 | 0.01815 | 0.11410 | 53 | 2.5 |
| NASNetMobile | tomatoVillage | 0.79978 | 0.76825 | 0.00660 | 0.09929 | 0.82453 | 0.78836 | 0.01485 | 0.12352 | 83 | 3.5 |
| ResNet101V2 | tomatoVillage | 0.82563 | 0.79410 | 0.00055 | 0.12222 | 0.85149 | 0.82471 | 0.00440 | 0.09901 | 50.5 | 4.5 |
| ResNet152V2 | tomatoVillage | 0.84158 | 0.80995 | 0.00440 | 0.10349 | 0.87239 | 0.83483 | 0.01760 | 0.09844 | 57.5 | 3.5 |
| ResNet50V2 | tomatoVillage | 0.83498 | 0.80572 | 0.00550 | 0.10196 | 0.85699 | 0.82987 | 0.01210 | 0.09980 | 52.5 | 3 |
| VGG16 | tomatoVillage | 0.83168 | 0.80292 | 0.00990 | 0.12369 | 0.86139 | 0.83633 | 0.01870 | 0.09466 | 101 | 4 |
| VGG19 | tomatoVillage | 0.82618 | 0.80987 | 0.00110 | 0.11067 | 0.87184 | 0.85418 | 0.00055 | 0.08572 | 94.5 | 18 |
| Xception | tomatoVillage | 0.79868 | 0.76573 | 0.00220 | 0.12968 | 0.83388 | 0.80679 | 0.01210 | 0.10640 | 63 | 2.5 |

# D. Appendix Augmentation

In this appendix all the results used to achieve the augmentation benchmark results for plant leaf disease classification are listed in Table D.1. In the table Augmentation Combined corresponds to augmented all, Augmentation Transform corresponds to augmented trans, Augmentation Color is represented as augmented color, Augmentation None simply as none, and Augmentation Noise is listed as augmented noise.

Table D.1: Full results obtained during augmentation experiments. Results originate from [133].

| Model | Dataset | Augmentation | Test Acc | Test F1 | Test Recall | Val Precision |
|---|---|---|---|---|---|---|
| MobileNetV3Large | novelPotato | augmented all | 98.69% | 98.97% | 98.69% | 98.69% |
| DenseNet121 | novelPotato | augmented all | 97.70% | 96.92% | 97.70% | 97.70% |
| DenseNet201 | novelPotato | augmented all | 98.03% | 97.95% | 98.03% | 98.03% |
| EfficientNetV2B0 | novelPotato | augmented all | 99.34% | 99.49% | 99.34% | 99.34% |
| EfficientNetV2B2 | novelPotato | augmented all | 98.36% | 97.89% | 98.36% | 98.36% |
| ResNet50V2 | novelPotato | augmented all | 96.39% | 96.13% | 96.39% | 96.39% |
| ResNet152V2 | novelPotato | augmented all | 96.72% | 95.91% | 96.72% | 96.72% |

Continued on next page

Table D.1 - continued from previous page

| Model | Dataset | Augmentation | Test Acc | Test F1 | Test Recall | Val Precision |
| --- | --- | --- | --- | --- | --- | --- |
| ConvNeXtSmall | novelPotato | augmented all | 99.02% | 98.96% | 99.02% | 99.02% |
| VGG19 | novelPotato | augmented all | 96.39% | 95.31% | 96.39% | 96.39% |
| MobileNetV3Large | pld | augmented all | 99.75% | 99.78% | 99.75% | 99.75% |
| DenseNet121 | pld | augmented all | 100.00% | 100.00% | 100.00% | 100.00% |
| DenseNet201 | pld | augmented all | 99.75% | 99.72% | 99.75% | 99.75% |
| EfficientNetV2B0 | pld | augmented all | 99.75% | 99.78% | 99.75% | 99.75% |
| EfficientNetV2B2 | pld | augmented all | 100.00% | 100.00% | 100.00% | 100.00% |
| ResNet50V2 | pld | augmented all | 99.26% | 99.28% | 99.26% | 99.26% |
| ResNet152V2 | pld | augmented all | 100.00% | 100.00% | 100.00% | 100.00% |
| ConvNeXtSmall | pld | augmented all | 99.75% | 99.72% | 99.75% | 99.75% |
| VGG19 | pld | augmented all | 99.75% | 99.78% | 99.75% | 99.75% |
| MobileNetV3Large | novelPotato | augmented color | 99.02% | 99.23% | 99.02% | 99.02% |
| DenseNet121 | novelPotato | augmented color | 96.72% | 96.22% | 96.72% | 96.72% |
| DenseNet201 | novelPotato | augmented color | 96.39% | 96.13% | 96.39% | 96.39% |
| EfficientNetV2B0 | novelPotato | augmented color | 98.36% | 98.71% | 98.36% | 98.36% |
| EfficientNetV2B2 | novelPotato | augmented color | 98.36% | 97.89% | 98.36% | 98.36% |
| ResNet50V2 | novelPotato | augmented color | 94.10% | 92.90% | 94.10% | 94.10% |
| ResNet152V2 | novelPotato | augmented color | 92.79% | 92.55% | 92.79% | 92.79% |

Continued on next page

**Table D.1 - continued from previous page**

| Model | Dataset | Augmentation | Test Acc | Test F1 | Test Recall | Val Precision |
|---|---|---|---|---|---|---|
| ConvNeXtSmall | novelPotato | augmented color | 99.02% | 99.23% | 99.02% | 99.02% |
| VGG19 | novelPotato | augmented color | 92.46% | 90.38% | 92.46% | 92.46% |
| MobileNetV3Large | pld | augmented color | 99.51% | 99.51% | 99.51% | 99.75% |
| DenseNet121 | pld | augmented color | 99.26% | 99.34% | 99.26% | 99.26% |
| DenseNet201 | pld | augmented color | 99.75% | 99.78% | 99.75% | 99.75% |
| EfficientNetV2B0 | pld | augmented color | 99.51% | 99.56% | 99.51% | 99.51% |
| EfficientNetV2B2 | pld | augmented color | 100.00% | 100.00% | 100.00% | 100.00% |
| ResNet50V2 | pld | augmented color | 98.77% | 98.65% | 98.77% | 98.77% |
| ResNet152V2 | pld | augmented color | 98.77% | 98.74% | 98.77% | 98.77% |
| ConvNeXtSmall | pld | augmented color | 100.00% | 100.00% | 100.00% | 100.00% |
| VGG19 | pld | augmented color | 99.26% | 99.21% | 99.26% | 99.26% |
| MobileNetV3Large | novelPotato | augmented noise | 96.39% | 95.77% | 96.39% | 96.39% |
| DenseNet121 | novelPotato | augmented noise | 96.07% | 95.18% | 96.07% | 96.07% |
| DenseNet201 | novelPotato | augmented noise | 95.08% | 93.99% | 95.08% | 95.08% |
| EfficientNetV2B0 | novelPotato | augmented noise | 97.70% | 97.68% | 97.70% | 97.70% |
| EfficientNetV2B2 | novelPotato | augmented noise | 98.03% | 97.38% | 98.03% | 98.03% |
| ResNet50V2 | novelPotato | augmented noise | 92.79% | 91.12% | 92.79% | 92.79% |
| ResNet152V2 | novelPotato | augmented noise | 94.43% | 93.96% | 94.43% | 94.43% |

Continued on next page

**Table D.1 - continued from previous page**

| Model | Dataset | Augmentation | Test Acc | Test F1 | Test Recall | Val Precision |
|---|---|---|---|---|---|---|
| ConvNeXtSmall | novelPotato | augmented noise | 98.69% | 98.42% | 98.69% | 98.69% |
| VGG19 | novelPotato | augmented noise | 89.51% | 87.30% | 89.51% | 89.51% |
| MobileNetV3Large | pld | augmented noise | 99.01% | 98.93% | 99.01% | 99.01% |
| DenseNet121 | pld | augmented noise | 98.77% | 98.84% | 98.77% | 98.77% |
| DenseNet201 | pld | augmented noise | 99.01% | 99.06% | 99.01% | 99.01% |
| EfficientNetV2B0 | pld | augmented noise | 98.77% | 98.63% | 98.77% | 98.77% |
| EfficientNetV2B2 | pld | augmented noise | 99.51% | 99.45% | 99.51% | 99.51% |
| ResNet50V2 | pld | augmented noise | 99.26% | 99.28% | 99.26% | 99.26% |
| ResNet152V2 | pld | augmented noise | 98.02% | 98.11% | 97.78% | 98.02% |
| ConvNeXtSmall | pld | augmented noise | 99.75% | 99.72% | 99.75% | 99.75% |
| VGG19 | pld | augmented noise | 99.26% | 99.34% | 99.01% | 99.26% |
| MobileNetV3Large | novelPotato | augmented trans | 97.70% | 97.68% | 97.70% | 97.70% |
| DenseNet121 | novelPotato | augmented trans | 98.03% | 97.94% | 98.03% | 98.03% |
| DenseNet201 | novelPotato | augmented trans | 98.36% | 98.18% | 98.36% | 98.36% |
| EfficientNetV2B0 | novelPotato | augmented trans | 97.70% | 97.35% | 97.70% | 97.70% |
| EfficientNetV2B2 | novelPotato | augmented trans | 98.36% | 97.89% | 98.36% | 98.68% |
| ResNet50V2 | novelPotato | augmented trans | 95.41% | 93.80% | 95.41% | 95.41% |
| ResNet152V2 | novelPotato | augmented trans | 96.72% | 95.44% | 96.72% | 96.72% |

Continued on next page

**Table D.1 - continued from previous page**

| Model | Dataset | Augmentation | Test Acc | Test F1 | Test Recall | Val Precision |
|---|---|---|---|---|---|---|
| ConvNeXtSmall | novelPotato | augmented trans | 98.03% | 97.92% | 98.03% | 98.03% |
| VGG19 | novelPotato | augmented trans | 95.74% | 95.50% | 95.74% | 95.74% |
| MobileNetV3Large | pld | augmented trans | 100.00% | 100.00% | 100.00% | 100.00% |
| DenseNet121 | pld | augmented trans | 100.00% | 100.00% | 100.00% | 100.00% |
| DenseNet201 | pld | augmented trans | 100.00% | 100.00% | 100.00% | 100.00% |
| EfficientNetV2B0 | pld | augmented trans | 99.51% | 99.44% | 99.51% | 99.51% |
| EfficientNetV2B2 | pld | augmented trans | 100.00% | 100.00% | 100.00% | 100.00% |
| ResNet50V2 | pld | augmented trans | 100.00% | 100.00% | 100.00% | 100.00% |
| ResNet152V2 | pld | augmented trans | 99.75% | 99.78% | 99.75% | 99.75% |
| ConvNeXtSmall | pld | augmented trans | 99.75% | 99.78% | 99.75% | 99.75% |
| VGG19 | pld | augmented trans | 99.75% | 99.73% | 99.75% | 99.75% |
| MobileNetV3Large | novelPotato | | 96.07% | 95.18% | 96.07% | 96.07% |
| DenseNet121 | novelPotato | | 96.39% | 95.11% | 96.39% | 96.71% |
| DenseNet201 | novelPotato | | 97.05% | 97.13% | 97.05% | 97.37% |
| EfficientNetV2B0 | novelPotato | | 94.75% | 92.10% | 94.75% | 94.75% |
| EfficientNetV2B2 | novelPotato | | 97.70% | 97.41% | 97.70% | 97.70% |
| ResNet50V2 | novelPotato | | 95.08% | 93.95% | 95.08% | 95.39% |
| ResNet152V2 | novelPotato | | 93.44% | 91.97% | 93.44% | 93.44% |

Continued on next page

**Table D.1 - continued from previous page**

| Model | Dataset | Augmentation | Test Acc | Test F1 | Test Recall | Val Precision |
|---|---|---|---|---|---|---|
| ConvNeXtSmall | novelPotato | | 98.03% | 97.90% | 98.03% | 98.03% |
| VGG19 | novelPotato | | 93.77% | 93.11% | 93.77% | 93.77% |
| MobileNetV3Large | pld | | 99.26% | 99.15% | 99.26% | 99.26% |
| DenseNet121 | pld | | 99.26% | 99.17% | 99.26% | 99.26% |
| DenseNet201 | pld | | 99.75% | 99.72% | 99.75% | 99.75% |
| EfficientNetV2B0 | pld | | 99.26% | 99.19% | 99.26% | 99.26% |
| EfficientNetV2B2 | pld | | 99.51% | 99.44% | 99.51% | 99.51% |
| ResNet50V2 | pld | | 99.01% | 98.90% | 99.01% | 99.01% |
| ResNet152V2 | pld | | 99.26% | 99.23% | 99.26% | 99.26% |
| ConvNeXtSmall | pld | | 99.75% | 99.72% | 99.75% | 99.75% |
| VGG19 | pld | | 99.01% | 99.00% | 99.01% | 99.01% |

# E. Dataset Construction

The full list of plants and classes included in the PLDC-80 dataset. Each class has 3,500 training images.

```
PLDC-80
  └ cassava_cassava_Cassava Bacterial Blight (CBB)
  └ cassava_cassava_Cassava Brown Streak Disease (CBSD)
  └ cassava_cassava_Cassava Green Mottle (CGM)
  └ cassava_cassava_Cassava Mosaic Disease (CMD)
  └ cassava_cassava_Healthy
  └ cds_corn_Gray Leaf Spot_AND_plantvillage_Corn_Cercospora_leaf_spot Gray_leaf_spot
  └ cds_corn_Northern Leaf Blight_AND_plantvillage_Corn_Northern_Leaf_Blight
  └ cds_corn_Northern Leaf Spot
  └ diamos_pear_slug
  └ diamos_pear_spot
  └ fgvc8_apple_frog_eye_leaf_spot
  └ fgvc8_apple_healthy_AND_plantvillage_Apple___healthy
  └ fgvc8_apple_powdery_mildew
  └ fgvc8_apple_rust_AND_plantvillage_Apple___Cedar_apple_rust
  └ fgvc8_apple_scab_AND_plantvillage_Apple___Apple_scab
  └ paddy_rice_bacterial_leaf_blight
  └ paddy_rice_bacterial_leaf_streak
  └ paddy_rice_bacterial_panicle_blight
  └ paddy_rice_blast
  └ paddy_rice_brown_spot
  └ paddy_rice_dead_heart
  └ paddy_rice_downy_mildew
  └ paddy_rice_hispa
  └ paddy_rice_normal
  └ paddy_rice_tungro
  └ pdd271_Soybean_downy_mildew_135
```

└ pdd271_Mung_bean_brown_spot_246

└ pdd271_Sweet_potato_healthy_leaf_220

└ pdd271_Sweet_potato_magnesium_deficiency_227

└ pdd271_Sweet_potato_sooty_mold_224

└ pdd271_leek_gray_mold_disease_339

└ pdd271_leek_hail_damage_338

└ pdd271_radish_black_spot_disease_297

└ pdd271_radish_mosaic_virus_disease_295

└ pdd271_radish_wrinkle_virus_disease_293

└ plantvillage_Apple___Black_rot

└ plantvillage_Blueberry___healthy

└ plantvillage_Cherry_(including_sour)___Powdery_mildew

└ plantvillage_Cherry_(including_sour)___healthy

└ plantvillage_Corn_(maize)___Common_rust_

└ plantvillage_Corn_(maize)___healthy

└ plantvillage_Grape___Black_rot

└ plantvillage_Grape___Esca_(Black_Measles)

└ plantvillage_Grape___Leaf_blight_(Isariopsis_Leaf_Spot)

└ plantvillage_Grape___healthy

└ plantvillage_Orange___Haunglongbing_(Citrus_greening)

└ plantvillage_Peach___Bacterial_spot

└ plantvillage_Peach___healthy

└ plantvillage_Pepper,_bell___Bacterial_spot

└ plantvillage_Pepper,_bell___healthy

└ plantvillage_Potato___Early_blight

└ plantvillage_Potato___Late_blight

└ plantvillage_Raspberry___healthy

└ plantvillage_Soybean___healthy

└ plantvillage_Squash___Powdery_mildew

└ plantvillage_Strawberry___Leaf_scorch

└ plantvillage_Strawberry___healthy

└ plantvillage_Tomato___Bacterial_spot

└ plantvillage_Tomato___Early_blight

└ plantvillage_Tomato___Late_blight

└ plantvillage_Tomato___Leaf_Mold

└ plantvillage_Tomato___Septoria_leaf_spot

└ plantvillage_Tomato___Spider_mites Two └spotted_spider_mite

└ plantvillage_Tomato___Target_Spot

└ plantvillage_Tomato___Tomato_Yellow_Leaf_Curl_Virus

└ plantvillage_Tomato___Tomato_mosaic_virus

└ plantvillage_Tomato___healthy

└ sms_strawberry_angular_leafspot

└ sms_strawberry_leaf_spot

└ sms_strawberry_powdery_mildew_leaf

└ sugar_cane_Banded Chlorosis

└ sugar_cane_Brown Spot

└ sugar_cane_BrownRust

└ sugar_cane_Grassy shoot

└ sugar_cane_Healthy Leaves

└ sugar_cane_Pokkah Boeng

└ sugar_cane_Sett Rot

└ sugar_cane_Viral Disease

└ sugar_cane_Yellow Leaf

└ sugar_cane_smut

# F. Appendix PLDC-6 Datasets

## F.1 iBean

Table F.1: Information about the iBean Dataset.

| Plants | Beans |
|---|---|
| Diseases | 1. Healthy<br>2. Angular Leaf Spot<br>3. Bean Rust |
| Number of Classes | 3 |
| Number of Images | 1,296 |

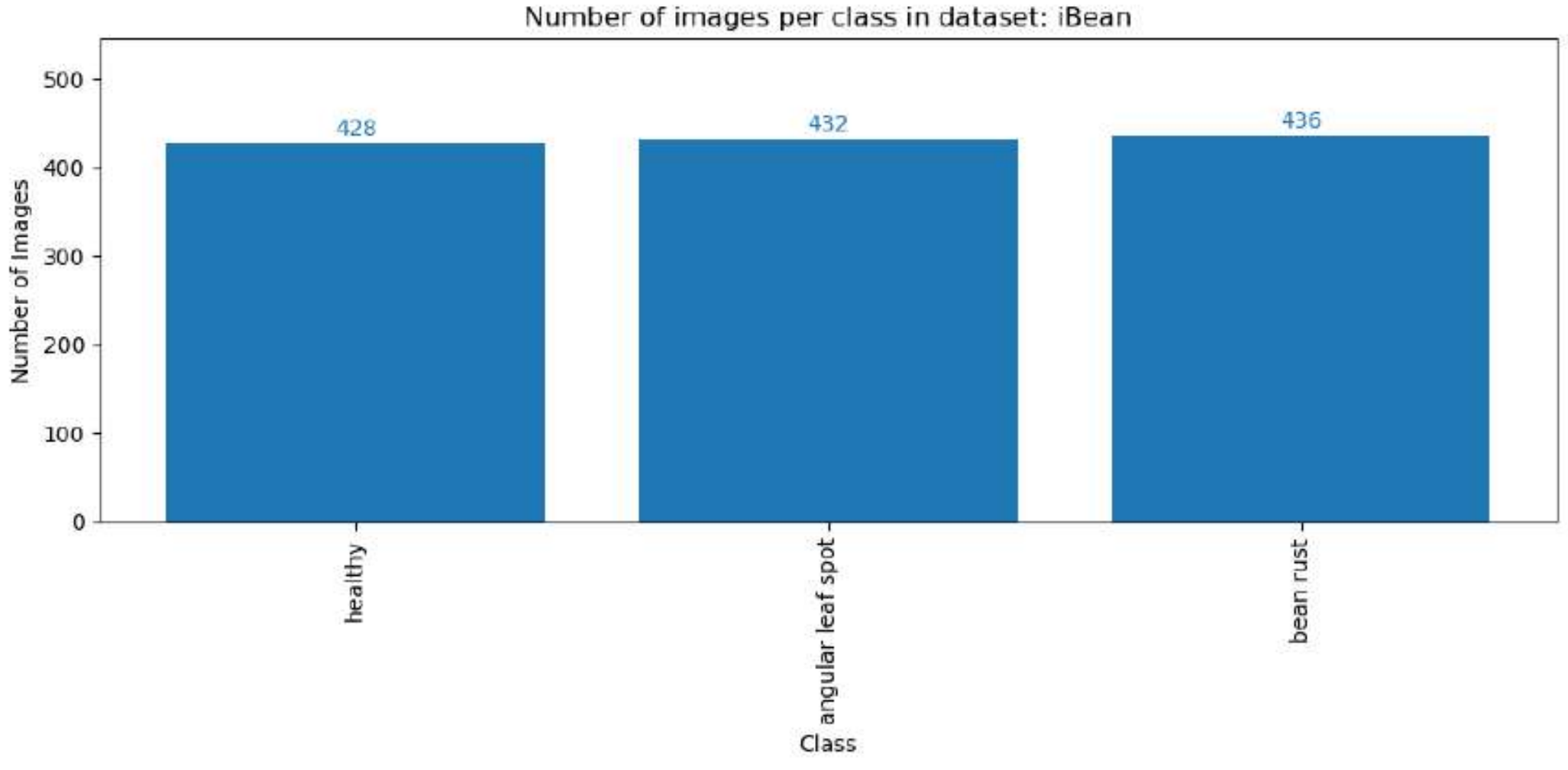

Figure F.1: Class Distribution of the iBean Dataset.

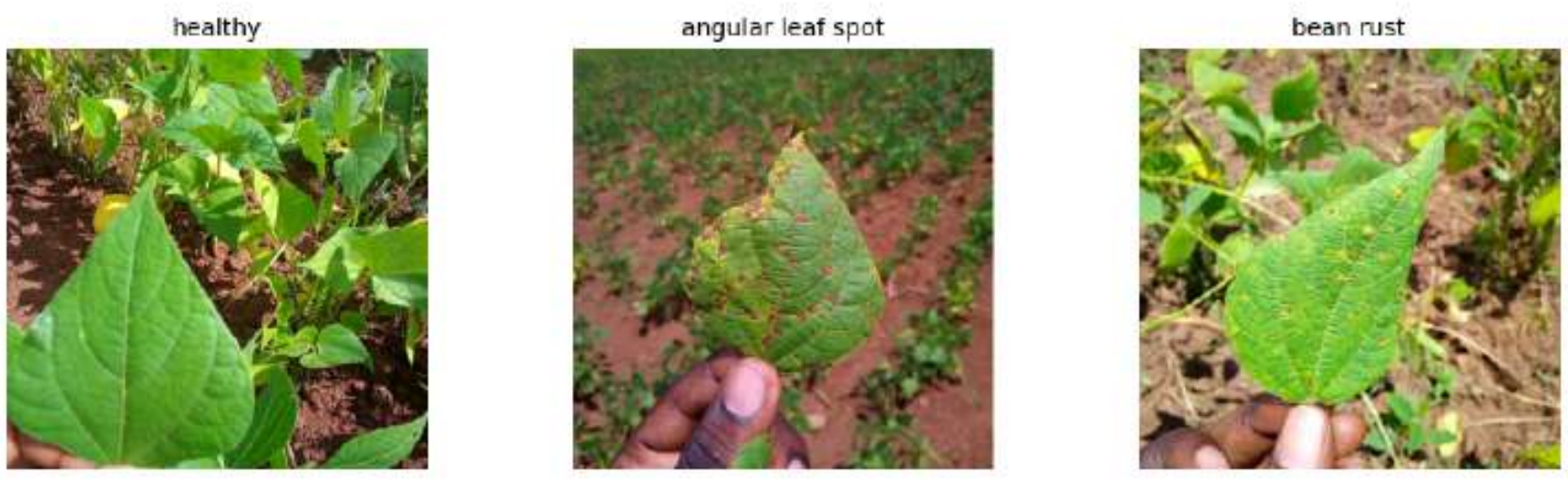

Figure F.2: A sample image of each class present in the iBean Dataset.

## F.2 Soybean

Table F.2: Information about the Soybean Dataset.

| Plants | Soybean images dataset |
|---|---|
| Diseases | 1. Healthy<br>2. Caterpillar<br>3. Diabrotica speciosa |
| Number of Classes | 3 |
| Number of Images | 6,410 |

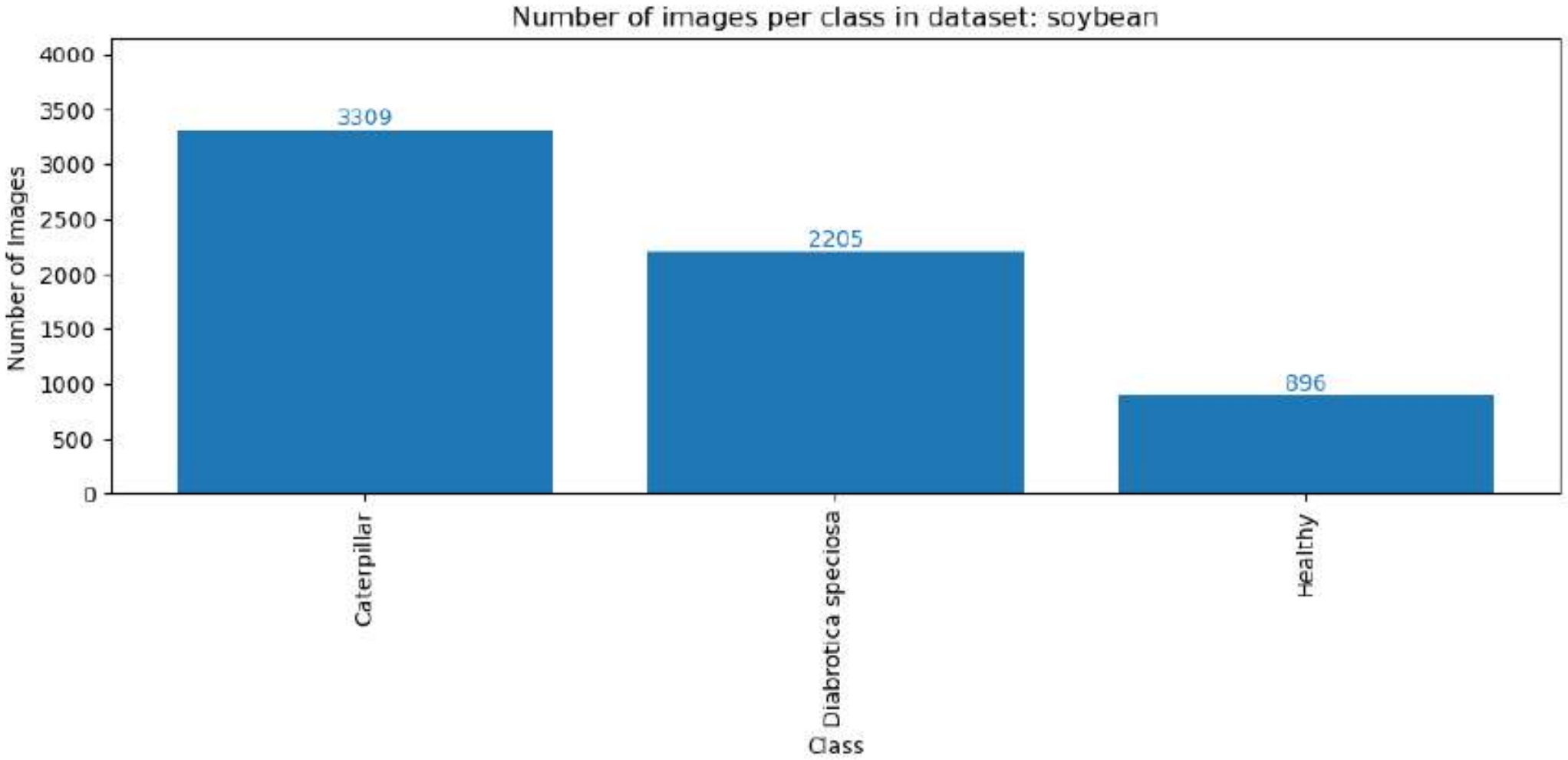

Figure F.3: Class Distribution of the Soybean Dataset.

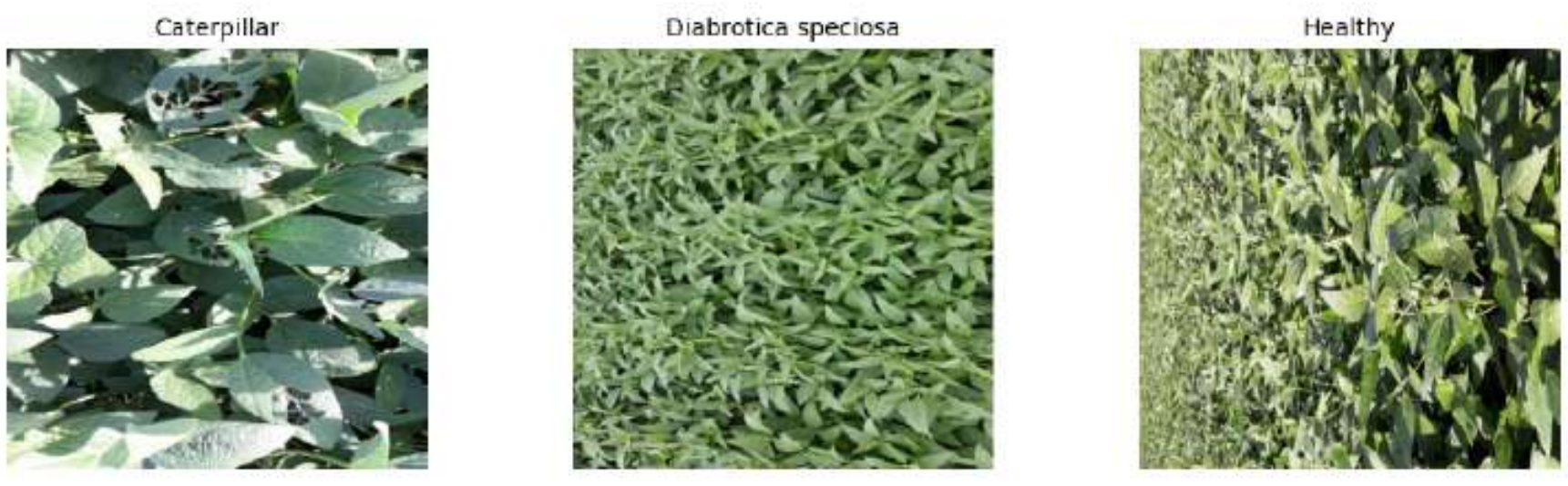

Figure F.4: A sample image of each class present in the Soybean Dataset.

## F.3 Classes PLDC-6

```
PLDC-6
  └ soybean_Healthy
  └ soybean_Caterpillar
  └ soybean_Diabrotica speciosa
  └ iBean_Healthy
  └ iBean_Angular Leaf Spot
  └ iBean_Bean Rust
```

To my friends and family.

# Acknowledgments

I want to thank my lab mates for helping me when I had any questions regarding this thesis or any of the other projects I worked on, my friends (both here in Korea and as well as back home and all over the world) and family for being there for me during the duration of this work. I also want to thank my professors here at CNU and outside that gave me valuable advice and always had an open ear. I also appreciate the high-performance GPU computing support of HPC-AI Open Infrastructure via GIST SCENT, without which many of the experiments in this thesis would not have been possible.